\documentclass[11pt, a4paper, twoside, openright]{book}

\usepackage[round]{natbib}  
\usepackage{style}
\usepackage{url}
\usepackage{quoting}  
\usepackage{caption}
\usepackage{subcaption} 
\usepackage{graphicx}
\usepackage{multirow}
\usepackage{amsfonts}
\usepackage{amsmath}
\usepackage{bm}  
\usepackage{relsize}
\usepackage{tabularx}
\usepackage{enumitem}
\usepackage{rotating}  
\usepackage[para,online,flushleft]{threeparttable}
\usepackage[ruled,vlined,noend,linesnumbered]{algorithm2e}
\usepackage{afterpage}
\usepackage{multicol}
\usepackage{arydshln}

\DeclareMathOperator*{\argminA}{arg\,min}

\newcommand{\MetaLoss}{\mathcal{M}}
\newcommand{\Loss}{\mathcal{L}}
\newcommand{\Task}{\mathcal{T}}
\newcommand{\Transpose}{\mathsf{T}}
\newcommand{\Dataset}{\mathcal{D}}
\newcommand{\Fitness}{\mathcal{F}}
\newcommand{\Warp}{\bm{\omega}}
\newcommand{\Update}{\mathrm{U}}
\newcommand{\Regularize}{\mathcal{R}}

\begin{document}

\frontmatter


\title{Meta-Learning Loss Functions for Deep Neural Networks}
\author{Christian Raymond}
\subject{Artificial Intelligence}

\abstract{
Humans can often quickly and efficiently solve new complex learning tasks given only a small set of examples. In contrast, modern artificially intelligent systems often require thousands or millions of observations in order to solve even the most basic tasks. Meta-learning aims to resolve this issue by leveraging past experiences from similar learning tasks to embed the appropriate inductive biases into the learning system. Historically methods for meta-learning components such as optimizers, parameter initializations, and more have led to significant performance increases. This thesis aims to explore the concept of meta-learning to improve performance, through the often-overlooked component of the loss function. The loss function is a vital component of a learning system, as it represents the primary learning objective, where success is determined and quantified by the system's ability to optimize for that objective successfully. 

In this thesis, we developed methods for meta-learning the loss function of deep neural networks. In particular, we first introduced a method for meta-learning symbolic model-agnostic loss function called \textit{Evolved Model Agnostic Loss} (EvoMAL). This method consolidates recent advancements in loss function learning and enables the development of interpretable loss functions on commodity hardware. Through empirical and theoretical analysis, we uncovered patterns in the learned loss functions, which later inspired the development of \textit{Sparse Label Smoothing Regularization} (SparseLSR), which is a significantly faster and more memory-efficient way to perform label smoothing regularization. Second, we challenged the conventional notion that a loss function must be a static function by developing \textit{Adaptive Loss Function Learning} (AdaLFL), a method for meta-learning adaptive loss functions. Lastly, we developed \textit{Neural Procedural Bias Meta-Learning} (NPBML) a task-adaptive few-shot learning method that meta-learns the parameter initialization, optimizer, and loss function simultaneously.
}

\phd


\maketitle

\chapter*{Acknowledgments}

I would first like to extend my gratitude to my amazing supervisors, Dr. Qi Chen, Prof. Bing Xue, and Prof. Mengjie Zhang, whom I have been working closely with for the last six years. Qi has always been incredibly supportive of me and my research. No matter how far down the rabbit hole or deep into the math I ventured, she was always willing to work through it with me. Her genuine research curiosity is something to be admired. Bing is a highly capable and extremely hard-working individual. Her guidance from the very start has consistently steered me in the right direction and helped me derive deep insights that I would have otherwise overlooked. Finally, I am thankful to Meng for his endless support and enthusiasm, which has motivated me and many others to pursue higher study. His leadership and influence will undoubtedly have a lasting impact on the field.

I would also like to express my thanks to staff members such as Yi, Marcus, Pondy, Bach, Andrew, Ying, and Fangfang who have offered support and invaluable discussions about research. Special thanks also go to the many friends I have made during my PhD, for creating a happy and enjoyable research environment. I would like to thank: Kaan, Hayden, Jordan, Yifan, Zhixing, Hengzhe, Junhao, Dylon, Qinyu, Nora, Ziyi, Jiabin, Qinglan, Peng, Joao, Harisu, Jesse, Carl, Rimas, and many others, whose presence made my time at the evolutionary computation research group unforgettable.

Last but not least, I would like to dedicate this thesis to my amazing parents, Gordon and Teresa, my partner, Janel, and the rest of my extended family $\heartsuit$. I am so grateful for their endless love, encouragement, and support, which has enabled me to take the trajectory I have taken. My parents' tireless support throughout my academic journey, from primary school to university, has been the driving force behind my success. Additionally, Janel has been my supportive backbone, offering love and encouragement, and motivating me to be the best person I can be.

\addtocontents{toc}{\vskip -0.1cm}  
\tableofcontents

\mainmatter
\pagestyle{fancy}

\chapter{Introduction}\label{chapter:introduction}

\textit{In this chapter, we introduce the central topic of this thesis --- meta-learning, and how it can be used to enhance the learning capabilities of modern artificial intelligence systems. Following this, we outline the high-level problem statement that will guide our research, supported by further discussions on meta-learning and loss functions. Finally, we conclude this chapter by detailing the research goals of this thesis.}

\section{Introduction}

Humans have an exceptional ability to learn new tasks from a very modest set of observations. We can often quickly adapt to new domains proficiently by building upon and utilizing past experiences of related tasks in combination with a small amount of information about the target domain \citep{lake2015human}. The utilization of shared task regularity learned from past experiences allows humans to achieve strong performance on a wide range of diverse tasks, demonstrating a relatively general form of intelligence. In contrast to this, modern \textbf{Artificial Intelligence} and \textbf{Machine Learning} systems are typically highly specialized. They often require thousands or even millions of observations to achieve competitive performance on a single task. This lack of sample-efficiency is a consequence of the ongoing treatment of end-to-end learning from scratch being considered the gold standard for learning \citep{collobert2011natural, yi2014learning}; pragmatically motivated by the ability to learn without requiring embedding substantial amounts of human knowledge and expertise into the learning process.

However, as pointed out by \cite{finn2018learning}, it makes little sense to have systems that learn each task \textit{de novo}, with a blank slate each time, as this would be like asking a (human) baby to learn to program a computer before having developed a basic understanding of concepts related to language, perception or reasoning. To narrow this gap between our intelligent systems and humans, it is evident that the notion of starting from scratch must be reconsidered in favor of a more sample-efficient approach to learning. A further limitation of conventional approaches to learning is that they typically have to \textit{a priori} assume a set of inductive biases, such as the class of searchable models, \textit{i.e.}, representational bias, or the learning rules used to find the set of model parameters, \textit{i.e.}, procedural bias \citep{gordon1995evaluation}. However, this approach is necessarily prone to sub-optimal performance as the set of inductive biases selected are seldom, if ever, the best ones for any given learning problem.

As such, an increasing amount of research has gone into investigating a learning paradigm called \textbf{Meta-Learning} \citep{hospedales2020meta, peng2020comprehensive, vanschoren2018meta, vilalta2002perspective}. Often informally referred to as the \textit{learning-to-learn} paradigm. Meta-learning aims to provide an alternative paradigm contrary to conventional AI approaches whereby intelligent systems leverage their past experiences over multiple learning episodes --- often (although not exclusively) covering a distribution of related tasks. Meta-learning utilizes past experiences to improve the future learning performances of a model, typically a deep neural network \citep{lecun2015deep, goodfellow2016deep}, by automatically selecting the suitable set of inductive biases (or a subset of them) for the given problem or family of related problems \citep{hospedales2020meta}.

Many meta-learning approaches have been proposed for optimizing various \textbf{Deep Learning} (DL) components \citep{goodfellow2016deep}. For example, early research on the topic explored using meta-learning for generating neural network learning rules \citep{schmidhuber1987evolutionary, schmidhuber1992learning, bengio1994use}, while more contemporary research has extended itself to learning activation functions \citep{ramachandran2017searching}, shared parameter initializations \citep{finn2017model}, and neural network architectures \citep{stanley2019designing} to whole learning algorithms from scratch \citep{real2020automl, co2021evolving} and many more \citep{houthooft2018evolved, ren2018learning, javed2019meta, liu2020evolving, li2020differentiable, baik2021meta, baik2023meta}.

\section{Problem Statement}

One key component that has been overlooked in the meta-learning literature until very recently is the \textbf{Loss Function} \citep{reed1999neural, friedman2001elements, goodfellow2016deep}. The loss function is a mathematical function that summarizes all the accuracies and inaccuracies of a complex machine learning system in a single number, a scalar value, allowing it to quantify its progress and learn. The system's success is determined and quantified by its ability to minimize that function \citep{reed1999neural}. A fundamental challenge in the selection process is ensuring that the chosen loss function faithfully encodes and represents the intended goal of the system. A poorly defined or goal-misaligned loss function is likely to result in unsatisfactory performance at no fault of the system --- instead, the fault is ours for badly specifying the goal.

In conventional approaches to designing intelligent systems, the selection of the loss function is made heuristically by selecting a suitable function from a modest set of applicable handcrafted loss functions \citep{janocha2017loss}. A limitation of this approach is that the set of handcrafted loss functions under consideration represents only a small fraction of the full space of possible loss functions. Furthermore, handcrafted loss functions are customarily designed with task generality in mind, \textit{i.e.}, large expansive classes of tasks in mind, but the system itself is only concerned with a single instantiation of that class. Consequently, when using these handcrafted loss functions in practice, there is a negative cost associated with this generality in terms of system performance, \textit{i.e.}, there exists a generality-specificity trade-off. 

Based on the aforementioned limitations of conventional approaches to loss function design and selection, meta-learning research has recently begun investigating the emergent topic of \textbf{Loss Function Learning} \citep{gonzalez2020improved, bechtle2021meta, hospedales2020meta}, which aims to infer task-specific loss functions directly from the data. Preliminary results using meta-learned loss functions show improved sample efficiency, convergence, and asymptotic performance across a number of tasks compared to handcrafted loss functions. Furthermore, the modularity of loss functions enables them to be trivially transferred to new models and tasks, a highly desirable characteristic not possible in other meta-learning regimes, where the learned component is a network architecture \citep{stanley2002evolving, elsken2019neural, stanley2019designing} or shared parameter initialization \citep{finn2017model, nichol2018first, antoniou2019train}.

Loss function learning has emerged as a promising area of research; however, several challenges and aspects have not yet been addressed. For example, many methods learn “black box” loss functions, which lack interpretability and require their parametric structure to be predetermined \citep{bechtle2021meta, gonzalez2021optimizing, gao2021searching}. Another limitation of existing methods is that the learned loss functions are assumed to be static throughout the learning process; however, this need not be the case, the loss function could adapt along with the base model. Finally, the role of loss function learning within the broader context of meta-learning warrants further investigation and analysis to better understand its full impact and utility to the field.

\section{Motivations}

In this section, we first elaborate on why we have chosen the paradigm of meta-learning to enhance the performance of deep neural networks. Subsequently, we provide the rationale behind why we have set our focus on the loss function by discussing the limitations inherent in the traditional design and the selection of them.

\subsection{Why Meta-Learning?}

\begin{figure}[t]
    \centering\captionsetup{justification=centering}
    \includegraphics[width=0.9\textwidth]{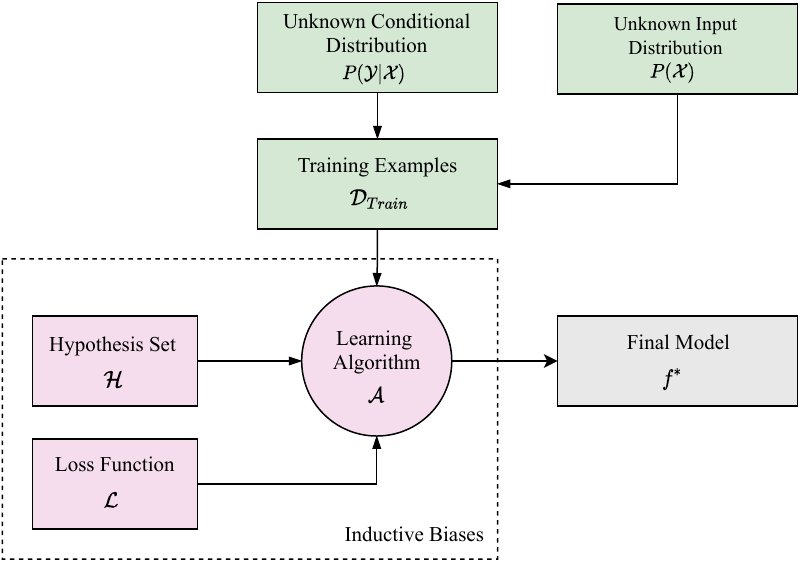}
    \caption{Conventional supervised learning problem setup.}
    \label{figure:conventional-learning-flowchart}
\end{figure}

In a prototypical supervised learning setup, we have inputs $x \in \mathcal{X}$ and outputs $y \in \mathcal{Y}$, which we use to construct a model/hypothesis $f: \mathcal{X} \rightarrow \mathcal{Y}$. Where $\mathcal{X}$ has some unknown input distribution $p(\mathcal{X})$, and we assume some unknown conditional distribution $p(\mathcal{Y}|\mathcal{X})$ which is what $f$ aims to learn. To learn $f$ we have at our disposal a dataset $\Dataset$, which is a set of independently and identically distributed (i.i.d.) examples drawn from some task $\Task$, which is constructed from a set of input-output pairs $\Dataset =\{(x_1, y_1), \dots, (x_n, y_n)\}$ where $n$ is the total number of instances. In the most simple case, $\Dataset$ is partitioned into two separate sets called the training set $\Dataset_{Train}$ and the testing set $\Dataset_{Test}$. In supervised learning, $\Dataset_{Train}$ is used for training $f$, while $\Dataset_{Test}$ is used to get an unbiased estimation of the performance of $f$ using out-of-sample observations. To construct the model $f$ using $\Dataset_{Train}$ a learning algorithm $\mathcal{A}$, a loss function $\Loss$, and hypothesis set $\mathcal{H}$ where $f \in \mathcal{H}$ must be selected. Figure \ref{figure:conventional-learning-flowchart} shows the summarized supervised learning setup where $\mathcal{A}(\Loss, \mathcal{H}, \Dataset_{Train}) \rightarrow f^{*}(x)$ .
\vspace{-2.5mm}
\subsubsection{Inductive Biases}

During the selection process of $\mathcal{A}$, $\Loss$, and $\mathcal{H}$, a set of assumptions are made about the data. Each of these assumptions is referred to as an \textbf{inductive bias} $\omega$, which in this context refers to an explicit or implicit assumption made by a learning algorithm $\mathcal{A}$ in order to perform induction --- that is, to generalize a finite set of observations into a general model $f$ of the domain \citep{abu2012learning}. There are two fundamental types of inductive biases, which we describe as follows:

\begin{itemize}

  \item \textbf{Representational (Declarative) Bias:} Specifies the representation of the space of hypotheses $\mathcal{H}$, which affects the size and functions contained in the search space \citep{gordon1995evaluation}. This in turn affects the set of feasible models that can be learned. For example, in the task of regression, one may constrain the final model to come from the set of linear functions such that $f_{\theta}(x) \in \mathcal{H} = \{\theta_{0} + \sum_{i}\theta_{i} \cdot x_{i} | \theta \in \mathbb{R}\}$, where $\theta$ is the model parameters.
  
  \item \textbf{Procedural (Algorithmic) Bias:} Determines the ordering of traversal over the solutions in the search space, as well as the selection of the final model $f^{*}$ \citep{gordon1995evaluation}. The most salient example of a procedural bias is the loss function, which quantifies the performance of a model throughout the learning process. For example, in the task of regression, the loss function used to optimize the weights $\theta$ could quantify the error as the absolute difference between the predictions and the ground truth labels $|y - f_{\theta}(x)|$, or alternatively the loss could be quantified as the squared difference $(y - f_{\theta}(x))^2$.
  
\end{itemize}

\begin{figure}[t]
    \centering\captionsetup{justification=centering}
    \includegraphics[width=0.9\textwidth]{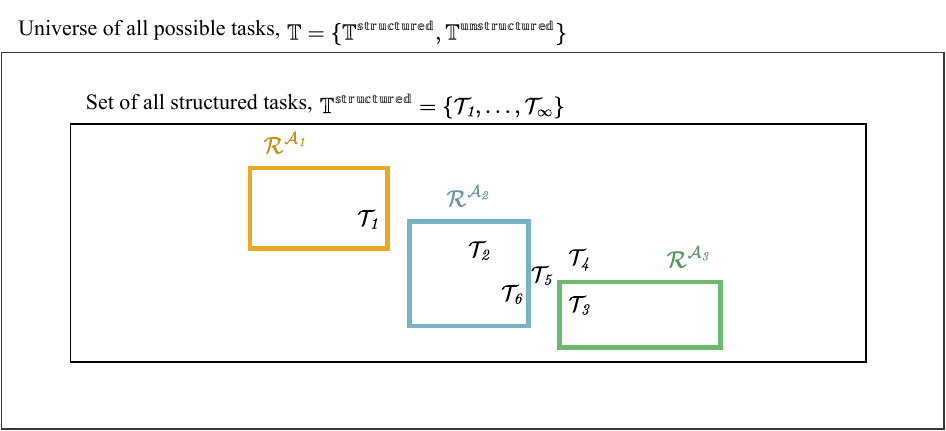}
    \caption{Each learning algorithm $\mathcal{A}$ covers a region of efficiency $\mathcal{R}_{\mathcal{A}}$ based on its inductive biases $\omega$. In this example, $\mathcal{A}_{1}$ can efficiently learn task $\Task_1$, while $\mathcal{A}_{2}$ can efficiently learn tasks $\Task_2$, and $\Task_6$, and finally $\mathcal{A}_3$ efficiently learn task $\Task_3$.}
    \label{fig:region-of-efficiency-1}
\end{figure}

\noindent
The selection of the inductive biases $\{\omega_1, \omega_2, \dots\}$ is \textit{extremely} consequential as it determines the region of efficiency of an algorithm, \textit{i.e}., the set of tasks $\{\Task_1, \Task_2, \dots \}$ that algorithm $\mathcal{A}$ can efficiently and effectively solve.  For example, in Figure \ref{fig:region-of-efficiency-1}, we visualize the space of all possible learning tasks $\mathbb{T}$ that an algorithm $\mathcal{A}$ can learn effectively over, where each algorithm has some region of efficiency denoted as $\mathcal{R_{A}} \in \mathbb{T}$. As illustrated, different algorithms have different regions of efficiency, which makes them better or worse at solving different tasks; for example, $\mathcal{A}_1$ can solve task $\Task_1$, but cannot solve tasks $\Task_2 - \Task_6$.

\subsubsection{No Free Lunch Theorem}

The \textit{No Free Lunch Theorems} (NFLT) by \cite{wolpert1997no} is a fundamental theory in machine learning that provides valuable insights into how we should reason about the selection of inductive biases. The NFLT tells us that across all possible learning tasks $\mathbb{T}$ no algorithm $\mathcal{A}$ can do better than another in expectation. If an algorithm gives better-than-average performance over one subset of problems, then there strictly must exist another subset of tasks where that algorithm performs worse than average. Crucially, all hope of finding more intelligent learning algorithms is not lost, as there are three important caveats to the NFLT \citep{ho2002simple}: 

\begin{itemize}

    \item \textbf{Conservation of Average Performance}: Crucially, the conservation of average performance in the NFLT only applies in expectation across all possible learning tasks $\mathbb{T}$. As such, the NFLT does not preclude the possibility of learning algorithms that excel on subsets of structured tasks $\mathbb{T}^{structured}$ that are profoundly important and relevant to humans, while performing worse on tasks of lesser importance such as random unstructured tasks $\mathbb{T}^{unstructured}$. 

    \item \textbf{Inductive Priors}: In theory selecting strong inductive biases that exploit the underlying structure of relevant learning tasks will improve performance. This is important because the problems we encounter in the real world are almost always highly structured; for example, there are regularities underlying vision and perception, different languages share some basic elements and linguistic structure, and the laws of physics apply to all things in the physical world. These crucial regularities underpin our ability to learn in the domains of computer vision, natural language processing, and robotics, respectively.

    \item \textbf{Performance and Sensitivity Tradeoffs}: In theory, if we know the underlying distribution of the tasks that we care about (e.g. if we know that we specifically care about task $\Task_{2} - \Task_{6}$ in Figure \ref{fig:region-of-efficiency-1}), then we can embed the appropriate inductive biases into our learning algorithm that would result in our learning algorithm having better performance than others on average with respect to those tasks. 
    
\end{itemize}

\vspace{5mm}\noindent\textbf{Open Problem}:
\textit{The NFLT tells us that if we know the underlying distribution of the tasks that we care about and we know what the appropriate inductive biases are that exploit the underlying structure, a better algorithm can be found for those tasks. However, the NFLT does not express how to identify the appropriate inductive biases for our learning algorithm.}

\subsubsection{Meta-Learning}

The field of meta-learning \citep{hospedales2020meta, peng2020comprehensive, vanschoren2018meta, vilalta2002perspective} provides a principled and intuitive method for tackling the challenges outlined in our discussion of the NFLT. Namely, identifying the relevant inductive biases that can significantly improve the performance of $\mathcal{A}$ across a set of learning tasks. Unlike the traditional guess-and-check method for selecting $\{\omega_1, \omega_2, \dots \}$, meta-learning seeks to create algorithms that can perform very efficiently over a distribution of related tasks $\{\Task_1, \Task_2, \dots \} \sim p(\Task)$. This is achieved by enabling these algorithms to learn at a meta-level --- essentially learning how to learn more efficiently and effectively. To learn how to learn most effectively, meta-learning algorithms are exposed to many related learning tasks, to identify underlying regularities between the tasks. The regularity learned from past experiences can then be embedded in $\mathcal{A}$ and leveraged to solve complex new learning tasks not seen previously more efficiently and effectively.   

\subsection{Why Loss Functions?}

Of particular consideration to this thesis is the design and selection process for determining which loss function $\Loss$ to use for a given learning algorithm $\mathcal{A}$. Careful consideration of the loss function is important as $\Loss$ determines both the learning path of $\mathcal{A}$ and the selection of the final model. The ultimate goal of a learning algorithm $\mathcal{A}$ is to find a model $f^*$ such that $\Loss$ is minimized. 
\begin{equation}\label{equation:learning-algorithm-goal}
f^{*} = \argminA_{f \in \mathcal{H}} \left[\Loss(y, f_{\theta}(x))\right]
\end{equation}
When constructing a learning algorithm a practitioner will typically select a handcrafted loss function from the literature that sufficiently matches the task under consideration. For example, if the task is a regression task the squared loss function may be selected. The practitioner may even try out multiple candidate loss functions to validate which loss function has the best performance empirically. In this subsection, we seek to demonstrate that the summarized methodology has numerous limitations and drawbacks.

\subsubsection{Limitations of Conventional Approaches to Loss Function Selection}

\afterpage{
    \begin{table}[t!]
    \centering
    \renewcommand{\arraystretch}{1.2}
    \captionsetup{justification=centering}
    \caption{Five loss functions for regression and their derivatives, where $e$ is the difference between the true and predicted label $e = (y - f_{\theta}(x))$ and $\delta$ is a tuning parameter.}
    \begin{threeparttable}
    \begin{tabular*}{0.9\textwidth}{l @{\extracolsep{\fill}} ccc}
        \hline
                                        & Loss Function & Derivative \\ \hline \hline
        Squared                         & $e^2$    & $2 \cdot e$ \\ 
        Pseudo-Huber \tnote{1}          & $\delta^2 \cdot [\sqrt{1 + (e/\delta)^2} - 1]$ & $e/(\delta \cdot \sqrt{e^2/\delta^2} + 1)$ \\
        Cauchy (Lorentzian) \tnote{2}   & $\log \cdot [1 + 0.5 \cdot (e/\delta)^2]$ & $(2 \cdot e)/(2\cdot\delta^2 + e^2)$ \\
        Geman-McClure \tnote{3}         & $e^2/(\delta + e^2)$ & $(2 \cdot \delta \cdot e)/(\delta + e^2)^2$ \\
        Welsh (Leclerc) \tnote{4}       & $\delta^2/2 \cdot [1 - \exp (-1 \cdot (e^2/2\cdot\delta^2))]$ & $0.5 \cdot e \cdot \exp^{-e^{2}/(2 \cdot \delta^2)}$ \\\hline
    \end{tabular*}
    
    \begin{tablenotes}\centering
    \scriptsize Loss Function References: \item[1] \citep{huber1964robust,charbonnier1994two} \item[2] \citep{black1996unification} \item[3] \citep{ganan1985bayesian} \item[4] \citep{dennis1978techniques,leclerc1989constructing}
    \end{tablenotes}

    \end{threeparttable}
    \label{table:loss-function-summary}
    \end{table}
    
    \begin{figure*}[t!]
    \centering
    \subfloat{{\includegraphics[width=1\textwidth]{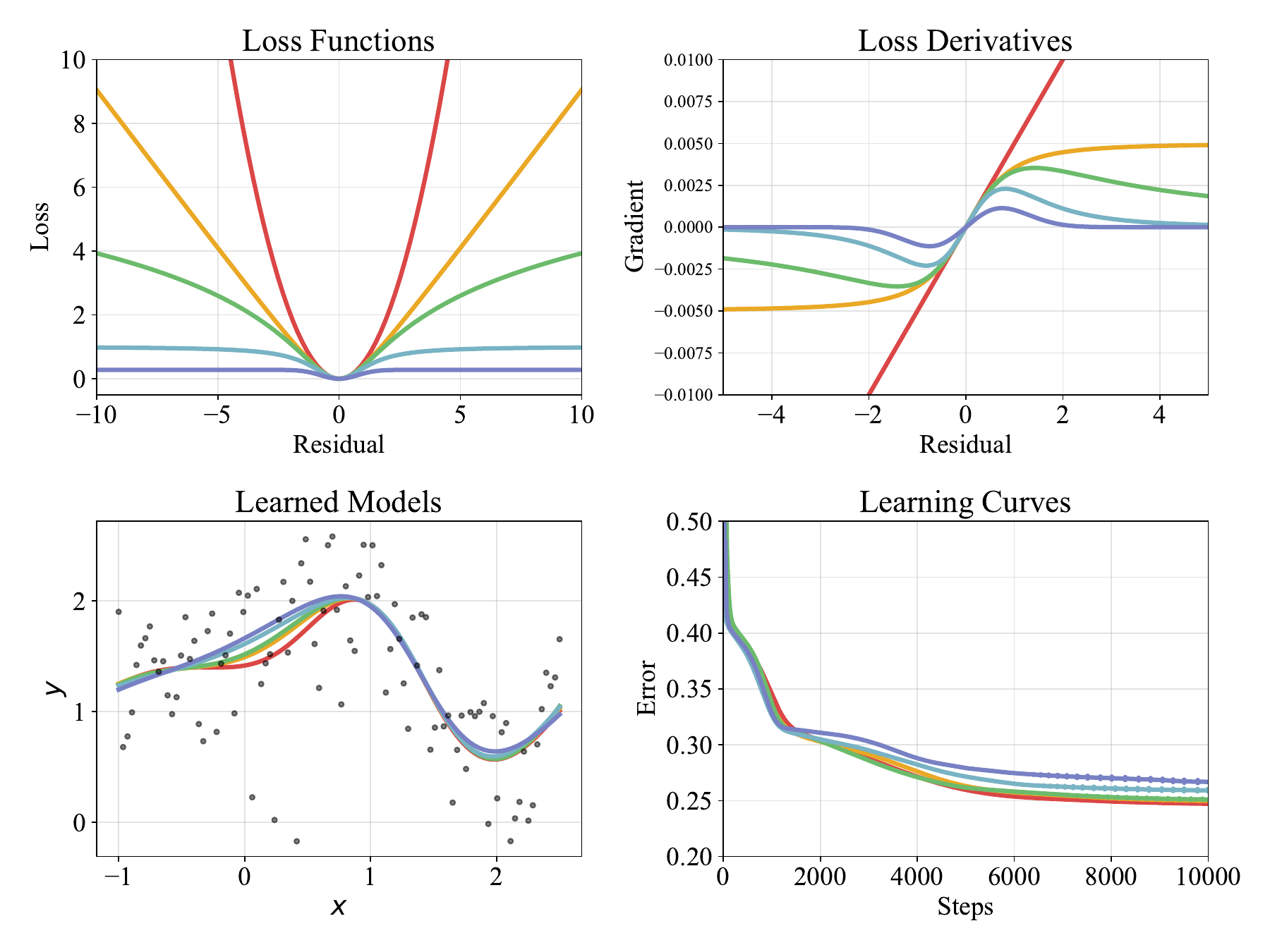}}}
    \qquad
    \subfloat{{\includegraphics[width=0.88\textwidth]{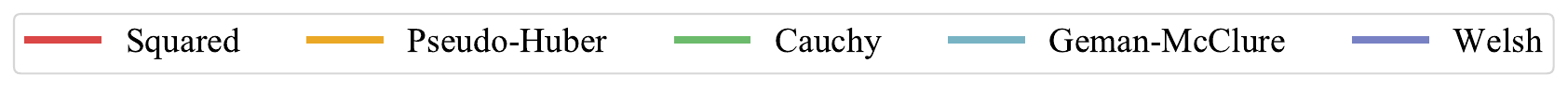}}}%
    \captionsetup{justification=centering}
    \caption{The loss functions and their corresponding derivatives (top), and the learned models with their corresponding learning curves (bottom) found using a set of simple feed-forward neural networks trained on a synthetic univariate regression task.}
    \label{figure:loss-function-summary}%
    \clearpage
\end{figure*}
}

The selection process of finding a suitable loss function is very consequential. To demonstrate this importance, we train a set of identical feed-forward neural networks on a simple regression task using an assortment of handcrafted loss functions used commonly in the machine learning literature. The selected loss functions and their corresponding derivatives are summarized in Table \ref{table:loss-function-summary} and are visualized in Figure \ref{figure:loss-function-summary}. As shown, the loss functions take on a distinct shape and scale; unsurprisingly, this results in different convergence behavior and a unique final model $f^{*}$. As demonstrated, the choice of loss function has a significant effect on the final learned model, even in the case of a simple univariate problem. Two key shortcomings of the trial-and-error (\textit{i.e.}, grid-search) approach to selecting loss functions are as follows:

\begin{itemize}

    \item \textbf{Search Space:} The set of handcrafted loss functions under consideration typically contains only a small fraction of the full space of possible loss functions for any given task. In the example shown in Figure \ref{figure:loss-function-summary}, only five loss functions were considered, but there exist vast quantities of loss functions in the literature \citep{janocha2017loss, barron2019general}. Moreover, even when considering just those presented in Table \ref{table:loss-function-summary}, many have a tuning parameter $\delta$, which needs to be searched for best performance to be achieved.
  
    \item \textbf{Task Specificity:} Handcrafted loss functions are prototypically designed with task-generality in mind, \textit{i.e.}, large expansive classes of tasks in mind; however, the practitioner is typically only concerned with one or a few instantiations of that class. As stated in the NFLT \citep{wolpert1997no} no algorithm $\mathcal{A}$ can do better than a random strategy in expectation --- this suggests that specialization to a subclass of tasks is in fact \textit{the only way} that improved performance can be achieved in general. As such, even in the idealized case, exhaustively searching the set of handcrafted loss functions would be prone to sub-optimal performance, since handcrafted loss functions are most commonly designed with task generality in mind. 
  
\end{itemize}

\subsubsection{Limitations of Conventional Approaches to Loss Function Design}

\begin{figure*}[t!]
    \centering
    \subfloat{{\includegraphics[width=0.45\textwidth]{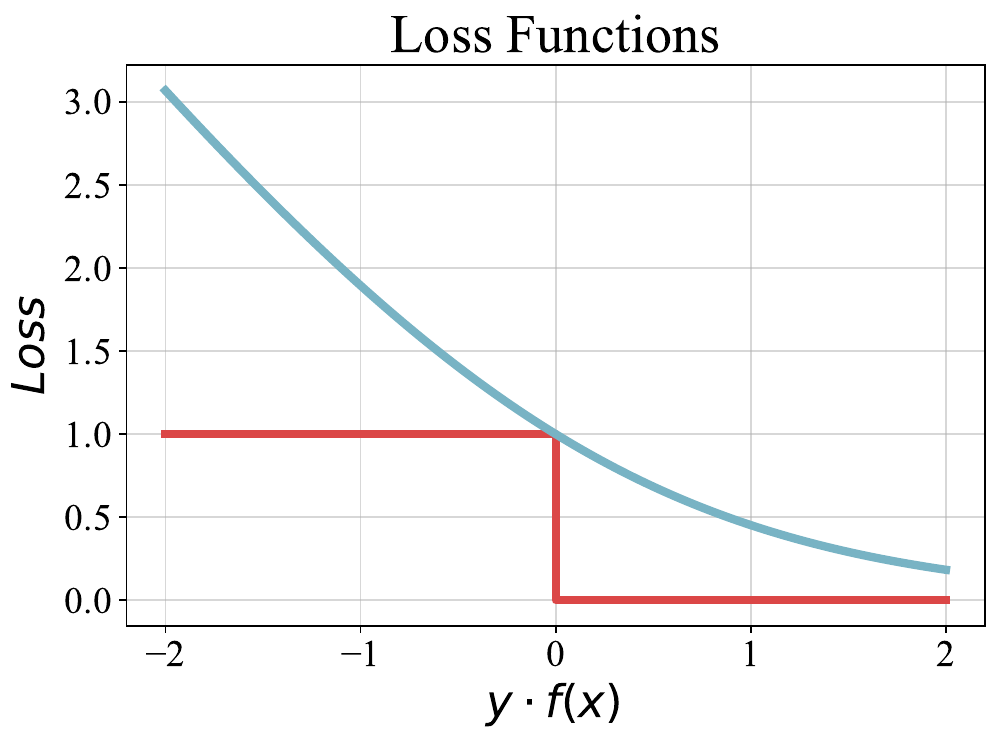}}}
    \qquad
    \subfloat{{\includegraphics[width=0.45\textwidth]{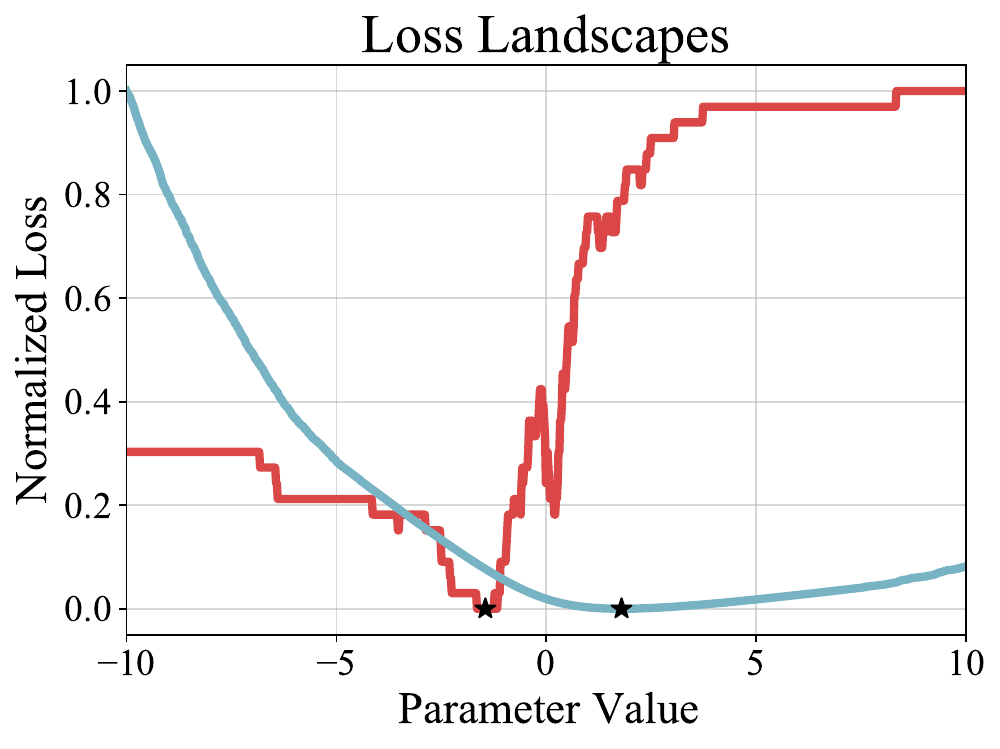}}}%
    \qquad
    \subfloat{{\includegraphics[width=0.35\textwidth]{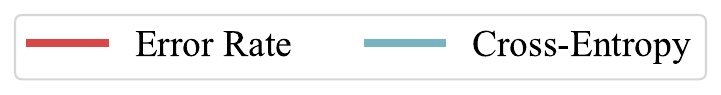}}}%
    \captionsetup{justification=centering}
    \caption{Visualizing the error rate and cross-entropy loss function (left), and comparing their loss landscapes right when using a simple logistic regression model trained on a synthetic classification task (right).}
    \label{fig:classification-loss-functions}%
    \clearpage
\end{figure*}

The loss function's role in a learning algorithm is to encode the system's end goal; thus, it is vital that the loss function faithfully encodes the system's end goal. Unfortunately, this is not trivial, as the loss function also directly influences the training dynamics of the model, as the geometry of the loss landscape is a function of the loss function, and the selected hypothesis set. There often exists a trade-off between how aligned the loss function is with the end goal and the training dynamics of the loss function as demonstrated in Figure \ref{fig:classification-loss-functions}. Further elaborating on the topics of goal alignment and training dynamics:

\begin{itemize}

    \item \textbf{Goal Alignment:} When the evaluation metric used to assess the performance of the learned model is not smooth or differentiable, an alternative function must be used as the loss function. In many cases, there is a misalignment between the true evaluation metric and its smooth differentiable approximation, as shown in Figure \ref{fig:classification-loss-functions} (right) where the optimal parameter value differs between the error rate and the log-loss, which results in sub-optimal learning performance. 
  
    \item \textbf{Training Dynamics:} A loss function is not simply a differentiable approximation of an evaluation metric, if this were the case many classification models would be directly optimizing differentiable approximations of the error rate evaluation metric (e.g. by Taylor expansion), for example. However, this is not the case since this would result in very poor training dynamics for the model as shown by the very difficult-to-optimize loss landscape of the error rate metric in Figure \ref{fig:classification-loss-functions} (right), which is non-convex and contains many flat spots rendering gradient-based methods useless. Consequently, loss functions that have better training dynamics such as the cross-entropy are used for training instead. 
  
\end{itemize}

\noindent
Based on the aforementioned limitations outlined in our discussion, we believe the loss function is a great candidate for being meta-learned. Traditional methods for designing and selecting loss functions often overlook task-specific information. In contrast, meta-learning enables the automatic development of highly performant loss functions tailored toward specific learning tasks. By meta-learning the loss function task-specific information about the data and model can be incorporated, thereby guiding the selection of inductive biases. Notably, unlike learned parameter initialization, the most popular component of a learning algorithm to meta-learn \citep{hospedales2020meta}, the loss function can be trivially transferred to new model architectures once meta-learned.

Automating the design and selection of the loss function opens up numerous new possibilities that challenge the conventional notion of what a loss function is and its role in a learning algorithm. For example, innovative work by \cite{antoniou2019learning} has shown that you can learn transductive loss functions, \textit{i.e.}, loss functions that learn without being given the ground truth labels. Another example is the work of \cite{bechtle2021meta}, who showed that you can shape the loss landscape by providing the loss function with additional task-related information. Finally, as will be shown in Chapter \ref{chapter:adalfl}, the loss function need not even be a static function, it can dynamically evolve throughout the learning process. 

\newpage

\section{Research Goals}

This thesis aims to enhance the learning capabilities of deep neural networks by advancing the emerging and nascent field of loss function learning. Its primary focus is on investigating how prior learning experiences can be leveraged to inform the automatic design and selection of the loss function, a fundamental component used in the training of machine learning models. The development of algorithms for meta-learning loss functions is expected to enhance the asymptotic performance and sample efficiency across both conventional supervised learning and few-shot learning tasks. To accomplish this goal, four key research objectives are established to guide the research:

\begin{enumerate}[leftmargin=*]

    \item \textbf{Develop a Method for Meta-Learning Symbolic Loss Functions:} \textit{Develop a new meta-learning framework for learning interpretable loss functions}. This goal aims to consolidate past research on the topic of loss function learning, combining gradient-based methods and evolution-based methods into one unified framework. When using the learned loss functions developed by the new framework, we expect to see improved sample efficiency and asymptotic performance compared to handcrafted loss functions and previous loss function learning techniques on a variety of different tasks and model architectures.

    \item \textbf{Analyze Meta-Learned Loss Functions: } \textit{Investigate how and why meta-learned loss functions can attain increased performance compared to their handcrafted counterparts}. This goal intends to locate and isolate the key underlying reasons for meta-learned loss functions success via further empirical and theoretical analysis. Key insights and discoveries about the optimization behaviors of learned loss functions are expected to be derived, facilitating future research into loss function learning, as well as the development of new more performant loss functions for training deep neural networks.

    \item \textbf{Develop a Method for Meta-Learning Adaptive Loss Functions:} \textit{Develop a meta-learning method for learning adaptive loss functions, which adapt throughout the learning process}. This goal aims to challenge the conventional notion that a loss function must be a static function. More specifically, we address the limitation that prior loss function learning techniques meta-learn static loss functions in an offline manner. Instead, we aim to meta-learn an adaptive loss function online, where the meta-learned loss function parameters are adapted in lockstep with the base model’s parameters. The expected outcome of this goal is the ability to learn adaptive loss functions which enhance the convergence, sample efficiency, and generalization of the models trained by them.

    \newpage
    
    \item \textbf{Develop a Method for Meta-Learning Neural Networks Procedural Biases:} \textit{Develop a task-adaptive method for simultaneously meta-learning the parameter initialization, optimizer, and loss function}. This objective aims to integrate advancements in meta-learning loss functions with the well-established and fast-moving research on meta-learning parameter initialization and gradient-based optimizers. These components are selected to be meta-learned as they are the primary components in neural networks where procedural biases are embedded. By leveraging prior knowledge from past learning experiences, we aim to instill robust and strong inductive biases which are expected to significantly enhance the performance and sample efficiency of models trained by our method compared to those trained with conventional training approaches.
    
\end{enumerate}

\section{Chapter Overview}

In this thesis, we explore the design and development of new state-of-the-art meta-learning methods for automatically designing loss functions for deep neural networks. 

\begin{itemize}[leftmargin=*]

    \item In \textbf{Chapter \ref{chapter:background}}, the relevant foundations and background for the topic of meta-learning are provided, where special emphasis is given to motivating meta-learning as an essential avenue for developing smarter more efficient artificially intelligent systems. Subsequently, we discuss three fundamental instantiations of meta-learning: (1) meta-learned initializations, (2) meta-learned optimizers, and (3) meta-learned loss functions. An understanding of these three key paradigms will equip the reader with the essential knowledge required to understand the context and content of the four key contributions chapters that constitute the core of this thesis.

    \item In \textbf{Chapter \ref{chapter:evomal}}, we develop a novel method for meta-learning symbolic model-agnostic loss functions called \textit{Evolved Model-Agnostic Loss} (EvoMAL). The newly proposed method consolidates recent advancements in gradient-based and evolution-based loss function learning and enables the development of interpretable loss functions on commodity hardware. The results and subsequent analysis show that meta-learned loss functions enhance the performance of deep neural networks compared to those trained by commonly used loss functions such as the cross-entropy loss for classification tasks and squared error loss for regression tasks. The contents of this chapter are based on the work presented initially in \citep{raymond2023fast} and then later developed further in \citep{raymond2023learning}.

    \item In \textbf{Chapter \ref{chapter:theory}}, we empirically and theoretically analyze the meta-learned loss functions from the preceding chapter. The analysis aims to address two fundamental questions: “\textit{Why do meta-learned loss functions perform better than handcrafted loss functions}” and “\textit{What are meta-learned loss functions learning}”. Through our analysis, clarity and insight are provided into the previously unknown underlying mechanisms of meta-learned loss functions. Our findings show that meta-learned loss functions often incorporate label smoothing regularization on classification tasks, a powerful strategy for avoiding overconfidence in deep neural networks. The findings of this chapter are utilized to develop a novel sparse variant of label smoothing regularization called \textit{Sparse Label Smoothing Regularization} (SparseLSR), which exhibits significantly enhanced speed and memory efficiency compared to its non-sparse counterpart. The contents of this chapter are based on the theoretical analysis presented in \citep{raymond2023learning}.

    \item In \textbf{Chapter \ref{chapter:adalfl}}, we develop a method called \textit{Adaptive Loss Function Learning} (AdaLFL) for meta-learning adaptive loss functions that evolve throughout the learning process. In contrast to prior loss function learning techniques that meta-learn static loss functions, our method instead meta-learns an adaptive loss function that updates the parameters of the loss function in lockstep with the parameters of the base model. This effectively mitigates the short-horizon bias prevalent in loss function learning, resulting in further improved performance relative to handcrafted loss functions and offline loss function learning techniques. The contents of this chapter are based on the work presented in \citep{raymond2023online}.

    \item In \textbf{Chapter \ref{chapter:npbml}}, we develop \textit{Neural Procedural Bias Meta-Learning} (NPBML) a task-adaptive method for simultaneously meta-learning the parameter initialization, optimizer, and loss function. This ambitious contribution builds upon our work in previous chapters by combining meta-learned loss functions with the well-established research on meta-learned parameter initialization and gradient-based optimizers. The results show that by meta-learning the procedural biases of neural networks we can induce robust inductive biases into our learning algorithm, thereby tailoring it towards a specific distribution of learning tasks. Consequently, this approach yields impressive learning performance across a diverse set of few-shot learning tasks. The contents of this chapter are based on the work presented in \citep{raymond2024meta}.

    \item Finally, in \textbf{Chapter \ref{chapter:conclusion}}, we conclude the thesis by discussing the objectives and milestones achieved and summarize how they have contributed to the fields of loss function learning and meta-learning. Following this, a retrospective discusses key findings from our research and experience. These insights are intended not only to encapsulate the essence of our work but also to serve as a source of inspiration and motivation for guiding future research directions into meta-learning.

\end{itemize}

\chapter{Background and Literature Review}\label{chapter:background}

\textit{This chapter provides a comprehensive introduction to meta-learning, introducing basic concepts and definitions. Following this, we position meta-learning with respect to related fields such as transfer learning, multi-task learning, and automated machine learning. Subsequently, we present an overview of the current meta-learning landscape using a modern taxonomy approach which decomposes meta-learning approaches along three key axes: meta-objective, meta-optimizer, and meta-representation. Finally, we introduce the subject of loss function learning and provide a structured overview of existing literature in this area.}

\section{Machine Learning}\label{section:machine-learning}

Machine Learning (ML) is the dominant approach to Artificial Intelligence (AI) in the 21st century \citep{jordan2015machine}. ML is concerned with developing intelligent algorithms that learn patterns from data to predict and uncover critical insights. In contrast to traditional algorithms, ML algorithms do not require explicit instructions to be programmed for a given task; instead, the algorithms learns to solve a task automatically by constructing a model (\textit{i.e.}, a set of instructions) directly from data \citep{abu2012learning}. More formally, \citep{mitchell1997machine} defines a ML algorithm as follows:

\begin{quoting}%
    \noindent\textbf{Definition 1 - Machine Learning:} \textit{A computer program is said to learn from experience $\mathbb{E}$ with respect to some task $\Task$ and performance metric $\mathcal{L^P}$, if its performance on $\Task$ as quantified by $\mathcal{L^P}$, improves with experience $\mathbb{E}$.} 
\end{quoting}

\noindent
The premise of learning from data is to use a set of observations (\textit{i.e.}, samples or instances) to uncover an underlying pattern or process --- this is a very broad premise. As such, it is challenging to summarize all ML algorithms into a single framework. As a result, several learning paradigms have arisen to deal with different tasks and assumptions \citep{domingos2015master}. The predominant elementary paradigms of ML are supervised learning, unsupervised learning, and reinforcement learning \citep{friedman2001elements}.

\begin{itemize}

  \item In \textbf{supervised learning}, a model is trained to map a set of input observations to their corresponding output target values, such that the underlying pattern is learned \citep{russell2002artificial}. Supervised learning is historically the most common ML paradigm, with prototypical manifestations of supervised learning being tasks such as classification and regression.
  
  \item In \textbf{unsupervised learning}, a model is trained on data with no pre-assigned labels. As a consequence, unsupervised learning is concerned with the discovery of inherent patterns within the data \citep{hinton1999unsupervised}. Common unsupervised learning tasks include but are not limited to, clustering and manifold learning.
  
  \item In \textbf{reinforcement learning}, the desired output of the model is not directly provided; instead, feedback is provided to the model in terms of reward and punishment based on how the model's actions affect the environment \citep{kaelbling1996reinforcement}. Reinforcement learning has seen many applications in robotics and game-playing systems such as DeepMind's AlphaGo \citep{silver2016mastering} and AlphaStar \citep{vinyals2019grandmaster}.
  
\end{itemize}

\noindent
This thesis concerns how meta-learning techniques can be applied to supervised learning tasks to improve performance --- this area of research is often referred to as \textbf{meta-supervised learning} \citep{hospedales2020meta}, as it sits at the union of meta-learning and supervised learning. Note that the literature on meta-learning is not limited to meta-supervised learning; there also exists an extensive and exciting amount of research on the topic of meta-reinforcement learning \citep{beck2023survey}. However, this lies outside the scope of this thesis, hence; all subsequent discussion and formalization of meta-learning are presented within the context of meta-supervised learning, even when not explicitly stated as such.

\section{Meta-Learning}\label{section:meta-learning}

Meta-learning is notoriously difficult to define, having been used in various terminologically inconsistent ways throughout historical and contemporary literature. This section introduces one general definition for meta-learning adapted from \citep{hospedales2020meta} and \citep{vilalta2002perspective}, which we deem to be the most relevant with respect to the topic of loss function learning. Following this, the basic meta-learning terminology and notation, which will be used for the remainder of this thesis is presented, as well as some additional discussion demarcating meta-learning from related fields including transfer learning, multi-task learning, and automated machine learning. To begin, we present a definition for meta-learning adapted from \citep{hospedales2020meta}, which is as follows:

\begin{quoting}%
    \noindent\textbf{Definition 2 - Meta-Learning:} \textit{The field concerned with the development of self-adapting learning algorithms that learn to solve new tasks from utilizing past experiences of solving similar related tasks. During base learning a set of inner learning algorithms are used to solve a set of tasks, minimizing some inner objective. During meta-learning, an outer algorithm updates the inner learning algorithms such that the models learn to improve on some pre-determined outer objective.} 
\end{quoting}

\noindent
As defined thus far, many conventional algorithms such as random search or grid search could categorically be distinguished as meta-learning algorithms. However, importantly meta-learning requires explicitly defined meta-level objectives and end-to-end optimization of the inner algorithms with respect to that objective. Meta-learning is prototypically conducted on multiple learning tasks sampled from a task family, leading to a base learning algorithm that performs well on new tasks sampled from that same family. However, in some limiting cases, only a single task may be used (\textit{i.e.}, single-task meta-learning). In the following sections, we introduce these concepts more formally.

\subsection{Problem Setup and Terminology}\label{subsec:problem-setup}

\subsubsection{Conventional Supervised Learning Setup}

\begin{figure}[t!]
    \centering\captionsetup{justification=centering}
    \includegraphics[width=0.6\textwidth]{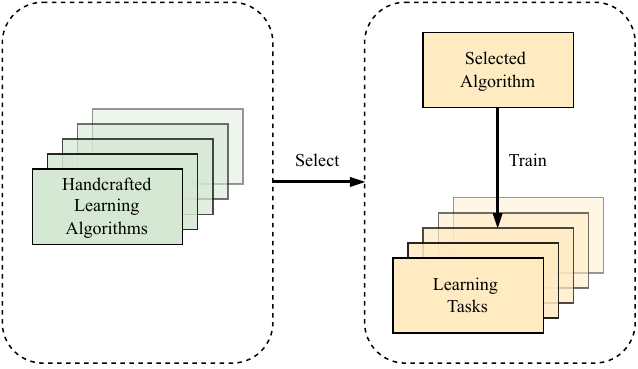}
    \caption{The conventional approach to designing and selecting a learning algorithm for a set of learning tasks in machine learning. Researchers manually design the learning algorithm, which is then deployed to solve new learning tasks.}
    \label{fig:vanilla-learning-process}
\end{figure}

In conventional approaches to supervised learning, a learning algorithm $\mathcal{A}$ (\textit{e.g.}, a neural network, gradient boosted trees or linear model) aims to generalize a finite set of observations, \textit{i.e.}, a dataset $\Dataset$, into a general model $f_{\theta}(x)$ of the domain, where $\theta$ is the model parameters. Formally, we define $\mathcal{A}$ as being a composition of components $\mathcal{A} = \{\mathbb{C}_1, \mathbb{C}_2, \dots\}$, where components are themselves just a set of inductive biases, $\{\mathbb{I}_1, \mathbb{I}_2, \dots\}$. Optimization of the model parameters $\theta$ now occurs by minimizing some loss function $\Loss$ as follows: 
\begin{equation}\label{eq:general-conventional-learning}
\theta^{*} = \argminA_{\theta} \Big[ \Loss(\Dataset, \theta(\mathcal{A})) \Big].
\end{equation}
Crucially, we make the dependence of $\theta$ on algorithm $\mathcal{A}$, or more precisely, the set of \textit{fixed} inductive biases $\{\mathbb{I}_1, \mathbb{I}_2, \dots\}$ explicit. In conventional approaches to supervised learning the inductive biases must be \textit{a priori} assumed before training, see Figure \ref{fig:vanilla-learning-process}. However, selecting the correct biases is not trivial, and the assumptions made by the set of chosen inductive biases drastically affect both the training dynamics and the final learned model. A poorly chosen set of inductive biases will result in a commensurately poor-performing algorithm, while in the inverse case, a well-selected set of inductive biases will result in a strong-performing algorithm.

\subsubsection{Conventional Meta-Learning Setup}

\begin{figure}[t!]
    \centering\captionsetup{justification=centering}
    \includegraphics[width=0.85\textwidth]{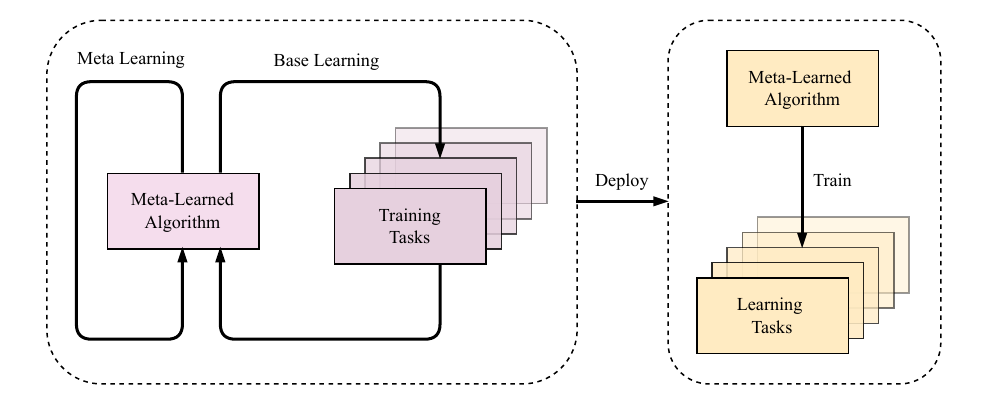}
    \caption{In meta-learning, the learning algorithm is automatically designed (the meta-training phase) by learning to learn over a task distribution $p(\Task)$, \textit{i.e.}, a set of related learning tasks, before being deployed (the meta-testing phase) to solve unseen tasks sampled from the same task distribution.}
    \label{fig:meta-learning-process}
\end{figure}

In contrast to the conventional learning setup, meta-learning seeks to improve the training dynamics and final learned model by learning one or more components in $\mathcal{A}$ rather than assuming they are fixed, as shown in Figure \ref{fig:meta-learning-process}. This is achieved by splitting the learning process into two distinct phases, \textit{meta-training} and \textit{meta-testing}. In the meta-training phase, meta-learning occurs via casting the problem as a bilevel optimization problem. A bilevel optimization refers to a hierarchical optimization problem, where one optimization problem contains another optimization problem as a constraint \citep{colson2007overview, bard2013practical}. The meta-training phase can be formalized as follows, where $\Loss^{meta}$ and $\Loss^{base}$ refer to the outer and inner objectives respectively. 
\begin{align}
\label{eq:prototypical-meta-training-phase}
\begin{split}
    \mathbb{C}^{*}_{\phi} &= \argminA_{\phi} \mathbb{E}_{\Task_i \sim p(\Task)} \Big[ \Loss^{meta}(\Task_{i}, \theta_{i}^{*}(\mathbb{C}_{\phi})) \Big]\\
    s.t.\;\;\;\;\; \theta_{i}^{*}(\mathbb{C}_{\phi})  &= \argminA_{\theta} \Big[ \Loss^{base}(\Dataset_{i}, \theta_{i}(\mathcal{A}_{i}(\mathbb{C}_{\phi})) \Big]
\end{split}
\end{align}
\noindent
In the meta-training phase, the outer level optimization attempts to embed meta-knowledge into the learned component $\mathbb{C}_{\phi}$, where $\phi$ are the learnable meta parameters. This is achieved by optimizing some meta-level loss function $\Loss^{meta}$ to produce a set of performant base models in the inner optimization which can effectively and efficiently solve a set of related tasks $\{\Task_1, \Task_2, \dots\}$. A task is defined as a dataset $\Dataset_i$ and a corresponding objective $\Loss^{base}$ which is sampled from a task distribution, \textit{i.e.}, $\Task_i = \{\Dataset_i, \Loss^{base}\} \sim p(\Task)$. Following this, the meta-testing phase follows a nearly identical setup to Equation \eqref{eq:general-conventional-learning}, substituting in $\mathbb{C}_{\phi}$ as an additional argument, utilizing the past experience of solving related tasks to more effectively solve a new unseen related task $\Task_{new} \sim p(\Task)$.

\begin{equation}\label{eq:prototypical-meta-testing-phase}
\theta^{*} = \argminA_{\theta} \Big[ \Loss^{base}(\Dataset_{new}, \theta_{i}(\mathcal{A}_{i}(\mathbb{C}_{\phi})) \Big]
\end{equation}

\noindent
The specific benefits of meta-learning are dependent on two fundamental choices. Firstly, what is the intended goal of meta-learning? For example, the goal may be to improve generalization, perform few-shot learning, or enhance domain adaptation capabilities. Secondly, which component $\mathbb{C}_{\phi}$ should be meta-learned to achieve this goal. Some existing examples in the literature include parameter initializations, optimizers and learning rules, and loss functions. We will return to discuss these topics later in detail in Sections \ref{section:meta-objective} and \ref{section:meta-representation}, respectively.

\subsubsection{Task Perspective of Meta-Learning}

\begin{figure}[t!]
    \centering\captionsetup{justification=centering}
    \includegraphics[width=0.9\textwidth]{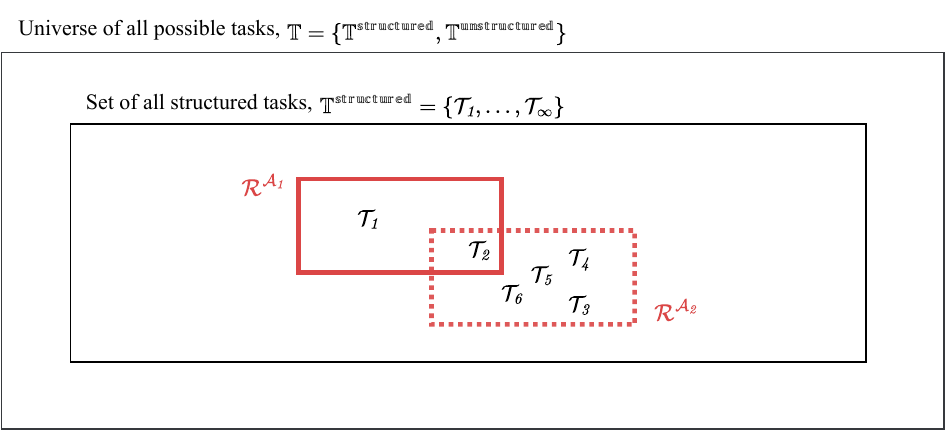}
    \caption{Each learning algorithm $\mathcal{A}$ covers a region of efficiency $\mathcal{R}^{\mathcal{A}}$ based on its inductive biases $\{\mathbb{I}_1, \mathbb{I}_2, \dots \}$. In this example, the learning algorithm $\mathcal{A}_{1}$ can efficiently learn tasks $\Task_1$ and $\Task_2$, while learning algorithm $\mathcal{A}_{2}$ can efficiently learn tasks $\Task_2 \dots \Task_5$.}
    \label{fig:task-regions-diagram}
\end{figure}

A high-level task perspective is given to illustrate the effects of meta-learning. In Figure \ref{fig:task-regions-diagram} we represent the space of all possible learning tasks $\mathbb{T}$, that an algorithm $\mathcal{A}$ can learn effectively over. Each algorithm has a limited region of effectiveness $\mathcal{R}^{\mathcal{A}} \in \mathbb{T}$, which favors certain inductive biases of others, where $\mathcal{A}$ can never be made to learn effectively over all tasks in $\mathbb{T}$ as long as the set of inductive biases $\{\mathbb{I}_1, \mathbb{I}_2, \dots \}$ stays fixed. The goal of meta-learning, in essence, is to create a clear principled methodology for \textit{dynamically} changing some of the inductive biases such that the region of efficiency $\mathcal{R}^{\mathcal{A}}$ shifts to more effectively learn over some task distribution $p(\Task)$ that we care about.

In Figure \ref{fig:task-regions-diagram}, we illustrate this by showing an algorithm $\mathcal{A}$, which has an efficiency region $\mathcal{R}^{\mathcal{A}_{\mathrm{1}}}$, which contains tasks $\Task_1$ and $\Task_2$. However, we would like to instead find some alternative algorithm with efficiency region $\mathcal{R}^{\mathcal{A}_{\mathrm{2}}}$ that can effectively learn tasks $\Task_2, \dots, \Task_5$ instead. One approach to solving this problem is to use meta-learning to embed $\mathcal{A}$ with a set of inductive biases favoring the desired distribution of tasks, such that the region of efficiency shifts from $\mathcal{R}^{\mathcal{A}_{\mathrm{1}}}$ to $\mathcal{R}^{\mathcal{A}_{\mathrm{2}}}$. 

\noindent
Conventional approaches to learning typically require inefficient exhaustive searches over many candidate algorithms and inductive bias combinations to find the matching region of efficiency for the desired task. In contrast, meta-learning offers a computationally tractable methodology for dynamically selecting $\mathcal{A}$ and its corresponding region $\mathcal{R}^{\mathcal{A}}$ in an end-to-end principled way. Note, one requisite to using meta-learning is having a suitably flexible and expressive representation (base learner) to begin with for learning, such that it is possible to make changes in $\mathbb{C}$, which would result in a successful shift of $\mathcal{R}^{\mathcal{A}}$. Consequently, most meta-learning research has been developed for DNNs since they offer a highly flexible and expressive representation for learning \citep{hornik1989multilayer}.

\subsection{Differentiating Related Fields}\label{subsec:differentiating-meta-learning}

The topic of meta-learning is often mistakenly conflated with other related paradigms such as transfer learning, multi-task learning, and automated machine learning. Accordingly, it is beneficial to delineate the key similarities and differences between meta-learning and its proximally closest paradigms.  

\subsubsection{Transfer Learning}

Transfer learning (TL) is a field concerned with utilizing past experiences gained from one or more source tasks and applying them to a new target task to improve aspects of the learning process, such as the sample-efficiency or accuracy \citep{michalski1983theory}. The most common instantiation of TL is parameter transfer, which involves learning a set of model parameters on a source model, which are then directly transferred to a target model as the initial parameters \citep{thrun2012learning,bengio2012deep}. This process yields a pre-trained (or partially trained) model, which can be further fine-tuned using the available data on the target task.

In TL, the learned component is extracted via vanilla (conventional) learning on the source task without the use of a meta-objective. In contrast, the learned component would be obtained in meta-learning by solving an outer optimization problem \citep{hospedales2020meta}. Importantly, meta-learning seeks methodology to derive better adaptation across vastly different tasks in $p(\Task)$ by explicitly aiming to learn the regularity and relationship between tasks, while in TL, transferring between vastly different tasks often results in negative (failed) transfer. Finally, meta-learning tends to be concerned with a much wider range of meta-representations (described further in Section \ref{section:meta-representation}) than just model parameters.

\subsubsection{Multi-Task Learning}

Multi-task learning (MTL) is an approach to inductive transfer that aims to learn several related tasks jointly to improve the performance \citep{caruana1997multitask}. Joint training is commonly achieved by employing a shared representation to learn multiple tasks simultaneously, exploiting regularities and differences across training tasks. Solving multiple tasks simultaneously induces a regularizing effect as it promotes the learning algorithm $\mathcal{A}$ to perform well on multiple tasks \citep{baxter2000model}. MTL has been shown to improve learning efficiency and inference performance for task-specific models compared to models trained in isolation \citep{zhang2021survey}.

Analogous to TL, a salient differentiating factor between meta-learning and MTL is that MTL problems are formed as single-level optimization without any meta-objective. Furthermore, the central goal of MTL is to solve a fixed finite number of known tasks, where they do not learn from prior experience. In contrast, the point of meta-learning is often to solve an infinite number of unseen future tasks from $p(\Task)$. While different, as pointed out in \citep{vanschoren2018meta} and \citep{peng2020comprehensive}, MTL can be successfully combined with meta-learning techniques to learn how to prioritize among multiple tasks.

\subsubsection{Automated Machine Learning}

Automated machine learning (AutoML) is a field concerned with making easier via automation the process of applying ML to real-world problems \citep{vanschoren2018meta, yao2018taking, hutter2019automated}. AutoML covers the complete pipeline from handling raw data to the development and deployment of an ML model. AutoML is commonly used to automate parts of the learning process that are typically done manually instead of being learned, such as data preparation and algorithm selection.

AutoML often focuses on different parts of the learning process compared to what meta-learning is typically concerned with, such as data cleaning, data augmentation, and algorithm selection \citep{hospedales2020meta}. However, AutoML, due to its inherently general and all-encompassing definition, is a very broad field of research. AutoML is sometimes known to make use of end-to-end optimization and multi-level optimization using a meta-objective; thus, in some respects, meta-learning can be seen as an intersecting specialization of AutoML. Techniques such as random search \citep{bergstra2012random} and Bayesian hyperparameter optimization \citep{snoek2012practical} used frequently in AutoML are typically not considered meta-learning.

\subsection{Meta-Learning Taxonomy}

Existing methods for meta-learning can be categorized in various ways. This thesis will use the taxonomy proposed in \citep{hospedales2020meta}, which partitions approaches into three independent axes, which are as follows:

\begin{itemize}

    \item \textbf{Meta-Objective:} The first axis is the goal (\textit{i.e.}, the “why?”) of meta-learning. The selection of $\Loss^{meta}$ determines the intentions of the meta-learning systems. Most commonly, the meta-objective in meta-learning is: few-shot learning \citep{snell2017prototypical, boudiaf2020transductive}, single-task and multi-task learning \citep{li2019feature, bechtle2021meta}, or fast adaptation and asymptotic performance \citep{gonzalez2020improved}. Further discussion of the selection of meta-objectives and their respective consequences is presented in Section \ref{section:meta-objective}.

    \item \textbf{Meta-Optimizer:} The second axis is the choice of meta-optimizer (\textit{i.e.}, the “how?”) used to learn $\mathbb{C}_{\phi}$. In practice, the meta optimizer is usually some form of gradient-based \citep{finn2017model,grefenstette2019generalized} or evolutionary-based learning \citep{houthooft2018evolved,al2019survey} method. In Section \ref{section:meta-optimizer}, we discuss the pros and cons of both approaches and give a brief overview of popular methods in the literature.
  
    \item \textbf{Meta-Representation:} The final axis is concerned with the component(s) $\mathbb{C}$ (\textit{i.e.}, the “what?”) that we want to meta-learn. Examples of meta-representation include gradient-based optimizers \citep{bengio1994use}, parameter initializations \citep{finn2017model,nichol2018first}, neural network architectures \citep{stanley2002evolving}, and many more. In Section \ref{section:meta-representation}, we give a brief overview of three of the most common meta-representations.
  
\end{itemize}

\section{Meta-Objective}\label{section:meta-objective}

In this section, we discuss the meta-objective, which in the context of meta-learning represents the goal of the learning system. The meta-objective is defined by the outer objective functions $\Loss^{meta}$ and the associated data flow between inner and outer optimizations steps. Typically, $\Loss^{meta}$ is defined as the task-relevant performance metric (or a differentiable approximation) computed on either the meta-training set $\Dataset^{train}$ or a held-out validation set $\Dataset^{valid}$. Using $\Loss^{meta}$ to derive a signal, the goal of the meta-learning system is now dependent on the design of the data flow between inner and outer optimization steps, which can produce several different effects. Three of the most common designs are as follows:

\subsection{Multi and Single-Task Learning}

Multi-task meta-learning refers to when the goal of the meta-learning system is to solve any task $\Task$ drawn from a given task distribution $p(\Task)$ \citep{finn2017model, snell2017prototypical}. In multi-task meta-learning, the inner optimization typically requires multiple tasks $\Dataset=\{\Task_1, \Task_2, \dots, \Task_m\}$ sampled from $p(\Task)$ to use at meta-training time such that meta-generalization can occur successfully. In contrast, single-task meta-learning refers to exclusively solving one specific task better. In single-task meta-learning, the inner optimization only uses one task, \textit{i.e.}, $\Dataset=\{\Task_1\}$, for both meta-training and meta-testing \citep{andrychowicz2016learning, gonzalez2020improved}. Note that the dataset will typically be partitioned into training and testing folds similar to conventional learning (non-meta-learning) datasets to quantify the base-generalization performance to unseen out-of-sample instances.

These two meta-objectives tend to have different assumptions and value propositions attached to them. In multi-task meta-learning, a task distribution $p(\Task)$ is required. Sampling multiple tasks from a task distribution and explicitly solving them enables the learned component $\mathbb{C}_{\phi}$ to better generalize to new unseen tasks not considered in the meta-training phase \citep{hospedales2020meta}. While in single-task meta-learning $\mathbb{C}_{\phi}$ is less likely to generalize successfully, especially when $p(\Task)$ represents a wide task distribution, \textit{i.e.}, tasks share some commonalities but are quite different in many respects. 

However, there are two key challenges in multi-task meta-learning. The first is that you do not always have the data resources available to perform meta-training, \textit{i.e.}, do not always have multiple large related datasets available. Second, training over multiple tasks at meta-training time is both memory and compute-heavy, as it often involves having multiple base models in memory being trained simultaneously while using high-order gradients. There are fewer memory and compute issues in single-task meta-learning as only a single task is used in the meta-training phase. In addition, conventional datasets can be used, which are more freely available. 

\subsection{Many and Few-Shot Learning}

Another common goal in meta-learning is to improve the base learner's one-shot, few-shot, or many-shot learning performance \citep{koch2015siamese,ravi2017optimization}. This is achieved via the inner optimization process using only a limited number of samples per task to compute the loss. Note, that few-shot learning and its variants are typically restricted to meta-learned parameter initializations and other meta-representations that store internal representations of the task in memory \citep{finn2017model,koch2015siamese}. Therefore, meta-representations such as loss functions and activation functions cannot achieve few-shot learning capabilities on their own.

\subsection{Fast Adaptation and Asymptotic Performance}

When the base loss $\Loss^{base}$ is computed at the very end of the inner optimization, meta-training improves the generalization and final asymptotic performance on the base task \citep{liu2021loss, li2021autoloss}. In contrast, if $\Loss^{base}$ is computed after only a few inner optimization step, meta-training promotes fast adaptation, \textit{i.e.}, improved convergence and sample-efficiency \citep{gonzalez2020improved, bechtle2021meta}. These two meta-objectives lie along a continuum, where more steps result in more preference towards improved generalization and final asymptotic performance, and fewer steps result in more preference towards fast adaptation. In practice, partial training sessions are the most prevalent in the literature due to reduced computational overhead compared to performing a full-duration training session. In addition, partial training sessions can often provide a reliable and robust approximation of the final predictive performance of a model.

\section{Meta-Optimizer}\label{section:meta-optimizer}

In this section, we discuss the meta-optimizer, which in the context of meta-learning refers to the learning strategy for how the meta-objective is optimized to learn $\mathbb{C}_{\phi}$. In meta-learning there are two dominant optimization strategies for learning, these being gradient-based learning and evolution-based learning. The following sections give an overview of both optimization strategies and a summary of their respective strengths and weaknesses. 

\subsection{Gradient-Based Learning}

\begin{figure}[t!]
\centering\captionsetup{justification=centering}

   \includegraphics[width=0.85\textwidth]{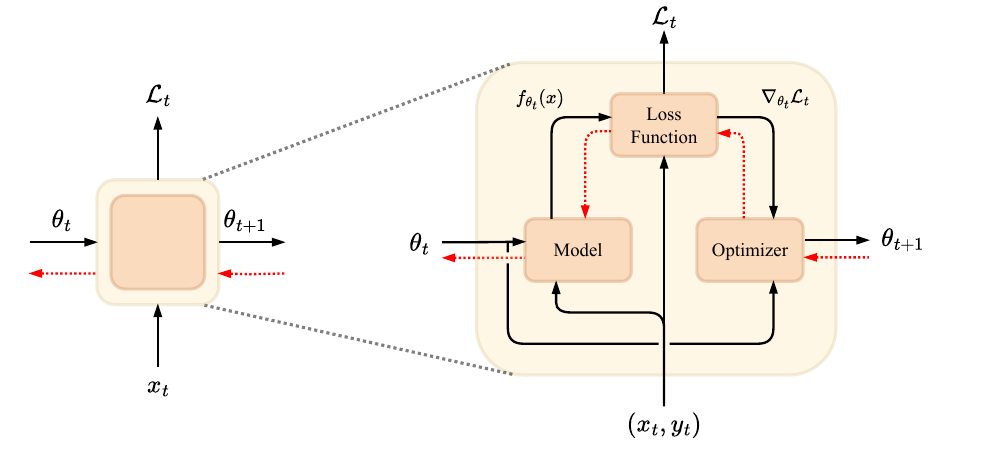}
   \caption{Decomposing the forward and backward pass of a single time-step in unrolled differentiation. As shown the new model parameters $\theta_{t+1}$ is dependent on the optimizer, loss function, and previous model parameters $\theta_{t}$.}

\label{fig:unrolled-differentiation-illustration-2}
\end{figure}

The most common approach to meta-optimization is gradient-based learning \citep{hospedales2020meta}. This requires computing the gradient of the outer objective $\nabla\Loss^{meta}$, which is connected via the chain rule to the model parameter $\theta$ as illustrated in Figures \ref{fig:unrolled-differentiation-illustration-1} and \ref{fig:unrolled-differentiation-illustration-2}. Many methods use some variant of gradient descent to optimize the meta parameters. Some examples of gradient-based meta-learning techniques include unrolled differentiation \citep{grefenstette2019generalized} and implicit differentiation \citep{rajeswaran2019meta}. Using gradient-based learning as the meta-optimizer has the following key advantages: 

\begin{figure}[t!]
\centering\captionsetup{justification=centering}

   \includegraphics[width=0.95\textwidth]{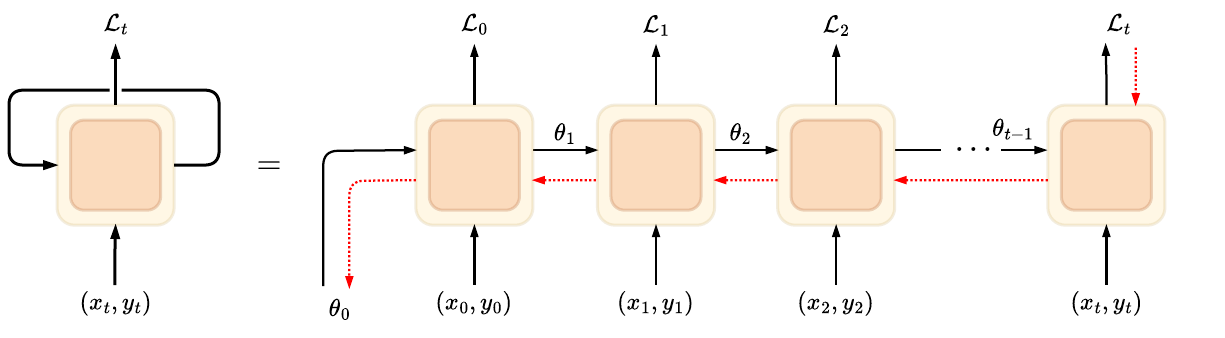}
   \caption{Illustrating unrolled differentiation, a gradient-based meta-optimization method that unrolls the optimization path to compute the gradient of the loss with respect to any arbitrary hyperparameter(s).}

\label{fig:unrolled-differentiation-illustration-1}
\end{figure}

\begin{itemize}

  \item \textbf{Performance Scalability:} Compared to evolution-based methods, gradient-based methods tend to be far more scalable due to not having to maintain a population of candidate solutions. Additionally, $\Dataset^{train}$ can be loaded in batches (\textit{i.e.}, online learning) without bringing the whole dataset into memory, enabling big data to be used efficiently. 
  
  \item \textbf{Compatibility:} Most meta-learning research happens within the context of DNNs; naturally, this makes the use of gradient-based methods a convenient choice. Gradient-based methods are easy to integrate into existing frameworks in an end-to-end way. 

\end{itemize}

\noindent
While very popular, gradient-based methods have some salient drawbacks that are undesirable when designing a meta-learning system. Two of the key drawbacks are as follows:

\begin{itemize}

  \item \textbf{Gradient Degradation:} A common issue when using gradient-based approaches in meta-learning is the inevitable gradient degradation problem, whose severity increases with the number of inner loop optimization steps similar to backpropogation through time in recurrent neural networks. 
  
  \item \textbf{Differentiability Constraint:} Most methods cannot handle non-differentiable and non-smooth meta-objectives. As a result, either a differentiable approximation must be used, which can be detrimental to the performance, or expensive reinforcement learning paradigms such as policy gradients \citep{houthooft2018evolved} must be used.

\end{itemize}

\noindent
Gradient-based techniques have been successfully applied to meta-learning parameter initializations \citep{finn2017model, nichol2018first, rajeswaran2019meta} and gradient-based optimizers \citep{andrychowicz2016learning, wichrowska2017learned}, and many more meta-representations \citep{ramachandran2017searching, liu2020evolving}. They have also shown impressive results in the few-shot learning domain \citep{koch2015siamese, finn2017model}. 

\subsection{Evolution-Based Learning}

Another approach to meta-optimization commonly seen in the meta-learning literature is evolution-based learning, which comes from the subfield of AI called evolutionary computation (EC) \citep{al2019survey}. EC refers to a family of population-based metaheuristics inspired by nature and is typically partitioned into three dominant learning paradigms \citep{xue2015survey}: evolutionary algorithms (EAs), swarm intelligence (SI), and other techniques. Examples of ECs techniques include genetic algorithms \citep{mitchell1998introduction}, particle swarm optimization \citep{kennedy1995particle}, differential evolution \citep{storn1997differential}, and covariance matrix adaptation evolutionary strategy \citep{hansen2003reducing}, which is illustrated in Figure \ref{fig:cmaes-illustration}. Evolution-based learning has the following advantages: 

\begin{figure}[t!]
\centering\captionsetup{justification=centering}

\begin{subfigure}[b]{0.4\textwidth}\centering
   \includegraphics[width=1\textwidth]{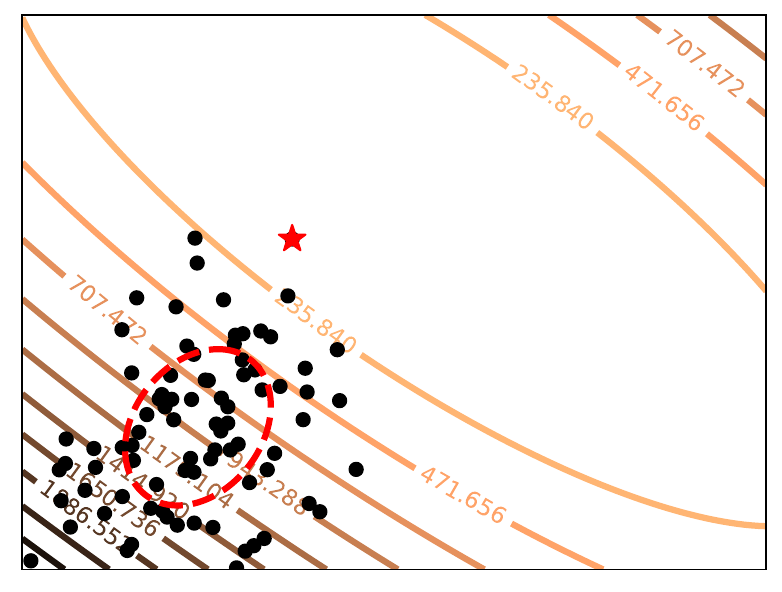}
   \caption{Generation 1}
\end{subfigure}
\begin{subfigure}[b]{0.4\textwidth}\centering
   \includegraphics[width=1\textwidth]{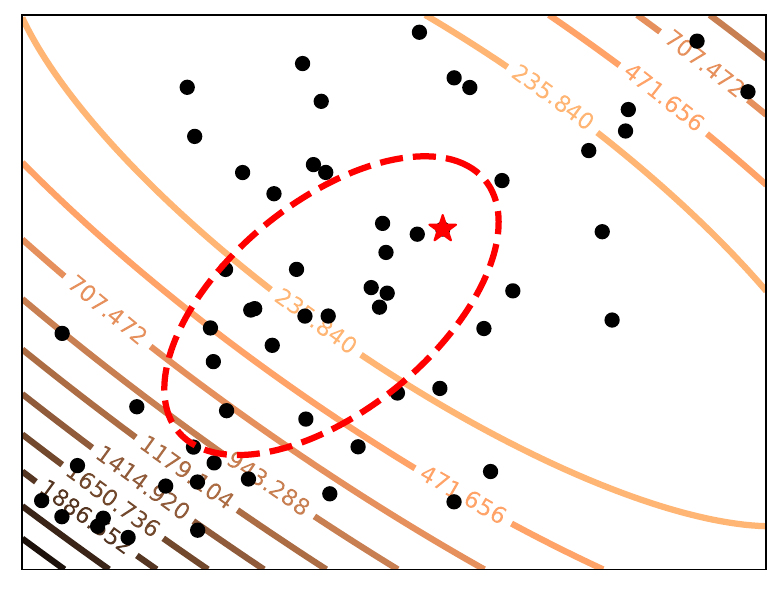}
   \caption{Generation 5}
\end{subfigure}

\begin{subfigure}[b]{0.4\textwidth}\centering
   \includegraphics[width=1\textwidth]{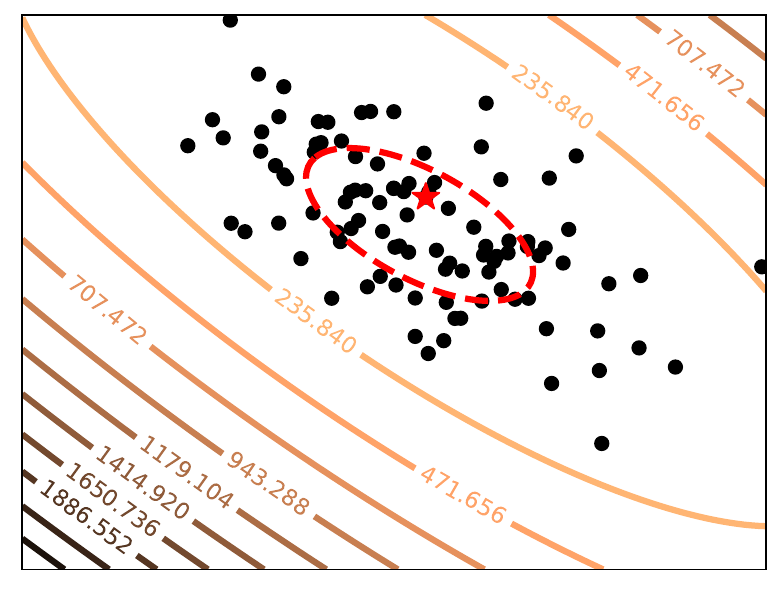}
   \caption{Generation 10}
\end{subfigure}
\begin{subfigure}[b]{0.4\textwidth}\centering
   \includegraphics[width=1\textwidth]{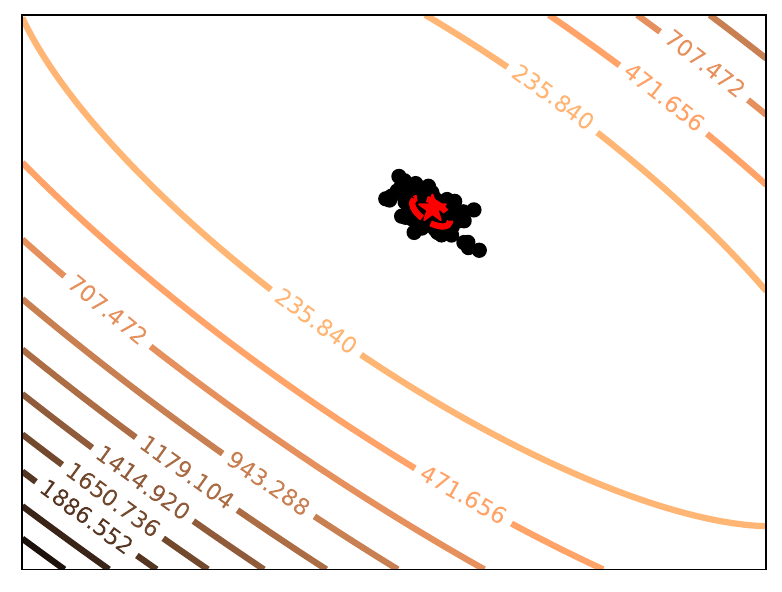}
   \caption{Generation 15}
\end{subfigure}
\caption{Illustrating CMA-ES, a popular evolution-based approach that maintains a population of solutions (shown as black dots). These solutions are evolved over several generations to locate the minimum of a convex optimization problem.}
\label{fig:cmaes-illustration}
\end{figure}

\begin{itemize}

  \item \textbf{Gradient-Free:} Evolution-based methods do not rely on backpropagation, which helps circumvent the gradient degradation issues and the costly high-order gradient computation often seen in gradient-based meta-optimization methods.
  
  \item \textbf{Flexibility:} Evolution-based methods are highly flexible as they can optimize for any base model $f_{\theta}(x)$, \textit{i.e.}, not just limited to neural network-based methods. Furthermore, they can optimize for any meta-objective $\Loss^{meta}$ with no differentiability constraint. Consequently, performance metrics such as the accuracy can be optimized directly instead of using a differentiable approximation.
  
  \item \textbf{Population-Based:} By maintaining a diverse population of solutions, meta-heuristics can avoid local optima that often plague gradient-based methods. Additionally, population-based EC methods can often learn multiple candidate solutions in a single execution of the algorithm.

\end{itemize}

\noindent
Although promising, evolution-based learning methods also have some notable limitations and disadvantages, which are as follows:

\begin{itemize}

  \item \textbf{Computationally Expensive:} As the number of learnable model parameters $\phi$ increases, the corresponding size of the population must also increase to ensure effective search capabilities. Consequently, population-based methods often have intractably high computational costs when solving large-scale optimization problems.
  
  \item \textbf{Hyperparameter Sensitivity:} There is often high sensitivity to the hyperparameters, e.g., crossover, mutation, and selection strategies, and their corresponding probabilities in genetic algorithms. This requires careful hyperparameter optimization to achieve state-of-the-art performance, which can be very time-consuming.
  
  \item \textbf{Fitting Ability:} Evolution-based techniques tend to be highly effective at locating promising regions of the search space with their global search ability. However, their local search ability is generally inferior to gradient-based methods, especially for large models containing potentially millions of meta-parameters. Evolution-based techniques lack theoretical guarantees of convergence to a local optimum. 

\end{itemize}

\noindent
Evolution-based techniques have been successfully applied to meta-learning optimizers \citep{cao2019learning}, network architectures \citep{stanley2002evolving}, and loss functions \citep{gonzalez2020improved} in meta-supervised learning. They are also commonly used in meta-reinforcement learning \citep{houthooft2018evolved,bechtle2021meta} as models are typically smaller, and inner optimizations are long and non-differentiable. 

\section{Meta-Representation}\label{section:meta-representation}

This section discusses the meta-representation, which refers to what component(s) of a learning algorithm are meta-learned. The decision about what component(s) are meta-learned is crucial as it directly influences the sorts of meta-knowledge $\mathbb{I}$ that can be obtained and embedded into $\mathcal{A}$, \textit{i.e.}, determines the space of possible outcomes for meta-learning. This section provides an overview of three prevalent meta-representations frequently encountered in the literature: parameter initializations, optimizers and learning rules, and loss functions.

\subsection{Parameter Initializations}

The parameter initialization of a model refers to the initial values selected for the learnable weights and biases of a neural network before training begins. The parameter initialization plays a critical role in a learning algorithm as it determines the starting point of a model prior to being optimized. Proper initialization is essential for achieving robust generalization and convergence, and for avoiding issues such as vanishing or exploding gradients, which can hinder the training process and degrade model performance. A good initialization sets the stage for efficient learning by providing the neural network with a suitable starting point from which it can effectively learn meaningful representations from the data. Formally we define a parameter initialization as follows:
\begin{quoting}
    \noindent\textbf{Definition 3 - Parameter Initialization:} \textit{In machine learning, the parameter initialization $\theta_0$ represents the initial set of values assigned to the parameters of a model before the training process begins.} 
\end{quoting}
Traditional approaches to parameter initialization in DNNs have historically focused on general-purpose initializations that aim to ensure stable and efficient training of neural networks. The most commonly used initialization method is random initialization, where the weights and biases are initialized randomly from a predefined distribution, typically, a uniform $\theta_0 \sim \mathcal{U}(-u, +u)$ or normal distribution $\theta_0 \sim \mathcal{N}(\mu, \sigma)$. One limitation of this approach is that you need to determine the parameters of the distribution, \textit{i.e.} the minimum and maximum range $u$ or standard deviation $\sigma$. Several techniques have been proposed to address this limitation, most notably, Xavier Glorot \citep{glorot2010understanding} and Kaiming He \citep{he2015delving} initialization.

Xavier Glorot initialization is a popular approach to initializing the weights of sigmoidal and hyperbolic tangent networks. Xavier Glorot initialization aims to keep the variance of activations and gradients consistent across layers, promoting stable training and mitigating issues like vanishing or exploding gradients. This is achieved by initializing the minimum and maximum range $u$ or standard derivation $\sigma$ as follows:
\begin{multicols}{2}\noindent
    \begin{equation}
        u = \sqrt{\frac{6}{\text{fan}^{in} + \text{fan}^{out}}}
    \end{equation}
    \begin{equation}
        \sigma = \sqrt{\frac{2}{\text{fan}^{in} + \text{fan}^{out}}}
    \end{equation}
\end{multicols}
\noindent
where $\text{fan}_{in}$ and $\text{fan}_{out}$ refer to the number of inputs and outputs in a layer, respectively. Another popular technique for initializing the parameters of a neural network is Kaiming He initialization, which is specifically designed for rectified linear networks. Kaiming He initialization initializes the weights similarly to Xavier Glorot initialization, except it only considers the number of inputs to a layer:
\begin{multicols}{2}\noindent
    \begin{equation}
        u = \sqrt{\frac{3}{\text{fan}^{in}}}
    \end{equation}
    \begin{equation}
        \sigma = \sqrt{\frac{1}{\text{fan}^{in}}}
    \end{equation}
\end{multicols}
\noindent
While Xavier Glorot and Kaiming He initialization are effective for training neural networks, they are general-purpose methods that do not incorporate prior knowledge about specific tasks. Consequently, these strategies necessitate learning from scratch, which requires many training instances and takes time and large amounts of computational resources to perform successfully.

An alternative approach is to use pre-trained neural network weights through transfer learning \citep{thrun2012learning,bengio2012deep}, where a model trained on one domain (e.g. ImageNet) and then adapted for use in a new target domain (e.g. CIFAR-10). This can greatly enhance the performance of a neural network and can reduce the number of instances needed to learn a satisfactory model. However, a key limitation of transfer learning is that the pre-trained model is not explicitly optimized for cross-task generalization \citep{hospedales2020meta}. Consequently, transfer learning approaches do not learn representations primed for few-shot learning or fast adaptation, as such many instances are still needed for training.

\subsubsection{Meta-Learning Parameter Initializations}

Contemporary approaches to meta-learning have spent significant amounts of time researching meta-learned parameter initializations. The most well-known and foundational method for meta-learning the parameter initialization is Model-Agnostic Meta-Learning (MAML) by \cite{finn2017model}. MAML aims to meta-learn the parameter initialization of a model via solving a bilevel optimization problem 
\begin{align}
\label{eq:maml-bilevel-optimization}
\begin{split}
    \bm{\theta}^{*} &= \argminA_{\bm{\theta}} \mathbb{E}_{\Task_i \sim p(\Task)} \left[\Loss^{meta}\left(\Task_i, f_{\theta^{*}_{i}(\bm{\theta})}\right)\right] \\ 
    s.t.\;\;\;\;\; \theta_{i}^{*}(\bm{\theta})  &= \argminA_{\theta_{i}} \left[\Loss^{base}(y_{i}, f_{\theta_{i}}(x_{i}))\right]
\end{split}
\end{align}
In MAML a model is explicitly trained to quickly adapt to new learning tasks by training over a distribution of tasks $p(\Task)$ and minimizing the expected performance. This is typically performed via unrolled differentiation, where each task's trajectory is unrolled and differentiated (see Figure \ref{fig:unrolled-differentiation-illustration-1}). Unlike transfer learning, MAML can learn a highly performant model at meta-testing time with only a few gradient steps, and a very limited number of instances available (e.g. 1 or 5 instances per class).

Many papers have extended MAML through alternative meta-optimization procedures, such as first-order gradient-based methods \citep{nichol2018first}, implicit differentiation \citep{rajeswaran2019meta}, hessian free optimization \citep{song2019maml}, and trajectory-agnostic meta-optimization \citep{flennerhag2018transferring}. Additionally, researchers have also investigated decoupling the meta-learned parameterization initializations from the high-dimensional space of the model parameter, via partitioning $\theta$ into subspaces \citep{lee2018gradient} or meta-learning in the latent representation \citep{rusu2019metalearning}.

\subsection{Optimizers and Learning Rules}

The optimizer in ML represents the learning rule used for optimizing a model's parameters $\theta$ via minimization (or maximization) of the loss function. It constitutes an essential component of all learning algorithms as it is the primary learning rule that steers the learning process through the search space, \textit{i.e.}, the optimizer determines the learning trajectory. An effective optimizer is crucial for obtaining strong learning performance, as it accelerates the training process of a DNN, and ensures that the model converges to a favorable region of the search space with good generalization performance. Formally we define an optimizer as follows:

\begin{quoting}%
    \noindent\textbf{Definition 4 - Optimizer:} \textit{In machine learning an optimizer is an iterative learning rule used to solve learning tasks of the form $\argminA_{\theta} \Loss(f_{\theta}(x))$. The optimizer aims to minimize a given loss function $\Loss$ by updating the model parameters $\theta_t$ in each iteration of the learning algorithm.} 
\end{quoting}

\noindent
In DNNs, the search direction of an optimizer is most commonly defined by the negative gradient of the loss function at the current point $\theta_t$ in weight space $-\nabla_{\theta_{t}} \Loss\left(f_{\theta_{t}}(x)\right)$. The foundational method for gradient-based optimization in neural networks is \textit{Stochastic Gradient Descent} (SGD):
\begin{equation}\label{eq:sgd-optimizer}
    \theta_{t+1} \leftarrow \theta_{t} - \alpha \cdot \nabla_{\theta_{t}} \Loss\left(f_{\theta_{t}}(x)\right)
\end{equation}
which takes the current model parameters $\theta_t$ at time-step $t$, and subtracts the gradient of the loss with respect to the model parameters. Finally, the gradient is scaled by the learning rate $\alpha$, which controls the step size taken at each step. SGD serves as the foundation for many of the more advanced gradient-based optimizers. For example, \textit{SGD with Momentum} \citep{polyak1964some} is a straightforward extension to SGD, accelerating learning by considering past gradients when performing each update step.
\begin{equation}
v_{t} \leftarrow \alpha \cdot \nabla_{\theta_{t}} \mathcal{L}\left(f_{\theta_{t}}(x)\right) + \gamma \cdot v_{t-1}
\end{equation}
\begin{equation}
\theta_{t+1} \leftarrow \theta_{t} - v_{t}
\end{equation}
where $v_{t}$ represents the velocity, which is defined by the current gradient plus the previous velocity $v_{t-1}$ scaled by the momentum hyperparameter $\gamma$ (typically set to 0.9). Subsequent advancements in gradient-based optimization have aimed to mitigate the reliance on the learning rate $\alpha$ in SGD. Some of the methods for achieving this are AdaGrad \citep{duchi2011adaptive}, AdaDelta \citep{zeiler2012adadelta}, RMSProp \citep{hinton2012neural}, and Adam \citep{kingma2014adam}. In the interest of brevity, we will only describe the latter algorithm, Adam, as it incorporates many of the mechanisms found in the aforementioned methods and is also the most commonly used optimizer in contemporary literature.

\textit{Adaptive Moment Estimation} (Adam) proposed by \cite{kingma2014adam} is one of the most popular gradient-based optimizers used for training neural networks. Adam similar to AdaDelta and RMSprop keeps track of an exponentially decaying average of past squared gradients $v_{t}$. In addition, Adam also keeps track of the exponentially decaying average of past gradients $m_{t}$, which is similar to the concept of momentum mentioned previously.
\begin{equation}
m_{t} = \beta_{1} \cdot m_{t-1} + (1 - \beta_1) \cdot \nabla_{\theta_{t}}\mathcal{L}\big(f_{\theta_{t}}(x)\big) 
\end{equation}
\begin{equation}
v_{t} = \beta_{2} \cdot v_{t-1} + (1 - \beta_2) \cdot \nabla_{\theta_{t}}\mathcal{L}\big(f_{\theta_{t}}(x)\big)^{2}
\end{equation}
Where $m_{t}$ and $v_{t}$ are approximations of the first moment (the mean) and the second moment (the uncentered/raw variance) of the gradients respectively (which is where the name comes from) and $\beta_{1}$ and $\beta_{2}$ are the decay rates for the first and second-moment estimations. The new terms $m_{t}$ and $v_{t}$ can then be substituted into the RMSprop learning rule giving the final Adam learning rule (for parsimony we sidestep the topic of bias correction):
\begin{equation}
\theta_{t+1} \leftarrow \theta_{t} - \frac{\alpha}{\sqrt{v_{t}} + \epsilon} \cdot m_{t}
\end{equation}
Contemporary approaches to designing handcrafted optimizers have evolved to address the diverse challenges of training DNNs more efficiently and effectively. These techniques have significantly advanced the field of optimization and have led to improvements in the design of general-purpose optimizers. As discussed in Chapter \ref{chapter:introduction}, the conservation of average performance implies that there are limits on performance gains in general --- perhaps general-purpose optimizers are already approaching these limits. However, by specializing in increasingly smaller subsets of tasks, there is the potential to create increasingly more powerful optimizers in the future.

\subsubsection{Meta-Learning Optimizers}

Some of the earliest research into meta-learning was by \cite{bengio1994use}, who explored meta-learning gradient-based optimizers (referred to as learning rules). Although the work showed intriguing results, it wasn’t until recently that the field colloquially known as Learning to Optimize (L2O) \citep{chen2022learning} started to gain significant traction. L2O is a paradigm that aims to meta-learn the (base) optimizer of a DNN. The principal goal of L2O is to obviate the laborious process of hand-engineering and selecting an optimizer by instead automatically inferring the optimizer directly from data. There are two predominant approaches to meta-learning optimizers (1) memory-based approaches, and (2) preconditioning matrices. We summarize both as follows:

The first approach to meta-learning an optimizer involves directly parameterizing a learning rule using the memory of a recurrent neural network \citep{andrychowicz2016learning, li2016learning, ravi2017optimization, chen2017learning, metz2019understanding}. For example, in \citep{andrychowicz2016learning} the authors noticed that the SGD learning rule from Equation \eqref{eq:sgd-optimizer} resembles the update for the cell state
in a Long Short-Term Memory (LSTM) cell \citep{hochreiter1997flat}:
\begin{equation}
C_t = F_t \odot C_{t-1} + I_{t} \odot \widetilde{C}_t
\end{equation}
where the forget gate $F_t = 1$, the previous cell state $C_{t-1} = \theta_{t-1}$, the input gate $I_{t} =  \alpha$, and the input modulation $\widetilde{C}_t = -\nabla_{\theta_{t-1}}\Loss_{t}$. Thus, an LSTM network could be meta-learned to train a  DNN as opposed to a traditional optimizer.

Memory-based methods are popular as they can theoretically represent any learning rule since recurrent neural networks are a universal function approximator \citep{schafer2006recurrent}. They can also scale to longer learning processes by using truncated backpropagation through time \citep{jaeger2002tutorial}, where backpropagation is performed over only a subset of steps, typically around 20 \citep{metz2019understanding}. However, these methods lack the important inductive bias regarding what constitutes a reasonable learning rule. Consequently, they can be challenging to train and may not generalize well, as their parameter updates do not guarantee convergence.

The second more recent approach to meta-learning an optimizer is to leverage Preconditioned Gradient Descent (PGD) by meta-learning a parameterized matrix (sometimes referred to as a warp) that preconditions gradients during training \citep{li2017meta, lee2018meta, park2019meta, flennerhag2019meta}. This modifies the SGD learning rule from Equation \eqref{eq:sgd-optimizer} to
\begin{equation}
    \theta_{t+1} \leftarrow \theta_{t} - \alpha \cdot P_{\omega} \cdot \nabla_{\theta_{t}} \Loss\left(f_{\theta_{t}}(x)\right),
\end{equation}
where $P_{\omega}$ is the meta-learned preconditioning matrix. These approaches are akin to second-order and natural gradient descent methods \citep{amari2000methods, wright2006numerical}. Unlike memory-based approaches, preconditioning methods inherit the useful inductive biases of gradient descent, thus ensuring convergence guarantees and making them powerful and easier to train. However, a drawback is that the preconditioning matrices cannot be readily transferred to new neural network architectures once meta-learned. Furthermore, the optimizer is embedded in the network architecture in some parameterizations such as \citep{lee2018meta, flennerhag2019meta}, which is undesirable in some cases.

\subsection{Loss Functions}\label{section:loss-function-learning}

The loss function $\Loss$ is a mathematical function that takes as arguments the true label $y$ and the predicted label of the model $f_{\theta}(x)$ with parameters $\theta$ and outputs a value referred to as the \textit{loss} \citep{reed1999neural}. The value of the loss quantifies the associated cost of an accurate or inaccurate prediction and serves as the primary goal for the learning algorithm $\mathcal{A}$ when optimizing the model parameters $\theta$. Thus, the learning algorithm's success is primarily determined and quantified by its ability to minimize the loss function successfully. Formally we define a loss function as follows:

\begin{quoting}%
    \noindent\textbf{Definition 5 - Loss Function:} \textit{In machine learning, a loss function $\Loss:\mathbb{R}^2 \rightarrow \mathbb{R}^1$ is a function defined over a singular instance, \textit{i.e.}, input-output pair $(x, y)$, which takes as arguments the true label $y$ and the predicted label of the model $f_{\theta}(x)$ and returns a loss value associated with its corresponding accuracy or inaccuracy.} 
\end{quoting}

\noindent
In the broader context of ML, it is common to see terms such as \textit{loss function}, \textit{cost function}, and \textit{objective function} used interchangeably, without any real consequence. For example, foundational ML textbooks \citep{bishop2006pattern} and \citep{goodfellow2016deep} explicitly state that they do not differentiate between these terms. However, in loss function learning, it is crucial to delineate between the concepts to make it clear what the learned component $\mathbb{C}$ in the meta-learning system corresponds to exactly. For clarity, we illustrate the differences by examining the squared error loss function, commonly used for the task of regression. 
\begin{equation}\label{eq:example-loss-function}
Loss\;Function = \Big(y- f_{\theta}(x)\Big)^2
\end{equation}
Equation \eqref{eq:example-loss-function} represents a loss function in its current form, as it is defined over only a singular instance. However, in ML, it is often desirable to take the loss across a set of $n$ instances, such as the whole training dataset $\Dataset^{train}$, or a subset of $\Dataset^{train}$ referred to as a batch. Therefore, we introduce the concept of a \textit{cost function}, which consists of evaluating the loss function across a set of instances. This produces a vector of $n$ loss values, which is typically aggregated into a scalar, this occurs via a \textit{reduction} operator; most commonly, the mean across the summation of the output vector.
\begin{equation}\label{eq:example-cost-function}
Cost\;Function = \underbrace{\frac{1}{n}\sum_{i=1}^{n}}_{Reduction}\underbrace{\vphantom{\sum_{i=1}^{n}}\Big(y_i- f_{\theta}(x_i)\Big)^2}_{Loss\;Function}
\end{equation}
Finally, we introduce the more general concept of an \textit{objective function}, which represents any function that is minimized or maximized for any ML task, including those not related to supervised learning. An example of an objective function is the Ridge regression \citep{hoerl1970ridge} objective function shown in Equation \eqref{eq:example-objective-function}, which aims to minimize both Equation \eqref{eq:example-cost-function} and an additional regularization penalty on $\theta$, where $\lambda$ is a tuning parameter controlling the regularization strength. 
\begin{equation}\label{eq:example-objective-function}
Objective\;Function = \underbrace{\vphantom{\sum_{j=1}^{p}}\frac{1}{n}\sum_{i=1}^{n}\Big(y_i- f_{\theta}(x_i)\Big)^2}_{Cost\;Function} +
\underbrace{\frac{\lambda}{n}\sum_{j=1}^{p}\theta^2_j}_{Regularizer}
\end{equation}

\subsubsection{Meta-Learning Loss Functions}

Loss function learning is a new and emerging subfield of meta-learning that aims to leverage past learning experiences to infer a highly performant loss function, denoted as $\MetaLoss_{\phi}$, directly from the data \citep{gonzalez2020improved, bechtle2021meta, gao2023meta}. The loss function is a central component of any learning algorithm, as it determines both the learning path of $\mathcal{A}$ and the selection of the final model $f_{\theta}(x)$. Therefore, any improvements made to conventional handcrafted loss functions will likely have a broad impact, positively affecting many different models in various domains and applications. Initial research into loss function learning has shown promise; however, existing methods have significant issues and limitations that need to be resolved for meta-learned loss functions to become a desirable alternative to conventional handcrafted loss functions. 

\subsubsection{Meta-Training Phase:}

The meta-training phase in loss function learning is a bilevel optimization problem. The goal of the outer optimization is to meta-learn a performant loss function $\MetaLoss_{\phi}$ across a set of related tasks $\{\Task_1, \Task_2, \dots\} \sim p(\Task)$, minimizing the average meta loss $\Loss^{meta}$, where $\Loss^{meta}$ is selected based on the desired task. The inner optimization directly uses $\MetaLoss_{\phi}$ to learn the base model parameters $\theta_{i}$ for each of the base models $f_{\theta_i}$. Formally, we define the meta-training phase of loss function learning as follows:
\begin{align}
\begin{split}
    \MetaLoss^{*}_{\phi} &= \argminA_{(\MetaLoss\;\vee\;\phi)} \mathbb{E}_{\Task_i \sim p(\Task)}\left[\Loss^{meta}\left(\Task_i, f_{\theta^{*}_{i}(\MetaLoss_{\phi})}\right)\right] \\ 
    s.t.\;\;\;\;\; \theta^{*}_{i}(\MetaLoss_{\phi})  &= \argminA_{\theta_{i}} \left[\MetaLoss_{\phi}(y_{i}, f_{\theta_{i}}(x_{i}))\right]
\end{split}
\end{align}

\noindent
This is similar to Equation \eqref{eq:prototypical-meta-training-phase} shown earlier; however, we have made one amendment, which is to make explicit that you can optimize the structure of the meta-loss function $\MetaLoss$ and keep $\phi$ fixed or optimize just $\phi$ using a fixed structure for $\MetaLoss$. In loss function learning, existing methods such as ML$^{3}$ \citep{bechtle2021meta}, Taylor-GLO \citep{gonzalez2021optimizing}, and TP-CMA-ES \citep{gao2021searching}, optimize the meta-parameters $\phi$ and keep $\MetaLoss$ fixed, while methods such as CSE-Autoloss \citep{liu2021loss}, AutoLoss-Zero \citep{li2021autoloss}, and NLFS \citep{morgan2024evolving, morgan2024neural}, attempt to optimize $\MetaLoss$ but keep $\phi$ fixed. Finally, GLO \citep{gonzalez2020improved} and EvoMAL \citep{raymond2023learning}, both $\MetaLoss$ and $\phi$ are optimized simultaneously.

\subsubsection{Meta-Testing Phase:}

The outer optimization can be omitted in the meta-testing phase, leaving only the inner optimization step, which follows a near-identical setup to previously. Except this time substituting in the best-learned loss function $\MetaLoss_{\phi}$ which can now be used in place of a handcrafted loss function $\Loss^{base}$ to solve new unseen related tasks $\Task_{new} \sim p(\Task)$ as follows:
\begin{equation}\label{eq:meta-testing-phase}
\theta^{*} = \argminA_{\theta} \left[\MetaLoss_{\phi}(y, f_{\theta}(x))\right]
\end{equation}
At meta-testing time $\MetaLoss_{\phi}$ can be used as a drop-in replacement for handcrafted loss functions. The respective learning algorithm $\mathcal{A}$, does not need to make any alterations to the optimization process to accommodate the meta-learned loss function.

\section{Related Work}\label{section:lfl-related-work}

The concept of meta-learning a loss function is relatively new, with only a small handful of methods existing in the literature. For the remainder of this section, we seek to summarize the existing approaches that are central to the topic of loss function learning, following which we analyze their relative advantages and disadvantages.

\begin{itemize}

    \item \cite{gonzalez2020improved} proposed \textit{Genetic Loss Function Optimization} (GLO), which is a method for meta-learning loss functions via a two-phase learning process. In the first phase, Genetic Programming (GP) was used to discover a set of symbolic loss functions. Subsequently, in the second phase, Covariance Matrix Adaptation Evolution Etrategy (CMA-ES) is used to perform local search and optimize the coefficients. The second phase enabled them to achieve improved performance compared to not using any local search mechanisms. Their results showed that neural networks trained with meta-learned loss functions developed by GLO outperformed the standard cross-entropy loss on two image classification tasks. The most notable limitation of GLO is the number of partial training sessions required to learn $\MetaLoss_{\phi}$ which is intractable for large-scale optimization problems.
  
    \item \cite{liu2021loss} proposed a convergence-simulation driven evolutionary search algorithm, called \textit{CSE-Autoloss}, which uses GP to meta-learn a set of loss functions for object detection. Their contributions were a newly designed search space specifically for object detection and two new filtering mechanisms: convergence property verification and model simulation, which they used to reduce the evolutionary process's runtime significantly. However, a key limitation of their approach is that they omit local-search strategies to reduce the runtime, which, as shown in \citep{gonzalez2020improved}, reduces the performance of the meta-learned loss functions.

    \clearpage

    \item \cite{li2021autoloss} proposed a method for meta-learning loss functions called \textit{AutoLoss-Zero}, which can meta-learn loss functions for computer vision tasks. Similar to both \citep{gonzalez2020improved} and \citep{liu2021loss} they used a GP-based search algorithm to search over a custom-designed search space, notably including a newly introduced aggregation-based function. They also proposed a novel loss rejection protocol and a gradient-equivalence-checking strategy, reducing the evolutionary process's runtime significantly. Note, similar to \citep{liu2021loss} they omit local-search strategies, which greatly reduces the performance of the meta-learned loss function $\MetaLoss_{\phi}$. Furthermore, their method is not task and model-agnostic.

    \item \cite{morgan2024neural} advanced GP-based loss function learning by proposing \textit{Neural Loss Function Search} (NLFS), which extended the design of the search space using a NASNet-inspired approach \citep{zoph2018learning}. This approach optimizes cells containing free-floating nodes referred to as hidden state nodes. The updated search space effectively minimized bloat and redundancy in the learned loss functions. Additionally, the authors explored enhancements to the underlying genetic programming algorithm, including surrogate fitness function evaluations and regularized evolution using a mutation-only aging genetic algorithm. Using their proposed technique they proposed three new loss functions which were empirically shown to perform better than the cross-entropy loss.
  
    \item \cite{bechtle2021meta} proposed \textit{Meta-Learning via Learned Loss} (ML$^3$), the first gradient-based framework for meta-learning parametric loss functions. The learned loss functions showed improved convergence and inference performance on both supervised and reinforcement learning tasks compared to top handcrafted loss functions. In the ML$^3$ framework, they represent $\MetaLoss_{\phi}$ as a two-hidden layer feedforward neural network and optimize parameters $\phi$ via unrolled differentiation \citep{grefenstette2019generalized}. Their results showed that you could meta-learn performant loss functions using comparably efficient gradient-based training methods. 
  
    \item \cite{gonzalez2021optimizing} proposed a follow-up method using CMA-ES again, but instead of optimizing GP expression trees, they optimized multivariate Taylor polynomials. Their loss function learning approach, called \textit{TaylorGLO}, was shown to perform well on a range of computer vision tasks, finding loss functions that outperformed the standard cross-entropy loss and the loss functions meta-learned previously by GLO. In addition, reducing the problem to only learning $\phi$ instead of $\MetaLoss_{\phi}$ significantly reduced the number of partial training sessions corresponding to a reduced computational cost.

    \clearpage
  
    \item \cite{gao2021searching} similar to \cite{gonzalez2021optimizing} proposed using CMA-ES to discover novel loss functions in the Taylor polynomial function space. Their newly proposed loss function learning method was explicitly designed to improve a base model’s robustness to noisy labels. In addition to the unique angle the paper took, they proposed a domain randomization strategy to evaluate the performance during the meta-learning stage to make the learned loss function generalize to different network architectures more effectively.
    
    \item \cite{gao2022loss} also explored using the Taylor polynomial function space, but instead of using CMA-ES they used the implicit function theorem (IFT) to approximate gradients to optimize $\phi$. The method was termed \textit{Implicit Taylor Loss Network} ITL-Net. Their newly proposed loss function learning method was used to improve the robustness to distribution shift. Results showed improved performances compared to the standard cross-entropy loss on a set of diverse domain generalization tasks. In addition, although not stated explicitly, the use of gradient-based optimization methods enabled loss functions to be meta-learned with a significantly reduced run time compared to \citep{gonzalez2021optimizing} and \citep{gao2021searching}, which relied on evolution-based method CMA-ES.

    \item \cite{antoniou2019learning} proposed a few-shot learning-specific method for loss function learning called \textit{Self-Critique and Adapt} (SCA). This method extended the optimization-based meta-learning technique MAML \citep{finn2017model} by replacing the inner optimization's loss function with a fully transductive, label-free loss function. The learned loss function $\MetaLoss_{\phi}$ in SCA was represented as a dilated convolutional neural network, where the meta-parameters $\phi$ were optimized in the outer optimization alongside the parameter initialization using unrolled differentiation. Their results demonstrated that meta-learned loss functions are a promising direction for enhancing the performance of optimization-based few-shot learning methods.

    \item \cite{baik2021meta} proposed a technique called \textit{Meta-Learning with Task-Adaptive Loss Function} (MeTAL), a method similar to the aforementioned SCA, which aimed to enhance the few-shot learning performance of MAML by using a meta-learned loss function represented as a set of feedforward neural networks. The novel innovation of MeTAL was to make the learned loss function task-adaptive through the use of a feature-wise linear modulation \citep{perez2018film, dumoulin2018feature}, which linearly modulated the loss function to each new task. The results showed that the optimal loss function may differ between tasks in the context of few-shot learning. By employing a task-adaptive loss function, the proposed method achieved strong few-shot learning performance across a range of well-established benchmarks.
  
\end{itemize}

\clearpage

\section{Chapter Summary}

This chapter provided an overview of machine learning and highlighted several key limitations inherent in conventional supervised learning. Based on these limitations, the paradigm of meta-learning, or learning to learn, was introduced as a promising area of research in Section \ref{section:machine-learning}. Following this, a taxonomy for meta-learning was introduced, discussing meta-learning approaches along three independent axes: meta-objective, meta-optimizer, and meta-representation. This taxonomy served as a framework that guided the subsequent discussions in Sections \ref{section:meta-objective}, \ref{section:meta-optimizer}, and \ref{section:meta-representation}, which discussed existing approaches along with their respective advantages and disadvantages. Finally, Section \ref{section:lfl-related-work} provided an overview of existing meta-learning approaches to loss function learning.
\chapter{Meta-Learning Symbolic Loss Functions}\label{chapter:evomal}

\textit{In this chapter, a new framework is developed for meta-learning loss functions called Evolved Model-Agnostic Loss (EvoMAL), which aims to meta-learn loss functions that significantly improve upon the performance of the models trained by them. The newly proposed method consolidates recent advancements in gradient-based and evolution-based loss function learning, enabling the development of interpretable loss functions on commodity hardware. The experimental results show that the meta-learned loss functions discovered by our framework outperform the \textit{de facto} standard cross-entropy loss (classification) and squared error loss (regression), as well as state-of-the-art loss function learning methods on a diverse range of neural network architectures and datasets.}

\section{Chapter Overview}

Deep neural networks are typically trained through the backpropagation of gradients originating from a handcrafted and manually selected loss function \citep{rumelhart1986learning}. A significant drawback of this approach is that loss functions have traditionally been designed with task-generality in mind, \textit{i.e.}, large expansive classes of tasks in mind, but the system itself is only concerned with a single instantiation or small subset of that class. As shown by the \textit{No Free Lunch Theorems} \citep{wolpert1997no}, specialization to a subclass of tasks is essential for improved performance. Therefore, the conventional method of heuristically selecting a loss function from a modest set of handcrafted options should be reconsidered in favor of a more principled, data-informed approach. 

The emerging field of loss function learning provides an alternative approach that aims to leverage task-specific information and past experiences to infer loss functions directly from data. Initial methods have shown promise in improving deep neural network training. However, they often rely on parametric representations such as neural networks \citep{bechtle2021meta} or Taylor polynomials \citep{gonzalez2021optimizing,  gao2021searching, gao2022loss}, which impose unnecessary assumptions and constraints on the learned loss function's structure. Non-parametric alternative to this use a two-stage process, which infers the loss function structure and parameters using genetic programming and covariance matrix adaptation \citep{gonzalez2020improved}. A method that quickly becomes intractable for large-scale optimization problems. Subsequent work \citep{liu2021loss,li2022autoloss} has attempted to address this issue; however, they crucially omit the optimization stage, which is known to produce sub-optimal performance.

This chapter aims to resolve these issues through a newly proposed framework called \textit{Evolved Model-Agnostic Loss} (EvoMAL), which meta-learns non-parametric symbolic loss functions via a hybrid neuro-symbolic search approach. The newly proposed framework aims to resolve the limitations of past approaches by combining genetic programming \citep{koza1992genetic} with unrolled differentiation \citep{wengert1964simple, domke2012generic, deledalle2014stein, maclaurin2015gradient} --- a gradient-based local-search procedure. This unifies two previously divergent lines of research on loss function learning, which prior to this method, exclusively used either a gradient-based or an evolution-based approach.

This contribution innovates existing approaches to loss function learning by introducing the first computationally tractable approach to optimizing symbolic loss functions, thereby enhancing the algorithm's scalability and performance. Furthermore, unlike prior approaches, the proposed framework is both task and model-agnostic, as it can be applied to any learning algorithm trained with a gradient descent style procedure and is compatible with different model architectures. This branch of general-purpose loss function learning algorithms provides a new powerful meta-learning paradigm for improving a neural network's performance, which has until recently not been explored.

\vspace{-2mm}\subsection{Contributions:}\vspace{-1mm}

The key contributions of this chapter are as follows:

\begin{itemize}

    \item We propose a new task and model-agnostic search space and a corresponding search algorithm for meta-learning interpretable symbolic loss functions.

    \item We demonstrate a simple transition procedure for converting expression tree-based symbolic loss functions into gradient trainable loss networks.

    \item We utilize the new loss function representation to integrate the first computationally tractable approach to optimizing symbolic loss functions into the framework.

    \item We evaluate the performance of the proposed framework by performing the first-ever comparison of state-of-the-art loss function learning techniques.

    \item We conduct experiments to assess the ability of meta-learned loss function to transfer to new unseen tasks and base models at meta-testing time.
    
\end{itemize}

\section{Evolved Model-Agnostic Loss}
\label{sec:method}

In this section, a novel hybrid neuro-symbolic search approach named \textit{Evolved Model-Agnostic Loss (EvoMAL)} is proposed, which consolidates and extends past research on the topic of loss function learning. The proposed method has three key stages, which we discuss in detail in Sections \ref{sec:loss-function-discovery}, \ref{sec:loss-function-transition-procedure}, and \ref{sec:loss-function-optimization}, respectively. First, the structure of the symbolic loss functions are inferred using genetic programming. Following this, a transition procedure is presented which transforms the expression tree-based representation into a new novel representation. Finally, utilizing this new representation the weights/coefficients are optimized via unrolled differentiation. This three-stage process enables the discovery of highly performant and interpretable loss functions.

\subsection{Problem Setup}

The goal of loss function learning is to meta-learn a loss function $\MetaLoss_{\phi}$ with parameters $\phi$, over a distribution of tasks $p(\Task)$ at meta-training time. Each \textit{task} is defined as a set of input-output pairs $\Task = \{(x_{1}, y_{1}), \dots, (x_{N}, y_{N})\}$, with multiple tasks composing a \textit{meta-dataset} $\Dataset = \{\Task_{1}, \Task_{2}, \dots\}$. Then, at meta-testing time the learned loss function $\MetaLoss_{\phi}$ is used in place of a traditional loss function to train a base learner, e.g. a classifier or regressor, denoted by $f_{\theta}(x)$ with parameters $\theta$ on a new unseen task from $p(\Task)$. In this chapter, we constrain the selection of base learners to models that learn via a gradient descent style procedures such that we can optimize weights $\theta$ as follows:
\begin{equation}
\theta_{t+1} = \theta_{t} - \alpha \nabla_{\theta_{t}} \MetaLoss_{\phi}(y, f_{\theta_{t}}(x)).
\label{eq:gradient-decent-setup}
\end{equation}
The proposed framework, EvoMAL, is a new approach to offline loss function learning, a meta-learning paradigm concerned with learning new and performant loss functions that can be used as a drop-in replacement for a prototypical handcrafted loss function such as the squared loss for regression or cross-entropy loss for classification \citep{gonzalez2020improving,gao2023meta}. Offline loss function learning follows a conventional offline meta-learning setup, which partitions the learning into two sequential phases: meta-training and meta-testing. For clarity, we further outline both stages in detail.

\subsubsection{Meta-Training Phase}

This phase is formulated as a bilevel optimization problem, where the goal of the outer optimization is to meta-learn a performant loss function $\MetaLoss_{\phi}$ minimizing the average task loss $\Loss^{meta}$ (the meta-objective) across a distribution of tasks $p(\Task)$, where $\Loss^{meta}$ is selected based on the desired task (typically, the cross-entropy loss for classification, and mean squared error for regression). The inner optimization uses $\MetaLoss_{\phi}$ as the base loss function to train the base model parameters $\theta$. Formally, the meta-training phase of EvoMAL is defined as 
\begin{align}
\label{eq:loss-function-learning-goal-new}
\begin{split}
    \MetaLoss_{\phi}^* &= \argminA_{\MetaLoss_\phi} \mathbb{E}_{\Task_{i} \sim p(\Task)}\left[\Loss^{meta}(y_{i}, f_{\theta^{*}_{i}}(x_{i}))\right]\\ 
    s.t.\;\;\;\;\; \theta^{*}_{i}(\phi)  &= \argminA_{\theta_{i}} \big[\MetaLoss_{\phi}(y, f_{\theta_{i}}(x))\big].
\end{split}
\end{align}

\subsubsection{Meta-Testing Phase}

In the testing phase, the best-performing loss function learned at meta-training time $\MetaLoss_{\phi}^*$ is used directly to optimize the base model parameters
\begin{equation}
\theta^{*} = \argminA_{\theta} \big[\MetaLoss_{\phi}^{*}(y, f_{\theta}(x))\big].
\end{equation}
In contrast to online loss function learning \citep{raymond2023online}, offline loss function learning does not require any alterations to the existing training pipelines to accommodate the meta-learned loss function, this makes the loss functions easily transferable and straightforward to use in existing code bases.

\subsection{Loss Functions Discovery}
\label{sec:loss-function-discovery}

To learn the symbolic structure of the loss functions in EvoMAL, we propose using Genetic Programming (GP), a powerful population-based technique that employs an evolutionary algorithm to directly search the set of primitive mathematical operations \citep{koza1992genetic}. In GP, solutions are composed of terminal and function nodes in a variable-length hierarchical expression tree-based structure. This symbolic structure is a natural and convenient way to represent loss functions, due to its high interpretability and trivial portability to new problems. Transferring a learned loss function from one problem to another requires very little effort, typically only a line or two of additional code. The task and model-agnostic loss functions produced by EvoMAL can be used directly as a drop-in replacement for handcrafted loss functions without requiring any new sophisticated meta-learning pipelines to train the loss on a per-task basis.

\subsubsection{Search Space Design}

\begin{table}[]
\captionsetup{justification=centering}
\caption{The set of searchable primitive mathematical operations used \\ in the loss function discovery phase of EvoMAL.}
\centering
\begin{tabular}{p{4.5cm}cc}
\hline
\textbf{Operator}  & \textbf{Expression}      & \textbf{Arity} \\ \hline \noalign{\vskip 1mm} 
Addition           & $x_1 + x_2$              & 2              \\
Subtraction        & $x_1 - x_2$              & 2              \\
Multiplication     & $x_1 * x_2$              & 2              \\
Division ($AQ$)    & $x_1 / \sqrt{1 + \smash{x_2^2}}$ & 2   \\
Minimum            & $\min(x_1, x_2)$         & 2              \\
Maximum            & $\max(x_1, x_2)$         & 2              \\ \noalign{\vskip 1mm}  \hline \noalign{\vskip 1mm} 
Sign               & sign$(x)$                & 1              \\
Square             & $x^2$                    & 1              \\
Absolute           & $|x|$                    & 1              \\
Logarithm          & $\log(|x| + \epsilon)$   & 1              \\
Square Root        & $\sqrt{\smash[b]{|x| + \epsilon}}$ & 1    \\
Hyperbolic Tangent & $\tanh(x)$               & 1              \\ \noalign{\vskip 1mm}  \hline 
\end{tabular}
\label{table:function-set}
\end{table}

To utilize GP, a search space containing promising loss functions must first be designed. When designing the desired search space, four key considerations are made --- first, the search space should superset existing loss functions such as the squared error in regression and the cross entropy loss in classification. Second, the search space should be dense with promising new loss functions while also containing sufficiently simple loss functions such that cross-task generalization can occur successfully at meta-testing time. Third, ensuring that the search space satisfies the key property of GP closure, \textit{i.e.}, loss functions will not cause \textit{NaN}, \textit{Inf}, undefined, or complex output. Finally, ensuring that the search space is both task and model-agnostic. With these considerations in mind, we present the function set in Table \ref{table:function-set}. Regarding the terminal set, the loss function arguments $f_{\theta}(x)$ and $y$ are used, as well as (ephemeral random) constants $+1$ and $-1$. Unlike previously proposed search spaces for loss function learning, we have made several necessary amendments to ensure proper GP closure, and sufficient task and model-generality. The salient differences are as follows: 

\begin{itemize}

  \item Prior work in \citep{gonzalez2020improved} uses unprotected operations: natural logarithm ($\log(x)$), square root ($\sqrt x$), and division $(x_1/x_2)$. Using these unprotected operations can result in imaginary or undefined output violating the GP closure property. To satisfy the closure property, we replace both the natural log and square root with protected alternatives, as well as replace the division operator with the analytical quotient ($AQ$) operator, a smooth and differentiable approximation to the division operator \citep{ni2012use}.
  
  \item Our proposed search space for meta-learning loss functions is both task and model-agnostic in contrast to \citep{liu2021loss} and \citep{li2022autoloss}, which use multiple aggregation-based and element-wise operations in the function set. These operations are suitable for object detection (the respective paper's target domain) but are not compatible when applied to other tasks such as tabulated and natural language processing problems. This greatly limits their application to the field and the generality of the proposed methods.
  
\end{itemize}

\subsubsection{Search Algorithm Design}

\begin{figure}[t]
\centering

    \includegraphics[height=9.5cm]{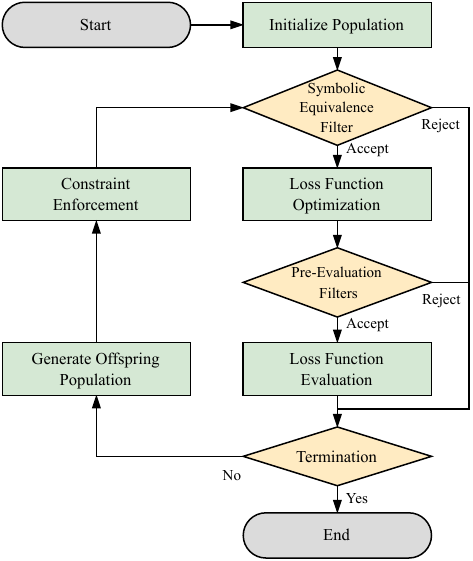}
    \captionsetup{justification=centering}
    \caption{An overview of the EvoMAL framework, showing the process for learning loss functions at meta-training time.}
    \label{fig:flowchart}

\end{figure}

The search algorithm used in the discrete outer optimization process of EvoMAL uses a prototypical implementation of GP. An overview of the algorithm is as follows:

\begin{itemize}[leftmargin=*]

    \item \textbf{Initialization}: To generate the initial population of candidate loss functions, 25 expressions are randomly generated using the \textit{Ramped Half-and-Half} method where the inner nodes are selected from the function set and the leaf nodes from the terminal set. 

\end{itemize}

\noindent
Subsequently, the main loop begins by performing the inner loss function optimization and evaluation process to determine each loss function's respective fitness (discussed in detail in Section \ref{sec:loss-function-evaluation}). Following this, a new offspring population of equivalent size is constructed via the selection, crossover, mutation, and elitism genetic operators.

\begin{itemize}[leftmargin=*]

    \item \textbf{Selection}: Selection of candidate loss functions from the population for crossover and mutation is achieved via a standard \textit{Tournament Selection}, which selects 4 loss functions from the population at random and returns the loss function with the best fitness.

    \item \textbf{Crossover}: Two loss functions are selected and then combined via a \textit{One-Point Crossover} with a probability of 70\%, which slices the two selected loss functions together to form a new loss function. 
  
    \item \textbf{Mutation}: A loss function is selected from the population, and a \textit{Uniform Mutation} is applied with a mutation rate of 25\%, which in place modifies a sub-expression at random with a newly generated sub-expression. 
  
    \item \textbf{Elitism}: To ensure that performance does not degrade elitism is employed to retain the top-performing loss functions with an elitism rate of 5\%. 
  
\end{itemize}

\noindent
The main loop is iteratively repeated up to 50 times until convergence, and the loss function with the best fitness is selected as the final learned loss function. For clarity, we include an overview of the outer optimization process in Figure \ref{fig:flowchart} and in Algorithm \ref{algorithm:genetic-programming}. Note, that to reduce the computational overhead of the meta-learning process, several time-saving measures, which we further refer to as filters, have been incorporated into the EvoMAL algorithm. This is discussed in detail in Section \ref{sec:loss-function-evaluation}.

\begin{algorithm}[t]

\SetAlgoLined
\DontPrintSemicolon
\SetKwInput{Input}{Input}
\BlankLine
\Input{
    $\mathcal{P} \leftarrow$ Population Size, $\mathcal{I} \leftarrow$ Number of Iterations \newline
    $\mathcal{C}r, \mathcal{M}r, \mathcal{E}r \leftarrow$ Crossover Rate, Mutation Rate, Elitism Rate\;
    \vspace{-2mm}
    \hrulefill
}

$population \longleftarrow$ $Initialize(\mathcal{P})$\;
\BlankLine
 
\For{$i \in \{0,\ ...\ , \mathcal{I}\}$}{

    \BlankLine
    $offspring \longleftarrow \{\emptyset\}$\;
    \BlankLine
	
	\For{$j \in \{0,\ ...\ , \mathcal{P}\}$}{
		
		\BlankLine
		$rand \sim \mathcal{U}(0,1)$
		\BlankLine
		
    	\If{$rand < \mathcal{C}r$}{
    	    $\MetaLoss^{(1)}, \MetaLoss^{(2)} \longleftarrow Selection(population)$\;
			$\MetaLoss^{(c)} \longleftarrow Crossover(\MetaLoss^{(1)}, \; \MetaLoss^{(2)})$ 		
		}
		
		\BlankLine
		
		\ElseIf{$rand < \mathcal{M}r + \mathcal{C}r$}{
		    $\MetaLoss^{(1)} \longleftarrow Selection(population)$\;
			$\MetaLoss^{(c)} \longleftarrow Mutation(\MetaLoss^{(1)})$ 		
		}
		\BlankLine
		$offspring \leftarrow offspring \cup \{\MetaLoss^{(c)}\}$\;
    }
    
    \BlankLine
    Transition Procedure (Section \ref{sec:loss-function-transition-procedure})\;
    Loss Optimization (Section \ref{sec:loss-function-optimization})\;
    Fitness Evaluation (Section \ref{sec:loss-function-evaluation})\;
    $population \longleftarrow offspring$\;
    \BlankLine
}
\Return{$\argminA_{\mathcal{F}}(population)$}

\caption{Outer Loss Function Discovery}
\label{algorithm:genetic-programming}
\end{algorithm}

\subsubsection{Constraint Enforcement}
\label{sec:constraints}

\begin{figure}[t]
\centering

    \includegraphics[width=11cm]{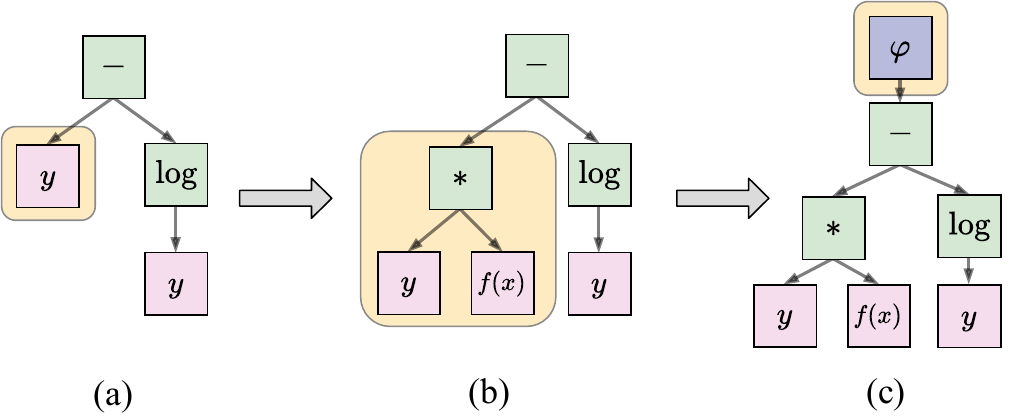}
    \captionsetup{justification=centering}
    \caption{Overview of the constraint enforcement procedure, where (a) is a constraint violating expression, (b) demonstrates enforcing the required arguments constraint, and (c) shows enforcing the non-negative output constraint.}
    
\label{fig:constraints}
\end{figure}

When using GP, the learned expressions can occasionally violate two important constraints of a loss function. \textbf{(1) Required Arguments Constraint:} A loss function must have as arguments $f_{\theta}(x)$ and $y$ by definition. \textbf{(2) Non-Negative Output Constraint:} A loss function should always return a non-negative output such that $\forall x,\forall y, \forall f_{\theta}[\MetaLoss(y, f_{\theta}(x)) \geq 0]$. To resolve these issues we propose two corresponding correction strategies, which we summarize in Figure \ref{fig:constraints} for clarity.

\begin{enumerate}

  \item \textbf{Required Arguments Constraint:} In \citep{gonzalez2020improved}, violating expressions were assigned the worst-case fitness, such that selection pressure would phase out those loss functions from the population. Unfortunately, this approach degrades search performance, as a subset of the population is persistently searching infeasible regions of the search space. To resolve this, we propose a simple but effective corrections strategy to violating loss functions, which randomly selects a terminal node and replaces it with a random binary node, \textit{i.e.}, function node with an arity of 2, with arguments $f_{\theta}(x)$ and $y$ in no predetermined order (required for non-associative operations).
  
  \item \textbf{Non-Negative Output Constraint:} An additional constraint enforced in the EvoMAL algorithm is that the learned loss function should always return a non-negative output such that $\MetaLoss:\mathbb{R}^2 \rightarrow \mathbb{R}_{0}^{+}$. This is achieved via all learned loss function's outputs being passed through an activation function $\varphi$, which was selected to be the smooth $\varphi^{Softplus}=ln(1+e^{x})$ activation.
  
\end{enumerate}

\newpage
\subsection{Loss Function Transition Procedure}
\label{sec:loss-function-transition-procedure}

\begin{figure}[t]
    \centering
    \includegraphics[width=\columnwidth]{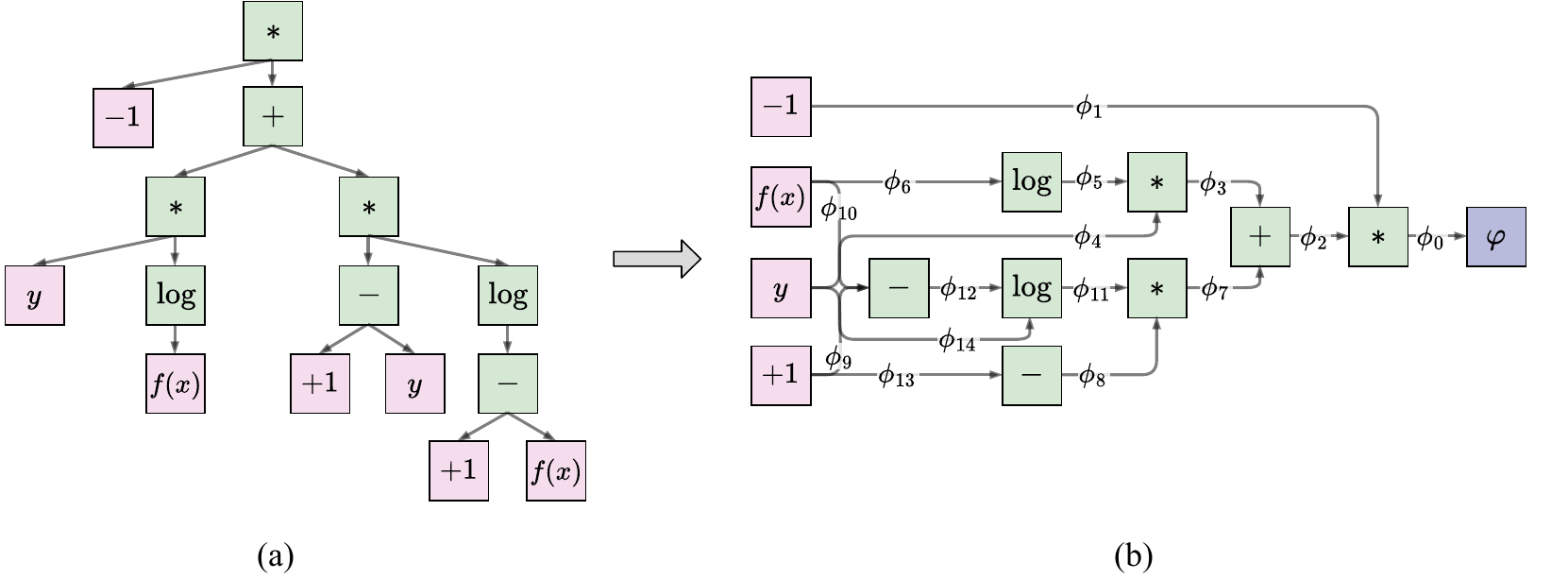} 
    \captionsetup{justification=centering}
    \caption{Overview of the transitional procedure used to covert the learned loss function $\MetaLoss$ shown in (a) into a trainable loss network $\MetaLoss_{\phi}^{\Transpose}$ shown in (b).}
    \label{fig:phase-transformer}
\end{figure}

To integrate unrolled differentiation into the EvoMAL framework, we must first transform the expression tree-based representation of $\MetaLoss$ into a compatible network-style representation. To achieve this, we propose the use of a transitional procedure that takes each loss function $\MetaLoss$, represented as a GP expression, and converts it into an automatic differentiation graph \citep{baydin2018automatic}, \textit{i.e.}, a weighted directed acyclic graph, as shown in Figure \ref{fig:phase-transformer}. First, a graph transpose operation $\MetaLoss^{\Transpose}$ is applied to reverse the edges such that they now go from the terminal (leaf) nodes to the root node. Following this, the edges of $\MetaLoss^{\Transpose}$ are parameterized by the learnable weights $\phi$, giving $\MetaLoss_{\phi}^\Transpose$, which we further refer to as a \textit{loss network} to delineate it clearly from its prior state. Finally, to initialize $\MetaLoss_{\phi}^\Transpose$, the weights are sampled from $\phi\sim\mathcal{N}(1, 1\mathrm{e}{-3})$, such that $\MetaLoss_{\phi}^\Transpose$ is initialized from its (near) original unit form, where the small amount of variance is to break network symmetry. The full transitional procedure is summarized in Algorithm \ref{alg:transition-procedure}. For computational efficiency, we use an adjacency list representation at the implementation level. This enables both the transpose and parameterization steps to occur simultaneously with a linear time and space complexity $\Theta(|\mathcal{V}| + |\mathcal{E}|)$ with respect to the number of vertices $\mathcal{V}$ and edges $\mathcal{E}$ (\textit{i.e.}, nodes and weights) in the learned loss function.

The newly proposed representation is an important and novel development as it bridges the expression tree-based representation commonly used in GP with an automatic differentiation graph-based representation. When represented as an automatic differentiation graph, well-established automatic differentiation libraries such as PyTorch \citep{paszke2017automatic}, TensorFlow \citep{abadi2016tensorflow}, and Jax \citep{jax2018github}, can be leveraged to optimize the parameters. This is valuable as alternative methods of differentiation such as manual, symbolic, or numerical differentiation suffer from numerous issues such as truncation and round-off error, expression swell, or poor scalability \citep{baydin2018automatic}.

\begin{algorithm}[]

\SetAlgoLined
\DontPrintSemicolon
\SetKwInput{Input}{Input}
\BlankLine
\Input{
    $\MetaLoss \leftarrow$ Loss function learned by GP\;
    \vspace{-2mm}
    \hrulefill
}

\BlankLine
$\mathcal{V},\mathcal{E} \leftarrow$ Initialize empty lists for nodes and edges\;
\BlankLine
\For {v $\in \MetaLoss$}{
    \For{$c \in child[v]$}{
        \BlankLine
        $\mathcal{V}[c]$ $\leftarrow$ $\mathcal{V}[c] \frown v$\;
        $\mathcal{E}[c]$ $\leftarrow$ $\mathcal{E}[c] \frown \phi \sim \mathcal{N}(1, 1\mathrm{e}{-3})$\;
        \BlankLine
    }
}
\Return $\MetaLoss_{\phi}^{\Transpose} \leftarrow (\mathcal{V}, \mathcal{E})$\;
\BlankLine

\caption{Loss Function Transition Procedure}
\label{alg:transition-procedure}
\end{algorithm}

\subsection{Loss Function Optimization}
\label{sec:loss-function-optimization}

\begin{algorithm}[t]
\SetAlgoLined
\DontPrintSemicolon
\SetKwInput{Input}{Input}
\BlankLine
\Input{
    $\MetaLoss \leftarrow$ Loss function learned by GP\newline
    $\mathcal{S}^{meta} \leftarrow$ Number of meta gradient steps\newline
    $\mathcal{S}^{base} \leftarrow$ Number of base gradient steps\;
    \vspace{-2mm}
    \hrulefill
}

\BlankLine
$\MetaLoss_{\phi}^{\Transpose} \leftarrow$ Transpose and parameterize edges of $\MetaLoss$\;
$\phi_0 \leftarrow$ Initialize loss network weights $\MetaLoss_{\phi}^{\Transpose}$\;
\For{$i \in \{0, ... , \mathcal{S}^{meta}\}$}{
    \For{$j \in \{0, ... , |\Dataset^{train}|\}$}{
        $\theta_0 \leftarrow$ Initialize parameters of base learner\;
        \For{$t \in \{0, ... , \mathcal{S}^{base}\}$}{
            $X_j$, $y_j$ $\leftarrow$ Sample task $\Task_{j} \sim p(\Task)$\;
            $\smash{\MetaLoss^{learned} \leftarrow \MetaLoss_{\phi_i}^{\Transpose}(y_j, f_{\theta_t}(X_j))}$\;
            $\smash{\theta_{t+1} \leftarrow \theta_{t} - \alpha \nabla_{\theta_t} \MetaLoss^{learned}}$\;
        }
        $\Loss^{task_{j}} \leftarrow \Loss^{meta}(y_j, f_{\theta_{t+1}}(X_j))$\;
    }
    $\phi_{i+1} \leftarrow \phi_{i} - \eta \nabla_{\phi_i} \sum_{j} \Loss^{task_{j}}$\;
}

\BlankLine

\caption{Loss Function Optimization}
\label{alg:meta-training}
\end{algorithm}

To optimize the weights of our learned loss functions, we employ unrolled differentiation a popular meta-optimization method for calculating exact gradients of the validation loss with respect to high-order variables (such as hyper-parameters, or in our work $\phi$) by propagating derivatives backward through the entire learning trajectory. For simplicity, we constrain the description of loss function optimization to the vanilla backpropagation case where the meta-training set $\Dataset^{train}$ contains one task, \textit{i.e.}, $|\Dataset^{train}| = 1$; however, the full process where $|\Dataset^{train}| > 1$ is given in Algorithm \ref{alg:meta-training}. 

\subsubsection{Unrolled Differentiation}

To learn the weights $\phi$ of the loss network $\MetaLoss_{\phi}^\Transpose$ at meta-training time with respect to base learner $f_{\theta}(x)$, we first use the initial values of $\phi_t$ to produce a base loss value $\MetaLoss^{learned}$ based on the forward pass of $f_{\theta_t}(x)$.
\begin{equation}
\MetaLoss^{learned} = \MetaLoss_{\phi_t}^{\Transpose}(y, f_{\theta_t}(x))
\label{eq:inner-loss}
\end{equation}
\noindent
Using $\MetaLoss^{learned}$, the weights $\theta_t$ are optimized by taking a predetermined number of inner base gradient steps $\mathcal{S}^{base}$, where at each step a new batch is sampled and a new base loss value is computed. Similar to the findings in \citep{bechtle2021meta}, we find $\mathcal{S}^{base}=1$ is usually sufficient to obtain good results. Each step is computed by taking the gradient of the loss value with respect to $\theta$, where $\alpha$ is the base learning rate. 
\begin{equation}
\begin{split}
\theta_{t+1}
& = \theta_t - \alpha \nabla_{\theta_t} \MetaLoss_{\phi_t}^{\Transpose}(y, f_{\theta_t}(x)) \\
& = \theta_t - \alpha \nabla_{\theta_t} \mathbb{E}_{\mathsmaller{X}, y} \big[ \MetaLoss_{\phi_t}^{\Transpose}(y, f_{\theta_t}(x)) \big]
\end{split}
\label{eq:inner-backward}
\end{equation}
\noindent
where the gradient computation can be decomposed via the chain rule into the gradient of $\MetaLoss_{\phi_t}^{\Transpose}$ with respect to the product of the base learner predictions $f_{\theta_t}(x)$ and the gradient of $f$ with parameters $\theta_t$.
\begin{equation}
\theta_{t+1} = \theta_t - \alpha \nabla_{f} \MetaLoss_{\phi_t}^{\Transpose}(y, f_{\theta_t}(x)) \nabla_{\theta_t}f_{\theta_t}(x)
\label{eq:inner-backward-decompose}
\end{equation}
Following this, $\theta_t$ has been updated to $\theta_{t+1}$ based on the current loss network weights; $\phi_t$ now needs to be updated to $\phi_{t+1}$ based on how much learning progress has been made. Using the new base learner weights $\theta_{t+1}$ as a function of $\phi_t$, we utilize the concept of a \textit{task loss} $\Loss^{meta}$ to produce a meta loss value $\Loss^{task}$ to optimize $\phi_t$ through $\theta_{t+1}$.
\begin{equation}
\Loss^{task} = \Loss^{meta}(y, f_{\theta_{t+1}}(x))
\label{eq:outer-loss}
\end{equation}
where $\Loss^{meta}$ is selected based on the respective application --- for example, the mean squared error loss for the task of regression or the categorical cross-entropy loss for multi-class classification. Optimization of the loss network loss weights $\phi$ now occurs by taking the gradient of $\Loss^{meta}$ with respect to $\phi_t$, where $\eta$ is the meta learning rate.
\begin{equation}
\begin{split}
\phi_{t+1}
& = \phi_t - \eta \nabla_{\phi_t}\Loss^{meta}(y, f_{\theta_{t+1}}(x)) \\
& = \phi_t - \eta \nabla_{\phi_t} \mathbb{E}_{\mathsmaller{X}, y} \big[ \Loss^{meta}(y, f_{\theta_{t+1}}(x)) \big]
\end{split}
\label{eq:outer-backward}
\end{equation}
where the gradient computation can be decomposed by applying the chain rule as shown in Equation (\ref{eq:outer-backward-decompose}) where the gradient with respect to the loss network weights $\phi_t$ requires the new model parameters $\theta_{t+1}$. 
\begin{equation}
\phi_{t+1} = \phi_t - \eta \nabla_{f}\Loss^{meta} \nabla_{\theta_{t+1}} f_{\theta_{t+1}} \nabla_{\phi_t}\theta_{t+1}(\phi_t) 
\label{eq:outer-backward-decompose}
\end{equation}
This process is repeated for a predetermined number of meta gradient steps $S^{meta}=250$, which was selected via cross-validation. Following each meta gradient step, the base learner weights $\theta$ are reset back to $\theta_0$ (line 5 of Algorithm \ref{alg:meta-training}). Note that Equations (\ref{eq:inner-backward} -- \ref{eq:inner-backward-decompose}) and (\ref{eq:outer-backward} -- \ref{eq:outer-backward-decompose}) are performed via automatic differentiation.

\begin{sidewaysfigure}[]

    \includegraphics[width=\textwidth]{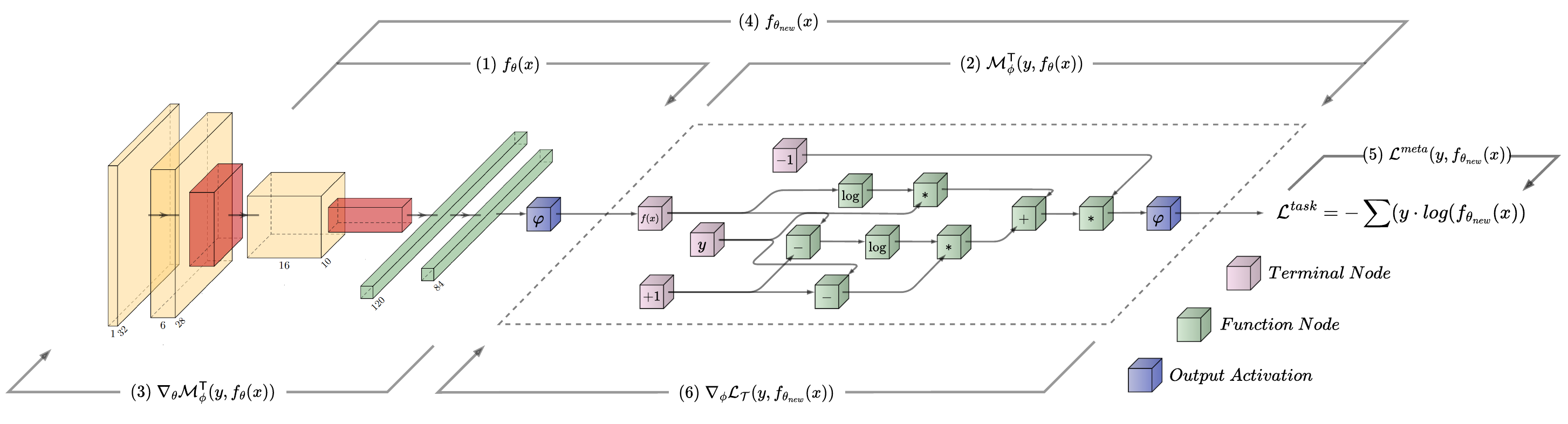}
    \vspace*{+3mm}
    \captionsetup{justification=centering}
    \caption{An overview of the EvoMAL algorithm at \textit{meta-testing} time, where the base network $f_{\theta}(x)$, shown (left) as the popular LeNet-5 architecture, is trained using the loss network $\MetaLoss_{\phi}^{\Transpose}$ (right) found at \textit{meta-training} time as the loss function.}

\label{fig:network}
\end{sidewaysfigure}

\subsection{Loss Function Evaluation}
\label{sec:loss-function-evaluation}

\begin{algorithm}[t]
\SetAlgoLined
\DontPrintSemicolon
\SetKwInput{Input}{Input}
\BlankLine
\Input{
    $\MetaLoss_{\phi}^{\Transpose} \leftarrow$ Loss function learned by EvoMAL\newline
    $S^{testing} \leftarrow$ Number of base testing gradient steps\;
    \vspace{-2mm}
    \hrulefill
}

\BlankLine
\For{$j \in \{0, ... , |\Dataset|\}$}{
    $\theta_{0} \leftarrow$ Initialize parameters of base learner\;
    $X_{j}$, $y_{j}$ $\leftarrow$ Sample task $\Task_{j} \sim p(\Task)$\;
    \For{$t \in \{0, ... , S^{testing}\}$}{
        $\Loss^{learned} \leftarrow \MetaLoss_{\phi}^{\Transpose}(y_{j}, f_{\theta_{t}}(X_{j}))$\;
        $\theta_{t+1} \leftarrow \theta_{t} - \alpha \nabla_{\theta_{t}} \Loss^{learned}$\;
    }
}
$\Fitness \leftarrow \frac{1}{|\Dataset|}\sum_{j}\Loss^{PM}(y_{j}, f_{\theta_{t+1}}(X_{j}))$

\BlankLine

\caption{Loss Function Evaluation}
\label{alg:meta-testing}
\end{algorithm}

To derive the fitness $\Fitness$ of $\MetaLoss_{\phi}^{\Transpose}$, a conventional training procedure is used as summarized in Algorithm \ref{alg:meta-testing}, where $\MetaLoss_{\phi}^{\Transpose}$ is used in place of a traditional loss function to train $f_{\theta}(x)$ over a predetermined number of base gradient steps $S^{testing}$. This training process is identical to training at meta-testing time as shown in Figure \ref{fig:network}. The final inference performance of $\MetaLoss_{\phi}^{\Transpose}$ is assigned to $\Fitness$, where any differentiable or non-differentiable performance metric $\Loss^{PM}$ can be used. For our experiments, we calculate $\Fitness$ using the error rate for the classification tasks and mean squared error for the regression tasks.
\begin{equation}
\Fitness = \frac{1}{|\Dataset|}\sum_{i}\Loss^{PM}(y_{i}, f_{\theta_{i}}(X_{i}))
\label{eq:fitness}
\end{equation}
Optimizing and evaluating a large number of candidate loss functions can become prohibitively expensive. Fortunately, in the case of loss function learning, several techniques can be employed to reduce significantly the computational overhead of the otherwise costly meta-learning process. 

\subsubsection{Symbolic Equivalence Filter}

For the GP-based symbolic search, a loss archival strategy based on a key-value pair structure with $\Theta(1)$ lookup is used to ensure that symbolically equivalent loss functions are not reevaluated. Two expression trees are said to be symbolically equivalent when they contain identical operations (nodes) in an identical configuration \citep{wong2006algebraic, kinzett2009numerical, javed2022simplification}. Loss functions that are identified by the symbolic equivalence filter skip both the loss function optimization and evaluation stage and are placed directly in the offspring population with their fitness cached.

\subsubsection{Poor Training Dynamics Filter}

Following loss function optimization, the candidate solution's fitnesses are evaluated. This costly fitness evaluation can be obviated in many cases since many of the loss functions found, especially in the early generations, are non-convergent and produce poor training dynamics. We use the loss rejection protocol employed in \citep{li2022autoloss} as a filter to identify candidate loss functions that should skip evaluation and be assigned the worst-case fitness automatically.

The loss rejection protocol takes a batch of $B$ randomly sampled instances from $\Dataset^{train}$ and using an untrained network $f_{\theta_{0}}(x)$ produces a set of predictions and their corresponding true target values $\{(\hat{y_b}, y_b)\}_{b=1}^{B}$. As minimizing the proper loss function $\MetaLoss_{\phi}^{\Transpose}$ should correspond closely with optimizing the performance metric $\Loss^{PM}$, a correlation $g$ between $\MetaLoss_{\phi}^{\Transpose}$ and $\Loss^{PM}$ can be calculated as
\begin{equation}
g(\MetaLoss_{\phi}^{\Transpose}) = \sum_{b=1}^{B} \Big[\Loss^{PM}(\hat{y_{b}}, y_{b}) - \Loss^{PM}(\hat{y_{b}^{*}}(\MetaLoss_{\phi}^{\Transpose}), y_{b})\Big],
\end{equation}
where $\hat{y_{b}^{*}}$ is the network predictions optimized with the candidate loss function $\MetaLoss_{\phi}^{\Transpose}$. Importantly, optimization is performed directly to $\hat{y_{b}^{*}}$, as opposed to the network parameters $\theta$, thus omitting any base network computation (\textit{i.e.}, no training of the base network).
\begin{equation}
\hat{y_{b}^{*}}(\MetaLoss_{\phi}^{\Transpose}) = \arg \min_{\hat{y_{b}^{*}}} \MetaLoss_{\phi}^{\Transpose}(\hat{y_{b}^{*}}, y_{b})
\end{equation}
A large positive correlation indicates that minimizing $\MetaLoss_{\phi}^{\Transpose}$ corresponds to minimizing the given performance metric $\Loss^{PM}$ (assuming both $\MetaLoss_{\phi}^{\Transpose}$ and $\Loss^{PM}$ are for minimization). In contrast to this, if $g\leq0$, then $\MetaLoss_{\phi}^{\Transpose}$ is regarded as being unpromising and should be assigned the worst-case fitness and not evaluated. The underlying assumption here is that if a loss function cannot directly optimize the labels, it is unlikely to be able to optimize the labels through the model weights $\theta$ successfully.

\subsubsection{Gradient Equivalence Filter}

Many of the candidate loss functions found in the later generations, as convergence is approached, have near-identical gradient behavior (\textit{i.e.}, functionally equivalence). To address this, the gradient equivalence checking strategy from \citep{li2022autoloss} is employed as another filter to identify loss functions that have near-identical behavior to those seen previously. Using the prediction from previously, the gradient norms are computed.
\begin{equation}
\{\parallel\nabla_{\hat{y}_{b}}\MetaLoss_{\phi}^{\Transpose}\parallel_{2}\}_{b=1}^{B}
\end{equation}
If, for all of the $B$ samples, two-loss functions have the same gradient norms within two significant digits, they are considered functionally equivalent, and their fitness is cached.

\subsubsection{Partial Training Sessions}

For the remaining loss functions whose fitness evaluation cannot be fully obviated, we compute the fitness $\Fitness$ using a truncated number of gradient steps $\mathcal{S}^{testing}=500$. As noted in \citep{grefenstette1985genetic, jin2011surrogate}, performance at the beginning of training is correlated with the performance at the end of training; consequently, we can obtain an estimate of what $\Fitness$ would be by performing a partial training session of the base model. Preliminary experiments with EvoMAL showed minimal short-horizon bias \citep{wu2018understanding}, and the ablation study found in \citep{gonzalez2021optimizing} indicated that $500$ gradient steps during loss function evaluation are a good trade-off between final base-inference performance and meta-training time. In addition to significantly reducing the runtime of EvoMAL, reducing the value of $\mathcal{S}^{testing}$ has the effect of implicitly optimizing for the convergence and sample-efficiency, as mentioned in \citep{hospedales2020meta}.

\section{Background and Related Work}
\label{sec:background}

The method we propose in this chapter addresses the problem of meta-learning a general-purpose model-agnostic (base) loss function, \textit{i.e.}, offline loss function learning. Several approaches have recently been proposed to accomplish this task, and a relevant trend among these methods is that they often fall into one of the following two key categories.

\subsection{Gradient-Based Approaches}

Gradient-based approaches predominantly aim to learn a loss function $\MetaLoss$ through the use of a meta-level neural network to improve on various aspects of the training. For example, in \citep{grabocka2019learning,huang2019addressing}, differentiable surrogates of non-differentiable performance metrics are learned to reduce the misalignment problem between the performance metric and the loss function. Alternatively, in \citep{houthooft2018evolved,wu2018learning,antoniou2019learning,kirsch2019improving,bechtle2021meta,collet2022loss,leng2022polyloss,raymond2023online}, loss functions are learned to improve sample efficiency and asymptotic performance in supervised and reinforcement learning, while in \citep{balaji2018metareg,barron2019general,li2019feature,gao2022loss}, they improved on the robustness of a model.

While the aforementioned approaches have achieved some success, they have notable limitations. The most salient limitation is that they \textit{a priori} assume a parametric form for the loss functions. For example, in \citep{bechtle2021meta} and \citep{psaros2022meta}, it is assumed that the loss functions take on the parametric form of a two hidden layer feed-forward neural network with 50 nodes in each layer and ReLU activations. However, such an assumption imposes a bias on the search, often leading to an over-parameterized and sub-optimal loss function. Another limitation is that these approaches often learn black-box (sub-symbolic) loss functions, which is not ideal, especially in the meta-learning context where \textit{post hoc} analysis of the learned component is crucial, before transferring the learned loss function to new unseen problems at meta-testing time.

\subsection{Evolution-Based Approaches}

A promising alternative paradigm is to use evolution-based methods to learn $\MetaLoss_{\phi}$, favoring their inherent ability to avoid local optima via maintaining a population of solutions, their ease of parallelization of computation across multiple processors, and their ability to optimize for non-differentiable functions directly. Examples of such work include \citep{gonzalez2021optimizing} and \citep{gao2021searching}, which both represent $\MetaLoss_{\phi}$ as parameterized Taylor polynomials optimized with covariance matrix adaptation evolutionary strategies (CMA-ES). These approaches successfully derive interpretable loss functions, but similar to previously, they also assume the parametric form via the degree of the polynomial.

To resolve the issue of having to assume the parametric form of $\MetaLoss_{\phi}$, another avenue of research first presented in \citep{gonzalez2020improved} investigated the use of genetic programming (GP) to learn the structure of $\MetaLoss_{\phi}$ in a symbolic form before applying CMA-ES to optimize the parameterized loss. The proposed method was effective at learning performant loss functions and clearly demonstrated the importance of local-search. However, the method had intractable computational costs as using a population-based method (GP) with another population-based method (CMA-ES) resulted in a significant expansion in the number of evaluations at meta-training time, hence it needing to be run on a supercomputer in addition to using a truncated number of training steps.

Subsequent work in \citep{liu2021loss} and \citep{li2022autoloss} reduced the computational cost of GP-based loss function learning approaches by proposing time-saving mechanisms such as rejection protocols, gradient-equivalence-checking, convergence property verification, and model optimization simulation. These methods successfully reduced the wall-time of GP-based approaches; however, both works omit the use of local-search strategies, which is known to cause sub-optimal performance when using GP \citep{topchy2001faster,smart2004applying,zhang2005learning}. Furthermore, neither method is task and model-agnostic, limiting their utility to a narrow set of domains and applications.

\section{Experimental Setup}
\label{sec:experiment-setup}

In this section, the performance of EvoMAL is evaluated. A wide range of experiments are conducted across seven datasets and numerous popular network architectures, with the performance contrasted against a representative set of benchmark methods. All the algorithms are implemented in \texttt{DEAP} \citep{fortin2012deap}, \texttt{PyTorch} \citep{paszke2019pytorch} and \texttt{Higher} \citep{grefenstette2019generalized}, and the relevant code can be found at \href{https://github.com/Decadz/Evolved-Model-Agnostic-Loss}{https://github.com/Decadz/Evolved-Model-Agnostic-Loss}.

\subsection{Benchmark Methods}

The selection of benchmark methods is intended to showcase the performance of the newly proposed algorithm against the current state-of-the-art. Additionally, the selected methods enable direct comparison between EvoMAL and its derivative methods, which aids in validating the effectiveness of hybridizing the approaches into one unified framework.
\begin{itemize}

  \item \textbf{Baseline} -- Directly using $\Loss^{meta}$ as the loss function, \textit{i.e.}, using the squared error loss (regression) or cross-entropy loss (classification) and a prototypical training loop (no meta-learning). 
  
  \item \textbf{ML$^3$ Supervised} -- Gradient-based method proposed by \cite{bechtle2021meta}, which uses a parametric loss function defined by a two hidden layer feed-forward network trained with unrolled differentiation, \textit{i.e.}, the method shown in Section \ref{sec:loss-function-optimization}.
  
  \item \textbf{TaylorGLO} -- Evolution-based method proposed by \cite{gonzalez2021optimizing}, which uses a third-order Taylor-polynomial representation for the meta-learned loss functions, optimized via covariance matrix adaptation evolution strategy.
  
  \item \textbf{GP-LFL} -- A proxy method used to aggregate previous GP-based approaches for loss function learning \textit{without} any local-search mechanisms, using an identical setup to EvoMAL excluding Section \ref{sec:loss-function-optimization}.
  
\end{itemize}
\noindent
Where possible, hyper-parameter selection has been standardized across the benchmark methods to allow for a fair comparison. For example, in TaylorGLO, GP-LFL, and EvoMAL we use an identical population size $=25$ and number of generations = $50$. For unique hyper-parameters, the suggested values from the respective publications are utilized.

\subsection{Benchmark Tasks}

Regarding the problem domains, seven datasets have been selected. Three tabulated regression tasks are initially used: Diabetes, Boston Housing, and California Housing, all taken from the UCI's dataset repository \citep{asuncion2007uci}. Following this, analogous to the prior literature \citep{gonzalez2020improved, gonzalez2021optimizing, bechtle2021meta}, both MNIST \citep{lecun1998gradient} and CIFAR-10 \citep{krizhevsky2009learning} are employed to evaluate the benchmark methods. Finally, experiments are conducted on the more challenging but related domains of SVHN \citep{netzer2011reading} and CIFAR-100 \citep{krizhevsky2009learning}, respectively.

For the three tabulated regression tasks, each dataset is partitioned 60:20:20 for training, validation, and testing. Furthermore, to improve the training dynamics, both the features and labels are normalized. For the remaining datasets, the original training-testing partitioning is used, with 10\% of the training instances allocated for validation. In addition, data augmentation techniques consisting of normalization, random horizontal flips, and cropping are applied to the training data during meta and base training.

\subsection{Benchmark Models}

Regarding the base models, a diverse set of neural network architectures are utilized to evaluate the selected benchmark methods. For Diabetes, Boston Housing, and California Housing, a simple Multi-Layer Perceptron (MLP) taken from \citep{baydin2018hypergradient} with 1000 hidden nodes and ReLU activations are employed. For MNIST, Logistic Regression \citep{mccullagh2019generalized}, MLP, and the well-known LeNet-5 architecture \citep{lecun1998gradient} are used. While on CIFAR-10 AlexNet \citep{krizhevsky2012imagenet}, VGG-16 \citep{simonyan2014very}, AllCNN-C \citep{springenberg2014striving}, ResNet-18 \citep{he2016deep}, Preactivation ResNet-101 \citep{he2016identity}, WideResNet 28-10 \citep{zagoruyko2016wide} and SqueezeNet \citep{iandola2016squeezenet} are used. For the remaining datasets, WideResNet 28-10 is again used, as well as PyramidNet \citep{han2017deep} on CIFAR-100. All models are trained using stochastic gradient descent (SGD) with momentum. The model hyper-parameters are selected using their respective values from the literature in an identical setup to \citep{gonzalez2021optimizing, bechtle2021meta}.

Finally, due to the stochastic nature of the benchmark methods, we perform five independent executions of each method on each dataset + model pair. Furthermore, we control for the base initializations such that each method gets identical initial conditions across the same random seed; thus, any difference in variance between the methods can be attributed to the respective algorithms.

\section{Results and Analysis}
\label{sec:results}

The results and analysis are presented from two distinct angles:
\begin{itemize}

  \item First, we analyze the performance of the base models trained by our meta-learned loss functions. In Section \ref{sec:meta-testing}, we report and analyze the final inference meta-testing performance when both the trained and testing domains are identical. Following this in Section \ref{sec:meta-transfer}, we consider the case of loss function transfer, where the training and testing domains are different.
  
  \item Second, the performance of the loss function learning algorithms themselves is analyzed. We investigate the meta-training learning curves in Section \ref{sec:meta-training-curves}, the algorithmic runtime in Section \ref{sec:runtime-analysis}, and finally we close by performing an analysis of our novel search space in Section \ref{sec:search-space-analysis}. 
  
\end{itemize}

\noindent
In the following chapter, considerable time will be spent analyzing EvoMAL's meta-learned loss functions to better understand both empirically and theoretically the role our learned loss function plays.

\subsection{Meta-Testing Performance}
\label{sec:meta-testing}

A summary of the final inference testing results across the seven tested datasets is shown in Tables \ref{table:meta-testing-results-classification} and \ref{table:meta-testing-results-regression} respectively, where the same dataset and model pair are used for both meta-training and meta-testing. The results show that meta-learned loss functions consistently perform better than the baseline handcrafted loss functions. Large performance gains are made on the California Housing, MNIST, and CIFAR-10 datasets, while more modest gains are observed on Diabetes, WideResNet CIFAR-100 and SVHN, and worse performance on Boston Housing MLP and CIFAR-100 PyramidNet, a similar finding to that found in \citep{gonzalez2021optimizing}. 

Contrasting the performance of EvoMAL to the benchmark loss function learning methods, it is shown that EvoMAL consistently meta-learns more performant loss functions, with better performance on all task + model pairs except for on CIFAR-10 VGG-16, where performance is comparable to the next best method. Furthermore, compared to its derivative methods ML$^3$ and GP-LFL, EvoMAL successfully meta-learns loss functions on more complex tasks, \textit{i.e.}, CIFAR-100 and SVHN, whereas the other techniques often struggle to improve upon the baseline. These results empirically confirm the benefits of unifying existing approaches to loss function learning into one unified framework. Furthermore, the results clearly show the necessity for integrating local-search techniques in GP-based loss function learning.

Compared to prior research on loss function learning, our method exhibits a relatively smaller improvement when comparing the use of meta-learned loss functions to handcrafted loss functions. For example, prior research has reported increasing the accuracy of a classification model by up to 5\% in some cases when using meta-learned loss functions compared to the cross-entropy loss. However, with heavily tuned baselines, optimizing for both $\alpha$ and $\mathcal{S}^{testing}$, such performance gains were very hard to obtain. This suggests that a proportion of the performance gains reported previously by loss function learning methods likely come from an implicit tuning effect on the training dynamics as opposed to a direct effect from using a different loss function. Implicit tuning is not a drawback of loss function learning as a paradigm; however, it is essential to disentangle the effects, as we will do in the following chapter.

Finally, although the performance gains are at times modest, it is impressive that performance gains can be made by a straightforward change to the selection of the loss function. Many previous developments in neural networks have added many millions of parameters or incorporated complex training routines to achieve similar improvements in performance \citep{liu2017survey,gu2018recent}. Furthermore, our results show that performance gains can be made upon well-tuned models that already employ a multitude of regularization techniques. This suggests that meta-learned loss functions can learn a distinct form of regularization that existing techniques such as data augmentation, dropout, weight decay, batch norm, skip layers, hyper-parameter optimization, etc., fail to capture.

\begin{sidewaystable}
\centering
\captionsetup{justification=centering}
\caption{Classification results reporting the mean $\pm$ standard deviation final inference error rate across 5 independent executions of each algorithm on each task + model pair. Loss functions are directly meta-learned and applied to the same respective task.}
\begin{threeparttable}

\begin{tabular}{lccccc}
\hline
\noalign{\vskip 1mm}
Task and Model          & Baseline          & ML$^3$            & TaylorGLO         & GP-LFL            & EvoMAL (Ours)                     \\ \hline \noalign{\vskip 1mm}
\textbf{MNIST}          &                   &                   &                   &                   &                                   \\
Logistic \tnote{1}      & 0.0787$\pm$0.0009 & 0.0768$\pm$0.0061 & 0.0725$\pm$0.0013 & 0.0781$\pm$0.0052 & \textbf{0.0721$\pm$0.0017}        \\ 
MLP \tnote{2}           & 0.0247$\pm$0.0005 & 0.0201$\pm$0.0081 & \textbf{0.0151$\pm$0.0013} & 0.0156$\pm$0.0012 & 0.0152$\pm$0.0012        \\ 
LeNet-5 \tnote{3}       & 0.0203$\pm$0.0025 & 0.0135$\pm$0.0039 & 0.0137$\pm$0.0038 & 0.0115$\pm$0.0015 & \textbf{0.0100$\pm$0.0010}        \\ \hline \noalign{\vskip 1mm} 
\textbf{CIFAR-10}       &                   &                   &                   &                   &                                   \\ 
AlexNet \tnote{4}       & 0.1544$\pm$0.0012 & 0.1450$\pm$0.0028 & 0.1499$\pm$0.0075 & 0.1506$\pm$0.0047 & \textbf{0.1437$\pm$0.0033}        \\
VGG-16 \tnote{5}        & 0.0771$\pm$0.0023 & 0.0700$\pm$0.0006 & 0.0700$\pm$0.0022 & \textbf{0.0686$\pm$0.0014} & 0.0687$\pm$0.0016        \\
AllCNN-C \tnote{6}      & 0.0761$\pm$0.0015 & 0.0712$\pm$0.0043 & 0.0735$\pm$0.0030 & 0.0701$\pm$0.0022 & \textbf{0.0697$\pm$0.0010}        \\
ResNet-18 \tnote{7}     & 0.0658$\pm$0.0019 & 0.0584$\pm$0.0022 & 0.0546$\pm$0.0033 & 0.0818$\pm$0.0391 & \textbf{0.0528$\pm$0.0015}        \\
PreResNet \tnote{8}     & 0.0661$\pm$0.0015 & 0.0660$\pm$0.0016 & 0.0660$\pm$0.0027 & 0.0658$\pm$0.0023 & \textbf{0.0655$\pm$0.0018}        \\
WideResNet \tnote{9}    & 0.0548$\pm$0.0016 & 0.0549$\pm$0.0040 & 0.0493$\pm$0.0023 & 0.0489$\pm$0.0014 & \textbf{0.0484$\pm$0.0018}        \\
SqueezeNet \tnote{10}   & 0.0838$\pm$0.0013 & 0.0800$\pm$0.0012 & 0.0800$\pm$0.0025 & 0.0810$\pm$0.0016 & \textbf{0.0796$\pm$0.0017}        \\ \noalign{\vskip 1mm} \hline \noalign{\vskip 1mm} 
\textbf{CIFAR-100}      &                   &                   &                   &                   &                                   \\
WideResNet \tnote{9}    & 0.2293$\pm$0.0017 & 0.2299$\pm$0.0027 & 0.2347$\pm$0.0077 & 0.3382$\pm$0.1406 & \textbf{0.2276$\pm$0.0033}        \\
PyramidNet \tnote{11}   & \textbf{0.2527$\pm$0.0028} & 0.2792$\pm$0.0226 & 0.3064$\pm$0.0549 & 0.2747$\pm$0.0087 & 0.2664$\pm$0.0063        \\ \noalign{\vskip 1mm} \hline \noalign{\vskip 1mm} 
\textbf{SVHN}           &                   &                   &                   &                   &                                   \\
WideResNet \tnote{9}    & 0.0340$\pm$0.0005 & 0.0335$\pm$0.0003 & 0.0343$\pm$0.0016 & 0.0340$\pm$0.0015 & \textbf{0.0329$\pm$0.0013}        \\ \noalign{\vskip 1mm} \hline
\end{tabular}

\begin{tablenotes}\centering
\tiny Network architecture references: \item[1] \citep{mccullagh2019generalized} \item[2] \citep{baydin2018hypergradient} \item[3] \citep{lecun1998gradient} \item[4] \citep{krizhevsky2012imagenet} \item[5] \citep{simonyan2014very} \item[6] \citep{springenberg2014striving} \item[7] \citep{he2016deep}  \item[8] \citep{he2016identity} \item[9] \citep{zagoruyko2016wide} \item[10] \citep{iandola2016squeezenet} \item[11] \citep{han2017deep}
\end{tablenotes}

\end{threeparttable}
\label{table:meta-testing-results-classification}
\end{sidewaystable}

\begin{sidewaystable}
\centering
\captionsetup{justification=centering}
\caption{Regression results reporting the mean $\pm$ standard deviation final inference testing mean squared error across 5 executions of each algorithm on each task + model pair. Loss functions are directly meta-learned and applied to the same respective task.}
\begin{threeparttable}

\begin{tabular}{lccccc}
\hline
\noalign{\vskip 1mm}
Task and Model          & Baseline          & ML$^3$            & TaylorGLO         & GP-LFL            & EvoMAL (Ours)                     \\ \hline \noalign{\vskip 1mm}
\textbf{Diabetes}       &                   &                   &                   &                   &                                   \\
MLP      \tnote{1}      & 0.3829$\pm$0.0065 & 0.4278$\pm$0.1080 & 0.3620$\pm$0.0096 & 0.3746$\pm$0.0577 & \textbf{0.3573$\pm$0.0198}        \\ \hline \noalign{\vskip 1mm} 
\textbf{Boston}         &                   &                   &                   &                   &                                   \\
MLP \tnote{1}           & \textbf{0.1304$\pm$0.0057} & 0.2474$\pm$0.0526 & 0.1349$\pm$0.0074 & 0.1341$\pm$0.0082 & 0.1314$\pm$0.0131        \\ \hline \noalign{\vskip 1mm} 
\textbf{California}     &                   &                   &                   &                   &                                   \\
MLP \tnote{1}           & 0.2326$\pm$0.0019 & 0.2438$\pm$0.0189 & 0.1794$\pm$0.0156 & 0.2071$\pm$0.0609 & \textbf{0.1723$\pm$0.0048}        \\ \hline \noalign{\vskip 1mm}
\end{tabular}

\begin{tablenotes}\centering
\tiny Network architecture references: \item[1] \citep{baydin2018hypergradient}
\end{tablenotes}
\end{threeparttable}
\label{table:meta-testing-results-regression}
\end{sidewaystable}

\subsection{Loss Function Transfer}
\label{sec:meta-transfer}

\begin{sidewaystable}
\centering
\captionsetup{justification=centering}
\caption{Loss function transfer results reporting the mean $\pm$ standard deviation final inference testing error rate across 5 independent executions of each algorithm on each task + model pair. The best performance and those within its standard deviation are bolded. Loss functions are meta-learned on CIFAR-10 with the respective model and then transferred to CIFAR-100 using that same model.}
\begin{threeparttable}

\begin{tabular}{lccccc}
\hline
\noalign{\vskip 1mm}
Task and Model          & Baseline          & ML$^3$            & TaylorGLO         & GP-LFL            & EvoMAL (Ours)                 \\ \hline \noalign{\vskip 1mm}
\textbf{CIFAR-100}      &                   &                   &                   &                   &                               \\
AlexNet \tnote{1}       & \textbf{0.5262$\pm$0.0094} & 0.7735$\pm$0.0295 & 0.5543$\pm$0.0138 & 0.5329$\pm$0.0037 & 0.5324$\pm$0.0031             \\
VGG-16 \tnote{2}        & \textbf{0.3025$\pm$0.0022} & 0.3171$\pm$0.0019 & 0.3155$\pm$0.0021 & 0.3171$\pm$0.0041 & 0.3115$\pm$0.0038             \\
AllCNN-C \tnote{3}      & 0.2830$\pm$0.0021 & 0.2817$\pm$0.0032 & 0.4191$\pm$0.0058 & 0.2849$\pm$0.0012 & \textbf{0.2807$\pm$0.0028}             \\
ResNet-18 \tnote{4}     & 0.2474$\pm$0.0018 & 0.6000$\pm$0.0173 & 0.2436$\pm$0.0032 & 0.2373$\pm$0.0013 & \textbf{0.2326$\pm$0.0014}             \\
PreResNet \tnote{5}     & 0.2908$\pm$0.0065 & \textbf{0.2838$\pm$0.0019} & 0.2993$\pm$0.0030 & 0.2839$\pm$0.0025 & 0.2899$\pm$0.0024             \\
WideResNet \tnote{6}    & 0.2293$\pm$0.0017 & 0.2448$\pm$0.0063 & 0.2285$\pm$0.0031 & 0.2276$\pm$0.0028 & \textbf{0.2238$\pm$0.0017}             \\
SqueezeNet \tnote{7}    & 0.3178$\pm$0.0015 & 0.3402$\pm$0.0057 & 0.3178$\pm$0.0012 & 0.3343$\pm$0.0054 & \textbf{0.3166$\pm$0.0026}             \\ \noalign{\vskip 1mm} \hline \noalign{\vskip 1mm} 
\end{tabular}

\begin{tablenotes}\centering
\tiny Network architecture references: \item[1] \citep{krizhevsky2012imagenet} \item[2] \citep{simonyan2014very} \item[3] \citep{springenberg2014striving} \item[4] \citep{he2016deep} \item[5] \citep{he2016identity} \item[6] \citep{zagoruyko2016wide} \item[7] \citep{iandola2016squeezenet}
\end{tablenotes}
\end{threeparttable}
\label{table:meta-testing-transfer}
\end{sidewaystable}

To further validate the performance of EvoMAL, the much more challenging task of loss function transfer is assessed, where the meta-learned loss functions learned in one source domain are transferred to a new but related target domain. In our experiments, the meta-learned loss functions are taken directly from CIFAR-10 (in the previous section) and then transferred to CIFAR-100 with no further computational overhead, using the same model as the source. To ensure a fair comparison is made to the baseline, only the best-performing loss function found across the 5 random seeds from each method on each task + model pair are used. A summary of the final inference testing error rates when performing loss function transfer is given in Table \ref{table:meta-testing-transfer}.

The results show that even when using a single-task meta-learning setup where cross-task generalization is not explicitly optimized, meta-learned loss functions can still be transferred with some success. In regards to meta-generalization, it is observed that EvoMAL and GP-LFL transfer their relative performance the most consistently to new tasks. In contrast, ML$^3$, which uses a neural network-based representation, often fails to generalize to the new tasks, a finding similar to \citep{hospedales2020meta} which found that symbolic representations often generalize better than sub-symbolic representations. Another notable result is that on both AlexNet and VGG-16 in the direct meta-learning setup, large gains in performance compared to the baseline are observed, as shown in Table \ref{table:meta-testing-results-classification}. Conversely, when the learned loss functions are transferred to CIFAR-100, all the loss function learning methods perform worse than the baseline as shown in Table \ref{table:meta-testing-transfer}. These results suggest that the learned loss functions have likely been meta-overfitted to the source task and that meta-regularization is an important aspect to consider for loss function transfer.

\subsection{Meta-Training Performance}
\label{sec:meta-training-curves}

\begin{figure*}

    \centering
    \begin{subfigure}{0.33\textwidth}
        \centering
        \includegraphics[width=1\textwidth, height=5.5cm]{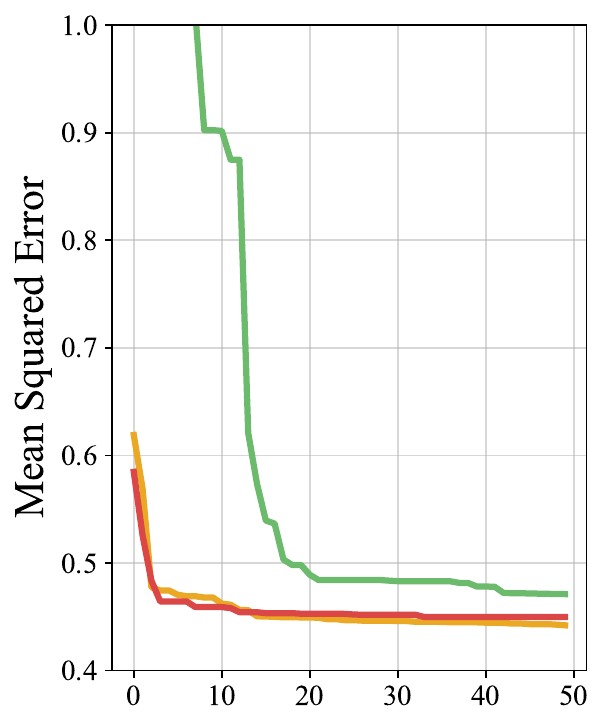}
        \caption{Diabetes MLP}
    \end{subfigure}%
    \hfill
    \begin{subfigure}{0.33\textwidth}
        \centering
        \includegraphics[width=1\textwidth, height=5.5cm]{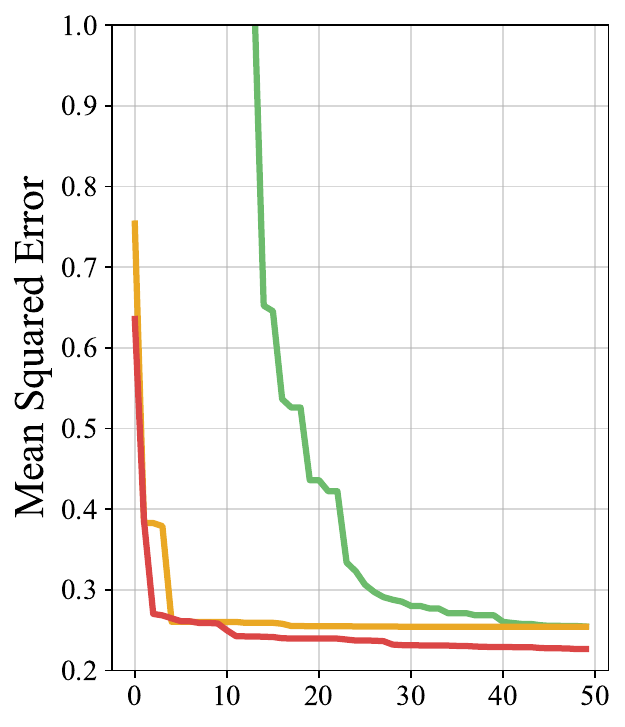}
        \caption{Boston MLP}
    \end{subfigure}%
    \hfill
    \begin{subfigure}{0.33\textwidth}
        \centering
        \includegraphics[width=1\textwidth, height=5.5cm]{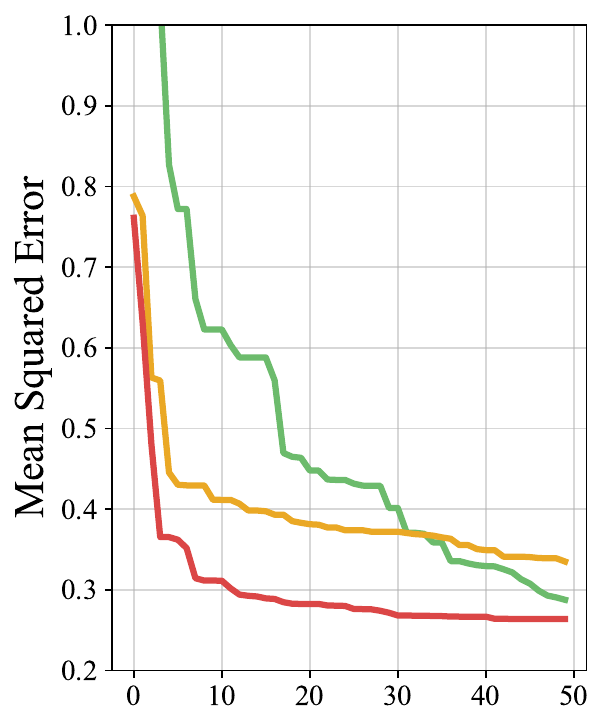}
        \caption{California MLP}
    \end{subfigure}%
    
    \vspace{3mm}
    
    \begin{subfigure}{0.33\textwidth}
        \centering
        \includegraphics[width=1\textwidth, height=5.5cm]{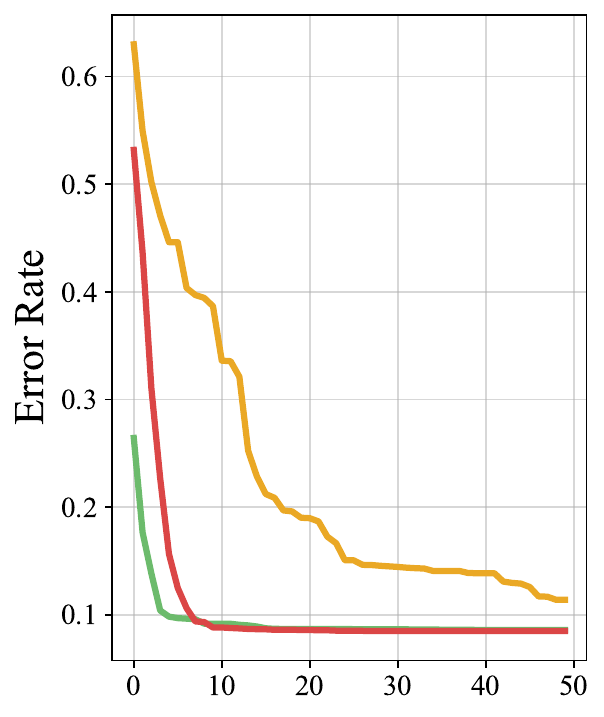}
        \caption{MNIST Logistic}
    \end{subfigure}%
    \hfill
    \begin{subfigure}{0.33\textwidth}
        \centering
        \includegraphics[width=1\textwidth, height=5.5cm]{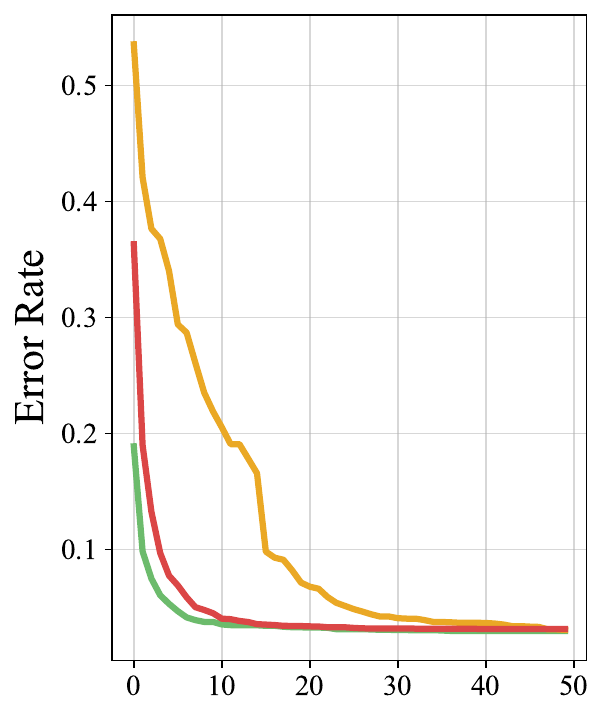}
        \caption{MNIST MLP}
    \end{subfigure}
    \hfill
    \begin{subfigure}{0.33\textwidth}
        \centering
        \includegraphics[width=1\textwidth, height=5.5cm]{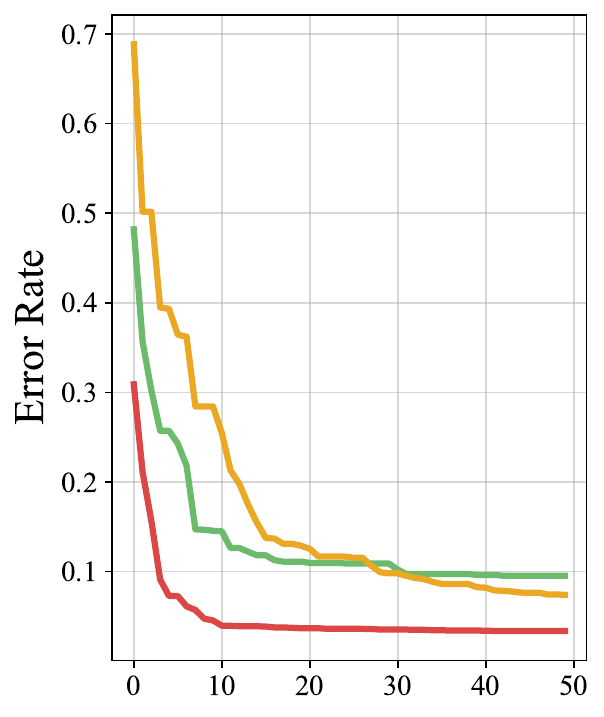}
        \caption{MNIST LeNet-5}
    \end{subfigure}%

    \vspace{3mm}
    \begin{subfigure}{0.33\textwidth}
        \centering
        \includegraphics[width=1\textwidth, height=5.5cm]{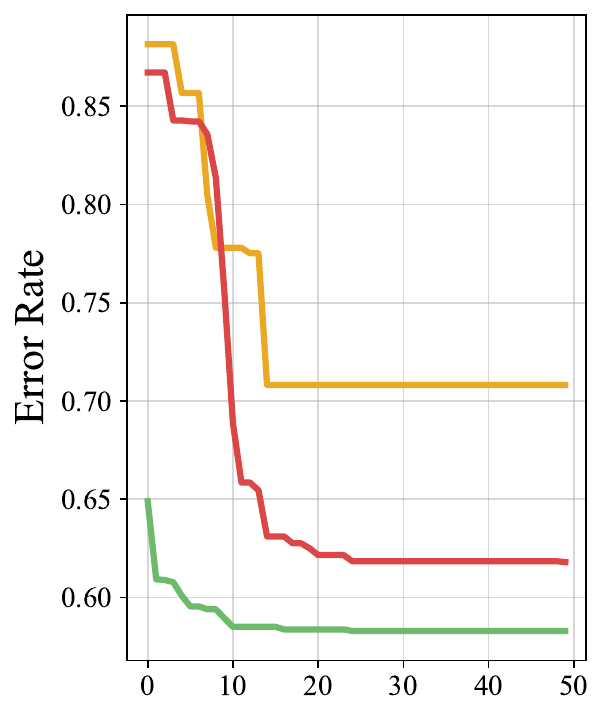}
        \caption{CIFAR-10 AlexNet}
    \end{subfigure}%
    \hfill
    \begin{subfigure}{0.33\textwidth}
        \centering
        \includegraphics[width=1\textwidth, height=5.5cm]{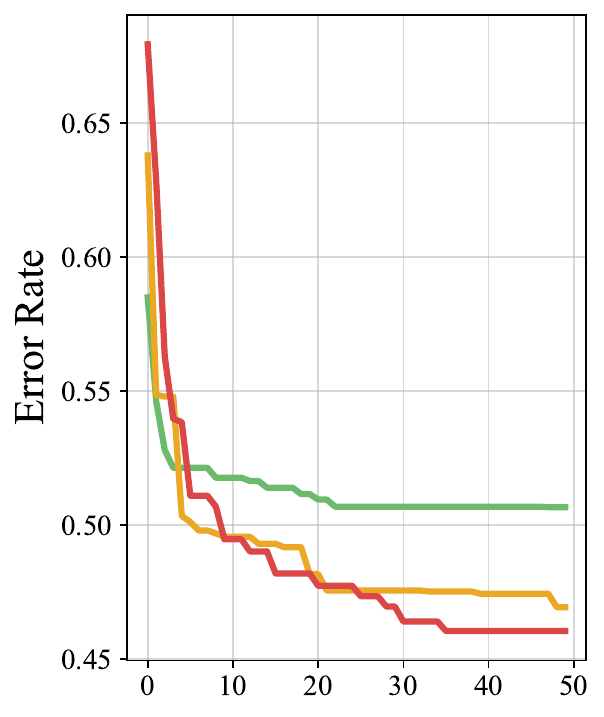}
        \caption{CIFAR-10 VGG-16}
    \end{subfigure}%
    \hfill
    \begin{subfigure}{0.33\textwidth}
        \centering
        \includegraphics[width=1\textwidth, height=5.5cm]{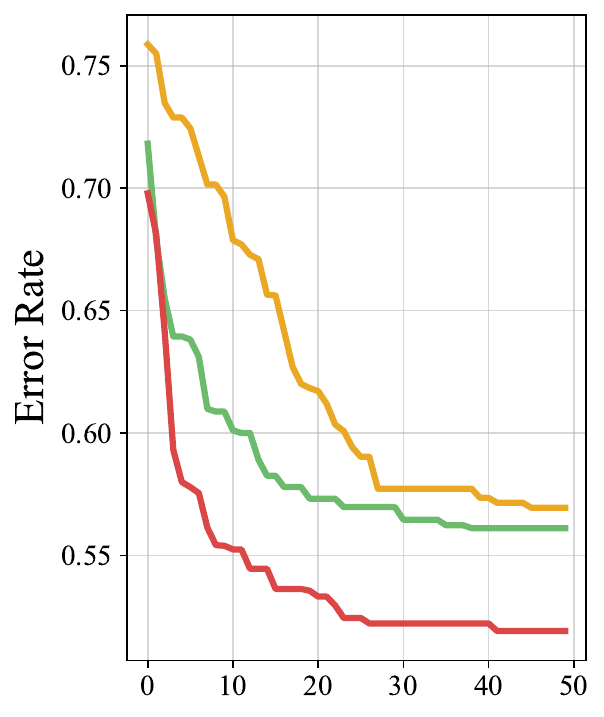}
        \caption{CIFAR-10 AllCNN-C}
    \end{subfigure}%
    \hfill
    \begin{subfigure}{\textwidth}
    \vspace{1mm}
        \centering
        \includegraphics[width=0.6\textwidth]{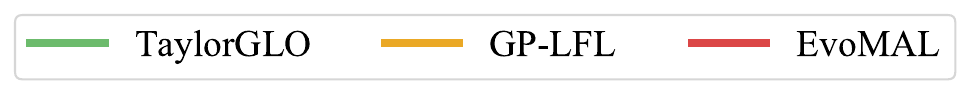}
    \end{subfigure}

    \captionsetup{justification=centering}
    \caption{Mean \textit{meta-training} learning curves across 5 independent executions of each algorithm, showing the fitness score (y-axis) against generations (x-axis), where a generation is equivalent to 25 evaluations.}
    \label{fig:meta-training-learning-curves-1}%
    
\end{figure*}

\begin{figure*}

    \ContinuedFloat

    \centering
    \captionsetup[subfigure]{justification=centering}
 
    \begin{subfigure}{0.33\textwidth}
        \centering
        \includegraphics[width=1\textwidth, height=5.5cm]{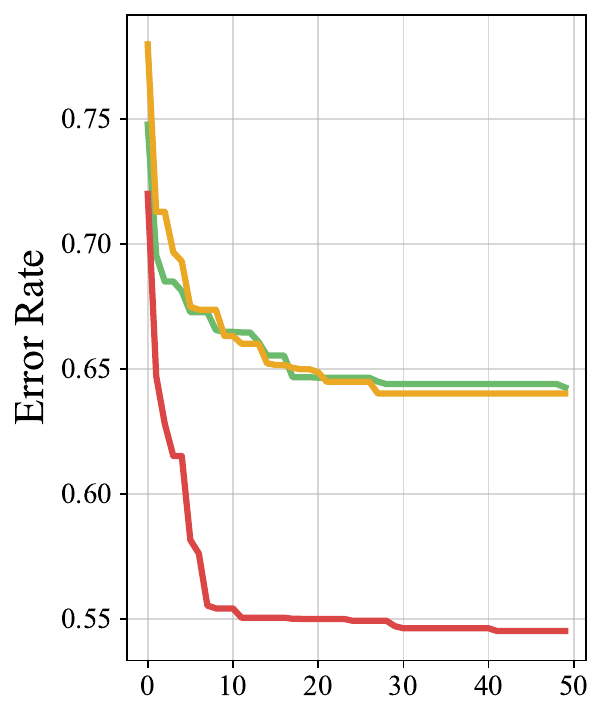}
        \caption{CIFAR-10 ResNet-18}
    \end{subfigure}
    \hfill
    \begin{subfigure}{0.33\textwidth}
        \centering
        \includegraphics[width=1\textwidth, height=5.5cm]{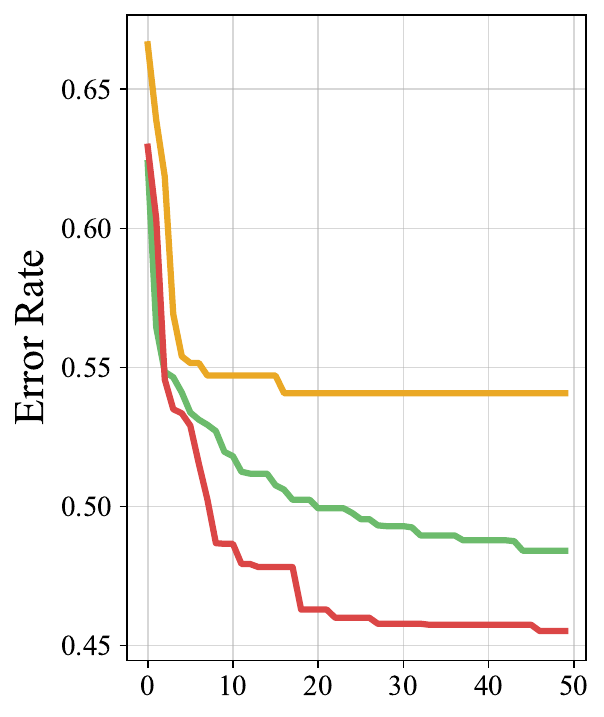}
        \caption{CIFAR-10 WRN 28-10}
    \end{subfigure}%
    \hfill
    \begin{subfigure}{0.33\textwidth}
        \centering
        \includegraphics[width=1\textwidth, height=5.5cm]{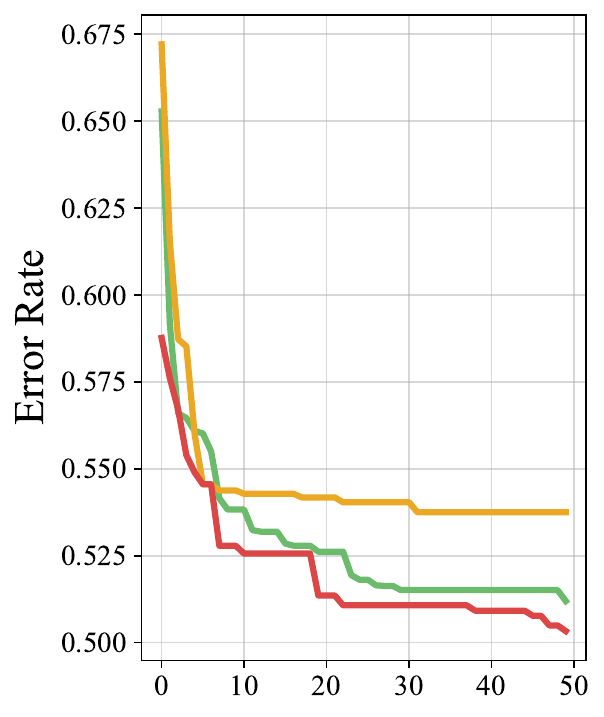}
        \caption{CIFAR-10 SqueezeNet}
    \end{subfigure}%
    \hfill

    \vspace{3mm}
    
    \begin{subfigure}{0.33\textwidth}
        \centering
        \includegraphics[width=1\textwidth, height=5.5cm]{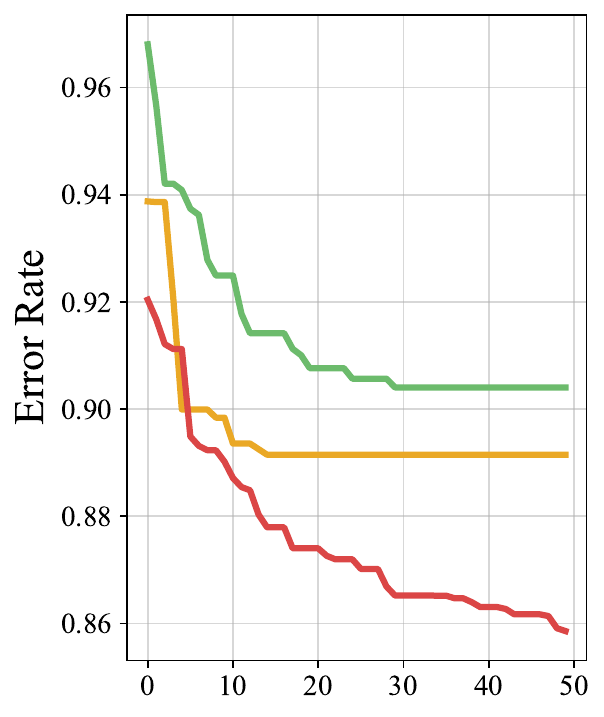}
        \caption{CIFAR-100 WRN 28-10}
    \end{subfigure}%
    \hfill
    \begin{subfigure}{0.33\textwidth}
        \centering
        \includegraphics[width=1\textwidth, height=5.5cm]{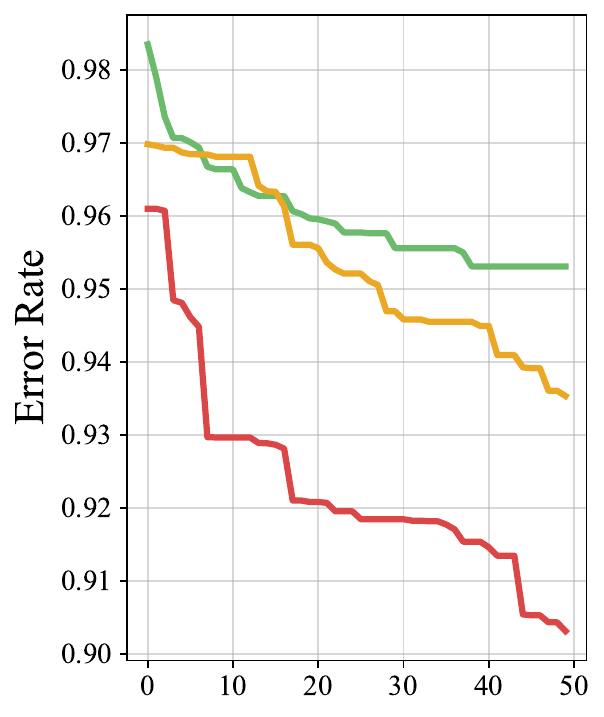}
        \caption{CIFAR-100 PyramidNet}
    \end{subfigure}%
    \hfill
    \begin{subfigure}{0.33\textwidth}
        \centering
        \includegraphics[width=1\textwidth, height=5.5cm]{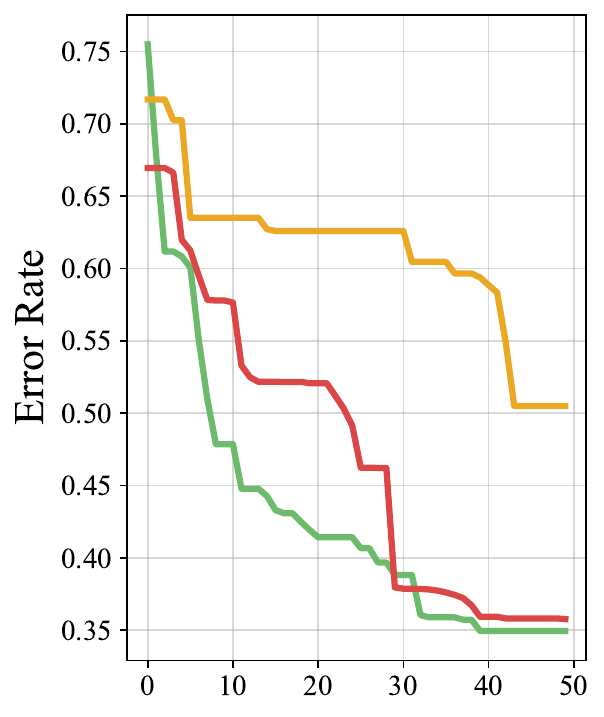}
        \caption{SVHN WRN 28-10}
    \end{subfigure}
    \begin{subfigure}{\textwidth}
    \vspace{1mm}
        \centering
        \includegraphics[width=0.6\textwidth]{chapter-3/figures/meta-training-legend.pdf}
    \end{subfigure}

    \captionsetup{justification=centering}
    \caption{Mean \textit{meta-training} learning curves across 5 independent executions of each algorithm, showing the fitness score (y-axis) against generations (x-axis), where a generation is equivalent to 25 evaluations.}
    \label{fig:meta-training-learning-curves-2}%

\end{figure*}

The meta-training learning curves are given in Figures \ref{fig:meta-training-learning-curves-1}, where the search performance of EvoMAL is compared to TaylorGLO and GP-LFL at each iteration and the performance is quantified by the fitness function using partial training sessions. Based on the results, it is very evident that adding local-search mechanisms into the EvoMAL framework dramatically increases the search effectiveness of GP-based loss function learning. EvoMAL is observed to consistently attain better performance than GP-LFL in a significantly shorter number of iterations. In almost all tasks, EvoMAL is shown to find better-performing loss functions in the first 10 generations compared to those found by GP-LFL after 50. 

Contrasting EvoMAL to TaylorGLO, it is generally shown again that for most tasks, EvoMAL produces better solutions in a smaller number of iterations. Furthermore, performance does not appear to prematurely converge on the more challenging tasks of CIFAR-100 and SVHN compared to TaylorGLO and GP-LFL. Interestingly, on both CIFAR-10 AlexNet and SVHN WideResNet, TaylorGLO can achieve slightly better final solutions compared to EvoMAL on average; however, as shown by the final inference testing error rates in Tables \ref{table:meta-testing-results-classification} and \ref{table:meta-testing-results-regression}, these do not necessarily correspond to better final inference performance. This discrepancy between meta-training curves and final inference is also observed in the inverse case, where EvoMAL is shown to have much better learning curves than both TaylorGLO and GP-LFL, \textit{e.g.} in CIFAR-10 AllCNN-C and ResNet-18. However, the final inference error rates of EvoMAL in Table \ref{table:meta-testing-results-classification} are only marginally better than those of TaylorGLO and GP-LFL. 

This phenomenon is likely due to a subset of the meta-learned loss functions implicitly tuning and optimizing for the base learning rate (discussed and analyzed further in the following chapter). Implicit learning rate tuning as a consequence of loss function learning can result in increased convergence capabilities, \textit{i.e.}, faster initial learning which results in better fitness when using partial training sessions, but does not necessarily imply a strongly generalizing and robustly trained model at meta-testing time when using full-length training sessions.

\subsection{Runtime Analysis}
\label{sec:runtime-analysis}

\begin{table}
\centering
\captionsetup{justification=centering}
\caption{Average runtime of the \textit{meta-training} process for each of the benchmark methods. Each algorithm is run on a single Nvidia RTX A5000, and the results are reported in hours.}
\begin{tabular}{p{4.5cm}cccc}
\hline
\noalign{\vskip 1mm}
Task and Model      & ML$^3$        & TaylorGLO     & GP-LFL        & EvoMAL                \\ \hline \noalign{\vskip 1mm}
\textbf{Diabetes}   &               &               &               &                       \\
MLP                 & 0.01          & 0.83          & 0.53          & 1.93                  \\ \noalign{\vskip 1mm} \hline \noalign{\vskip 1mm} 
\textbf{Boston}     &               &               &               &                       \\
MLP                 & 0.01          & 0.85          & 0.46          & 1.59                  \\ \noalign{\vskip 1mm} \hline \noalign{\vskip 1mm} 
\textbf{California} &               &               &               &                       \\
MLP                 & 0.01          & 0.94          & 0.84          & 2.06                  \\ \noalign{\vskip 1mm} \hline \noalign{\vskip 1mm} 
\textbf{MNIST}      &               &               &               &                       \\
Logistic            & 0.02          & 1.44          & 0.61          & 2.45                  \\
MLP                 & 0.03          & 2.06          & 0.77          & 5.31                  \\
LeNet-5             & 0.03          & 2.30          & 0.82          & 3.29                  \\ \noalign{\vskip 1mm} \hline \noalign{\vskip 1mm} 
\textbf{CIFAR-10}   &               &               &               &                       \\
AlexNet             & 0.03          & 3.75          & 1.04          & 5.90                  \\
VGG-16              & 0.12          & 4.67          & 0.89          & 9.12                  \\
AllCNN-C            & 0.12          & 4.55          & 0.90          & 8.77                  \\
ResNet-18           & 0.60          & 9.14          & 1.02          & 54.22                 \\
PreResNet           & 0.41          & 8.40          & 0.96          & 41.73                 \\
WideResNet          & 0.63          & 12.85         & 0.98          & 66.72                 \\
SqueezeNet          & 0.12          & 4.95          & 0.41          & 11.18                 \\\noalign{\vskip 1mm} \hline \noalign{\vskip 1mm} 
\textbf{CIFAR-100}  &               &               &               &                       \\
WideResNet          & 0.20          & 20.34         & 1.34          & 57.61                 \\
PyramidNet          & 0.20          & 24.83         & 1.32          & 49.89                 \\\noalign{\vskip 1mm} \hline \noalign{\vskip 1mm} 
\textbf{SVHN}       &               &               &               &                       \\
WideResNet          & 0.20          & 41.57         & 1.09          & 67.28                 \\\noalign{\vskip 1mm} \hline
\end{tabular}
\label{table:runtime}
\end{table}

The use of a two-stage discovery process by EvoMAL enables the development of highly effective loss functions, as shown by the meta-training and meta-testing results. Producing on average models with superior inference performance compared to TaylorGLO and ML$^3$ which only optimize the coefficients/weights of fixed parametric structures, and GP-LFL which uses no local-search techniques. However, this bi-level optimization procedure where both the model structure and parameters are inferred adversely affects the computational efficiency of the meta-learning process, as shown in Table \ref{table:runtime}, which reports the average runtime (in hours) of meta-training for each loss function learning method.

The results show that EvoMAL is more computationally expensive than ML$^3$ and GP-LFL and approximately twice as expensive as TaylorGLO on average. Although EvoMAL is computationally expensive, it should be emphasized that this is still dramatically more efficient than GLO \citep{gonzalez2020improved}, the bi-level predecessor to TaylorGLO, whose costly meta-learning procedure required a supercomputer for even very simple datasets such as MNIST. 

The bi-level optimization process of EvoMAL is made computationally tractable by replacing the costly CMA-ES loss optimization stage from GLO with a significantly more efficient gradient-based procedure. On CIFAR-10, using the relatively small network, PreResNet-20 GLO required 11,120 partial training sessions and approximately 171 GPU days of computation \citep{gonzalez2020improved, gonzalez2021optimizing} compared to EvoMAL, which only needed on average 1.7 GPU days. In addition, the reduced runtimes of EvoMAL can also be partially attributed to the application of time-saving filters, which enables a subset of the loss optimizations and fitness evaluations to be either cached or obviated entirely. 

\begin{figure}[t]
\centering

    \includegraphics[width=11cm]{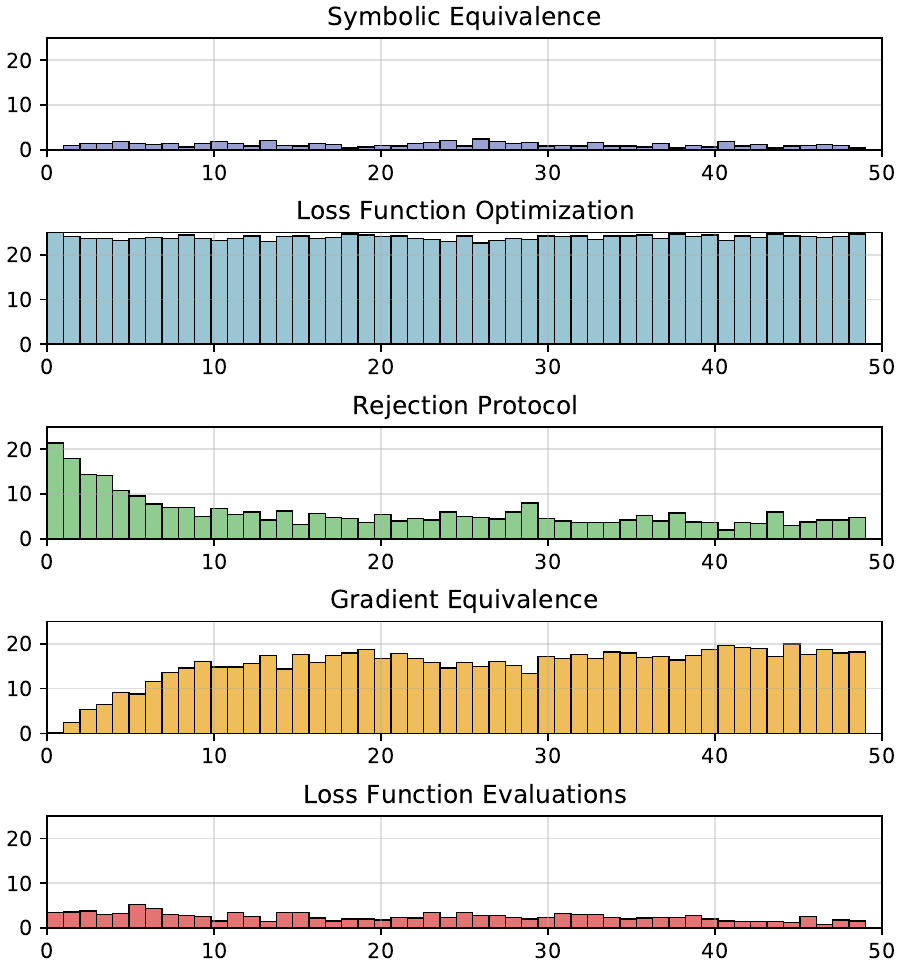}
    \captionsetup{justification=centering}
    \caption{Frequencies of occurrence of the time-saving filters, and the corresponding frequencies of loss optimization and evaluation throughout the symbolic search process. Reporting the average frequencies across all task + model pairs and independent executions of EvoMAL.}
    
\label{fig:meta-training-filters}
\end{figure}

To summarize the effects of the time-saving filters, a set of histograms is presented in Figure \ref{fig:meta-training-filters} showing the frequency of occurrence throughout the evolutionary process. Examining the symbolic equivalence filter results, it is observed that, on average, $\sim$10\% of the loss functions are identified as being symbolically equivalent at each generation; consequently, these loss functions have their fitness cached, and $\sim$90\% of the loss functions progress to the next stage and are optimized. Regarding the pre-evaluation filters, the rejection protocol initially rejects the majority of the optimized loss functions early in the search for being unpromising, automatically assigning them the worst-case fitness. In contrast, in the late stages of the symbolic search, this filter occurs incrementally less frequently, suggesting that convergence is being approached, further supported by the frequency of occurrence of the gradient equivalence filter, which caches few loss functions at the start of the search, but many near the end. Due to this aggressive filtering, only $\sim$25\% of the population at each generation have their fitness evaluated, which helps to reduce the runtime further. Note that the runtime of EvoMAL can be further reduced for large-scale optimization problems through parallelization to distribute loss function optimization and evaluation across multiple GPUs or clusters. 

\subsection{Search Space Analysis}
\label{sec:search-space-analysis}

\begin{figure}
\centering

    \includegraphics[width=0.8\columnwidth]{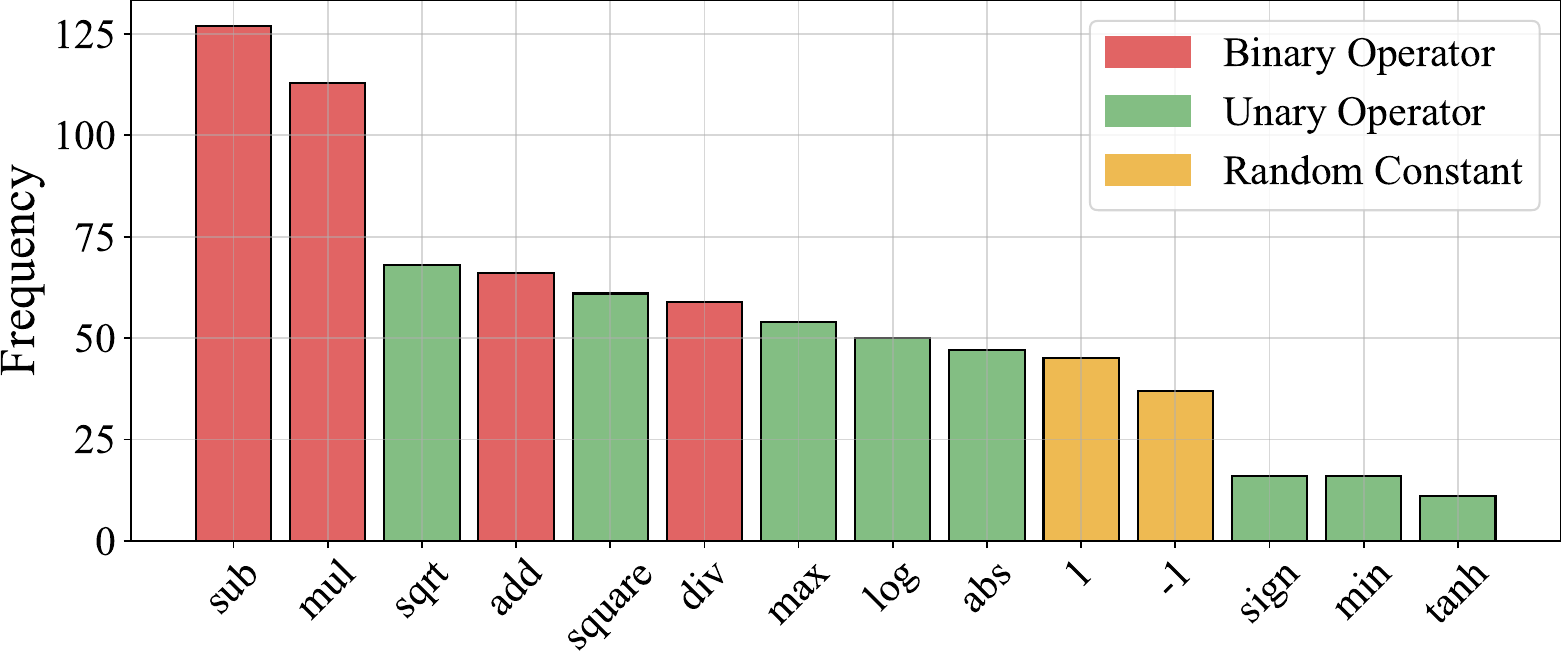}
    \captionsetup{justification=centering}
    \caption{Frequencies of the primitive mathematical operations in the symbolic loss functions discovered by EvoMAL across all datasets and random seeds.}
    
\label{figure:primitive-frequencies}
\end{figure}

To better understand the effects of the search space design on the learned loss functions developed by EvoMAL, we analyze the best-performing loss functions developed by EvoMAL. In particular, the frequencies of occurrence of the function nodes of the best-performing loss functions from all EvoMAL experiments are presented in Figure \ref{figure:primitive-frequencies}. It is observed that the most frequently recurring loss function sub-structure is the use of the subtraction or multiplication operator, with arguments $y$ and $f_{\theta}(x)$. This result is expected as these are commonly used building blocks in loss functions for computing the error in regression and classification, respectively. There is also high usage of the square root and square operators, which are commonly used for scaling the effects of large and small errors to either penalize or discount different groups of predictions (\textit{i.e.}, inliers and outliers). Finally, there is minimal usage of the $sign$, $min$, and $tanh$ operators, which empirically suggests that these function nodes are seldom important to the design of learned loss functions and that future studies may wish to consider removing them from the function set to increase the search efficiency.

\section{Chapter Summary}
\label{sec:conclusions}

This chapter presents a new framework for meta-learning symbolic loss function via a hybrid neuro-symbolic search approach called Evolved Model-Agnostic Loss (EvoMAL). The proposed technique uses genetic programming to learn a set of expression tree-based loss functions, which are subsequently transformed into gradient-trainable loss networks using a new novel transition procedure. The new representation enables the use of unrolled differentiation --- a fast and efficient gradient-based meta-optimization technique for enhancing the search capabilities of EvoMAL significantly. Unlike previous approaches, which stack evolution-based techniques, EvoMAL's computationally efficient local-search enables loss function learning on commodity hardware. 

The experimental results on classification and regression tasks confirm that EvoMAL consistently meta-learns loss functions that produce more performant models compared to those trained with conventional handcrafted loss functions and state-of-the-art loss function learning techniques. Furthermore, we pioneer the investigation of loss function transfer, where loss functions are learned in one domain and then transferred to another. The results show that even in a single-task meta-learning setup, where cross-task generalization is not explicitly optimized, meta-learned loss functions can still exhibit some success when transferred between tasks.

Many promising future research directions come as a consequence of this research. In terms of loss function learning, natural extensions of the work would be to meta-learn the loss function and other deep neural network components simultaneously, similar to the methods presented in \citep{ravi2017optimization, li2017meta, elsken2020meta, baik2021meta}. For example, the meta-learned loss functions could incorporate additional arguments such as the time-step or model weights, which would implicitly induce learning rate scheduling or weight regularization. Another example would be to combine neural architecture search with loss function learning, given that the current experiments in this chapter have used handcrafted neural network architectures that are biased towards the squared error loss or cross-entropy loss since they were designed to optimize for that specific loss function. Larger performance gains may be achieved using custom neural network architectures explicitly designed for meta-learned loss functions.

\chapter{Analysis of Meta-Learned Loss Functions}\label{chapter:theory}

\textit{In this chapter, empirical and theoretical analysis is performed on the meta-learned loss functions from the previous chapter, to answer the fundamental questions of “Why do meta-learned loss functions perform better than handcrafted loss functions”, and “What are meta-learned loss functions learning”. Notably, we demonstrate empirically that previous hypotheses for why meta-learned loss functions are so performant do not sufficiently explain the improvement in performance from using meta-learned loss functions. Instead, we show deep theoretical links with regularization techniques; namely, label smoothing regularization, as one of the causes. We utilize our theoretical findings to develop a new classification loss function called “sparse label smoothing regularization” which functions similarly to label smoothing regularization; but offers a notable improvement in computational efficiency, with the time and space complexity being reduced from linear to constant time with respect to the number of classes being considered.}

\section{Chapter Overview}

In the preceding chapter, Evolved Model-Agnostic Loss (EvoMAL) was developed for the task of loss function learning \citep{raymond2023learning, raymond2023fast}. This algorithmic contribution was a significant step forward in loss function learning as it demonstrated the first computationally feasible approach for meta-learning loss functions discrete symbolic structure followed by continuous optimization of the parameters. The use of a two-stage learning process is significant as both the discrete and continuous stages hold considerable importance. The discrete stage in EvoMAL uses genetic programming, which enables the development of interpretable and symbolic loss functions, while the continuous stage uses unrolled differentiation, which greatly improves the convergence and final performance. 

Without the continuous stage in EvoMAL, the loss function learning algorithm has poor search efficiency, and the learned loss functions have sub-optimal performance, as shown in Chapter \ref{chapter:evomal}. Alternatively, past research has explored omitting the discrete stage; however, this requires replacing the learned symbolic structure from genetic programming with parametric representations such as Taylor polynomials \citep{gonzalez2021optimizing,gao2021searching} or neural networks \citep{bechtle2021meta,psaros2022meta} which are very difficult or impossible to understand and analyze.

Thus far, the value of being able to understand and analyze the learned loss functions has not yet been demonstrated or proven. Therefore, in this chapter, we aim to do just that by performing a theoretical analysis of some of the learned loss functions developed by EvoMAL. In this analysis, we aim to elucidate: (1) why meta-learned loss functions perform better than handcrafted loss functions, and (2) what are meta-learned loss functions learning. The results of this analysis bring us fascinating insights into what learned loss functions are learning, and specifically, their deep connections to label smoothing regularization, something which has previously been hinted in \citep{gonzalez2020improved} and \citep{gonzalez2020effective}.

Inspired by the findings of our theoretical analysis a new novel loss function named \textit{Sparse Label Smoothing Regularization} is developed. The newly proposed loss function performs similarly to the cross-entropy with label smoothing regularization \citep{szegedy2016rethinking, muller2019does}; however, it is significantly faster to compute and utilizes much less memory. 

\subsection{Contributions}

The key contributions of this chapter are as follows:

\begin{itemize}

    \item We visualize and analyze some of the meta-learned loss functions developed by EvoMAL from the previous chapter, identifying key underlying trends and highlighting novel features of the learned loss functions.

    \item We critically examine the existing hypotheses for how and why meta-learned loss functions can improve generalization performance, and demonstrate empirically that they do not sufficiently explain the improvement in performance over handcrafted loss functions.

    \item We perform a comprehensive theoretical analysis of one of the learned loss functions developed by EvoMAL and identify its behavioral similarities with the cross-entropy loss and label smoothing regularization at two crucial point in the learning process.

    \item We develop a new classification loss function called Sparse Label Smoothing Regularization (SparseLSR) Loss, which has constant time and space complexity with respect to the number of classes when evaluating relative to traditional label smoothing regularization which has linear time and space complexity.
    
\end{itemize}

\section{Visualizing Meta-Learned Loss Functions}
\label{sec:loss-functions}

\begin{figure*}[t!]

    \centering
    \begin{subfigure}{1\textwidth}
      \centering
      \includegraphics[width=11cm]{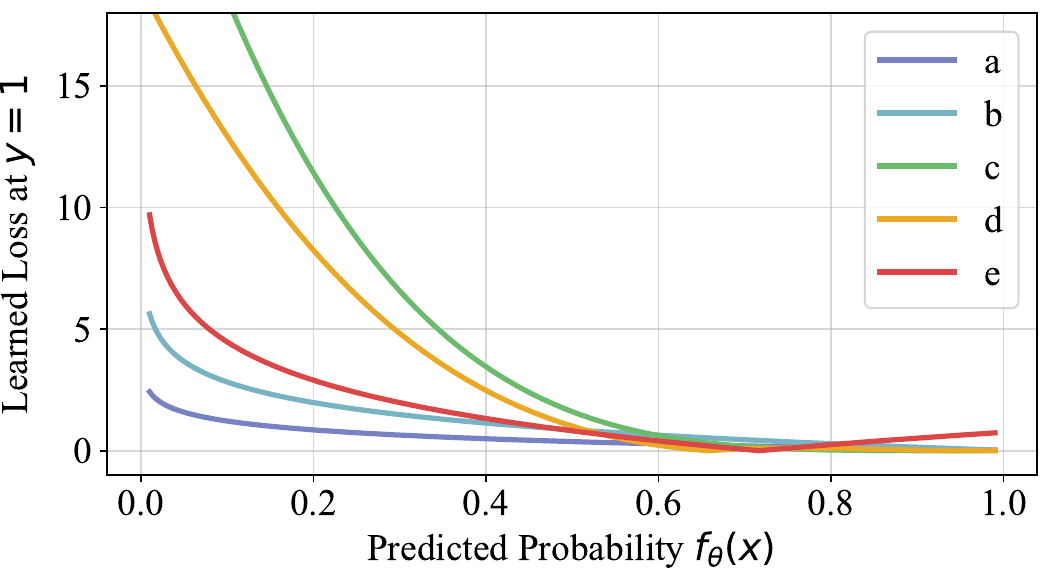}
    \end{subfigure}%
    
    \vspace{3mm}
    
    \begin{subfigure}{1\textwidth}
      \centering
      \includegraphics[width=11cm]{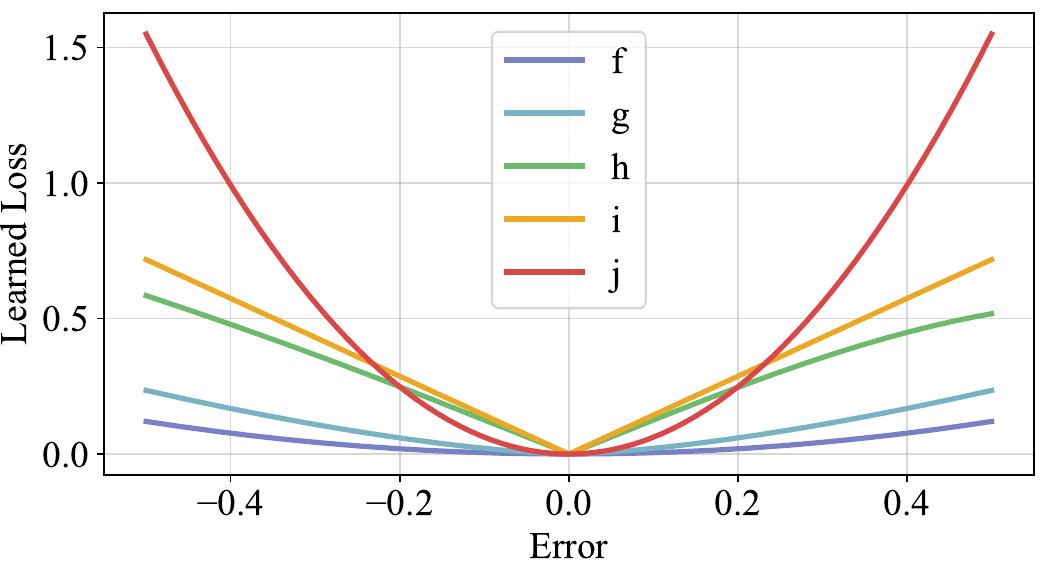}
    \end{subfigure}

    \captionsetup{justification=centering}
    \caption{Example loss functions meta-learned by EvoMAL. The top plot shows classification loss functions and the bottom plot shows regression loss functions.}
    \label{fig:meta-learned-loss-functions}

\end{figure*}

\begin{table}[t!]

\centering
\captionsetup{justification=centering}
\caption{Example loss functions meta-learned by EvoMAL, where solutions have been numerically and algebraically simplified. Furthermore, the parameters $\phi$ have also been omitted for improved clarity and parsimony.}

\begin{tabular}{l>{\centering\arraybackslash}p{0.85\columnwidth}}
\multicolumn{1}{c}{\textbf{}} &  \\ \hline \noalign{\vskip 1mm}

a. & $|\log(\sqrt{y \cdot f_{\theta}(x)} + \epsilon)|$ \\ 

b. & $\log(y \cdot f_{\theta}(x)) + \sqrt{\log(y \cdot f_{\theta}(x))}^{2}$ \\ \noalign{\vskip 1mm}

c. & $((y - f_{\theta}(x))/f_{\theta}(x))^4$ \\ \noalign{\vskip 1mm} 

d. & $y^3 + f_{\theta}(x)^3 + f_{\theta}(x) \cdot y^2 + f_{\theta}(x)^2 \cdot y$ \\ \noalign{\vskip 1mm}

e. & $|\log((y \cdot f_{\theta}(x))^2 + \epsilon)|$ \\ \noalign{\vskip 1mm} 

\hline \noalign{\vskip 1mm} \hline \noalign{\vskip 1mm}

f. & $(f_{\theta}(x) - \min(\max(y, -1), 1))^2$ \\ \noalign{\vskip 1mm}

g. & $|y \cdot (\sqrt{|y/\sqrt{1 + f_{\theta}(x)^2}|})|$ \\ \noalign{\vskip 1mm}

h. &  $|(y/(1 + \log(|y - 1|)^2)) - f_{\theta}(x)|$ \\ \noalign{\vskip 1mm}

i. & $\sqrt{|y \cdot (y - f_{\theta}(x))|}$ \\ \noalign{\vskip 1mm}

j. & $\sqrt{|y - f_{\theta}(x)|}$ \\ \noalign{\vskip 1mm} \hline

\end{tabular}

\label{table:meta-learned-loss-functions}
\end{table}

To gain some insight into why the meta-learned loss functions developed by EvoMAL are so performant, we begin our analysis by examining a subset of the unique and interesting loss functions found throughout our experiments in Chapter \ref{chapter:evomal}. In Figure \ref{fig:meta-learned-loss-functions}, examples of the meta-learned loss functions produced by EvoMAL are presented. The corresponding loss functions are also given symbolically in Table \ref{table:meta-learned-loss-functions}.

\subsection{Classification Loss Functions}

In regard to the classification loss functions meta-learned by EvoMAL, they appear to converge upon three classes of loss functions. First are cross-entropy loss variants such as loss functions a) and b), which closely resemble the cross-entropy loss functionally and symbolically. Second are loss functions that have similar characteristics to the parametric focal loss \citep{lin2017focal}, such as loss functions c) and d). These loss functions recalibrate how easy and hard samples are prioritized; in most cases, very little or no loss is attributed to high-confidence correct predictions, while significant loss is attributed to high-confidence wrong predictions. Finally, loss functions such as e) demonstrate unintuitive behavior, such as assigning more loss to confident and correct solutions relative to unconfident and correct solutions.

\subsection{Regression Loss Functions}

Analyzing the regression loss functions, several unique behaviors are observed, particularly around incorporating strategies for improving robustness to outliers. For example in f) the loss function takes on the form of the squared loss; however, it incorporates a thresholding operation via the $\max$ operator for directly limiting the size of the ground truth label. Alternatively, in loss functions g), i), and j), the square root operator is frequently used, which decreases the loss attributed to increasingly large errors. Finally, we observe that many of the learned loss functions, take on shapes comparable to well-known handcrafted loss functions; for example, h) appears to be a hybrid between the absolute loss and the Cauchy (Lorentzian) loss \citep{black1996unification}.

\subsection{General Observations about Learned Loss Functions}

In addition to the trends identified for classification and regression, respectively, there are several more noteworthy trends:

\begin{itemize}

    \item Structurally complex loss functions typically perform worse than simpler loss functions, potentially due to the increased meta or base optimization difficulties. This is an equivalent finding to what was empirically found when meta-learning activation functions \citep{ramachandran2017searching}.  
  
    \item Many of the learned loss functions in classification are asymmetric, producing different loss values for false positive and false negative predictions, often caused by exploiting $f_{\theta}(x)$ softmax output activation, where the summation of the base models outputs is required to equal to $1$. 

    \item In classification, the domain of the target y and model predictions $f_{\theta}(x)$ is known, \textit{i.e.}, $y \in \{0, 1\}$ and $f_{\theta}(x) \in (0, 1)$. This knowledge can be utilized to further simplify the loss functions. This is very useful as it enables equivalence relations to be drawn more easily between loss functions that may appear different but are the same.

    \item In regression, the most frequently recurring loss function sub-structure is the subtraction operator between $y$ and $f_{\theta}(x)$. In contrast, in classification, the most common sub-structure was the multiplication operator between $y$ and $f_{\theta}(x)$. This suggests that there are common task-dependent patterns in the design of learned loss functions, and these can be utilized to potentially improve search efficiency. 

\end{itemize}

\section{Examination of Existing Hypotheses}

Loss function learning as a paradigm has consistently shown to be an effective way of improving performance \citep{gonzalez2020improved,gonzalez2021optimizing,bechtle2021meta,raymond2023learning}; however, it is not yet fully understood what meta-learned loss functions are actually learning and why they are so performant compared to their handcrafted counterparts. In this section, we critically examine two of the existing hypotheses for why meta-learned loss functions are so performant.

\subsection{Hypothesis 1: Flatter Loss Landscapes}
\label{sec:loss-landscapes}

\begin{figure*}[t!]

    \centering
    \begin{subfigure}{1\textwidth}
      \centering
      \includegraphics[width=11cm]{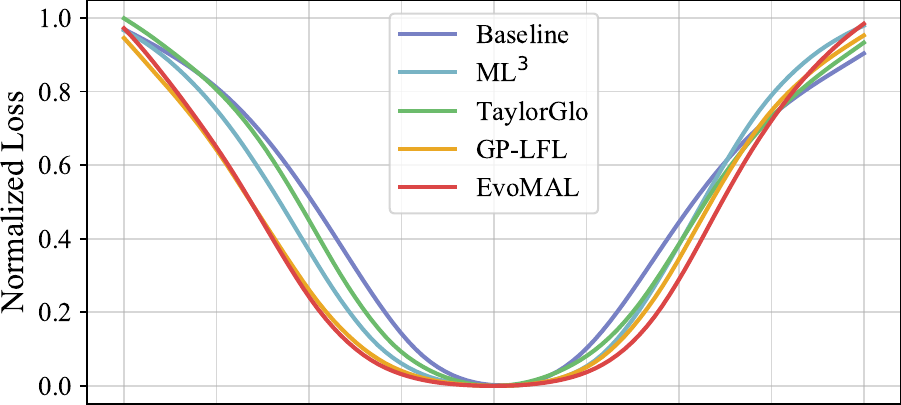}
    \end{subfigure}%
    
    \vspace{5mm}
    
    \begin{subfigure}{1\textwidth}
      \centering
      \includegraphics[width=11cm]{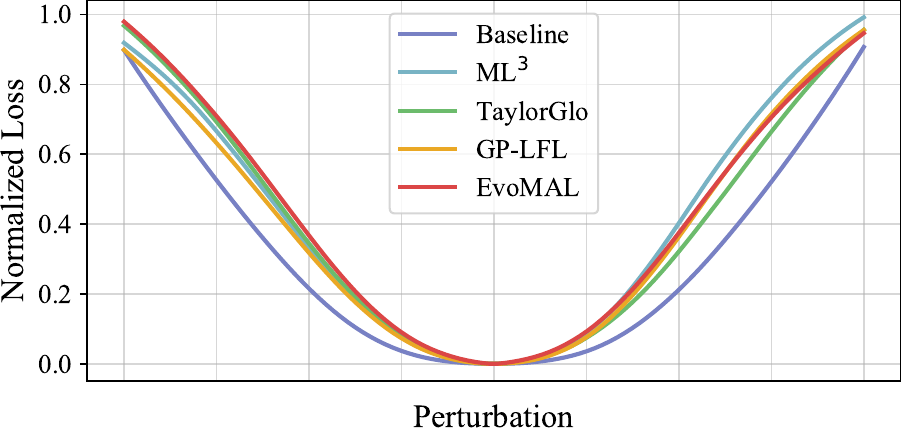}
    \end{subfigure}

    \captionsetup{justification=centering}
    \caption{The average 1D loss landscapes generated on the CIFAR-10 dataset, the top figure shows the loss landscapes generated on AllCNN-C, and the bottom figure shows the loss landscapes generated on AlexNet. The landscapes show the average (mean) loss taken across the 5 independent executions of each algorithm on each task+model pair.}
    \label{fig:loss-landscapes}%

\end{figure*}

In \cite{gonzalez2021optimizing} and \cite{gao2022loss}, it was found that the loss landscapes of models trained with learned loss functions produce flatter landscapes relative to those trained with the cross-entropy loss. The flatness of a loss landscape has been hypothesized to correspond closely to a model's generalization capabilities \citep{hochreiter1997flat, keskar2016large, li2018visualizing, chaudhari2019entropy}; thus, they conclude that meta-learned loss functions improve generalization. These findings are independently reproduced and are shown in Figure \ref{fig:loss-landscapes}. The loss landscapes are generated using the filter-wise normalization method \citep{li2018visualizing}, which plots a normalized random direction of the weight space $\theta$. 

The loss landscapes visualizations show that the loss functions developed by EvoMAL can produce flatter loss landscapes on average in contrast to those produced by the cross-entropy, ML$^3$, TaylorGLO, and GP-LFL, as shown in the top figure which shows the landscapes generated using AllCNN-C. In contrast to prior findings, we find that in some cases the meta-learned loss functions in our experiments show relatively sharper loss landscapes compared to those produced using the cross-entropy loss, as shown in the bottom figure in Figure \ref{fig:loss-landscapes}. These findings suggest that the relative flatness of the loss landscape does not fully explain why meta-learned loss functions can produce improved performance, especially since there is evidence that sharp minima can generalize well \textit{i.e.}, “\textit{flat vs sharp}” debate \citep{dinh2017sharp}.

\clearpage

\subsection{Hypothesis 2: Implicit Learning Rate Tuning}
\label{sec:implicit-tuning}

\begin{figure}
\centering

    \includegraphics[width=11cm]{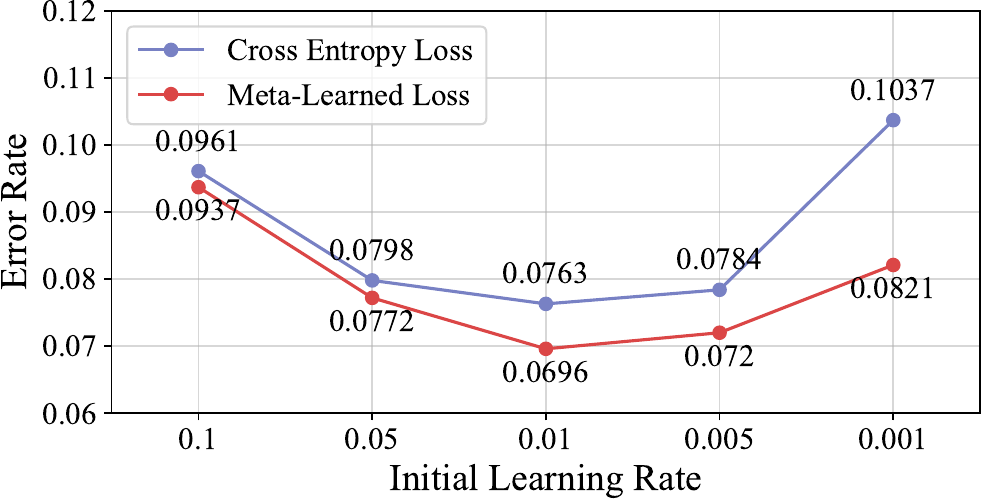}
    \captionsetup{justification=centering}
    \caption{Grid search comparing the average error rate of the baseline cross-entropy loss and EvoMAL on CIFAR-10 AllCNN-C across a set of base learning rate values, where 5 executions of each algorithm are performed on each learning rate value.}

\label{fig:learning-rate}
\end{figure}

Another explanation for why meta-learned loss functions improve performance over handcrafted loss functions is that they can implicitly tune the (base) learning rate for some suitably expressive representation of $\MetaLoss$, since $\exists\alpha\exists\phi:\theta - \alpha\nabla_{\theta}\Loss_{meta} \approx \theta - \nabla_{\theta}\MetaLoss_{\phi}^{\Transpose}$. Thus, performance improvement when using meta-learned loss functions may be the indirect result of a change in the learning rate, scaling the resulting gradient of the loss function. To validate the implicit learning rate hypothesis, a grid search is performed over the base-learning rate using the cross-entropy loss and EvoMAL on CIFAR-10 AllCNN-C. The results are shown in Figure \ref{fig:learning-rate}. 

The results show that the base learning rate $\alpha$ is a crucial hyper-parameter that influences the performance of both the baseline and EvoMAL. However, in the case of EvoMAL, it is found that when using a relatively small $\alpha$ value, the base learning rate is implicitly tuned, and the loss function learning algorithm achieves an artificially large performance margin compared to the baseline. Implicit learning rate tuning of a similar magnitude is also observed when using relatively large $\alpha$ values; however, the algorithm's stability is inconsistent, with some runs failing to converge. Finally, when a near-optimal $\alpha$ value is used, performance improvement is consistently better than the baseline. These results indicate two key findings: (1) meta-learned loss functions improve upon handcrafted loss functions and that the performance improvement when using meta-learned loss functions is not primarily a result of implicit learning rate tuning when $\alpha$ is tuned. (2) The base learning rate $\alpha$ can be considered as part of the initialization of the meta-learned loss function, as it determines the initial scale of the loss function.

\section{Theoretical Analysis}
\label{sec:theoretical-analysis}

To further investigate the questions of \textit{why do meta-learned loss functions perform better than handcrafted loss functions}, and \textit{what are meta-learned loss functions learning}, we perform a theoretical analysis inspired by the analysis performed in \citep{gonzalez2020effective}. The analysis reveals important connections between our learned loss functions and the cross-entropy loss and label smoothing regularization. 

\subsection{Learned Loss Function}\label{sec:deriving-learned}

In this theoretical analysis, we analyze the general form of loss functions a) and e) from Table \ref{table:meta-learned-loss-functions} in Section \ref{sec:loss-functions} (where $\epsilon=1e-7$ is a small constant):
\begin{equation}
\mathcal{M}^{a} = |\log(\sqrt{y \cdot f_{\theta}(x) + \epsilon})|
\end{equation}
\begin{equation}
\mathcal{M}^{e} = |\log((y \cdot f_{\theta}(x) + \epsilon)^2)|
\end{equation}
where $\mathcal{M}^{a}$ and $\mathcal{M}^{e}$ are equivalent upto a scaling factor $\phi_{0}$, since
\begin{equation}
|\log((y \cdot f_{\theta}(x) + \epsilon)^{\phi_{0}})| = \phi_{0} \cdot |\log(y \cdot f_{\theta}(x) + \epsilon )|.
\end{equation}
Therefore, the general form of the learned loss functions can be given as follows:
\begin{equation}
\mathcal{M}_{\phi} = |\log(y \cdot f_{\theta}(x) + \epsilon)|
\end{equation}
where, one further simplification can be made since the non-target learned loss (\textit{i.e.}, when $y=0$)  evaluates to $|\log(\epsilon)|$ which is not dependent on the base model predictions, and supplies a constant gradient. Consequently, the target label $y$ can be taken outside the $\log$ and $\epsilon$ can be dropped giving us the final learned loss function which we further refer to as the \textit{Absolute Cross-Entropy Loss} ($\Loss^{ACE}$) function:
\begin{equation}
\Loss^{ACE} = y \cdot |\log(f_{\theta}(x))|
\end{equation}
where the meta parameters $\phi_0$ and $\phi_1$ can be made explicit as they reveal key insights into why $\Loss^{ACE}$ can improve performance.
\begin{equation}\label{eq:learned-loss-function}
\Loss^{ACE} = \phi_{0} \cdot y \cdot |\log(\phi_{1} \cdot f_{\theta}(x))|
\end{equation}
Regarding $\phi_{0}$, it is straightforward to see that this corresponds to an explicit linear scaling of the loss function. This is identical in behavior to what the learning rate $\alpha$ is doing in stochastic gradient descent and makes explicit the relationship we established in Section \ref{sec:implicit-tuning}. As for $\phi_{1}$, as shown in the subsequent sections, this parameter enables $\Loss^{ACE}$ to regularize overconfident predictions in a similar fashion label smoothing regularization.

\subsection{Learning Rule Decomposition}\label{sec:decomposition}

First, we decompose our learning rule to isolate the contribution of the loss function. For simplicity, we will consider the simple case of using vanilla stochastic gradient descent (SGD). Repeating the learning rule shown in Equation \eqref{eq:gradient-decent-setup} from the previous chapter:
\begin{equation}
    \theta_{t+1} \leftarrow \theta_{t} - \alpha \nabla_{\theta_{t}}\Big[\Loss(y, f_{\theta_{t}}(x))\Big]
\end{equation}
where $\alpha$ is the (base) learning rate and $\Loss$ is the loss function, which takes as arguments the true label $y$ and the base model predictions $f_{\theta}(x)$. Consequently, the learning rule for the base model weights $\theta_{t}$ based on the  $i^{th}$ model output $f_{\theta}(x)_{i}$ and target $y_{i}$ can be described as follows:
\begin{equation}\label{eq:general-form}
\begin{split}
\theta_{t+1} 
& \leftarrow \theta_{t} - \alpha \frac{\partial}{\partial \theta_{t}} \Big[\Loss(y_{i}, f_{\theta_{t}}(x)_{i})\Big] \\
& = \theta_{t} - \alpha \bigg[\frac{\partial}{\partial f}\Loss(y_{i}, f_{\theta_{t}}(x)_{i}) \cdot \frac{\partial}{\partial \theta_{t}} f_{\theta}(x)_{i}\bigg] \\
& = \theta_{t} + \alpha \bigg[\underbrace{-\frac{\partial}{\partial f}\bigg(\Loss(y_{i}, f_{\theta_{t}}(x)_{i})\bigg)}_{\delta} \cdot \frac{\partial}{\partial \theta_{t}} f_{\theta_{t}}(x)_{i}\bigg]
\end{split}
\end{equation}
Given this general form, we can now substitute any loss function into Equation \eqref{eq:general-form} to give a unique expression $\delta$ which describes the behavior of the loss function with respect to the probabilistic output. Through substitution of the $\Loss^{ACE}$ into the SGD learning rule we get
\begin{equation}
\theta_{t+1} \leftarrow \theta_{t} + \alpha \bigg[\underbrace{-\frac{\partial}{\partial f} \bigg(y \cdot |\log(\phi_{1} \cdot f_{\theta}(x))| \bigg)}_{\delta_{\Loss^{ACE}}} \cdot \frac{\partial}{\partial \theta_{j}} f_{\theta}(x)_{i}\bigg],
\end{equation}
where the behavior $\delta_{\Loss^{ACE}}$ is defined as follows:
\begin{equation}
\begin{split}
\delta_{\Loss^{ACE}} 
&= - \frac{\partial}{\partial f} \bigg(y \cdot |\log(\phi_{1} \cdot f_{\theta}(x))|\bigg) \\
&= - y \cdot \frac{\log(\phi_{1} \cdot f_{\theta}(x))}{|\log(\phi_{1} \cdot f_{\theta}(x))|} \cdot \frac{\partial}{\partial f} \bigg(\log(\phi_{1} \cdot f_{\theta}(x))\bigg) \\
&= - y \cdot \frac{\log(\phi_{1} \cdot f_{\theta}(x))}{|\log(\phi_{1} \cdot f_{\theta}(x))|} \cdot \frac{1}{\phi_{1} f_{\theta}(x)} \cdot \frac{\partial}{\partial f} \bigg(\phi_{1} \cdot f_{\theta}(x)\bigg) \\
&= - y \cdot \frac{\log(\phi_{1} \cdot f_{\theta}(x))}{|\log(\phi_{1} \cdot f_{\theta}(x))|} \cdot \frac{1}{\phi_{1} f_{\theta}(x)} \cdot \phi_{1} \\
&= - \frac{\phi_{1} \cdot y \cdot \log(\phi_{1} \cdot f_{\theta}(x))}{\phi_{1} \cdot f_{\theta}(x) \cdot |\log(\phi_{1} \cdot f_{\theta}(x))|} \\
&= - \frac{y \cdot \log(\phi_{1} \cdot f_{\theta}(x))}{f_{\theta}(x) \cdot |\log(\phi_{1} \cdot f_{\theta}(x))|} \\
\end{split}
\end{equation}

\subsection{Dual Point Analysis}

Now that a framework for deriving the behavior of a loss function has been established, we can analyze $\Loss^{ACE}$ from two unique and significant points in the learning process: (1) initial learning behavior at the null epoch when learning begins under a random initialization, and (2) in the zero training error regime where a loss function's regularization behaviors can be observed when there is nothing new to learn from the training data.

\subsubsection{Behavior at the Null Epoch}\label{sec:null}

First, consider the case where the base model weights $\theta_{0}$ are randomly initialized (\textit{i.e.}, the null epoch before any learned has happened) such that the expected value of the $i^{th}$ output of the base model is given as
\begin{equation}
\forall k \in [1, \mathcal{C}], \text{where } \mathcal{C} > 2: \mathbb{E} [f_{\theta}(x)_{i}] = \mathcal{C}^{-1}
\end{equation}
where $\mathcal{C}$ is the number of classes. Note, that the output $f_{\theta}(x)_{i}$ represents the post-softmax output which converts the raw logits into probabilities. Hence, the predicted probabilities are uniformly distributed among the classes $\frac{1}{\mathcal{C}} = \mathcal{C}^{-1}$.

\subsubsection{Behavior at Zero Training Error}\label{sec:zero}

Second, the zero training error regime is considered, which explores what happens when there is nothing left to learn from the training data. This regime is of great interest as the regularization behavior (or lack thereof) can be observed and contrasted. The most obvious way to analyze the zero training error case is to substitute the true target $y_{i}$ for the predicted output $f_{\theta}(x)_{i}$; however, this can sometimes result in indeterminates (\textit{i.e.}, $0/0$). Therefore, we instead consider the case where we approach the zero training error regime, \textit{i.e.}, where we get $\epsilon$ close to the true label. Since all the outputs sum to $1$, let

\begin{equation}
\mathbb{E}[f_{\theta}(x)_{i}] =
\begin{cases}
\epsilon, & y_{i} = 0 \\
1 - \epsilon(\mathcal{C} - 1), & y_{i} = 1
\end{cases}
\end{equation}

\subsection{Relationship to Cross-Entropy Loss}

From Figure \ref{fig:meta-learned-loss-functions} in Section \ref{sec:loss-functions} it is shown that the absolute cross-entropy loss ($\Loss^{ACE}$) bears a close resemblance to the widely used and \textit{de facto} standard cross-entropy loss ($\Loss^{CE}$). However, as we will show they are not just similar, they are equivalent when using a default parameterization, \textit{i.e.}, $\phi_0, \phi_1 = 1$, at the null epoch and when approaching the zero training error regime.

\subsubsection{Cross-Entropy Loss Decomposition}

\noindent
First decomposing the cross-entropy learning rule via substitution of the loss into Equation \eqref{eq:general-form} we get the following rule: 
\begin{equation}
\theta_{t+1} \leftarrow \theta_{t} + \alpha \bigg[\underbrace{\frac{\partial}{\partial f} \bigg(y_{i} \cdot \log(f_{\theta}(x)_{i})\bigg)}_{\delta_{\Loss^{CE}}} \frac{\partial}{\partial \theta_{j}} f_{\theta}(x)_{i}\bigg]
\end{equation}
where the behavior is defined as
\begin{equation}
    \delta_{\Loss^{CE}} = \frac{y_{i}}{f_{\theta}(x)_{i}}.
\end{equation}

\subsubsection{Behavior at the Null Epoch}

Subsequently, behavior at the null epoch can then be defined for $\Loss^{CE}$ piecewise for target vs non-target outputs as
\begin{equation}
\begin{split}
\delta_{\Loss^{CE}} &=
\begin{cases}
\frac{0}{\mathcal{C}^{-1}}, & y_{i} = 0 \\
\frac{1}{\mathcal{C}^{-1}}, & y_{i} = 1
\end{cases} \\
&=
\begin{cases}
0, & y_{i} = 0 \\
\mathcal{C}, & y_{i} = 1
\end{cases}
\end{split}
\label{eq:ce-behavior-null}
\end{equation}
This shows that when $y_{i}=1$ the target output value is maximized, while the non-target output values remain the same; however, due to the base model's softmax output activation function, the target output value being maximized will in turn minimize the non-target outputs. Notably, this is identical behavior at the null epoch to $\Loss^{ACE}$,
\begin{equation}
\begin{split}
\delta_{\Loss^{ACE}} &= 
\begin{cases}
-\frac{0}{\mathcal{C}^{-1} \cdot |\log(\phi_{1} \cdot \mathcal{C}^{-1})|}, & y_{i} = 0 \\
-\frac{\log(\phi_{1} \cdot \mathcal{C}^{-1})}{\mathcal{C}^{-1} \cdot |\log(\phi_{1} \cdot \mathcal{C}^{-1})|}, & y_{i} = 1
\end{cases} \\
&= 
\begin{cases}
-\frac{0}{\mathcal{C}^{-1} \cdot |\log(\mathcal{C}^{-1})|}, & y_{i} = 0 \\
\frac{-1 \cdot \log(\mathcal{C}^{-1})}{\mathcal{C}^{-1} \cdot (-1 \cdot \log(\mathcal{C}^{-1}))}, & y_{i} = 1
\end{cases} \\
&= 
\begin{cases}
0, & y_{i} = 0 \\
\mathcal{C}, & y_{i} = 1
\end{cases}
\end{split}
\label{eq:learned-behavior-null}
\end{equation}

\subsubsection{Behavior at Zero Training Error}

On the other extreme, the behavior of the cross-entropy loss ($\Loss^{CE}$) when approaching zero training error can be defined piecewise for target vs non-target outputs as follows
\begin{equation}
\begin{split}
\delta_{\Loss^{CE}} &= 
\begin{cases}
y_{i}/\epsilon, & y_{i} = 0 \\
y_{i}/(1 - \epsilon(\mathcal{C} - 1)), & y_{i} = 1
\end{cases} \\
&=\lim_{\epsilon\rightarrow 0}
\begin{cases}
0, & y_{i} = 0 \\
1, & y_{i} = 1
\end{cases}
\end{split}
\label{eq:ce-behavior-zero}
\end{equation}
The target output value is again maximized, while the non-target output values are minimized due to the softmax. This behavior is replicated by $\Loss^{ACE}$:
\begin{equation}
\begin{split}
\delta_{\Loss^{ACE}} &= 
\begin{cases}
-\frac{0}{\epsilon \cdot |\log(\phi_{1} \cdot \epsilon)|}, & y_{i} = 0 \\
-\frac{\log(\phi_{1} \cdot (1 - \epsilon(\mathcal{C} - 1))))}{(1 - \epsilon(\mathcal{C} - 1)) \cdot |\log(\phi_{1} \cdot (1 - \epsilon(\mathcal{C} - 1)))|}, & y_{i} = 1
\end{cases} \\
&= 
\begin{cases}
-\frac{0}{\epsilon \cdot |\log(\phi_{1} \cdot \epsilon)|}, & y_{i} = 0 \\
\frac{-1 \cdot \log(\phi_{1} \cdot (1 - \epsilon(\mathcal{C} - 1))))}{(1 - \epsilon(\mathcal{C} - 1)) \cdot (-1 \cdot \log(\phi_{1} \cdot (1 - \epsilon(\mathcal{C} - 1))))}, & y_{i} = 1
\end{cases} \\
&=\lim_{\epsilon\rightarrow 0}
\begin{cases}
0, & y_{i} = 0 \\
1, & y_{i} = 1
\end{cases}
\end{split}
\label{eq:learned-behavior-zero}
\end{equation}

\subsubsection{Result Discussion}

\begin{figure*}[t!]

\begin{subfigure}{0.5\columnwidth}
\centering
\includegraphics[width=1\columnwidth]{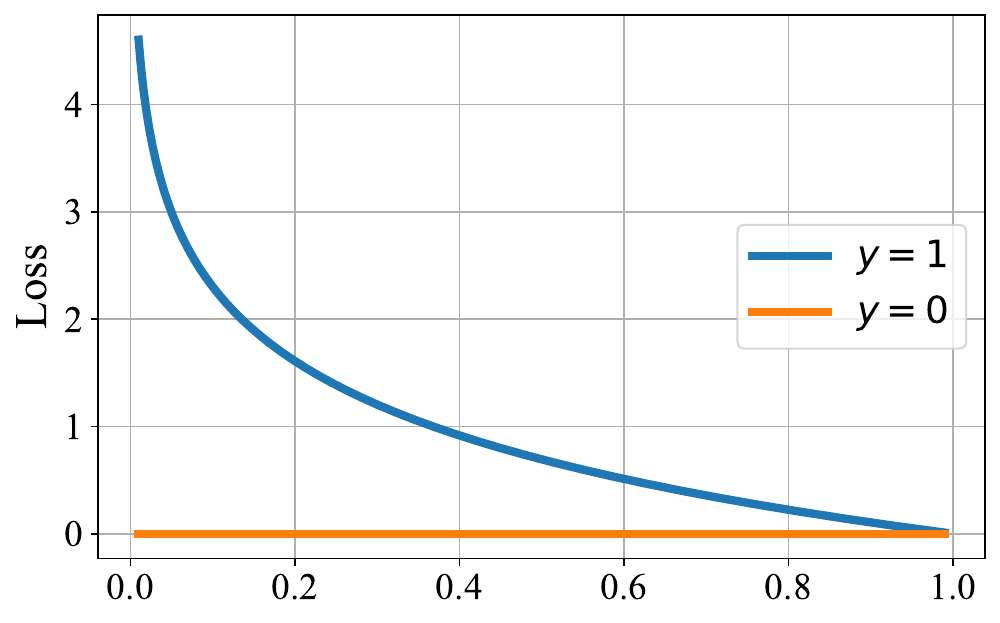}
\end{subfigure}
\begin{subfigure}{0.5\columnwidth}
\centering
\includegraphics[width=1\columnwidth]{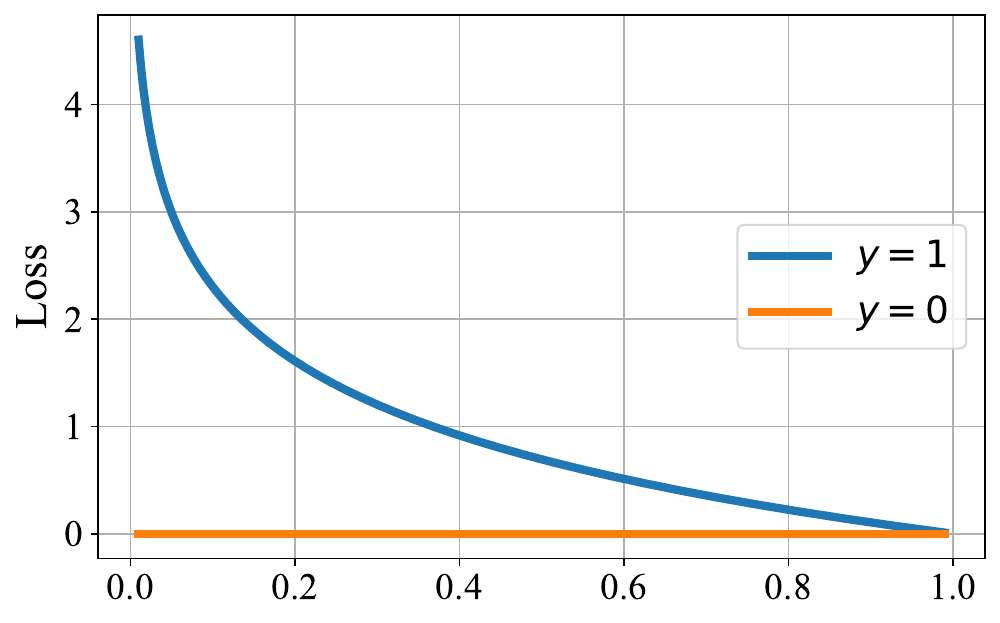}
\end{subfigure}

\captionsetup{justification=centering}
\caption{Contrasting the \textit{Cross-Entropy Loss} (left) to the \textit{Absolute Cross-Entropy Loss} (right), where the x-axis represents the model's predictions $f_{\theta}(x)$.}
\label{fig:learned-loss-function}
\end{figure*}

\noindent
These results show that in the extreme limits of training (\textit{i.e.}, very start and very end of training), the absolute cross-entropy loss with a default parameterization $\phi_{0}, \phi_{1} = 1$ is equivalent in learning behavior to the cross-entropy loss. At the null epoch, both loss functions will maximize the target and minimize the non-target (due to the softmax). Approaching zero training error, this behavior will continue, with the target output value being maximized and the non-target outputs minimized. This result can be validated visually as shown in Figure \ref{fig:learned-loss-function}. This is a notable result as it shows that EvoMAL was able to derive the cross-entropy loss function directly from the data.

\subsection{Relationship to Label Smoothing Regularization}
\label{sec:label-smoothing-regularization}

A key aspect of the absolute cross-entropy loss, particularly evident in loss function (e) in Figure \ref{fig:meta-learned-loss-functions}, is its perplexing tendency to counterintuitively push the base model's prediction $f_{\theta}(x)_i$ away from 1 as it approaches close to 1. Functionally, this phenomenon aims to prevent a model from becoming too overconfident in its predictions, which can lead to poor generalization on new unseen testing data. Mitigating overconfidence is crucial for developing robust deep neural networks that can generalize outside the training data.

This behavior is similar to \textit{Label Smoothing Regularization} ($\Loss^{LSR}$) \citep{szegedy2016rethinking, muller2019does}, a widely-used regularization technique that transforms hard targets into soft targets. Label smoothing adjusts target labels according to the formula $y_{i} \leftarrow y_{i}(1 - \xi) + \xi/\mathcal{C}$, where $1 > \xi > 0$ is the smoothing coefficient. By softening the target labels $\Loss^{LSR}$ reduces the confidence of the model, which consequently helps to mitigate overfitting and enhance generalization.

As shown in the forthcoming derivations, the absolute cross-entropy loss and label smoothing regularization ($\Loss^{LSR}$) exhibit similarities in their effects on model predictions. Both techniques maximize the target outputs and minimize the non-target outputs, particularly at the initial training stage (null epoch) and as the training error approaches zero. These effects are observed under the conditions where $\phi_0=1$ and $2 > \phi_1 > 1$

\subsubsection{Label Smoothing Regularization Loss Decomposition}

First decomposing the label smoothing learning rule via substitution of the loss into Equation \eqref{eq:general-form} we get the following rule:
\begin{equation}
\theta_{t+1} \leftarrow \theta_{t} + \alpha \bigg[\underbrace{\frac{\partial}{\partial f} \bigg(\left(y_{i}(1-\xi) + \frac{\xi}{\mathcal{C}}\right) \log(f_{\theta}(x)_{i})\bigg)}_{\delta_{\Loss^{LSR}}} \frac{\partial}{\partial \theta_{j}} f_{\theta}(x)_{i}\bigg]
\end{equation}
where the behavior is defined as follows:
\begin{equation}
\begin{split}
\delta_{\Loss^{LSR}} 
&= \frac{\partial}{\partial f} \left(\left(y_{i}(1-\xi) + \frac{\xi}{\mathcal{C}}\right) \log(f_{\theta}(x)_{i})\right)\\
&= \left(y_{i}(1-\xi) + \frac{\xi}{\mathcal{C}}\right) \cdot \frac{\partial}{\partial f} \left(\log(f_{\theta}(x)_{i})\right)\\
&= \left(y_{i}(1-\xi) + \frac{\xi}{\mathcal{C}}\right) \cdot \frac{1}{f_{\theta}(x)_i}\\
&= \frac{y_{i}(1 - \xi) + \xi/\mathcal{C}}{f_{\theta}(x)_{i}} \\
\end{split}
\end{equation}

\subsubsection{Behavior at the Null Epoch}

Behavior at the null epoch can then be defined for label smoothing regularization ($\Loss^{LSR}$) piecewise for target vs non-target outputs as
\begin{equation}\label{eq:label-smoothing-null-behavior}
\begin{split}
\delta_{\Loss^{LSR}} &= 
\begin{cases}
\frac{\xi/\mathcal{C}}{\mathcal{C}^{-1}}, & y_{i} = 0 \\
\frac{1 - \xi + \xi/\mathcal{C}}{\mathcal{C}^{-1}}, & y_{i} = 1
\end{cases} \\
&= 
\begin{cases}
\xi, & y_{i} = 0 \\
\mathcal{C} - \mathcal{C}\xi + \xi, & y_{i} = 1
\end{cases}
\end{split}
\end{equation}
where the $y_{i}=1$ case is shown to dominate $y_{i}=0$ since
\begin{equation}
\xi < \mathcal{C} - \mathcal{C}\xi + \xi.
\end{equation}
This result shows that when $y_{i}=1$ the target output value will be maximized, while the non-target output values will be minimized due to the base model's softmax activation function. This is as expected, both consistent with the behavior of the cross-entropy, Equation \eqref{eq:ce-behavior-null}, as well as with the absolute cross-entropy, where $\phi_1=1$, Equation \eqref{eq:learned-behavior-null}. Furthermore, as shown this behavior is also consistent with the behavior of the absolute cross-entropy where $2>\phi_1>1$, since
\begin{equation}
\begin{split}
\delta_{\Loss^{ACE}} &= 
\begin{cases}
-\frac{0}{\mathcal{C}^{-1} \cdot |\log(\phi_{1} \cdot \mathcal{C}^{-1})|}, & y_{i} = 0 \\
-\frac{\log(\phi_{1} \cdot \mathcal{C}^{-1})}{\mathcal{C}^{-1} \cdot |\log(\phi_{1} \cdot \mathcal{C}^{-1})|}, & y_{i} = 1
\end{cases} \\
&= 
\begin{cases}
-\frac{0}{\mathcal{C}^{-1} \cdot |\log(\phi_{1} \cdot \mathcal{C}^{-1})|}, & y_{i} = 0 \\
\frac{-1 \cdot \log(\phi_{1} \cdot \mathcal{C}^{-1})}{\mathcal{C}^{-1} \cdot (-1 \cdot \log(\phi_{1} \cdot \mathcal{C}^{-1}))}, & y_{i} = 1
\end{cases} \\
&= 
\begin{cases}
0, & y_{i} = 0 \\
\mathcal{C}, & y_{i} = 1
\end{cases}
\end{split}
\end{equation}

\subsubsection{Behavior at Zero Training Error}

As label smoothing regularization is a regularization technique that penalizes overconfidence, we expect to see noteworthy results when approaching zero training error. The regularization behavior of $\delta_{LSR}$ as the zero training error case is approached:
\begin{equation}\label{eq:label-smoothing-behavior}
\begin{split}
\delta_{\Loss^{LSR}} &= 
\begin{cases}
\frac{\xi/\mathcal{C}}{\epsilon}, & y_{i} = 0 \\
\frac{1 - \xi + \xi/\mathcal{C}}{1 - \epsilon(\mathcal{C} - 1)}, & y_{i} = 1
\end{cases} \\
&=\lim_{\epsilon\rightarrow 0}
\begin{cases}
+\infty, & y_{i} = 0 \\
1 - \xi + \frac{\xi}{\mathcal{C}}, & y_{i} = 1
\end{cases}
\end{split}
\end{equation}
where the $y_{i}=0$ case is shown to dominate $y_{i}=1$ since
\begin{equation}
+\infty > 1 - \xi + \frac{\xi}{\mathcal{C}}.
\end{equation}
Contrary to the prior results we can see that as the zero training error case is approached the non-target output is maximized while the target output is minimized. This result clearly demonstrates the underlying mechanism for how label smoothing regularization penalizes overconfidence and improves generalization. A noteworthy discovery that we put forward is that this underlying mechanism is also apparent in $\Loss^{ACE}$. Let $\phi_0=1$ and $2 > \phi_1 > 1$, then:
\begin{equation}
\begin{split}
\delta_{\Loss^{ACE}} &= 
\begin{cases}
-\frac{0}{\epsilon|\log(\phi_{1} \cdot \epsilon)|}, & y_{i} = 0 \\
-\frac{\log(\phi_{1}(1 - \epsilon(\mathcal{C} - 1) + \epsilon))}{(1 - \epsilon(\mathcal{C} - 1) + \epsilon)|\log(\phi_{1}(1 - \epsilon(\mathcal{C} - 1) + \epsilon))|}, & y_{i} = 1 \\
\end{cases} \\
&= 
\begin{cases}
-\frac{0}{\epsilon|\log(\phi_{1} \cdot \epsilon)|}, & y_{i} = 0 \\
\frac{-1 \cdot \log(\phi_{1}(1 - \epsilon(\mathcal{C} - 1) + \epsilon))}{(1 - \epsilon(\mathcal{C} - 1) + \epsilon)\cdot\log(\phi_{1}(1 - \epsilon(\mathcal{C} - 1) + \epsilon))}, & y_{i} = 1 \\
\end{cases} \\
&=\lim_{\epsilon\rightarrow 0}
\begin{cases}
0, & y_{i} = 0 \\
-1, & y_{i} = 1 \\
\end{cases}
\end{split}
\end{equation}

\subsubsection{Results Discussion}

The results show the functional equivalence in behavior between label smoothing regularization and the absolute cross-entropy loss where $2 > \phi_{1} > 1$. At the start of training where the model weights $\theta$ are randomized $\delta_{\Loss^{ACE}}$ is shown to be identical to $\delta_{\Loss^{LSR}}$ and $\delta_{\Loss_{CE}}$, maximizing the target and due to the softmax minimizing the non-target. In contrast, the results presented in Equation \ref{eq:learned-behavior-zero} show that when approaching zero training error $\delta_{\Loss^{ACE}}$ has the same unique behavior as $\delta_{\Loss^{LSR}}$, forcing the non-target outputs to be maximized while the target output is minimized. Contrary to a typical loss function, $\Loss^{LSR}$ and $\Loss^{ACE}$ aim to maximize the divergence between overly confident prediction between the predicted distribution and the true target distribution.

\begin{figure}[t!]
\centering

\begin{subfigure}{0.45\textwidth}
\includegraphics[width=1\textwidth]{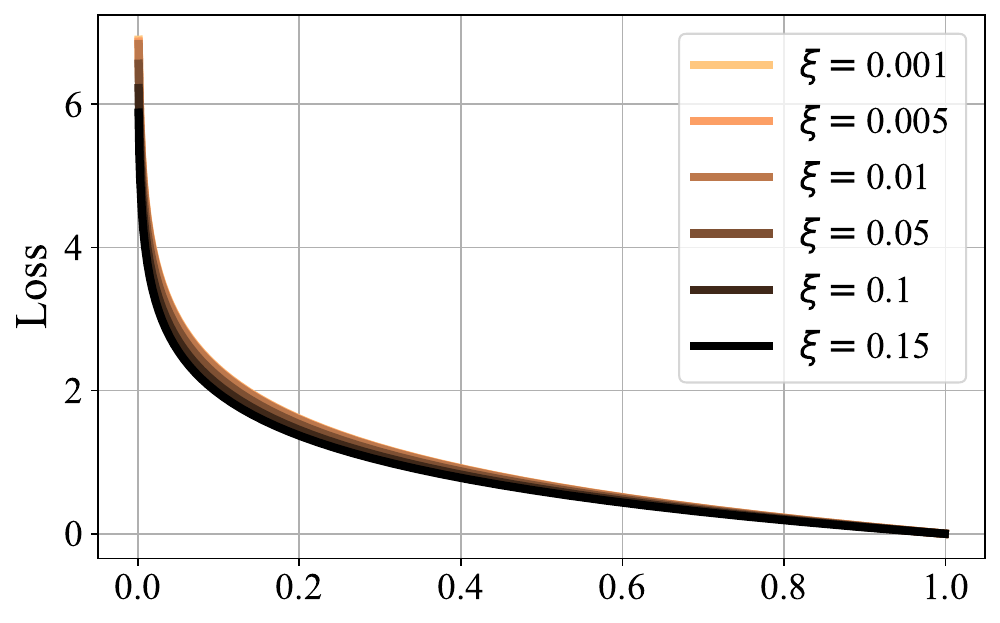}
\end{subfigure}%
\hspace{5mm}
\begin{subfigure}{0.45\textwidth}
\includegraphics[width=1\textwidth]{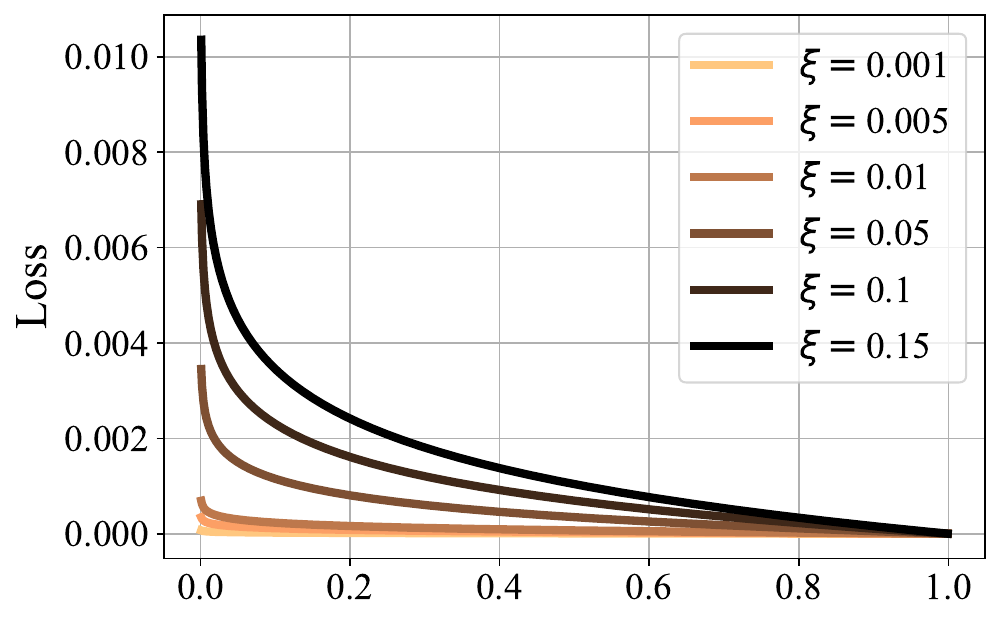}
\end{subfigure}%

\captionsetup{justification=centering}
\caption{Visualizing \textit{Label Smoothing Regularization Loss} with varying smoothing values $\xi$, where the left figure shows the target loss ($y_i = 1$), and the right figure shows the non-target loss ($y_i = 0$).}
\label{fig:label-smoothing-loss}
\end{figure}

The parity in behavior between $\Loss^{LSR}$ and $\Loss^{ACE}$ can be seen in Figures \ref{fig:varying-learned-loss-function} and \ref{fig:label-smoothing-loss}, respectively. This is a notable result as it shows that EvoMAL was able to discover label smoothing regularization directly from the data, with no prior knowledge of the regularization technique being given to the method. Furthermore, although similar in behavior, it is noteworthy that $\Loss^{ACE}$ is “\textit{sparse}” as it only needs to be computed on the target output \textit{i.e.}, constant time and space complexity $\Theta(1)$, with respect to the number of classes $\mathcal{C}$, as it is 0 for all non-target outputs, as shown in Equations \eqref{eq:learned-behavior-null} and \eqref{eq:learned-behavior-zero} when $y_i=0$. In contrast, label smoothing regularization must be applied to both the target output and all non-target outputs \textit{i.e.}, linear time and space complexity $\Theta(\mathcal{C})$, with respect to the number of classes $\mathcal{C}$, as shown in Equation \eqref{eq:label-smoothing-behavior}. Hence, the learned loss function can be computed significantly faster than the cross-entropy with label smoothing.

\subsection{Further Analysis}

\begin{figure}[t!]
\begin{subfigure}{0.5\columnwidth}
\centering
\includegraphics[width=1\columnwidth]{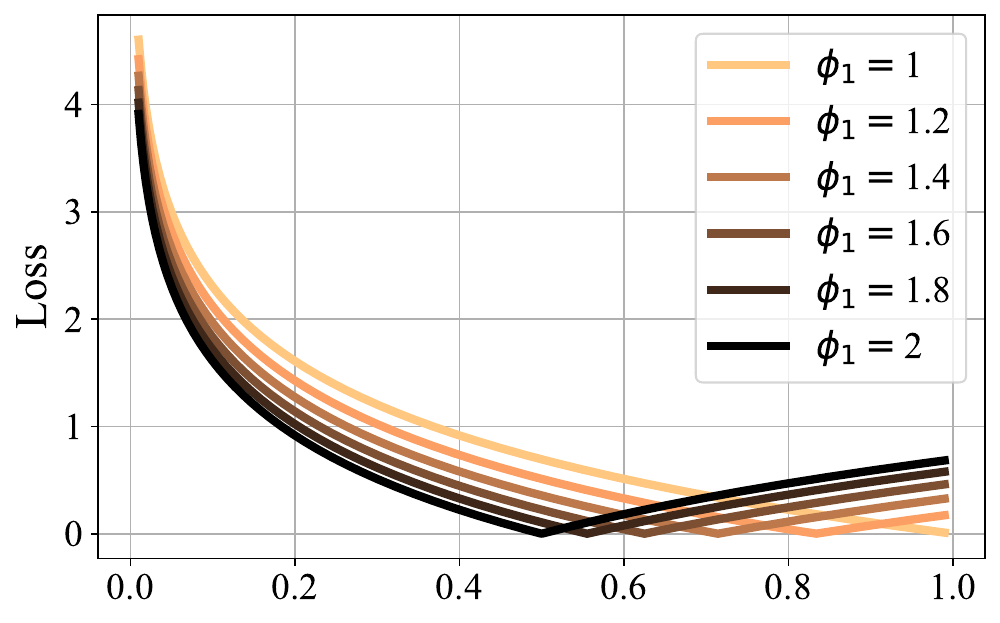}
\end{subfigure}
\begin{subfigure}{0.5\columnwidth}
\centering
\includegraphics[width=1\columnwidth]{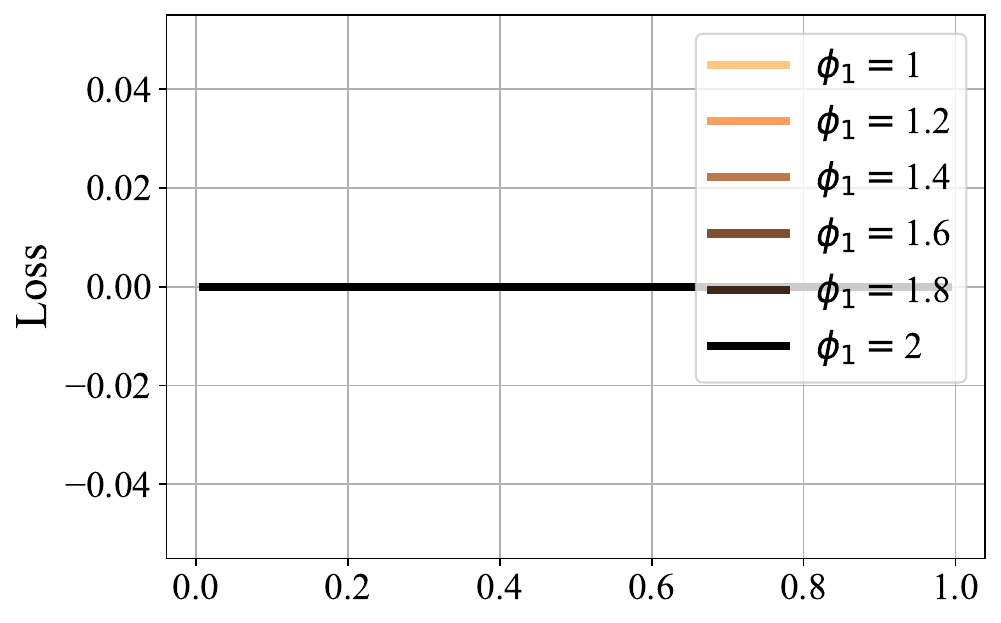}
\end{subfigure}
\captionsetup{justification=centering}
\caption{Visualizing the proposed \textit{Absolute Cross-Entropy Loss}, with varying values of $\phi_1$, where the left figure shows the target loss ($y_i = 1$), and the right figure shows the non-target loss ($y_i = 0$).}
\label{fig:varying-learned-loss-function}
\end{figure}

\begin{figure}[t!]
\centering

\begin{subfigure}{0.45\textwidth}
\includegraphics[width=1\textwidth]{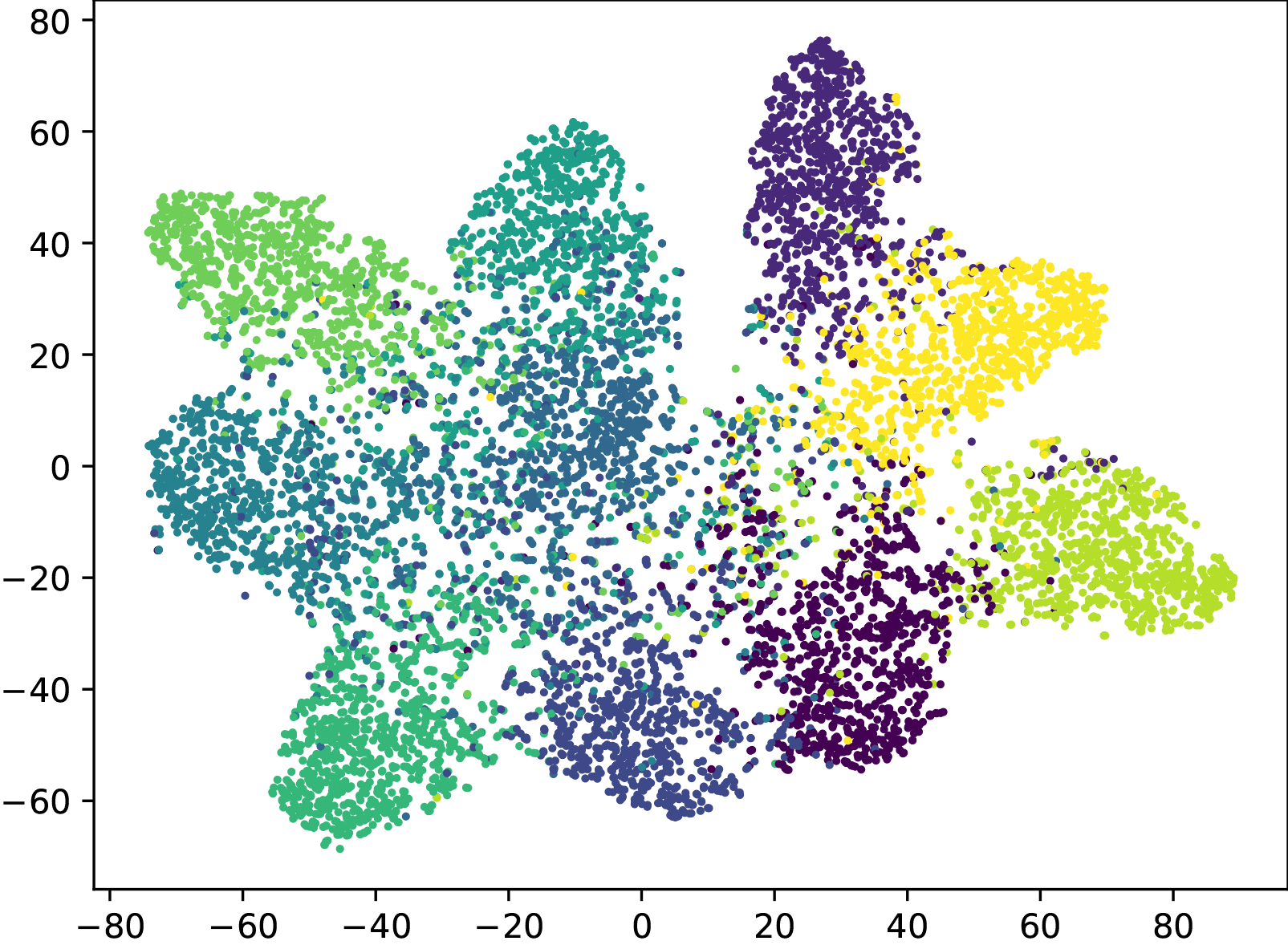}
\caption{Cross-Entropy -- 15.66\%.}
\end{subfigure}%
\hspace{5mm}
\begin{subfigure}{0.45\textwidth}
\includegraphics[width=1\textwidth]{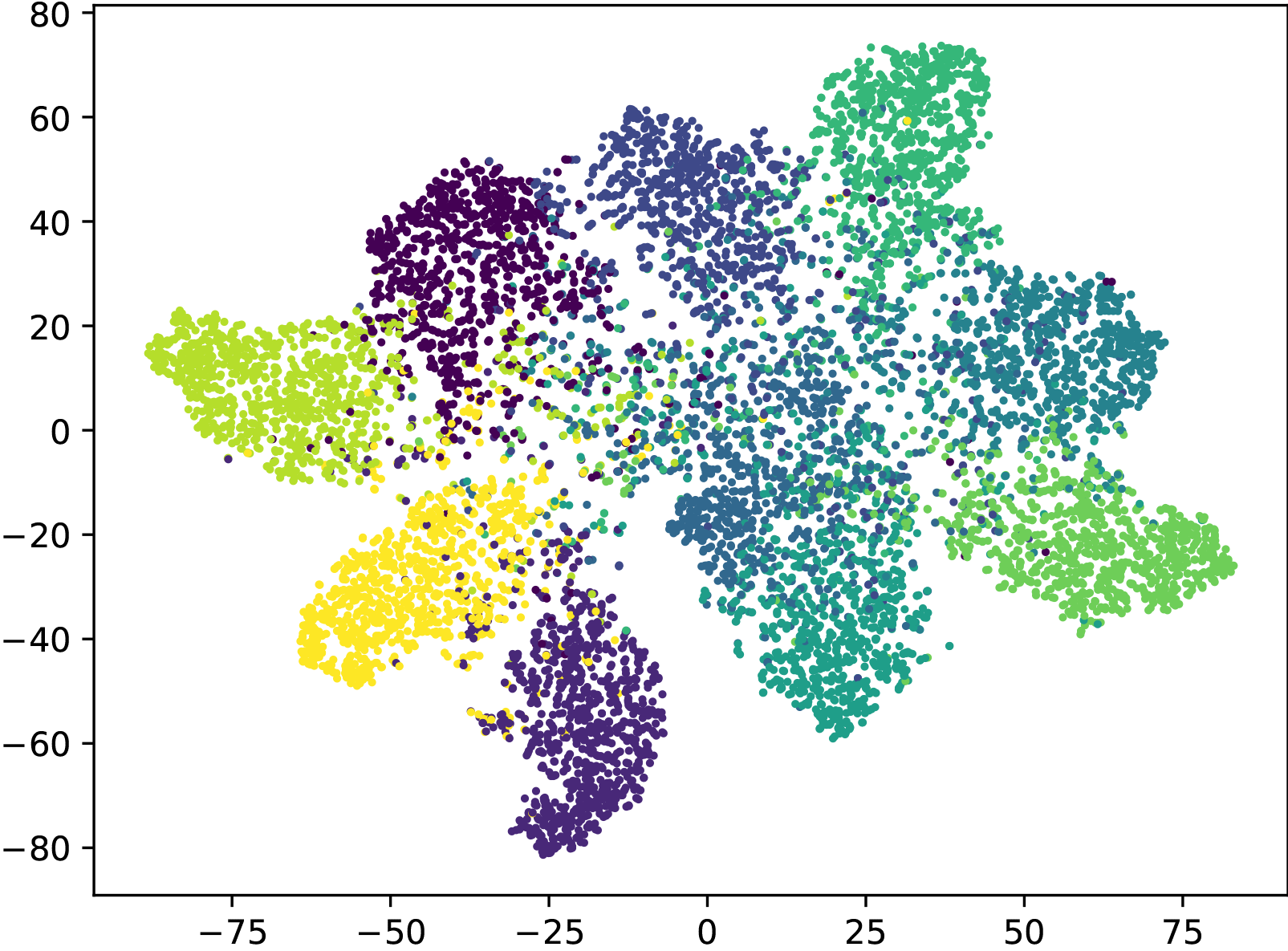}
\caption{Absolute CE ($\phi_{1}=1$) -- 15.43\%.}
\end{subfigure}%
\par\bigskip
\begin{subfigure}{0.45\textwidth}
\includegraphics[width=1\textwidth]{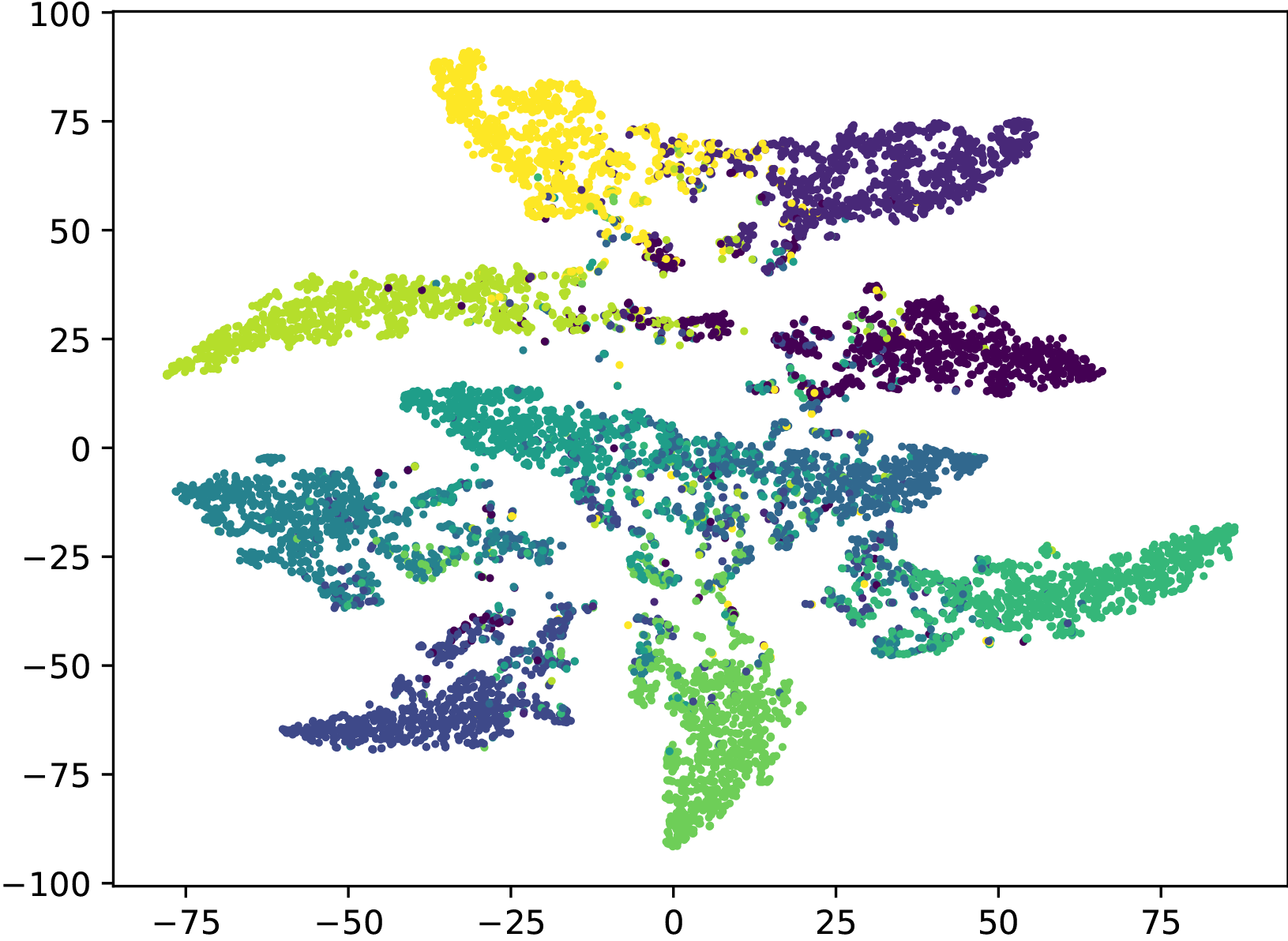}
\caption{Label Smoothing ($\xi = 0.1$) -- 14.99\%}
\end{subfigure}
\hspace{5mm}
\begin{subfigure}{0.45\textwidth}
\includegraphics[width=1\textwidth]{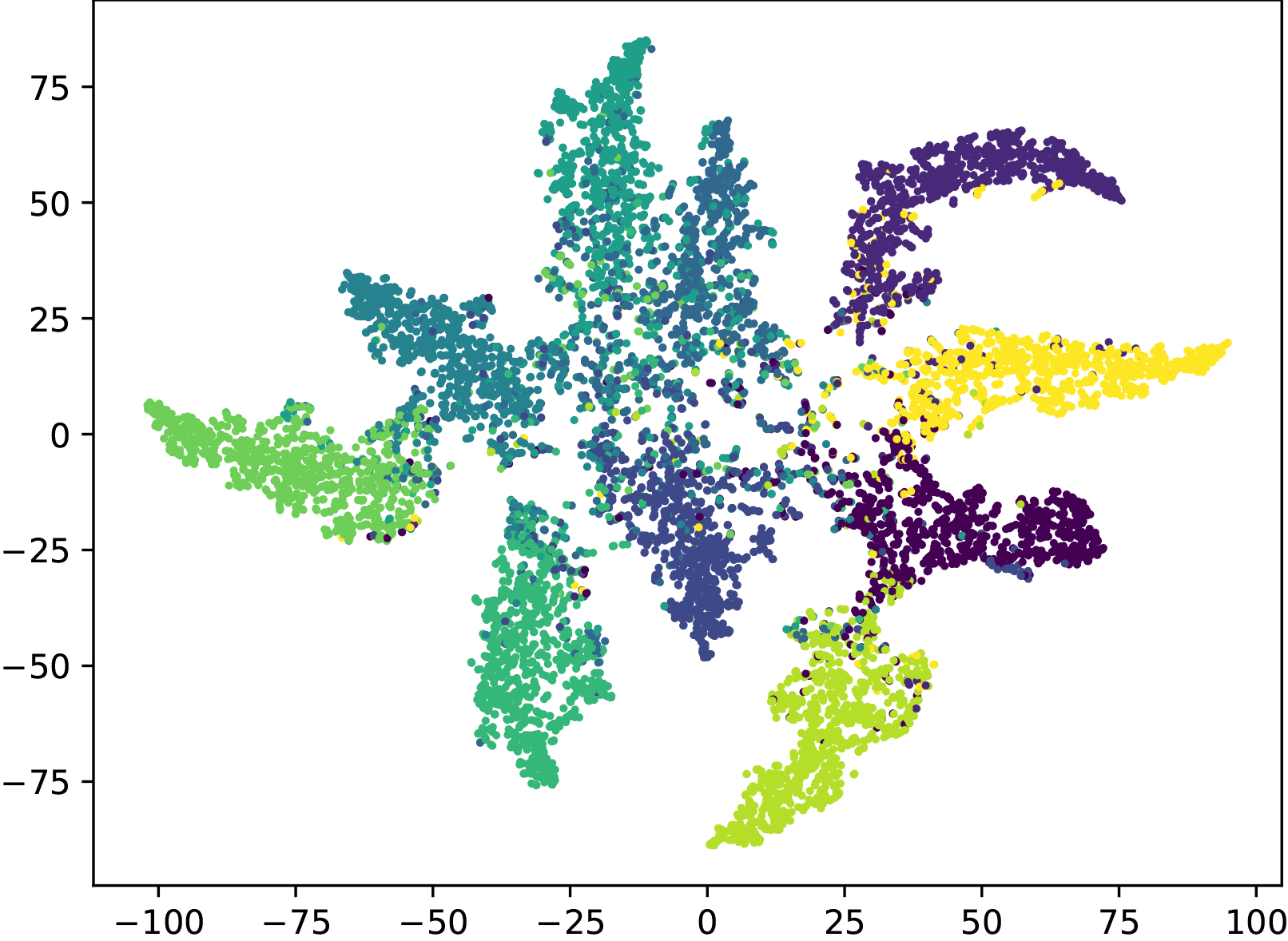}  
\caption{Absolute CE ($\phi_{1}=1.1$) -- 14.92\%}
\end{subfigure}

\captionsetup{justification=centering}
\caption{The penultimate layer representations on AlexNet CIFAR-10, using t-SNE for dimensionality reduction. Comparing \textit{Absolute Cross-Entropy Loss} (Absolute CE) where $\phi_1 = \{1, 1.1\}$, to \textit{Cross-Entropy Loss} and \textit{Label Smoothing Regularization Loss}. Color represents the class of an instance.}
\label{fig:learned-respresentations}
\end{figure}

To validate our theoretical findings we visualize and analyze the penultimate layer representations of AlexNet on CIFAR-10 using t-distributed Stochastic Neighbor Embedding (t-SNE) \citep{van2008visualizing}, when trained with the cross-entropy, label smoothing regularization, and the absolute cross-entropy with $\phi_{1}=\{1, 1.1\}$. The results are presented in Figure \ref{fig:learned-respresentations}. It shows that the representations learned by (a) the cross-entropy and (b) absolute cross-entropy when $\phi=1$ are visually very similar, supporting the findings in Sections \ref{sec:null} and \ref{sec:zero}. Furthermore, analyzing the learned representations of (c) label smoothing regularization where $\xi = 0.1$ and (d) the absolute cross-entropy where $\phi_{1} = 1.1$ it is shown that both the learned representations have better discriminative inter-class representations and tighter intra-class representations. This empirical analysis supports our findings from the theoretical analysis in Section \ref{sec:theoretical-analysis}, showing consistent regularization results between label smoothing regularization and the absolute cross-entropy loss when $\phi_{1} > 1$.

The results of our meta-learned loss function thus far have shown to be very positive; nonetheless, we would like to draw our readers' attention to two prominent limitations of the absolute cross-entropy loss function. Firstly, $\Loss^{ACE}$ has shown to be equivalent in behavior to $\Loss^{LSR}$ at the null epoch and when approaching zero training error, \textit{i.e.}, at the start and very end of training; however, for intermediate points, the behavior is not identical. Secondly, $\Loss^{ACE}$ is not smooth as shown in Figure \ref{fig:varying-learned-loss-function} when transitioning between minimizing and maximizing the target output. Consequently, this can cause training to become unstable, disrupting the learning process and negatively affecting the generalization of the base model. In the following section, we apply our findings and discoveries from the absolute cross-entropy loss to propose a new loss function that resolves both of these issues while maintaining the key properties of $\Loss^{ACE}$, specifically, its constant time and space complexity due to target-only loss computation.

\newpage

\section{Sparse Label Smoothing Regularization}
\label{sec:sparse-label-smoothing-regularization}

Label Smoothing Regularization (LSR) \citep{szegedy2016rethinking, goodfellow2016deep, muller2019does} is a popular and effective technique for preventing overconfidence in a classification model. This technique achieves regularization by penalizing overconfident predictions through adjusting the target label $y$ during training such that a $y_{i} \leftarrow y_{i}(1 - \xi) + \xi/\mathcal{C}$, where $1 > \xi > 0$ is the smoothing coefficient. LSR forces the development of more robust and generalizable features within the model, as it discourages the model from being overly confident in its predictions and forces it to consider alternative possibilities, ultimately improving generalization performance. The cross-entropy loss when combined with label smoothing regularization is defined as follows:
\begin{equation}
\resizebox{0.9\columnwidth}{!}{$
\begin{split}
    \Loss^{LSR}  
    &= - \sum_{i=1}^{\mathcal{C}} \left(y_{i} \cdot (1 - \xi) + \frac{\xi}{\mathcal{C}}\right) \cdot \log(f_{\theta}(x)_{i}) \\
    &= - \sum_{i=1, y_{i} = 1}^{\mathcal{C}} \left(y_{i} \cdot (1 - \xi) + \frac{\xi}{\mathcal{C}}\right) \cdot \log(f_{\theta}(x)_{i}) + \sum_{i=1, y_{i} = 0}^{\mathcal{C}} \left(y_{i} \cdot (1 - \xi) + \frac{\xi}{\mathcal{C}}\right) \cdot \log(f_{\theta}(x)_{i}) \\
    &= - \underbrace{\sum_{i=1, y_{i} = 1}^{\mathcal{C}} \left(1 - \xi + \frac{\xi}{\mathcal{C}}\right) \cdot \log(f_{\theta}(x)_{i})}_{\text{Target Loss}} + \underbrace{\sum_{i=1, y_{i} = 0}^{\mathcal{C}} \frac{\xi}{\mathcal{C}} \cdot \log(f_{\theta}(x)_{i})}_{\text{Non-Target Losses}} \\
\end{split}
$}
\label{eq:label-smoothing-regularization}
\end{equation}
In contrast to the cross-entropy loss which is typically implemented as a “\textit{sparse}” loss function, \textit{i.e.}, only computes the loss on the target output as the non-target loss is $0$, label smoothing regularization is a “\textit{non-sparse}” loss function as it has a non-zero loss for both the target and non-target outputs. Consequently, this results in having to do a scalar expansion on the target\footnote{In automatic differentiation/neural network libraries such as \texttt{PyTorch} the target labels are by default represented as a vector of target indexes $y \in \mathbb{Z}^{B} \cap [0, \mathcal{C}]$; therefore, non-sparse loss functions such as label smoothing regularization require one-hot encoding of the target prior to computing the loss function.}, or more commonly a vector to matrix (column) expansion when considering a batch of instances, by one-hot encoding the labels as follows:
\begin{equation}
\resizebox{0.8\columnwidth}{!}{$
y = 
\begin{bmatrix}
    y_1 \\
    y_2 \\
    \vdots \\
    y_{B} \\
\end{bmatrix} \in \mathbb{Z}^{B} \cap [0, \mathcal{C}]
\xrightarrow{\text{One-Hot Encode}}
\begin{bmatrix}
    y_{1, 1} & y_{2, 1} & \cdots & y_{\mathcal{C}, 1} \\
    y_{1, 2} & y_{2, 2} & \cdots & y_{\mathcal{C}, 2} \\
    \vdots & \vdots & \ddots & \vdots \\
    y_{1, B} & y_{2, B} & \cdots & y_{\mathcal{C}, B} \\
\end{bmatrix} \in \{0, 1\}^{\mathcal{C} \times B}
$}
\end{equation}
The label smoothing loss can subsequently be calculated using the one-hot encoded target matrix $y \in \{0, 1\}^{\mathcal{C} \times B}$ and the model predictions $f_{\theta}(x) \in (0, 1)^{\mathcal{C} \times B}$, where the summation of the resulting losses is taken across the $\mathcal{C}$ rows before being reduced in the $B$ batch dimension using some aggregation function such as the arithmetic mean. 

\pagebreak

\subsection{Derivation of Sparse Label Smoothing Regularization}

\begin{figure}[t!]
\centering

\begin{subfigure}{0.45\textwidth}
\includegraphics[width=1\textwidth]{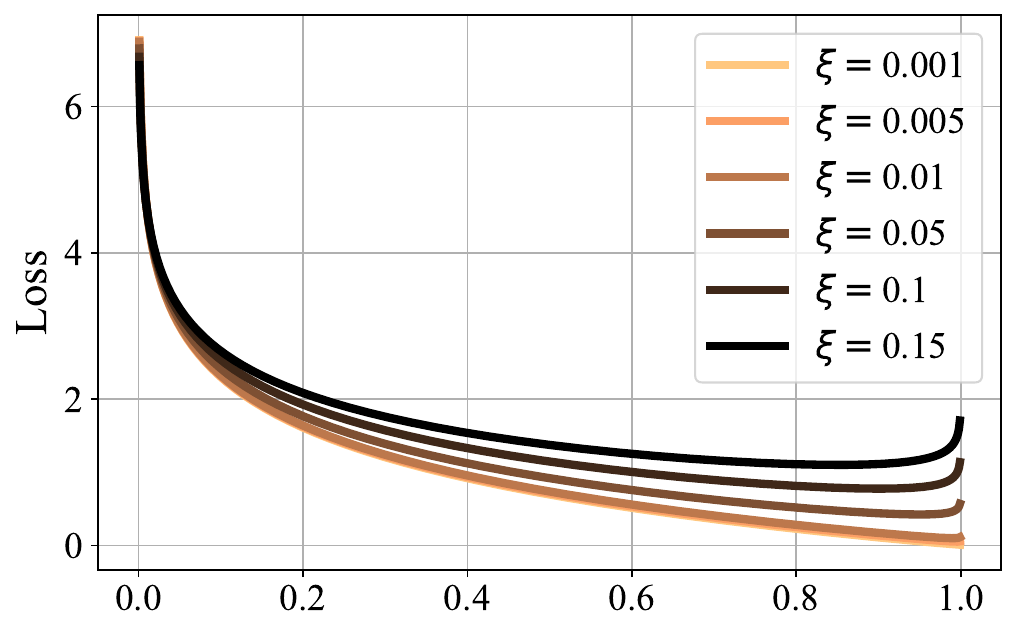}
\end{subfigure}%
\hspace{5mm}
\begin{subfigure}{0.45\textwidth}
\includegraphics[width=1\textwidth]{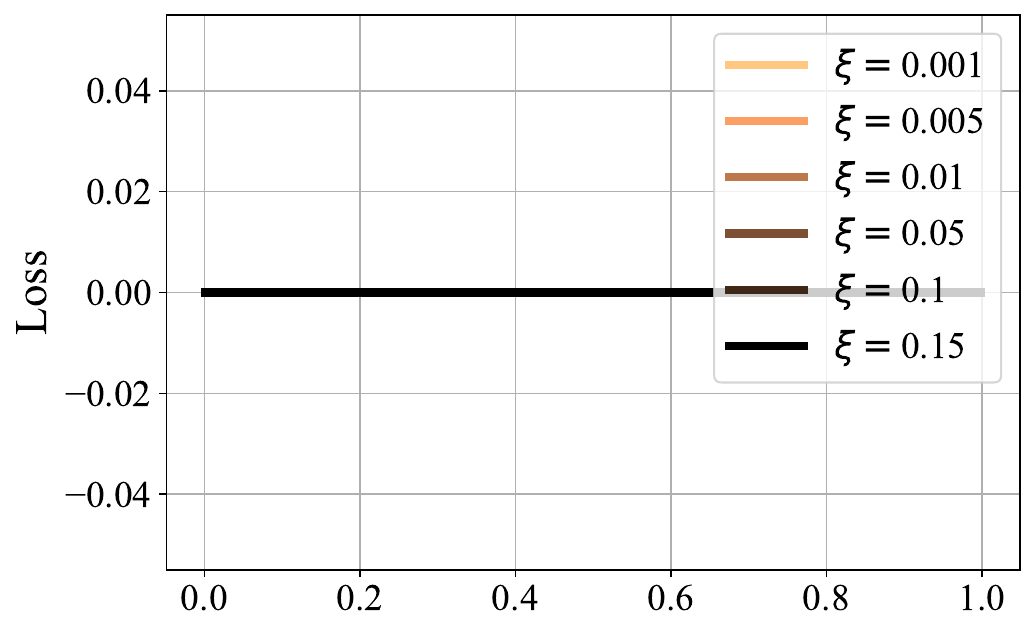}
\end{subfigure}%

\captionsetup{justification=centering}
\caption{Visualizing the proposed \textit{Sparse Label Smoothing Regularization} (SparseLSR) Loss with varying smoothing values $\xi$, where the left figure shows the target loss (\textit{i.e.}, $y_i = 1$), and the right figure shows the non-target loss (\textit{i.e.}, $y_i = 0$).}
\label{fig:fast-label-smoothing-loss}
\end{figure}

Due to its non-sparse nature, label smoothing regularization has a time and space complexity of $\Theta(\mathcal{C})$, as opposed to $\Theta(1)$ for the cross-entropy (ignoring the $B$ batch dimensions, as the concept of loss is defined on a singular instance). However, by utilizing an innovative technique inspired by the behavior of the absolute cross-entropy loss analyzed in Section \ref{sec:theoretical-analysis}, which we call the “\textit{redistributed loss trick}”, we show a way to compute what we further refer to as \textit{Sparse Label Smoothing Regularization} (SparseLSR) loss, which is as the name implies sparse, and has a time and space complexity of $\Theta(1)$.

The key idea behind sparse label smoothing regularization is to utilize the redistributed loss trick, which redistributes the expected non-target loss into the target loss, obviating the need to calculate the loss on the non-target outputs. The redistributed loss trick can retain near identical behavior due to the output softmax function redistributing the gradients back into the non-target outputs during backpropagation. The sparse label smoothing regularization loss is defined as follows:
\begin{equation}
\begin{split}
    \Loss^{SparseLSR}  
    &= - \sum_{i=1, y_{i} = 1}^{\mathcal{C}} \left(1 - \xi + \frac{\xi}{\mathcal{C}}\right) \cdot \log(f_{\theta}(x)_{i}) + \mathbb{E} \left[\sum_{j=1, y_{i} = 0}^{\mathcal{C}} \frac{\xi}{\mathcal{C}} \cdot \log(f_{\theta}(x)_{j})\right] \\
    &= - \sum_{i=1, y_{i} = 1}^{\mathcal{C}} \left(1 - \xi + \frac{\xi}{\mathcal{C}}\right) \cdot \log(f_{\theta}(x)_{i}) + \sum_{j=1, y_{j} = 0}^{\mathcal{C}} \frac{\xi}{\mathcal{C}} \cdot \mathbb{E} \left[\log(f_{\theta}(x)_{j})\right], \\
\end{split}
\end{equation}
where the expectation of the model's non-target output $\mathbb{E} \left[\log(f_{\theta}(x)_{j})\right]$ is approximated via a first-order Taylor-expansion, \textit{i.e.}, a linear approximation, which lets us rewrite the expectation in terms of $f_{\theta}(x)_j$.

\begin{equation}
\begin{split}
     \mathbb{E} \left[\log(f_{\theta}(x)_{j})\right]
     &\approx \mathbb{E} \left[\log(\mathbb{E}\left[f_{\theta}(x)_{j}\right]) + \frac{1}{\mathbb{E}\left[f_{\theta}(x)_{j}\right]}(f_{\theta}(x)_{j} - \mathbb{E}\left[f_{\theta}(x)_{j}\right])\right] \\
     &= \mathbb{E} \left[\log(\mathbb{E}\left[f_{\theta}(x)_{j}\right])\right] + \mathbb{E} \left[\frac{1}{\mathbb{E}\left[f_{\theta}(x)_{j}\right]}(f_{\theta}(x)_{j} - \mathbb{E}\left[f_{\theta}(x)_{j}\right])\right] \\
     &= \log(\mathbb{E}\left[f_{\theta}(x)_{j}\right]) + \frac{1}{\mathbb{E}\left[f_{\theta}(x)_{j}\right]}\mathbb{E}\left[(f_{\theta}(x)_{j} - \mathbb{E}\left[f_{\theta}(x)_{j}\right])\right] \\
     &= \log(\mathbb{E}\left[f_{\theta}(x)_{j}\right])
\end{split}
\label{eq:linear-approximation-of-log}
\end{equation}
As defined by the softmax activation function, the summation of the model's output predictions is $\sum_{i = 1}^{\mathcal{C}}f_{\theta}(x)_i = 1$; therefore, the expected value of the non-target output prediction $\mathbb{E}[f_{\theta}(x)_j]$ where $y_j = 0$ can be given as $1-f_{\theta}(x)_i$ where $y_i = 1$ normalized over the number of non-target outputs $\mathcal{C}-1$. Substituting this result back into our expression gives the following:
\begin{equation}
\begin{split}
    \Loss^{SparseLSR} 
    &\approx - \sum_{i=1, y_{i} = 1}^{\mathcal{C}} \left(1 - \xi + \frac{\xi}{\mathcal{C}}\right) \cdot \log(f_{\theta}(x)_{i}) + \sum_{j=1, y_{i} = 0}^{\mathcal{C}} \frac{\xi}{\mathcal{C}} \cdot \log(\mathbb{E}\left[f_{\theta}(x)_{j}\right]) \\
    &= - \sum_{i=1, y_{i} = 1}^{\mathcal{C}} \left(1 - \xi + \frac{\xi}{\mathcal{C}}\right) \cdot \log(f_{\theta}(x)_{i}) + \sum_{j=1, y_{i} = 0}^{\mathcal{C}} \frac{\xi}{\mathcal{C}} \cdot \log\left(\frac{1 - f_{\theta}(x)_i}{\mathcal{C}-1}\right) \\
\end{split}
\end{equation}
where the first conditional summation can be removed to make explicit that $\Loss^{SparseLSR}$ is only non-zero for the target output, \textit{i.e.}, where $y_i = 1$, and the second conditional summation can be removed to obviate recomputation of the non-target segment of the loss which is currently defined as the summation of a constant. The final definition of Sparse Label Smoothing Regularization Loss ($\Loss^{SparseLSR}$) is:
\begin{equation}\label{eq:sparse-label-smoothing-regularization}
    \Loss^{SparseLSR} \approx - \sum_{i=1}^{\mathcal{C}} y_{i}\left[\left(1 - \xi + \frac{\xi}{\mathcal{C}}\right) \cdot \log(f_{\theta}(x)_{i}) + \frac{\xi(\mathcal{C}-1)}{\mathcal{C}} \cdot \log\left(\frac{1 - f_{\theta}(x)_i}{\mathcal{C}-1}\right)\right] \\
\end{equation}
which has a corresponding derivative of
\begin{equation}
\resizebox{0.9\columnwidth}{!}{$
\begin{split}
\frac{\partial}{\partial f}\left[ \Loss^{SparseLSR} \right] 
&= - \sum_{i=1}^{\mathcal{C}} y_{i}\left[\frac{\partial}{\partial f}\left[\left(1 - \xi + \frac{\xi}{\mathcal{C}}\right) \cdot \log(f_{\theta}(x)_{i})\right] + \frac{\partial}{\partial f}\left[\frac{\xi(\mathcal{C}-1)}{\mathcal{C}} \cdot \log\left(\frac{1 - f_{\theta}(x)_i}{\mathcal{C}-1}\right)\right]\right] \\
&= - \sum_{i=1}^{\mathcal{C}} y_{i}\left[\left(1 - \xi + \frac{\xi}{\mathcal{C}}\right) \cdot \frac{1}{f_{\theta}(x)_i} + \frac{\xi(\mathcal{C}-1)}{\mathcal{C}} \cdot \frac{1}{f_{\theta}(x)_{i} - 1} \right] \\
&= - \sum_{i=1}^{\mathcal{C}} y_{i}\left[\frac{\mathcal{C}-\xi \cdot \mathcal{C} + \xi}{\mathcal{C} \cdot f_{\theta}(x)} + \frac{\xi(\mathcal{C}-1)}{C(f_{\theta}(x)-1)}\right] \\
&= - \sum_{i=1}^{\mathcal{C}} y_{i}\left[\frac{\mathcal{C} \cdot f_{\theta}(x) + \mathcal{C}(\xi - 1) - \xi}{\mathcal{C} \cdot f_{\theta}(x) \cdot (f_{\theta}(x) - 1)}\right] \\
\end{split}
$}
\end{equation}

\pagebreak

The proposed loss function $\Loss^{SparseLSR}$ is visualized in Figure \ref{fig:fast-label-smoothing-loss}. As shown it resolves the key limitations of $\Loss^{ACE}$; namely, the lack of smoothness and its divergent behavior from $\Loss^{LSR}$ for intermediate points between the null epoch and approaching zero training error. Importantly, $\Loss^{SparseLSR}$ maintains identical time and space complexity due to target-only loss computation, as shown in Figure \ref{fig:fast-label-smoothing-loss} where all non-target losses are zero.

Note, that the approximation of the non-target losses used in $\Loss^{SparseLSR}$ assumes that the remaining output probabilities are uniformly distributed among the non-target outputs. Although this is a crude assumption, as demonstrated in Section \ref{sec:sparse-label-smoothing-regularization-experiments} this does not affect the regularization performance. Intuitively, a small or even moderate amount of error in the approximation of the non-target loss in label smoothing regularization is insignificant, as we are approximating the value of an error term.

\subsection{Numerical Stability}

The proposed sparse label smoothing regularization loss is prone to numerical stability issues, analogous to the cross-entropy loss, when computing logarithms and exponentials (exponentials are taken in the softmax when converting logits into probabilities) causing under and overflow. In particular, the following expressions are prone to causing numerical stability issues:
\begin{equation}
\Loss^{SparseLSR} \approx - \sum_{i=1}^{\mathcal{C}} y_{i}\bigg[\left(1 - \xi + \frac{\xi}{\mathcal{C}}\right) \cdot \underbrace{\log\bigg(f_{\theta}(x)_{i}\bigg)}_{\substack{\text{numerically} \\ \text{unstable}}} + \frac{\xi(\mathcal{C}-1)}{\mathcal{C}} \cdot \underbrace{\log\left(\frac{1 - f_{\theta}(x)_i}{\mathcal{C}-1}\right)}_{\substack{\text{numerically} \\ \text{unstable}}}\bigg] 
\end{equation}
In order to attain numerical stability when computing $log(f_{\theta}(x)_i)$ the well known \textit{log-sum-exp trick} is employed to stably convert the pre-activation logit $z_i$ into a log probability which we further denote as $\widetilde{f_{\theta}}(x)_i$:
\begin{equation}
\begin{split}
    \widetilde{f_{\theta}}(x)_i
    &= \log(f_{\theta}(x)_i) \\ 
    &= \log\left(\frac{e^{z_{i}}}{\sum_{j=1}^{\mathcal{C}}e^{z_{j}}}\right) \\
    &= \log\left(e^{z_{i}}\right) - \log\left(\textstyle\sum_{j=1}^{\mathcal{C}}e^{z_{j}}\right) \\
    &= z_{i} - \underbrace{\left(\max(z) + \log\left(\textstyle\sum_{j=1}^{\mathcal{C}}e^{z_{j} - \max(z)}\right)\right)}_{\text{LogSumExp}} \\
\end{split}
\end{equation}
%
%
Regarding the remaining numerically unstable term, this can also be computed stably via the log-sum-exp trick; however, it would require performing the log-sum-exp operation an additional time, which would negate the time and space complexity savings over the non-sparse implementation of label smoothing regularization. Therefore, we propose to instead simply take the exponential of the target log probability to recover the raw probability and then add a small constant $\epsilon=1e-7$ to avoid the undefined $\log(0)$ case. The numerically stable sparse label smoothing loss is defined as follows:
\begin{equation}
\begin{split}
    \Loss^{SparseLSR} 
    &\approx - \sum_{i=1}^{\mathcal{C}} y_{i}\left[\left(1 - \xi + \frac{\xi}{\mathcal{C}}\right) \cdot \widetilde{f_{\theta}}(x)_{i} + \frac{\xi(\mathcal{C}-1)}{\mathcal{C}} \cdot \log\left(\frac{1 - e^{\widetilde{f_{\theta}}(x)_i} + \epsilon}{\mathcal{C}-1}\right)\right] \\
\end{split}
\end{equation}
%

\subsection{Time and Space Complexity Analysis}
\label{sec:time-and-space-complexity}

\begin{figure}
\centering

\begin{subfigure}{0.47\textwidth}
\includegraphics[width=1\textwidth]{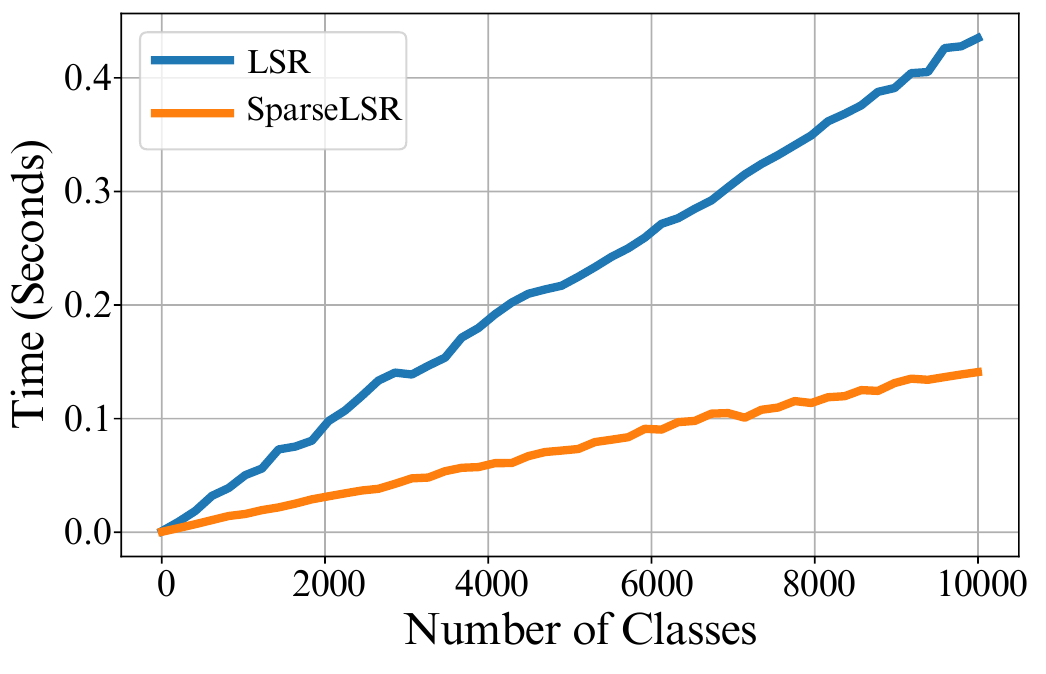}
\end{subfigure}%
\hspace{5mm}
\begin{subfigure}{0.47\textwidth}
\includegraphics[width=1\textwidth]{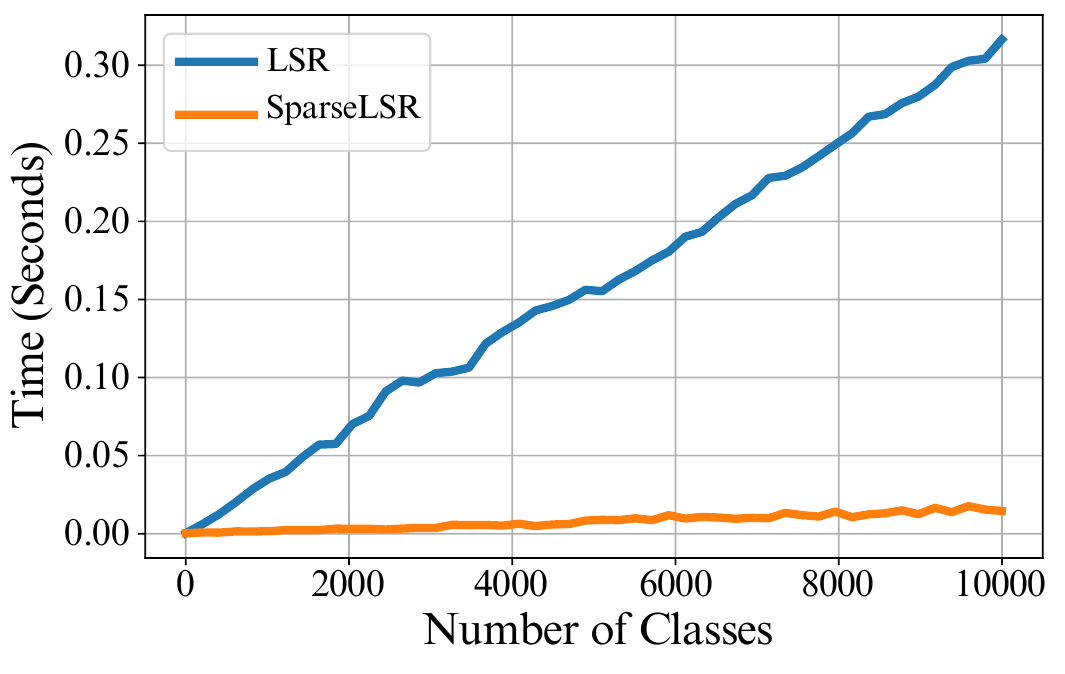}
\end{subfigure}%
\captionsetup{justification=centering}
\caption{Comparing the runtime of Label Smoothing Regularization (LSR) Loss and Sparse Label Smoothing Regularization (SparseLSR) Loss. The left figure shows the runtime with the log softmax operation and the right figure shows without.}
\label{fig:fastlsr-runtime}
\end{figure}

The proposed sparse label smoothing loss function is a sparse method for computing the prototypically non-sparse label smoothing loss given in Equation \ref{eq:label-smoothing-regularization}. Sparse label smoothing attributes zero losses for the non-target outputs. This makes it invariant to the number of classes $\mathcal{C}$; therefore, the loss function is only calculated and stored on the target while all non-target outputs/losses can be ignored, which gives a computational and storage complexity of $\Theta(1)$ compared to the $\Theta(\mathcal{C})$ for non-sparse label smoothing. To empirically validate the computational complexity of $\Loss^{SparseLSR}$ the runtime is tracked across a series of experiments using a batch of 100 randomly generated targets and logits where the number of classes is gradually increased $\mathcal{C} \in \mathbb{Z}: \mathcal{C} \in [3, 10000]$. The results are presented in Figure \ref{fig:fastlsr-runtime}, where the left figure shows the runtime with the log-softmax operation and the right figure shows without. 

The results confirm that the proposed sparse label smoothing loss function is significantly faster to compute compared to the non-sparse variant. When omitting the softmax-related computation, it is straightforward to see that the non-sparse label smoothing loss function scales linearly in the number of classes $\mathcal{C}$ while the sparse version does not. Although the time savings are admittedly modest in isolation, the loss function is frequently computed in the hundreds of thousands or even millions of times when training deep neural networks; therefore, the time savings when deployed at scale are notable. For example, let $f_{\theta}$ be trained for $1,000,000$ gradient steps for a $\mathcal{C}=10,000$ class classification problem with a batch size $\mathcal{B} = 100$. From Figure \ref{fig:fastlsr-runtime} the spare label smoothing loss function saves approximately $0.3$ seconds per batch compared to the non-sparse variant, so when run over a training session of $1,000,000$ gradient steps the time savings correspond to $(0.3 \times 1,000,000)/(60^2) \approx$ $83.3$ hours, which is a significant amount of time.

\subsection{Visualizing Learned Representations}
\label{sec:sparse-label-smoothing-regularization-experiments}

To further validate the parity of the regularization behavior of our sparse label smoothing loss function compared to non-sparse label smoothing we again visualize and analyze the penultimate layer representations of AlexNet on CIFAR-10 using t-SNE \citep{van2008visualizing}. The results are presented in Figure \ref{fig:fast-label-smoothing-regularization-1}, and they show that the learned representations between the two loss functions are very similar, with often near identical inter-class separations and inter-class overlaps between the same classes. These results empirically confirm that our sparse label smoothing loss function which redistributes the expected non-target loss into the target has faithfully retrained the original behavior of label smoothing regularization. 

A surprising result from our experiments was that for larger values of $\xi$, there is slightly improved inter-class separation and more compact intra-class representation when using the sparse version of label smoothing regularization. Consequently, our sparse label smoothing loss function has slightly improved final inference performance relative to non-sparse label smoothing, often showing a consistent $0.5-1\%$ improvement in error rate, which was not intentional or expected. We hypothesize that this is an effect caused by the model uniformly penalizing all the non-target outputs for overconfident target predictions during backpropagation, enabling the model to learn more robust decision boundaries and internal representations. In non-sparse label smoothing, non-target outputs are not penalized uniformly; instead, they are penalized on a case-by-case basis. 

\begin{figure*}
\centering

\begin{subfigure}{0.45\textwidth}
\includegraphics[width=1\textwidth]{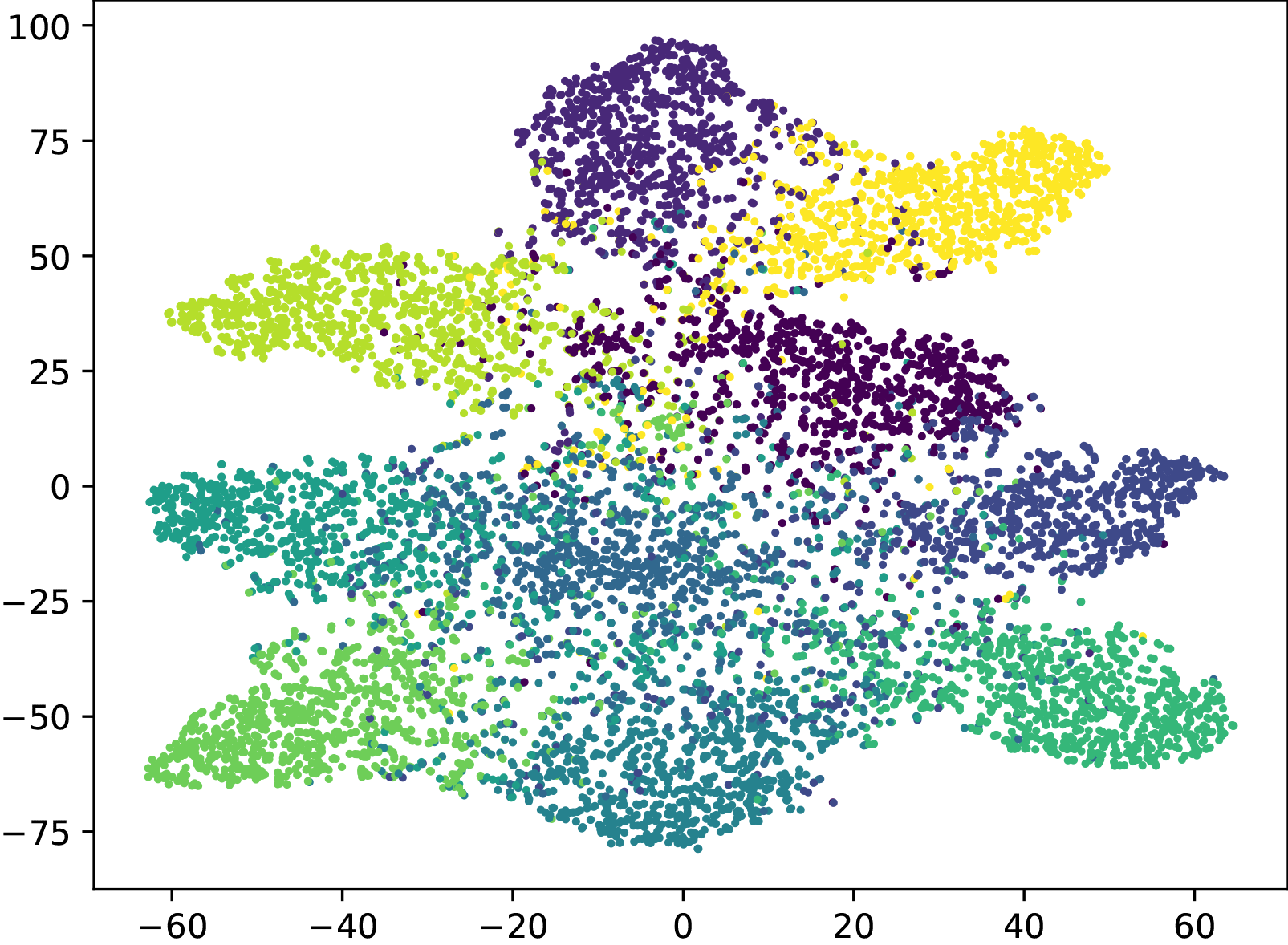}
\caption{LSR ($\xi=0$) -- 15.50\%.}
\end{subfigure}%
\hspace{5mm}
\begin{subfigure}{0.45\textwidth}
\includegraphics[width=1\textwidth]{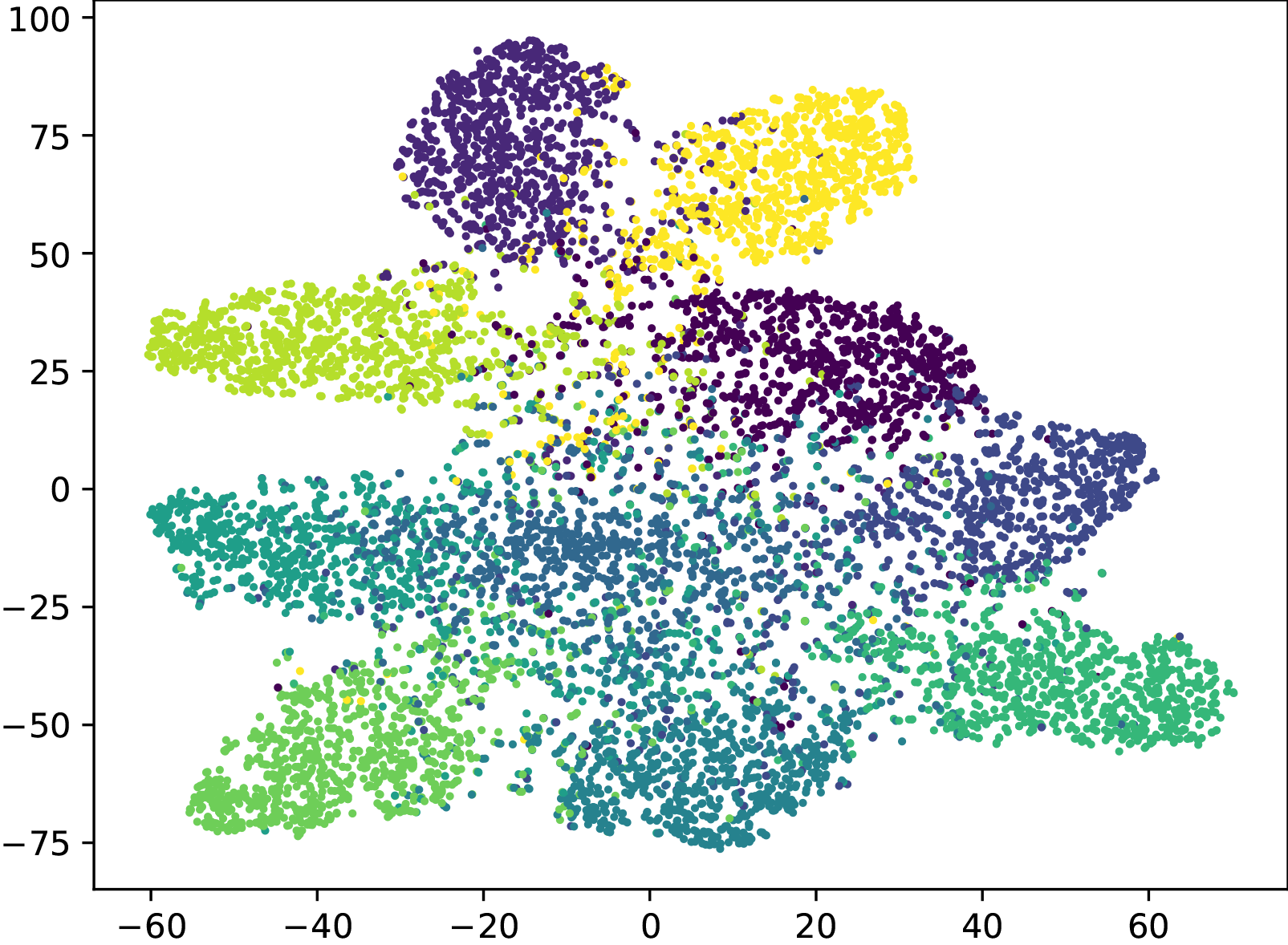}
\caption{SparseLSR ($\xi=0$) -- 15.23\%.}
\end{subfigure}%

\par\bigskip

\begin{subfigure}{0.45\textwidth}
\includegraphics[width=1\textwidth]{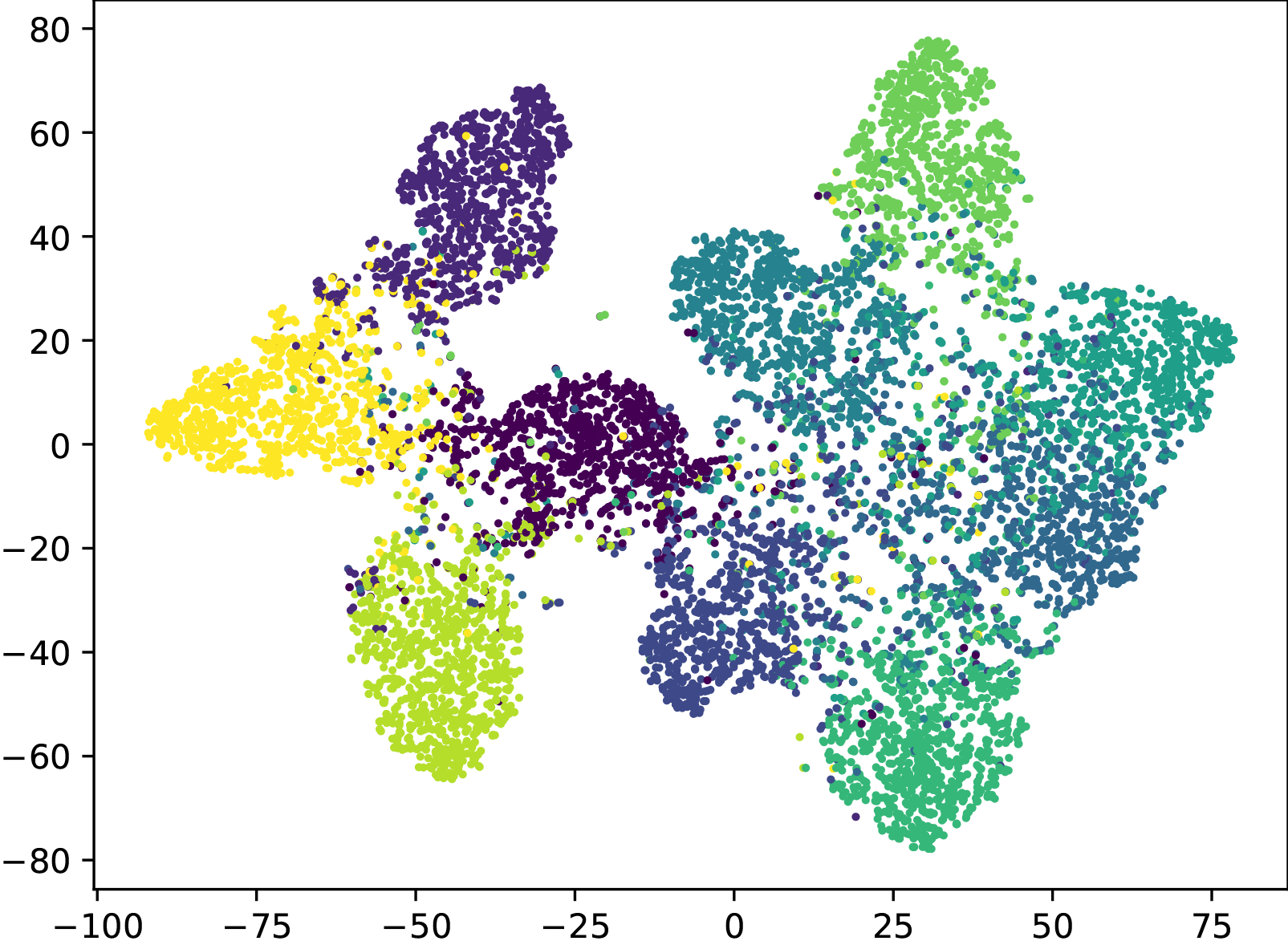}
\caption{LSR ($\xi=0.001$) -- 15.53\%.}
\end{subfigure}%
\hspace{5mm}
\begin{subfigure}{0.45\textwidth}
\includegraphics[width=1\textwidth]{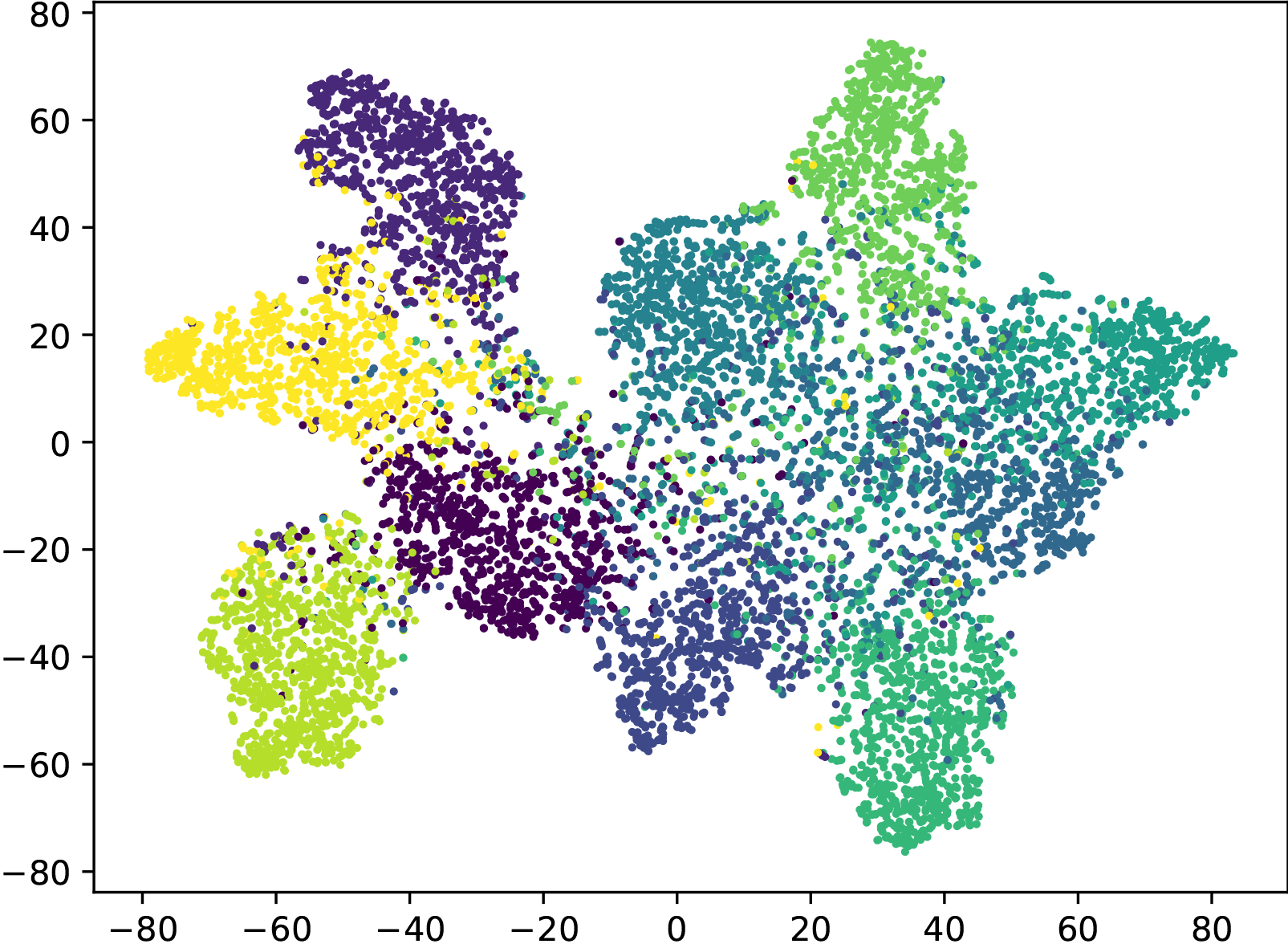}
\caption{SparseLSR ($\xi=0.001$) -- 15.47\%.}
\end{subfigure}%

\par\bigskip

\begin{subfigure}{0.45\textwidth}
\includegraphics[width=1\textwidth]{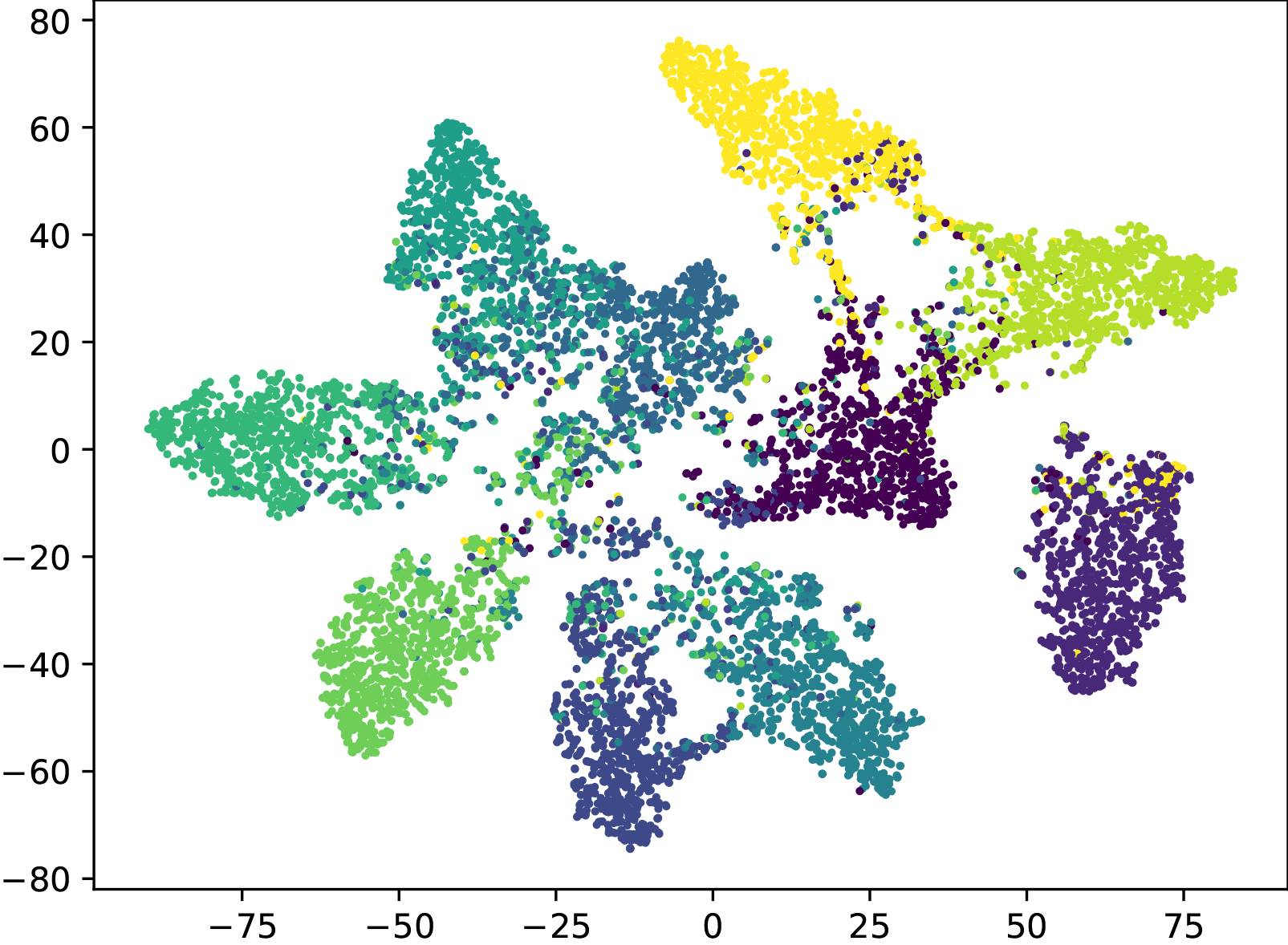}
\caption{LSR ($\xi=0.005$) -- 15.31\%.}
\end{subfigure}%
\hspace{5mm}
\begin{subfigure}{0.45\textwidth}
\includegraphics[width=1\textwidth]{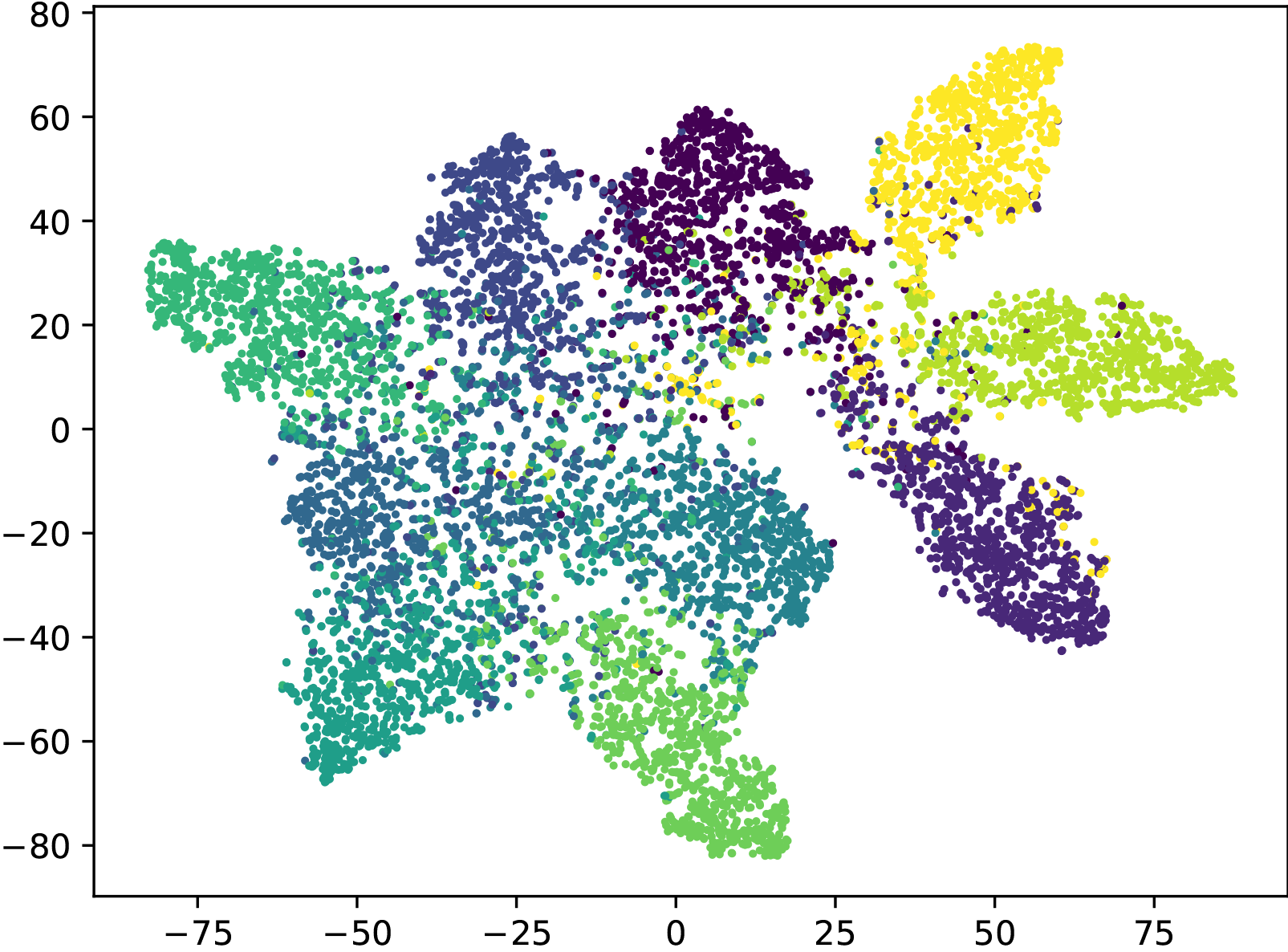}
\caption{SparseLSR ($\xi=0.005$) -- 15.66\%.}
\end{subfigure}%

\captionsetup{justification=centering}
\caption{Visualizing the penultimate layer representations on AlexNet CIFAR-10. Comparing \textit{Label Smoothing Regularization} (LSR) to our proposed \textit{Sparse Label Smoothing Regularization} (SparseLSR).}
\label{fig:fast-label-smoothing-regularization-1}
\end{figure*}

\begin{figure*}
\centering

\ContinuedFloat

\begin{subfigure}{0.45\textwidth}
\includegraphics[width=1\textwidth]{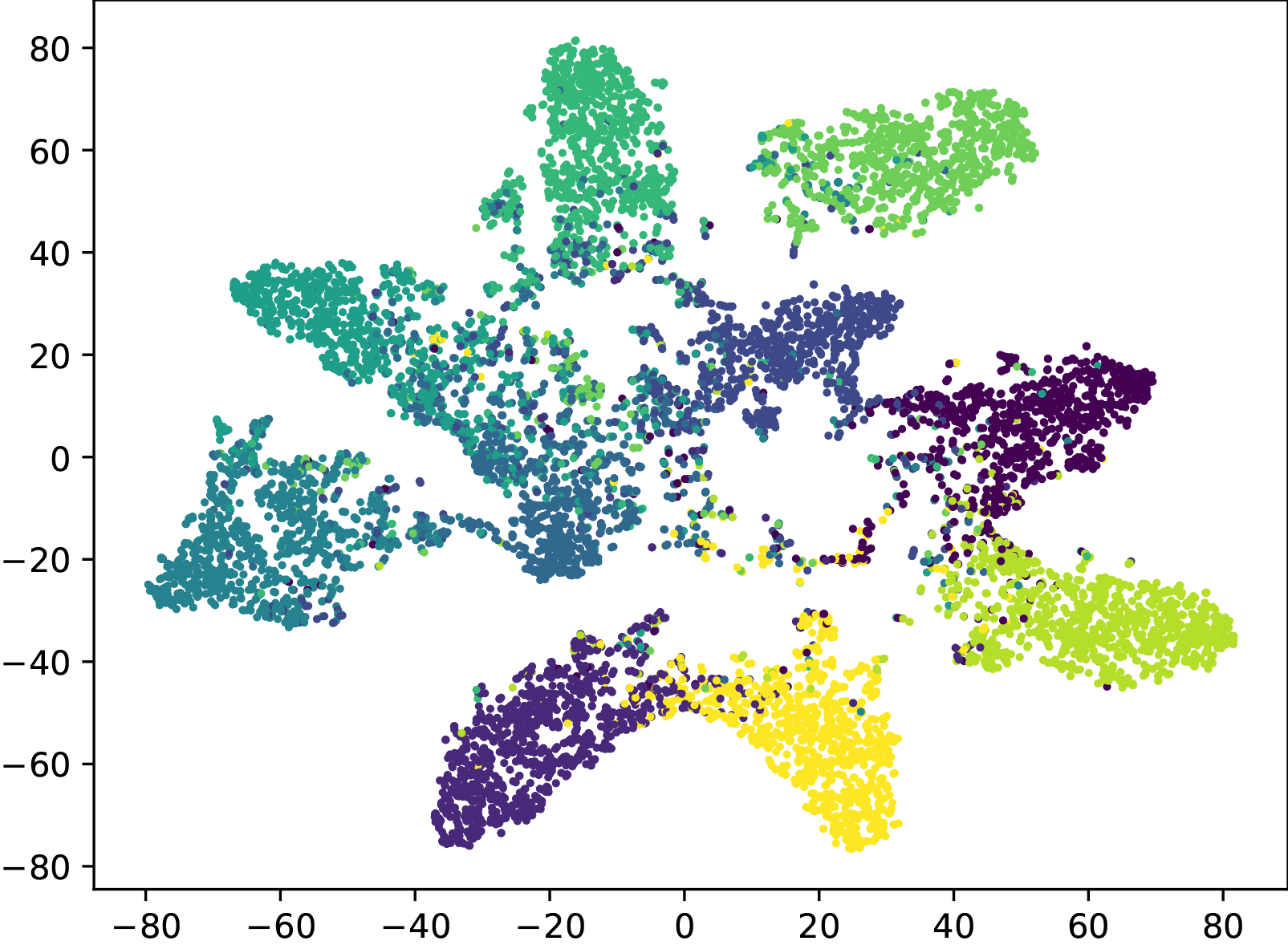}
\caption{LSR ($\xi=0.01$) -- 15.77\%.}
\end{subfigure}%
\hspace{5mm}
\begin{subfigure}{0.45\textwidth}
\includegraphics[width=1\textwidth]{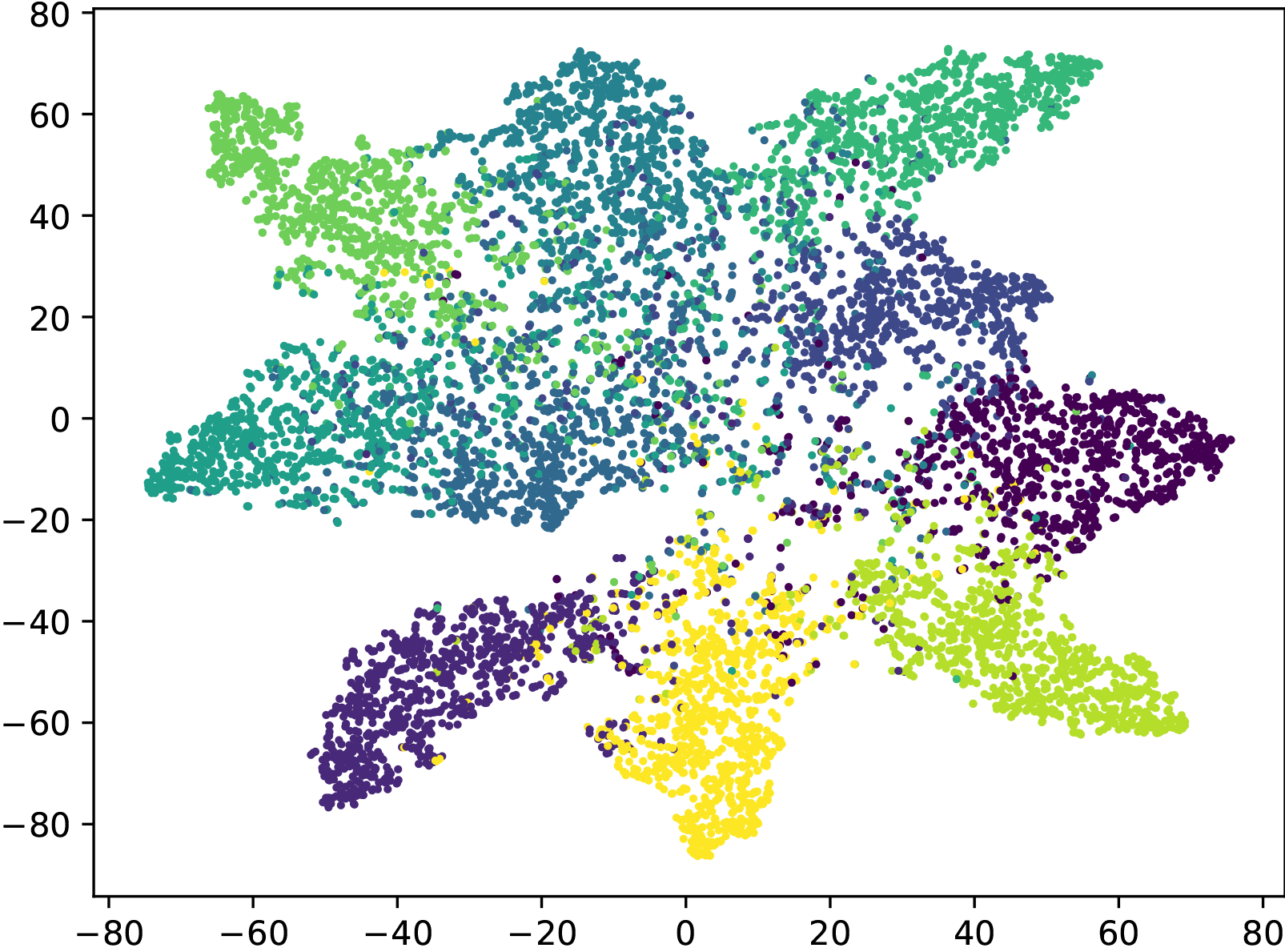}
\caption{SparseLSR ($\xi=0.01$) -- 15.42\%.}
\end{subfigure}%

\par\bigskip

\begin{subfigure}{0.45\textwidth}
\includegraphics[width=1\textwidth]{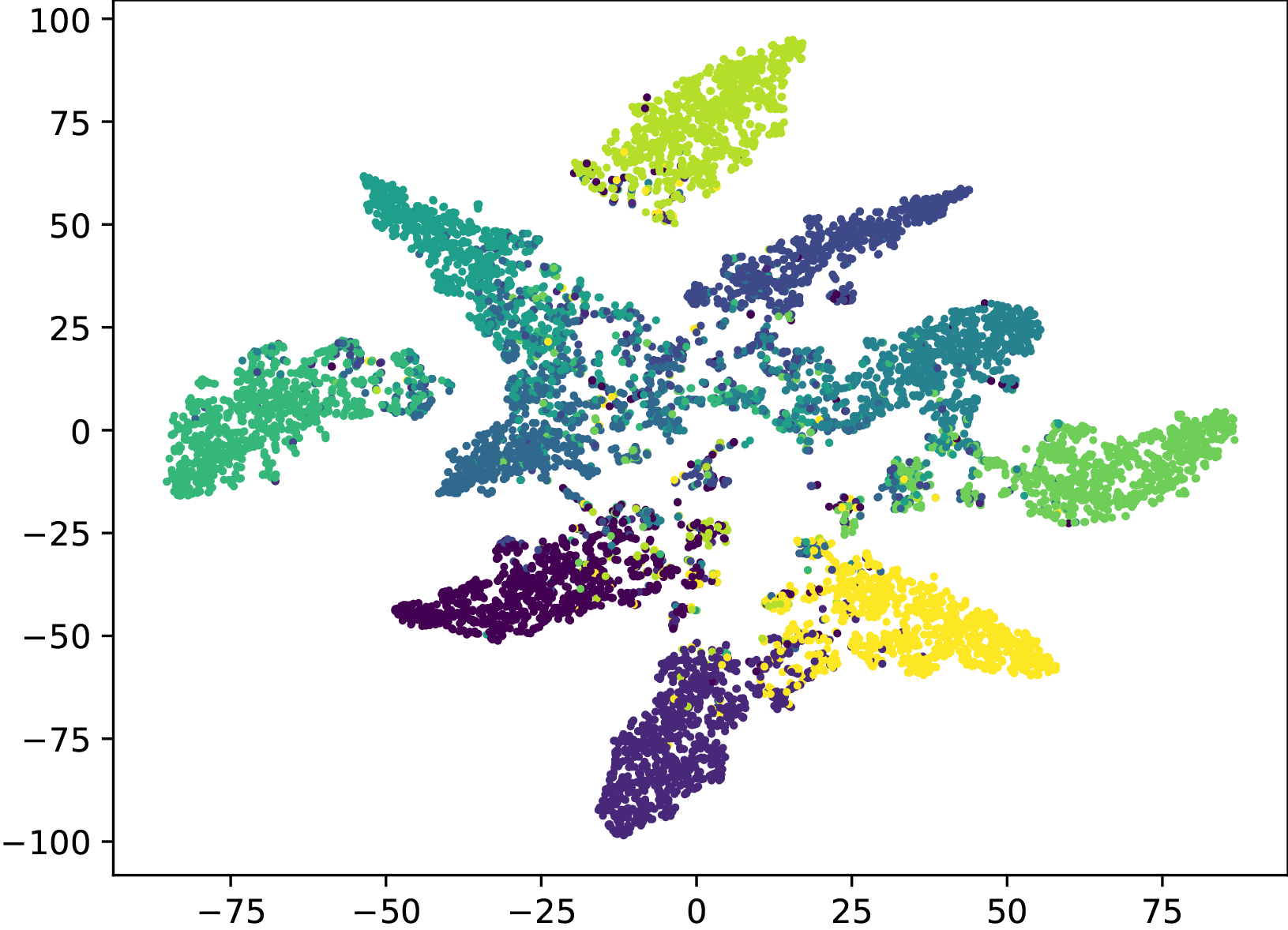}
\caption{LSR ($\xi=0.05$) -- 15.33\%.}
\end{subfigure}%
\hspace{5mm}
\begin{subfigure}{0.45\textwidth}
\includegraphics[width=1\textwidth]{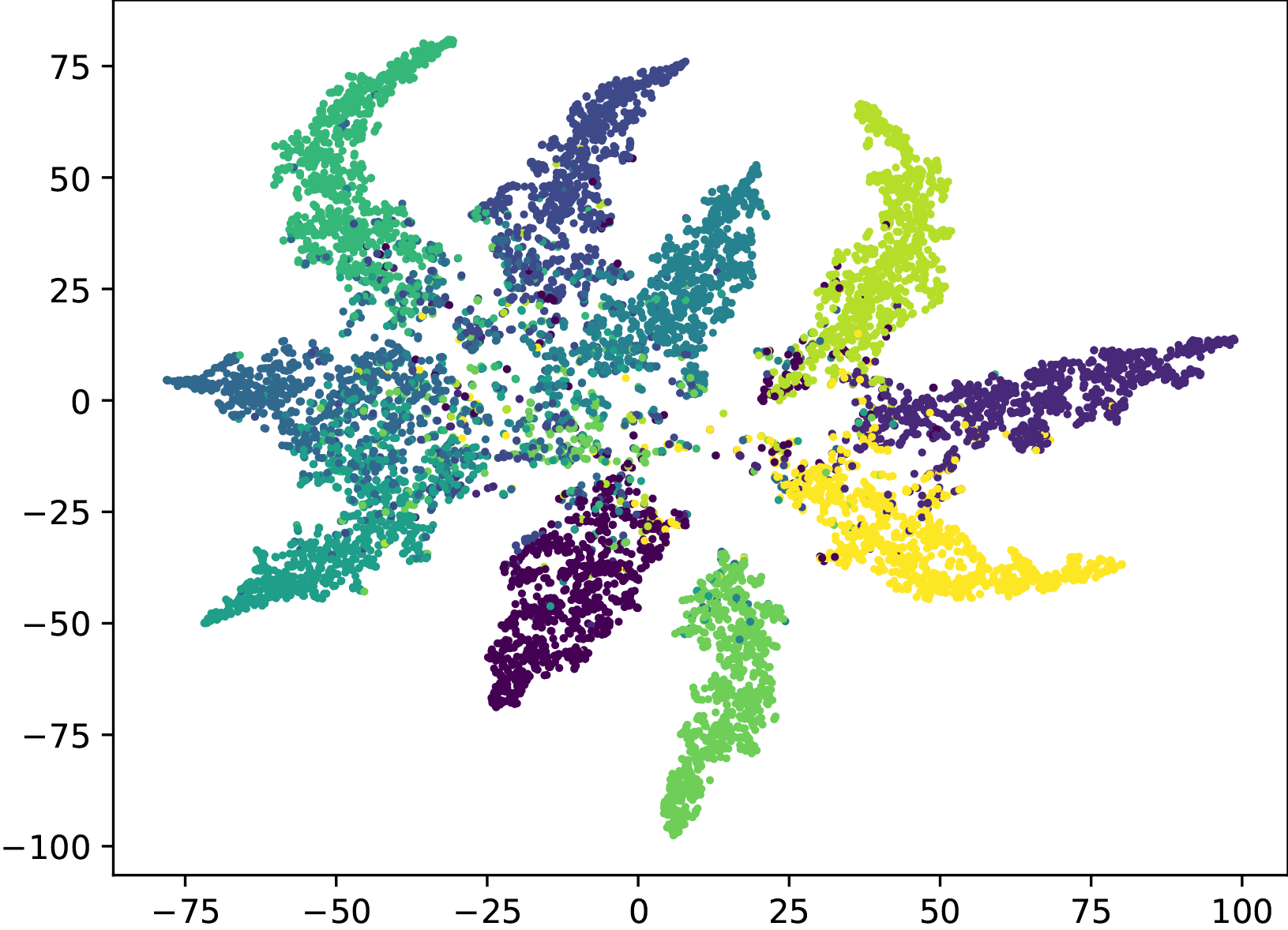}
\caption{SparseLSR ($\xi=0.05$) -- 14.74\%.}
\end{subfigure}%

\par\bigskip

\begin{subfigure}{0.45\textwidth}
\includegraphics[width=1\textwidth]{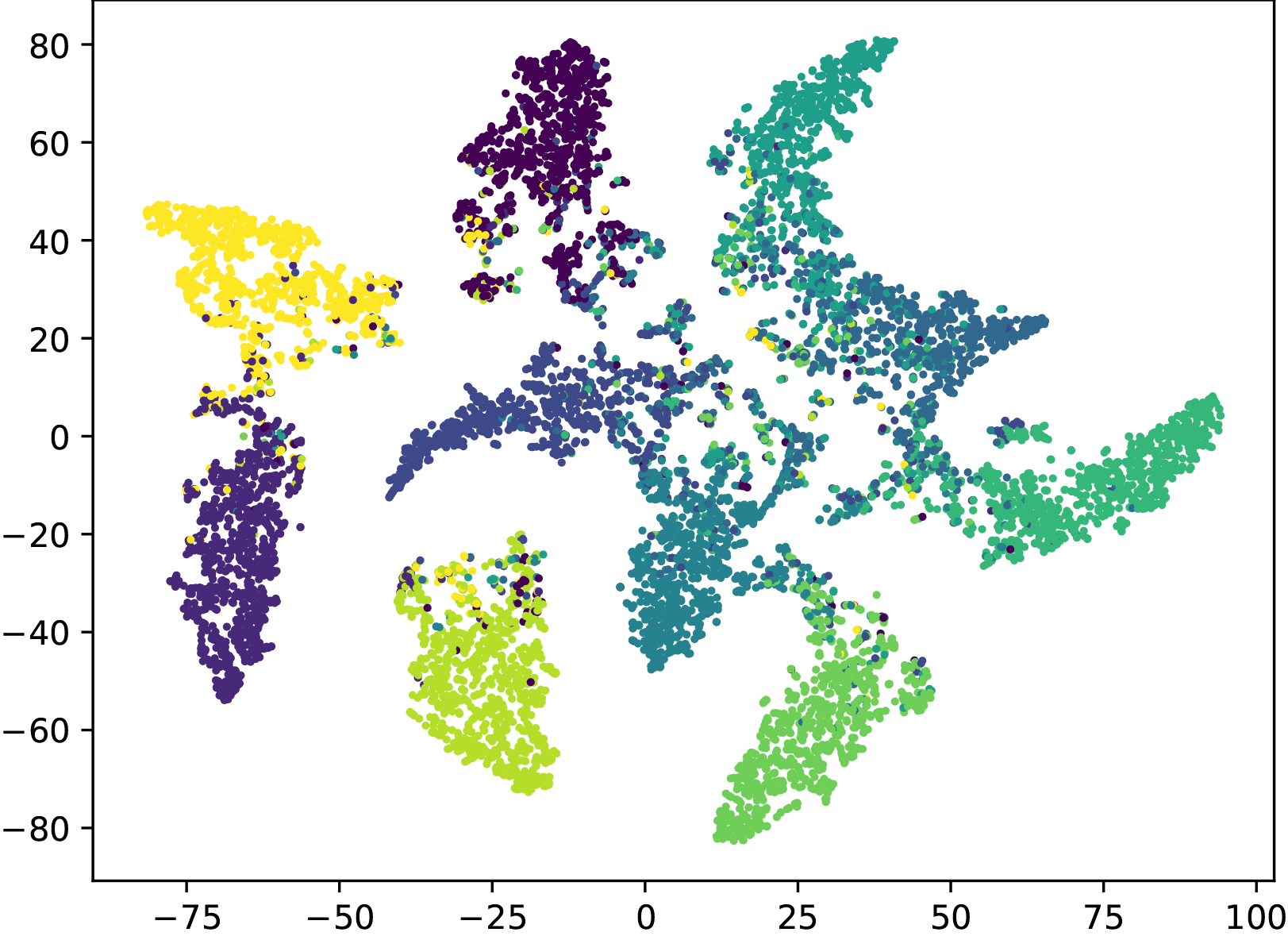}
\caption{LSR ($\xi=0.1$) -- 15.05\%.}
\end{subfigure}%
\hspace{5mm}
\begin{subfigure}{0.45\textwidth}
\includegraphics[width=1\textwidth]{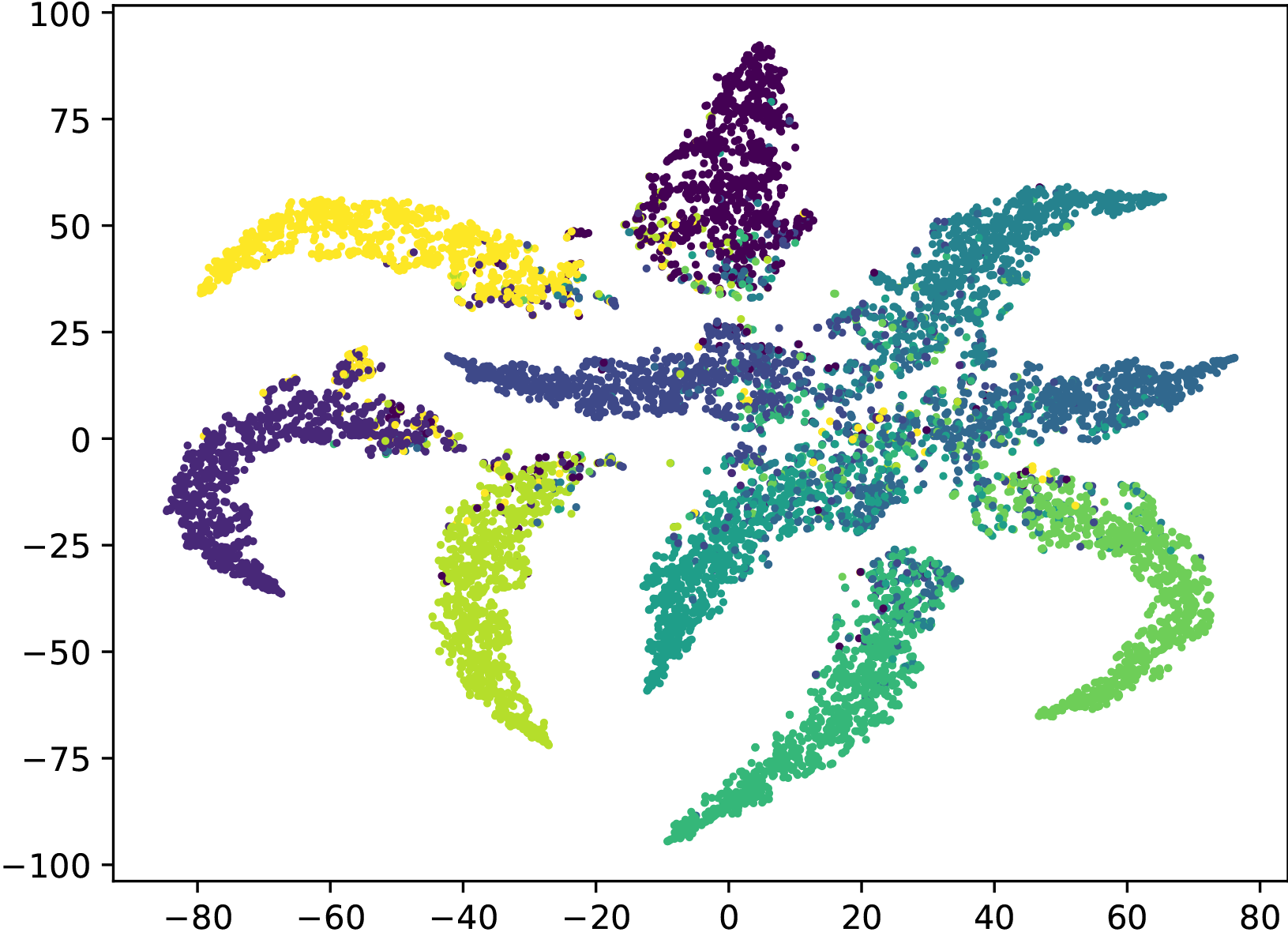}
\caption{SparseLSR ($\xi=0.1$) -- 14.68\%.}
\end{subfigure}%

\captionsetup{justification=centering}
\caption{Visualizing the penultimate layer representations on AlexNet CIFAR-10. Comparing \textit{Label Smoothing Regularization} (LSR) to our proposed \textit{Sparse Label Smoothing Regularization} (SparseLSR).}
\label{fig:fast-label-smoothing-regularization-2}
\end{figure*}

\begin{figure*}
\centering

\ContinuedFloat

\begin{subfigure}{0.45\textwidth}
\includegraphics[width=1\textwidth]{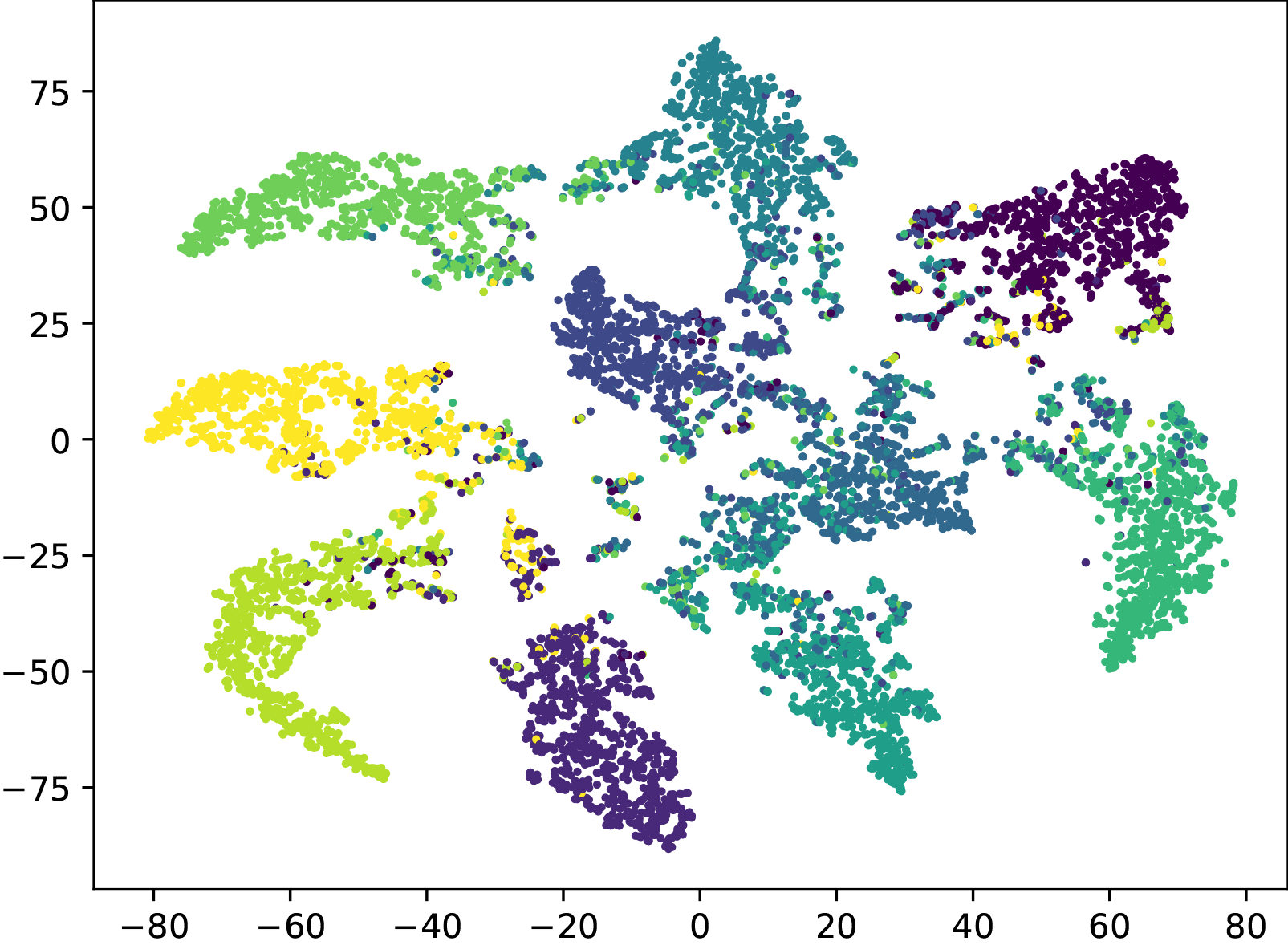}
\caption{LSR ($\xi=0.2$) -- 15.04\%.}
\end{subfigure}%
\hspace{5mm}
\begin{subfigure}{0.45\textwidth}
\includegraphics[width=1\textwidth]{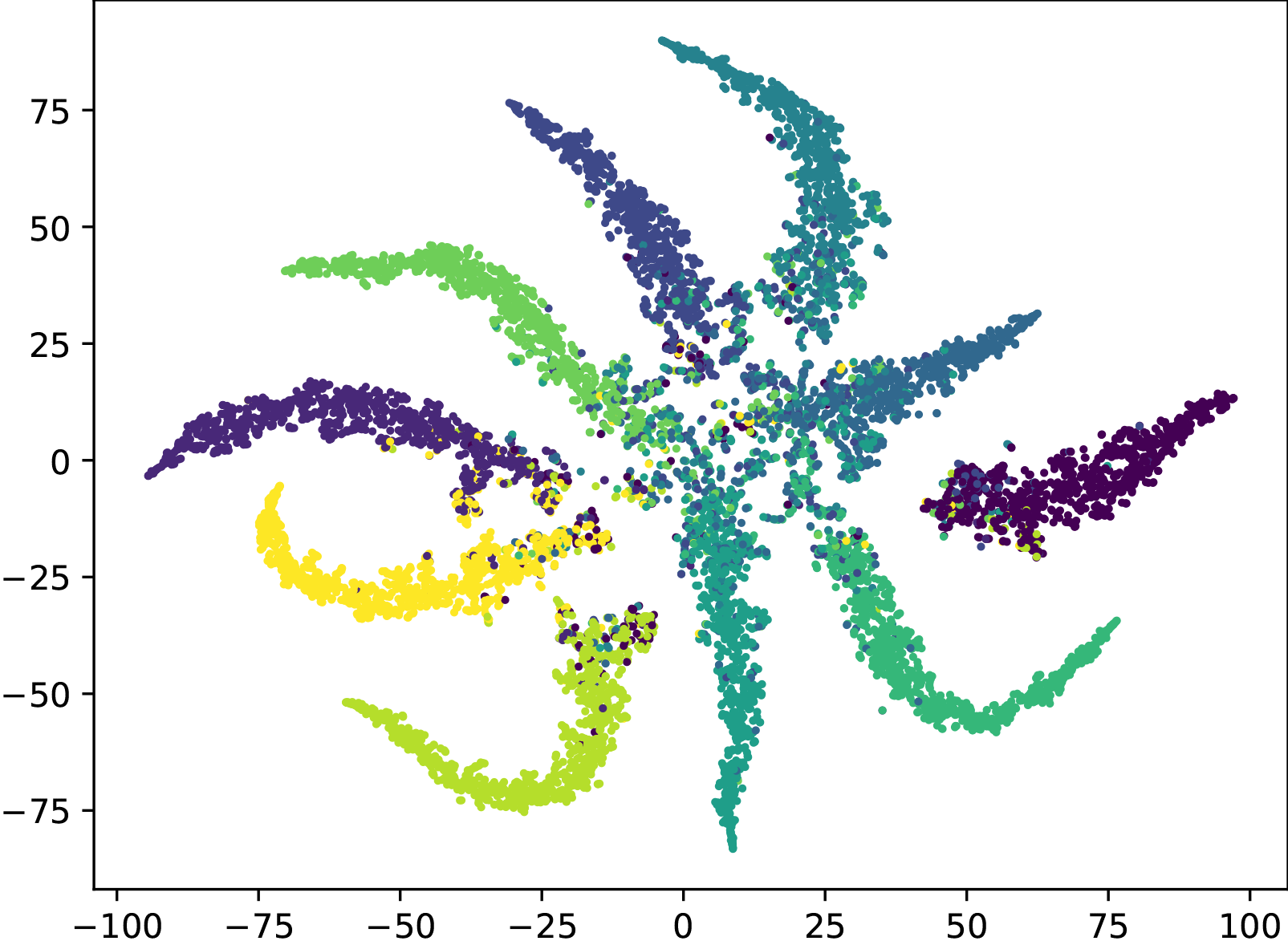}
\caption{SparseLSR ($\xi=0.2$) -- 14.41\%.}
\end{subfigure}%

\par\bigskip

\begin{subfigure}{0.45\textwidth}
\includegraphics[width=1\textwidth]{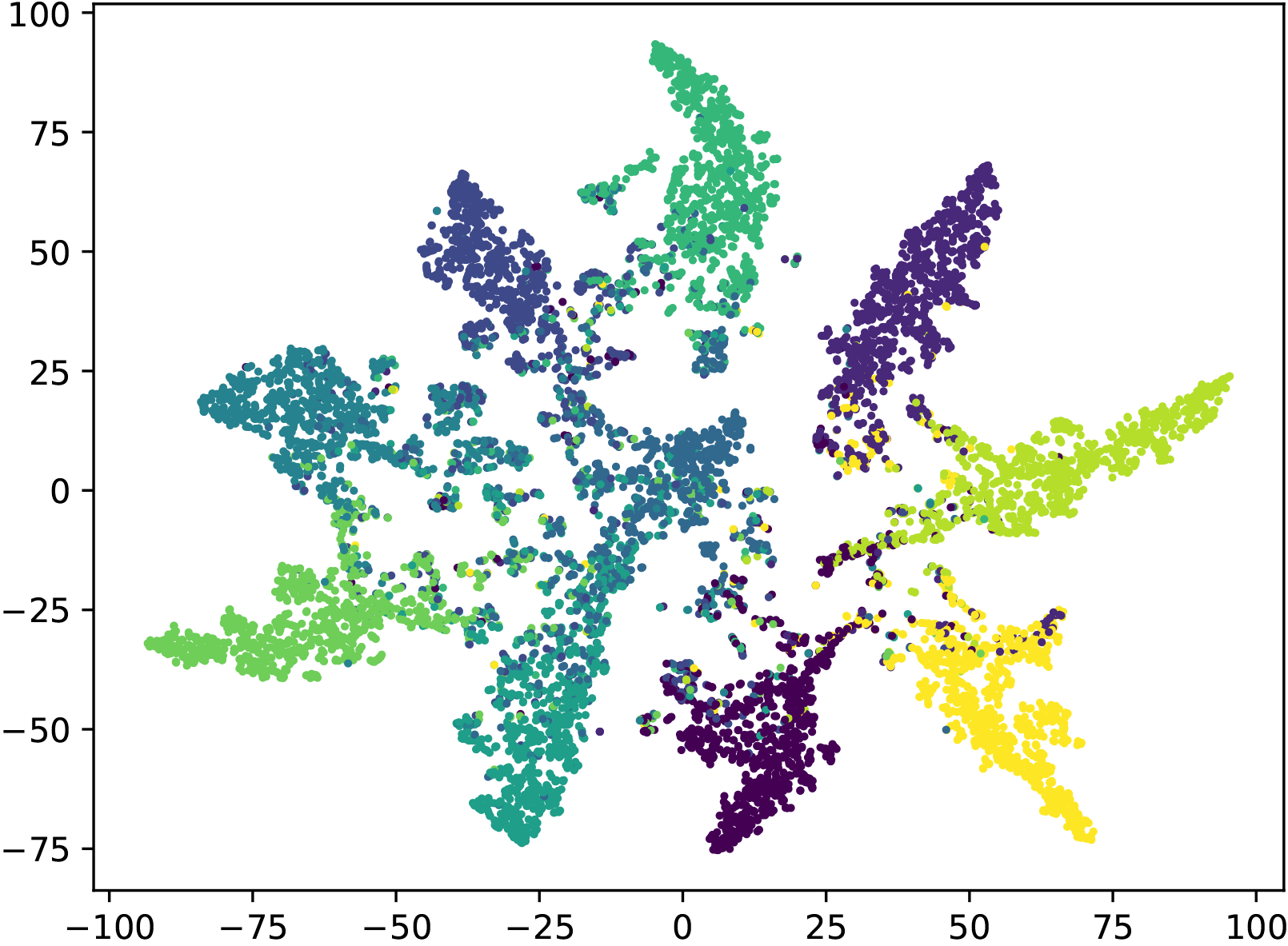}
\caption{LSR ($\xi=0.4$) -- 16.72\%.}
\end{subfigure}%
\hspace{5mm}
\begin{subfigure}{0.45\textwidth}
\includegraphics[width=1\textwidth]{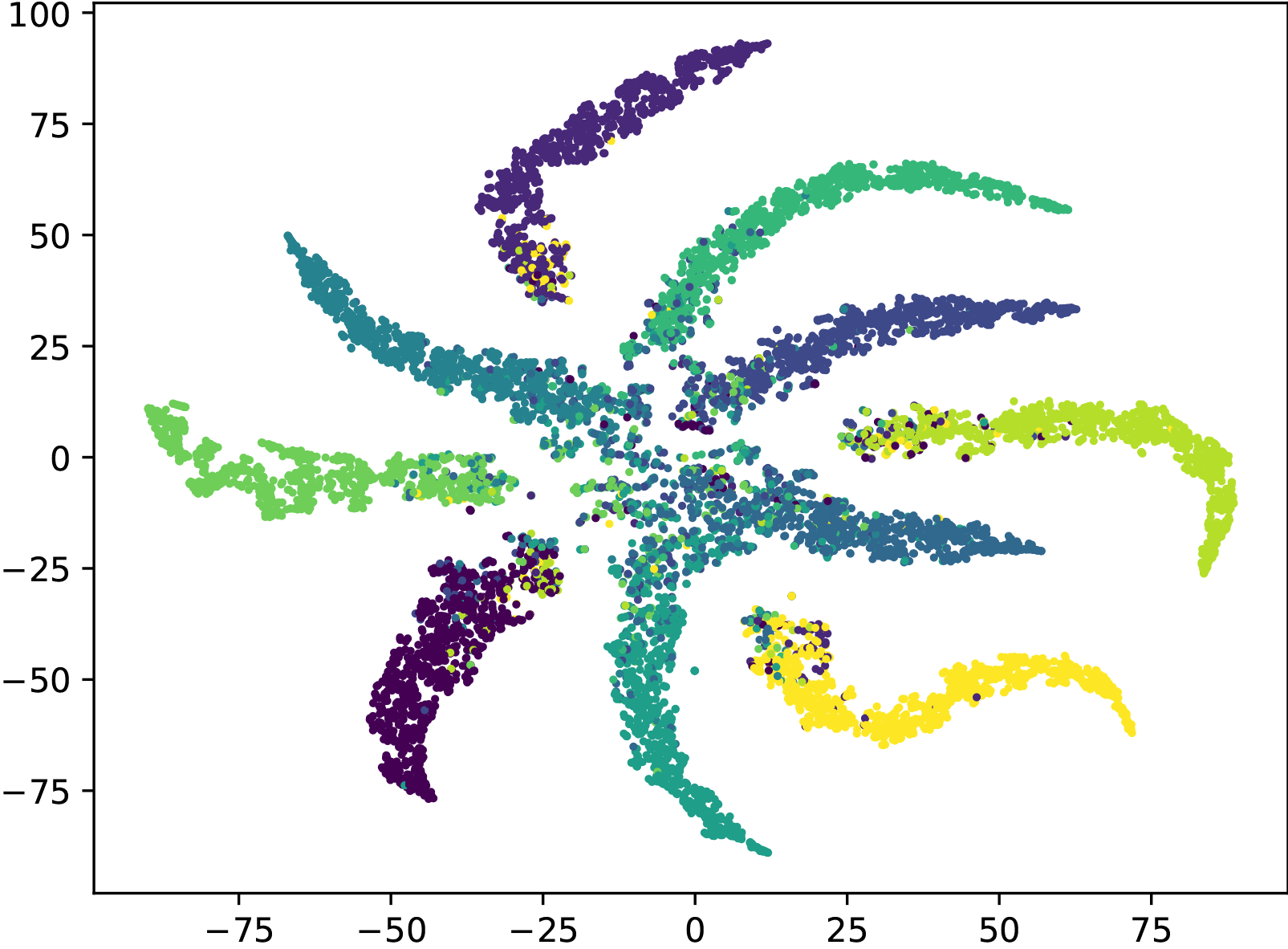}
\caption{SparseLSR ($\xi=0.4$) -- 14.31\%.}
\end{subfigure}%

\captionsetup{justification=centering}
\caption{Visualizing the penultimate layer representations on AlexNet CIFAR-10. Comparing Label Smoothing Regularization (LSR) to our proposed Sparse Label Smoothing Regularization (SparseLSR).}
\label{fig:fast-label-smoothing-regularization-3}
\end{figure*}

\begin{figure}
\centering
\includegraphics[width=0.5\textwidth]{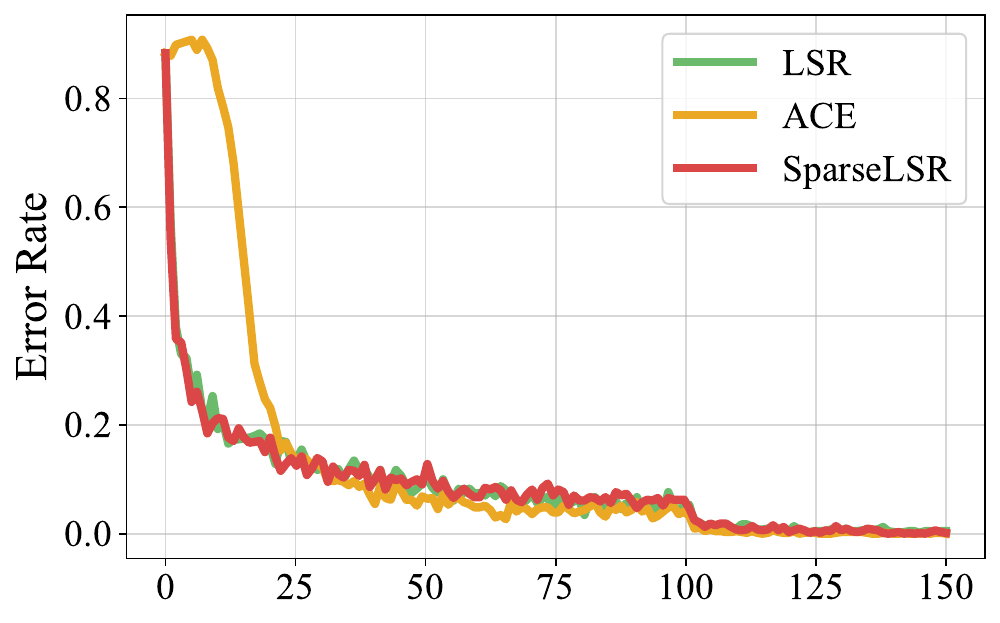}
\captionsetup{justification=centering}
\caption{The mean training curves on AlexNet CIFAR-10 when using Label Smoothing Regularization, Absolute Cross-Entropy Loss, and Sparse Label Smoothing Regularization. The x-axis shows the number of gradient steps in thousands.}
\label{fig:proposed-loss-learning-curves}
\end{figure}

\subsection{Training Dynamics}

Finally, one of the key motivations for designing sparse label smoothing regularization was to address the non-smooth behavior of the absolute cross-entropy loss when transitioning between minimizing and maximizing the target output, which causes poor training dynamics. To demonstrate the improved training dynamics of sparse label smoothing regularization we visualize the training learning curves in Figure \ref{fig:proposed-loss-learning-curves}, when using the different loss functions. The results show that the absolute cross-entropy loss often does not converge early in the training (likely due to it being non-smooth). In contrast, both sparse and non-sparse label smoothing regularization demonstrate smooth and rapid convergence, with their learning curves closely overlapping. These findings suggest that the linear approximation utilized in Equation \eqref{eq:linear-approximation-of-log} for sparse label smoothing does not adversely affect the learning behavior.

\subsection{Extensions}

\begin{figure}[t!]
\centering
\includegraphics[width=1\textwidth]{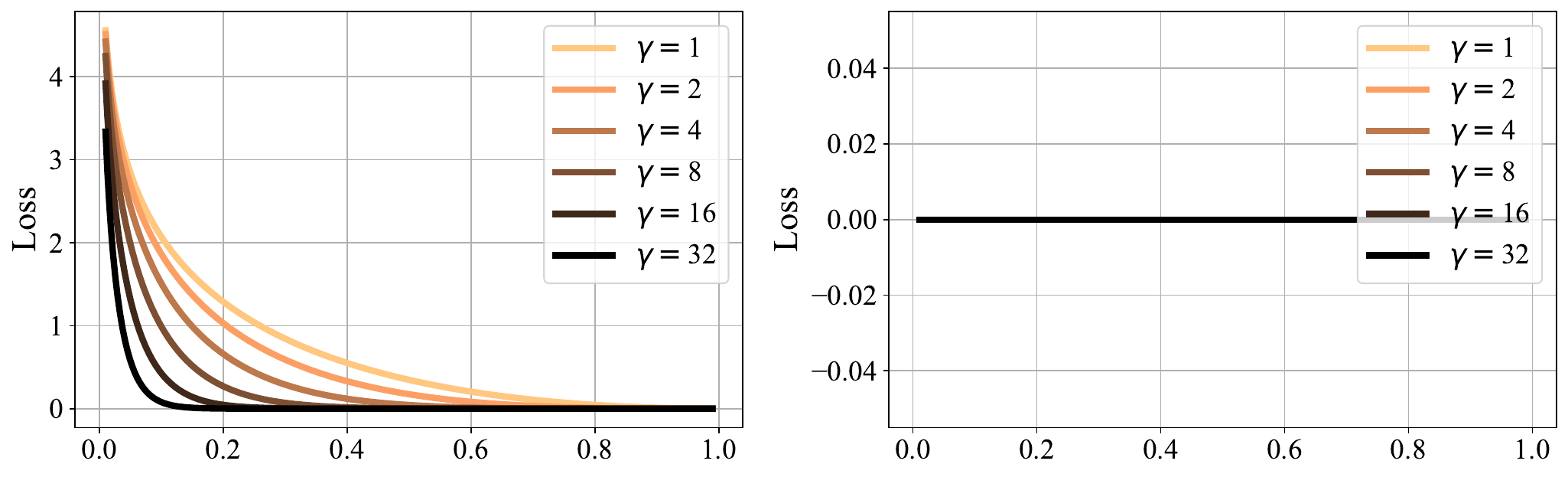}
\captionsetup{justification=centering}
\caption{Visualizing the \textit{Focal Loss} with varying focusing values $\gamma$, where the left figure shows the target loss (\textit{i.e.}, $y_i = 1$), and the right figure shows the non-target loss (\textit{i.e.}, $y_i = 0$).}
\label{fig:focal_loss}
\end{figure}

\begin{figure}[t!]
\centering
\includegraphics[width=1\textwidth]{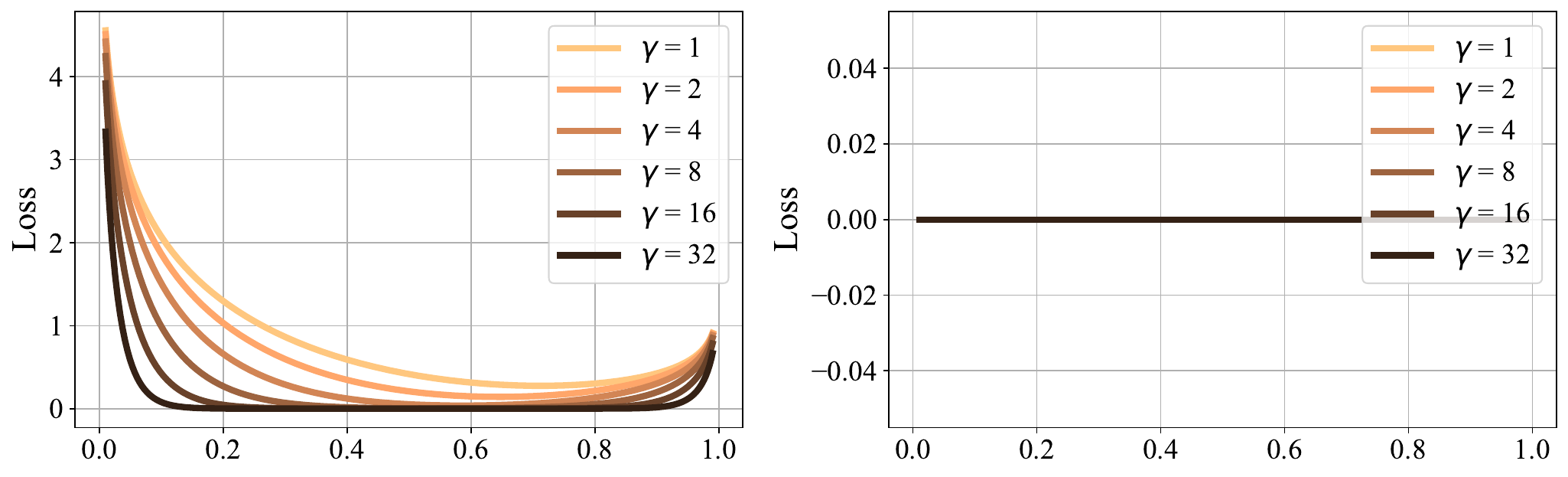}
\captionsetup{justification=centering}
\caption{Visualizing the proposed \textit{Focal Loss with Sparse Label Smoothing Regularization} (Focal+SparseLSR) with varying focusing values $\gamma$ and a smoothing coefficient value $\xi=0.2$, where the left figure shows the target loss (\textit{i.e.}, $y_i = 1$), and the right figure shows the non-target loss (\textit{i.e.}, $y_i = 0$).}
\label{fig:focal_sparse_label_smoothing_loss}
\vspace{2mm}
\end{figure}

The redistributed loss trick described in Section \ref{sec:sparse-label-smoothing-regularization} is general and can be applied straightforwardly to any softmax-based classification loss function. To illustrate this, we demonstrate how the Focal Loss, a popular classification loss function, can be extended to integrate sparse label smoothing regularization. The Focal Loss, introduced by \cite{lin2017focal}, extends the standard cross-entropy loss by incorporating a term $(1 - f_{\theta}(x)_i)^{\gamma}$, where $\gamma \geq 0$ is the ``focusing'' hyperparameter. The focusing term reduces the loss for well-classified examples, thereby emphasizing harder misclassified examples during training, as shown in Figure \ref{fig:focal_loss}. 
\begin{equation}
    \Loss^{Focal} = - \sum_{i=1}^{\mathcal{C}} y_{i} \bigg[(1 - f_{\theta}(x)_{i} )^{\gamma} \cdot \log\left(f_{\theta}(x)_{i}\right)\bigg]
\end{equation}
The focal loss can be combined with sparse label smoothing regularization by incorporating the terms $(1 - f(x)_i)^{\gamma}$ and its shifted reflection $(f(x)_i)^{\gamma}$ into Equation \ref{eq:sparse-label-smoothing-regularization} as follows:
\begin{multline}
    \Loss^{Focal+SparseLSR} = - \sum_{i=1}^{\mathcal{C}} y_{i}\bigg[\bigg(1 - f_{\theta}(x)_i \bigg)^{\gamma} \cdot \left(1 - \xi + \frac{\xi}{\mathcal{C}}\right) \cdot \log(f_{\theta}(x)_{i}) \\ + \bigg(f_{\theta}(x)_i \bigg)^{\gamma} \cdot \frac{\xi(\mathcal{C}-1)}{\mathcal{C}} \cdot \log\left(\frac{1 - f_{\theta}(x)_i}{\mathcal{C}-1}\right)\bigg] 
\end{multline}
As illustrated in Figure \ref{fig:focal_sparse_label_smoothing_loss}, the \textit{Focal Loss with Sparse Label Smoothing Regularization} (Focal+SparseLSR) can capture two key behaviors commonly seen in meta-learned loss functions. First, it can increase and decrease the relative loss assigned to well-classified versus poorly-classified examples. Second, it can penalize overly confident predictions through sparse label smoothing regularization. Importantly, $\Loss^{Focal+SparseLSR}$ can achieve both of these behaviors while remaining sparse, thereby benefiting from the computational and memory advantages outlined in Section \ref{sec:time-and-space-complexity}.

\section{Chapter Summary}

In conclusion, this chapter performed an empirical and theoretical analysis of some of the meta-learned loss functions from Chapter \ref{chapter:evomal}. Our analysis began with a brief overview of some of the meta-learned loss functions learned by EvoMAL, followed by a critical assessment of some of the previous hypotheses for “why meta-learned loss functions perform better than handcrafted loss functions” and “what are meta-learned loss functions learning”. In a quest to answer these questions, we performed a theoretical analysis of the \textit{Absolute Cross-Entropy} (ACE) Loss, one of the learned loss functions from EvoMAL, and the analysis revealed that it has identical behavior at the start and very end of training to Label Smoothing Regularization (LSR), a popular regularization technique for penalizing overconfident predictions, which explains why it was able to improve performance.

Although effective, the absolute cross-entropy loss function is not smooth and has divergent behavior to label smoothing regularization at intermediate points in the training. To resolve these issues a new novel loss function was proposed inspired by our theoretical findings called \textit{Sparse Label Smoothing Regularization} (SparseLSR). The proposed loss function is similar to non-sparse label smoothing regularization; however, it utilizes the redistributed loss trick which takes the expected non-target loss and redistributes it into the target loss, obviating the need to calculate the loss on the non-target outputs. Our experimental results show that the proposed loss function is significantly faster to compute, with the time and space complexity being reduced from linear to constant time with respect to the number of classes being considered. The \texttt{PyTorch} code for the proposed sparse label smoothing loss can be found at: \href{https://github.com/Decadz/Sparse-Label-Smoothing-Regularization}{https://github.com/Decadz/Sparse-Label-Smoothing-Regularization}
\newcommand{\Regularization}{\mathcal{R}}

\chapter{Meta-Learning Adaptive Loss Functions}\label{chapter:adalfl}

\textit{In this chapter, we challenge the conventional notion that a loss function must be a static function. More specifically, we address the limitation that prior loss function learning techniques meta-learn static loss functions in an offline fashion --- where the meta-objective only considers the very early stages of training. This induces a short-horizon bias on the discovery and selection of the loss function causing significant bias towards loss functions that perform well at the start of training but perform poorly at the end of training. To resolve this limitation we propose a new technique called Adaptive Loss Function Learning (AdaLFL), for adaptively updating the loss function online in lockstep with the base model weights. The experimental results show that our proposed method has promising performance relative to handcrafted loss functions and offline loss function learning techniques.}

\section{Chapter Overview}

\begin{figure}[t!]

    \centering
    \begin{subfigure}{0.5\textwidth}
        \centering
        \includegraphics[width=1\textwidth]{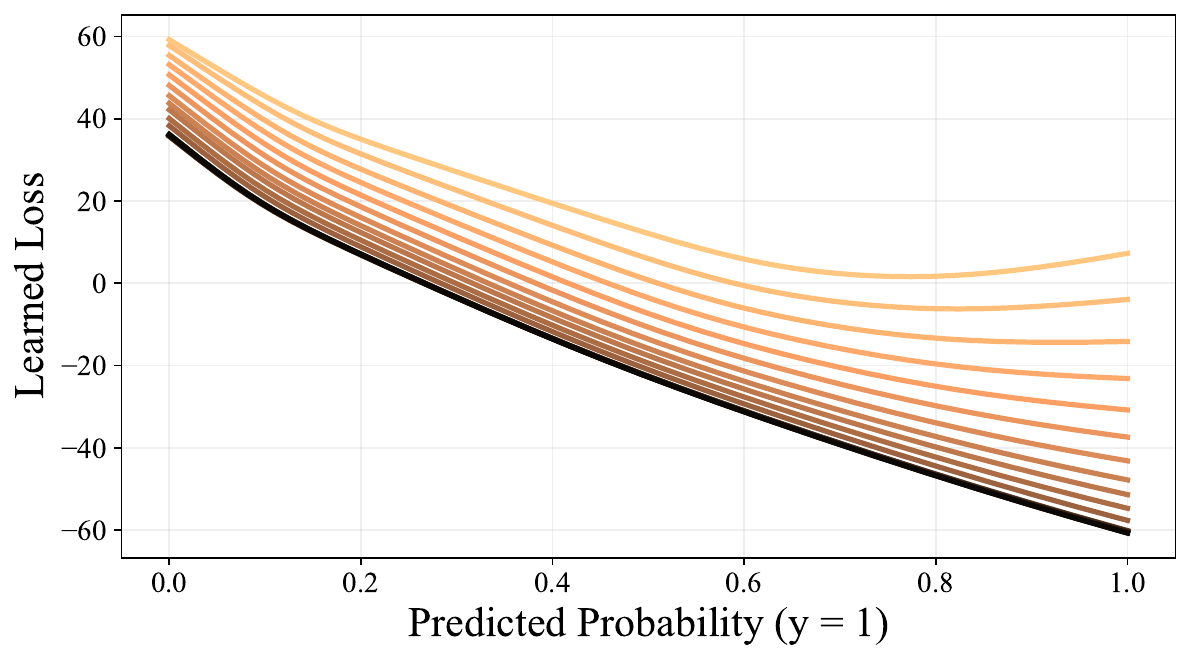}
    \end{subfigure}%
    \hfill
    \begin{subfigure}{0.5\textwidth}
        \centering
        \includegraphics[width=1\textwidth]{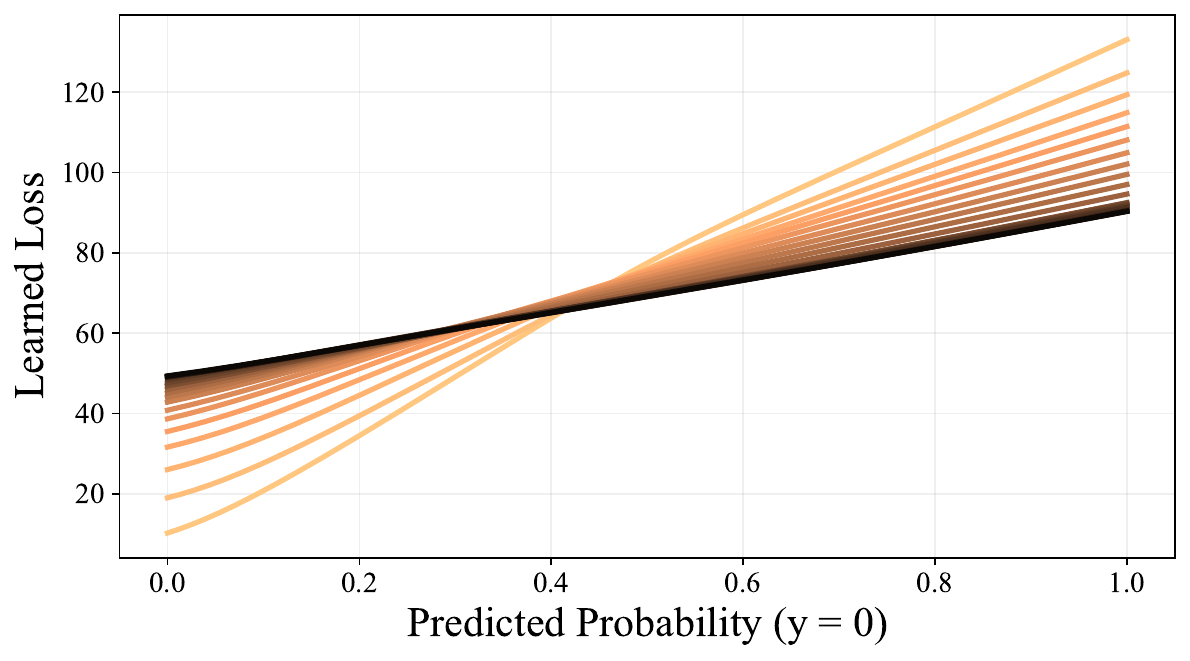}
    \end{subfigure}%

    \vspace{2mm}
    
    \begin{subfigure}{0.5\textwidth}
        \centering
        \includegraphics[width=1\textwidth]{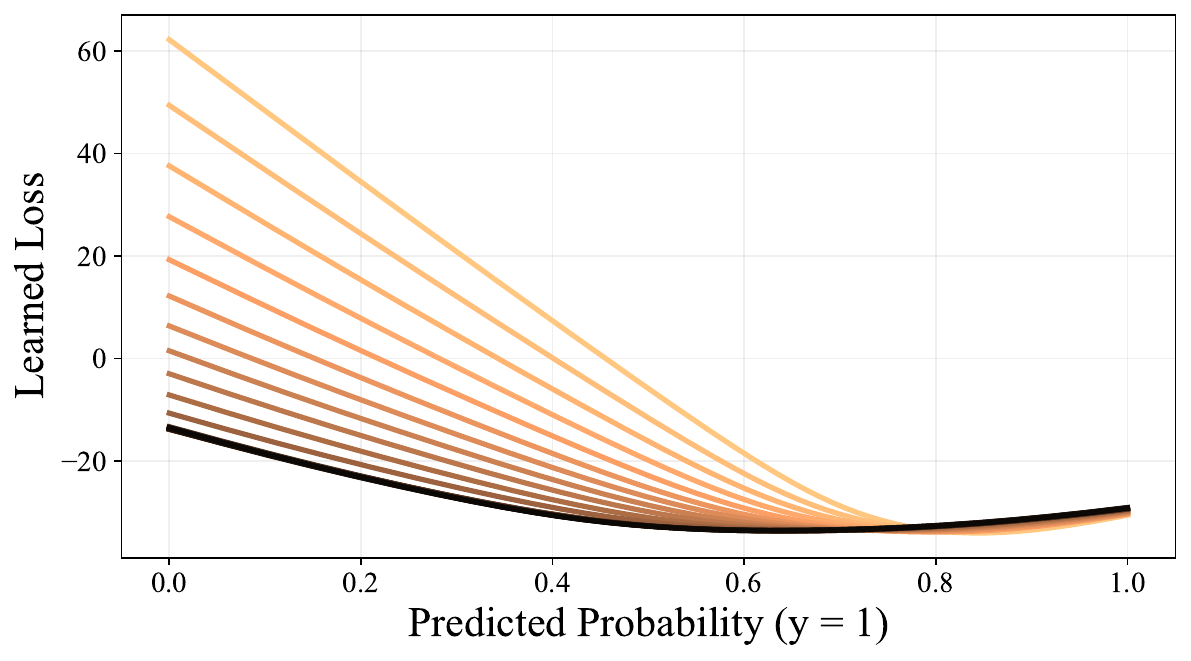}
    \end{subfigure}%
    \hfill
    \begin{subfigure}{0.5\textwidth}
        \centering
        \includegraphics[width=1\textwidth]{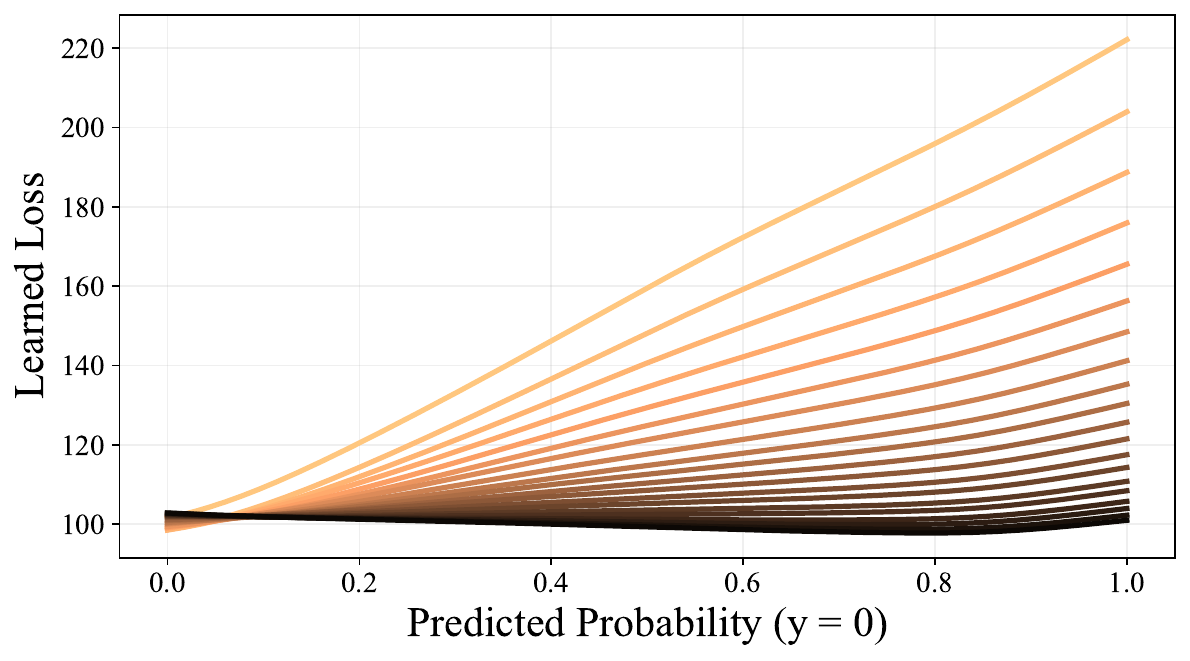}
    \end{subfigure}%
    
    \begin{subfigure}{0.5\textwidth}
        \centering
        \includegraphics[width=1\textwidth]{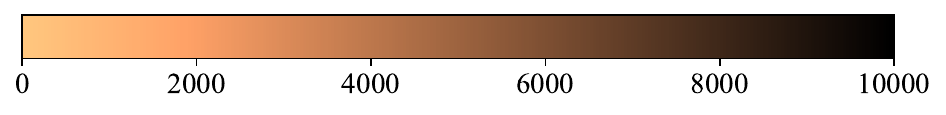}
    \end{subfigure}%

    \vspace{-2mm}

\captionsetup{justification=centering}
\caption{Example adaptive meta-learned loss functions generated by AdaLFL on the CIFAR-10 dataset, where each row represents a classification loss function, and the color represents the current gradient step.}
\label{fig:example-loss-function}

\end{figure}

Loss function learning aims to meta-learn a task-specific loss function, which yields improved performance when utilized in training compared to handcrafted loss functions. Initial approaches to loss function learning have shown promise at enhancing various aspects of deep neural network training, such as improving the convergence and sample efficiency \citep{gonzalez2020improved, bechtle2021meta}, as well as the generalization \citep{gonzalez2021optimizing, liu2021loss, li2022autoloss, leng2022polyloss}, and model robustness \citep{gao2021searching, gao2022loss}. However, \textit{one prevailing limitation} of the existing approaches to loss function learning is that they have thus far exclusively focused on learning a loss function in the offline meta-learning settings. 

In offline loss function learning, training is prototypically partitioned into two phases. In the first phase, the loss function is meta-learned by iteratively updating the learned loss function, where each update corresponds to taking one or a few training steps with the base model from a random initialization and computing the loss at the end of the training trajectory. Second, the base model is trained to completion using the learned loss function, which is now \textit{fixed} and is used in place of the conventional handcrafted loss function. Unfortunately, this methodology is prone to a severe short-horizon bias \citep{wu2018understanding} towards loss functions which are performant in the early stages of training but often have poor performance in the later stages.

To address the limitation of offline loss function learning, we propose a new technique for \textit{online} loss function learning called \textit{Adaptive Loss Function Learning} (AdaLFL). In the proposed technique, the learned loss function is represented as a small feed-forward neural network trained simultaneously with the base learning model. Unlike prior methods, AdaLFL can adaptively transform both the shape and scale of the loss function throughout the learning process to adapt to what is required at each stage of the learning process, as shown in Figure \ref{fig:example-loss-function}. In offline loss function learning, the central goal is to improve the performance of a model by specializing the loss function to a small set of related tasks. Online loss function learning naturally extends this general philosophy by specializing the loss function to each gradient step of a single task. 

\subsection{Contributions}

The key contributions of this chapter are as follows:

\begin{itemize}

    \item We propose a method that can effectively learn adaptive loss functions via online meta-learning by utilizing online unrolled differentiation to update the meta-learned loss function after each update to the base model.

    \item We address shortcomings in the design of neural network-based loss function parameterizations, which previously caused learned loss functions to be biased toward overly flat shapes resulting in poor training dynamics.

    \item Empirically, we demonstrate that models trained with our method have enhanced convergence capabilities and inference performance compared to handcrafted loss functions and offline loss function learning methods.

    \item Finally, we analyze the meta-learned loss functions, highlighting several key trends to explore why our adaptive meta-learned loss functions outperform traditional handcrafted loss functions.

\end{itemize}

\section{Adaptive Loss Function Learning}
\label{section:method}

In this chapter, we aim to automate the design and selection of the loss function and improve upon the performance of supervised machine learning systems. This is achieved via meta-learning an adaptive loss function that transforms both its shape and scale throughout the learning process. To achieve this, we propose \textit{Adaptive Loss Function Learning} (AdaLFL), an efficient task and model-agnostic approach for online adaptation of the base loss function. In what follows, we provide a brief problem setup in Section \ref{sec:adalfl-problem-setup}, followed by an overview of our learned loss function representation in Sections \ref{sec:adalfl-loss-function-representation} and \ref{sec:adalfl-smooth-leaky-relu}. Subsequently, in Sections \ref{sec:adalfl-offline-optimization} and \ref{sec:adalfl-online-optimization}, we discuss the initialization and meta-optimization of our adaptive loss functions, respectively.

\subsection{Problem Setup}
\label{sec:adalfl-problem-setup}

In a prototypical supervised learning setup, we are given a set of $N$ independently and identically distributed (i.i.d.) examples of form $\Dataset = \{(x_{1}, y_{1}), \dots, (x_{N}, y_{N})\}$, where $x_i \in X$ is the $i$th instance's feature vector and $y_i \in Y$ is its corresponding class label. We want to learn a mapping between $X$ and $Y$ using some base learning model, \textit{e.g.}, a classifier or regressor, $f_{\theta}: \mathcal{X} \rightarrow \mathcal{Y}$, where $\theta$ is the base model parameters. As done in the previous chapter, we constrain the selection of the base models to those amenable to a stochastic gradient descent (SGD) style training procedures such that optimization of model parameters $\theta$ occurs via optimizing some task-specific loss function $\Loss^{meta}$ as follows:
\begin{equation}
\theta_{t+1} = \theta_{t} - \alpha \nabla_{\theta_{t}} \Loss^{meta}(y, f_{\theta_{t}}(x))
\label{eq:setup}
\end{equation}
where $\Loss^{meta}$ is a handcrafted loss function, typically the cross entropy between the predicted label and the ground truth label in classification or the squared error in regression. The principal goal of AdaLFL is to replace this conventional handcrafted loss function $\Loss^{meta}$ with a meta-learned adaptive loss function $\MetaLoss_{\phi}$, where the meta-parameters $\phi$ are learned simultaneously with the base parameters $\theta$, allowing for online adaptation of the loss function. We formulate the task of learning $\phi$ and $\theta$ as a non-stationary bilevel optimization problem, where $t$ is the current time step
\small
\begin{align}
\begin{split}
    \phi_{t+1} &= \argminA_{\phi} \left[\Loss^{meta}(y, f_{\theta_{t+1}}(x))\right] \\ 
    s.t.\;\;\;\;\; \theta_{t+1}(\phi_{t})  &= \argminA_{\theta} \left[\MetaLoss_{\phi_{t}}(y, f_{\theta_{t}}(x))\right].
\end{split}
\end{align}
\normalsize
The outer optimization aims to meta-learn a performant loss function $\MetaLoss_{\phi}$ that minimizes the error on the given task. The inner optimization directly minimizes the learned loss value produced by $\MetaLoss_{\phi}$ to learn the base model parameters $\theta$.

\subsection{Loss Function Representation}
\label{sec:adalfl-loss-function-representation}

\begin{figure}[t!]

    \centering
    \begin{subfigure}{0.48\textwidth}
        \centering
        \includegraphics[width=1\textwidth]{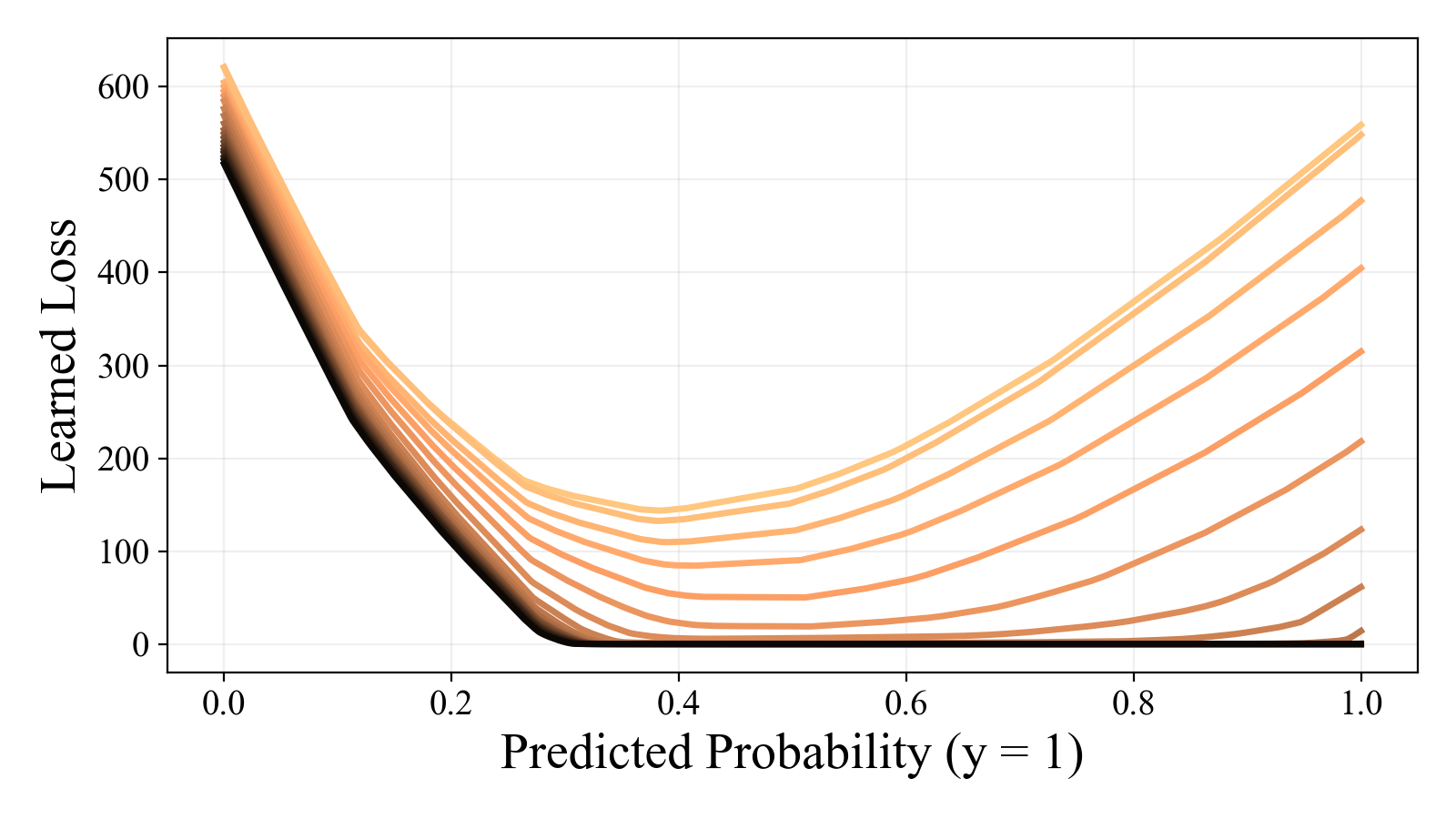}
    \end{subfigure}%
    \hspace{2mm}
    \begin{subfigure}{0.48\textwidth}
        \centering
        \includegraphics[width=1\textwidth]{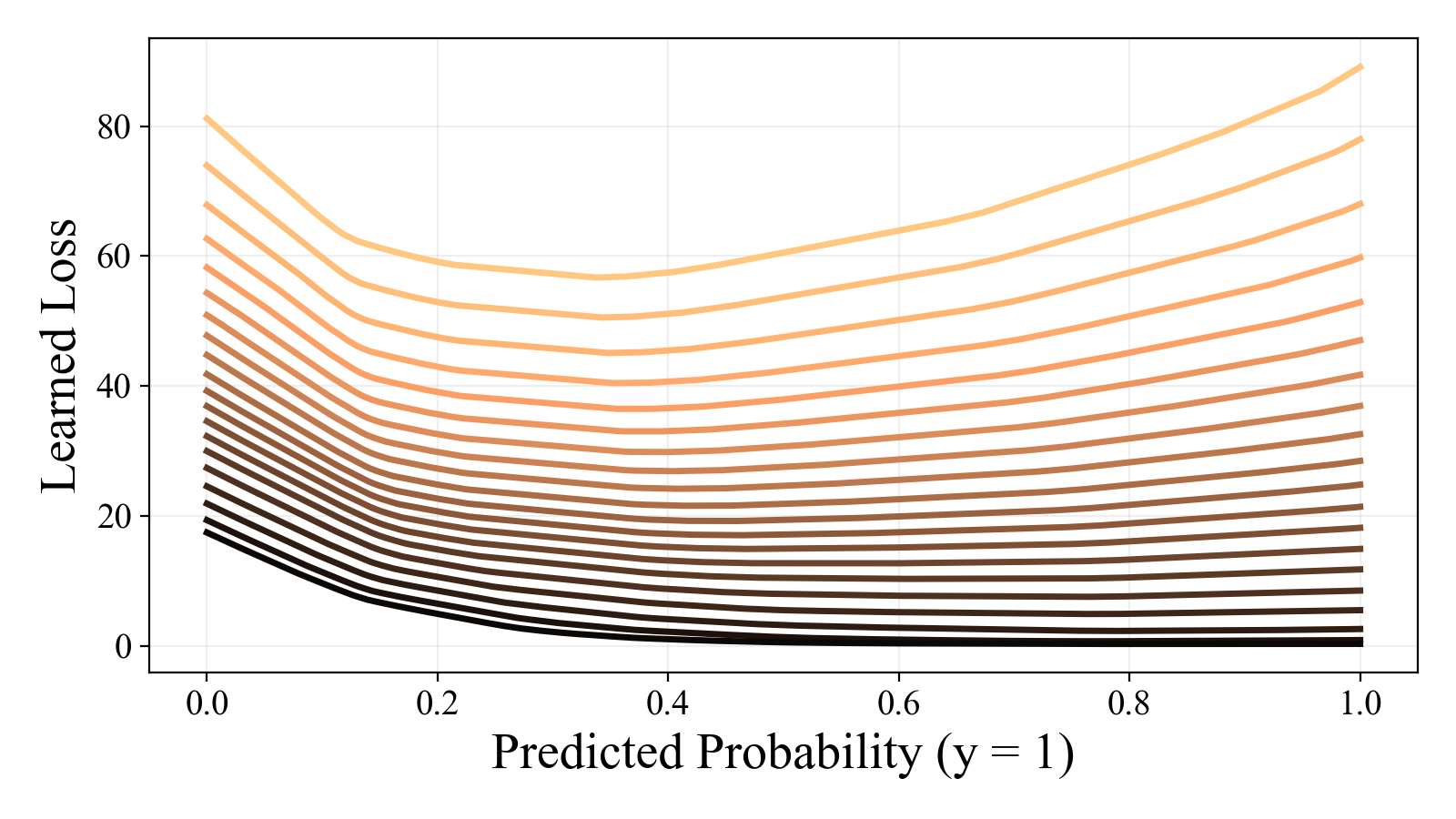}
    \end{subfigure}%

    \vspace{3mm}

    \begin{subfigure}{0.48\textwidth}
        \centering
        \includegraphics[width=1\textwidth]{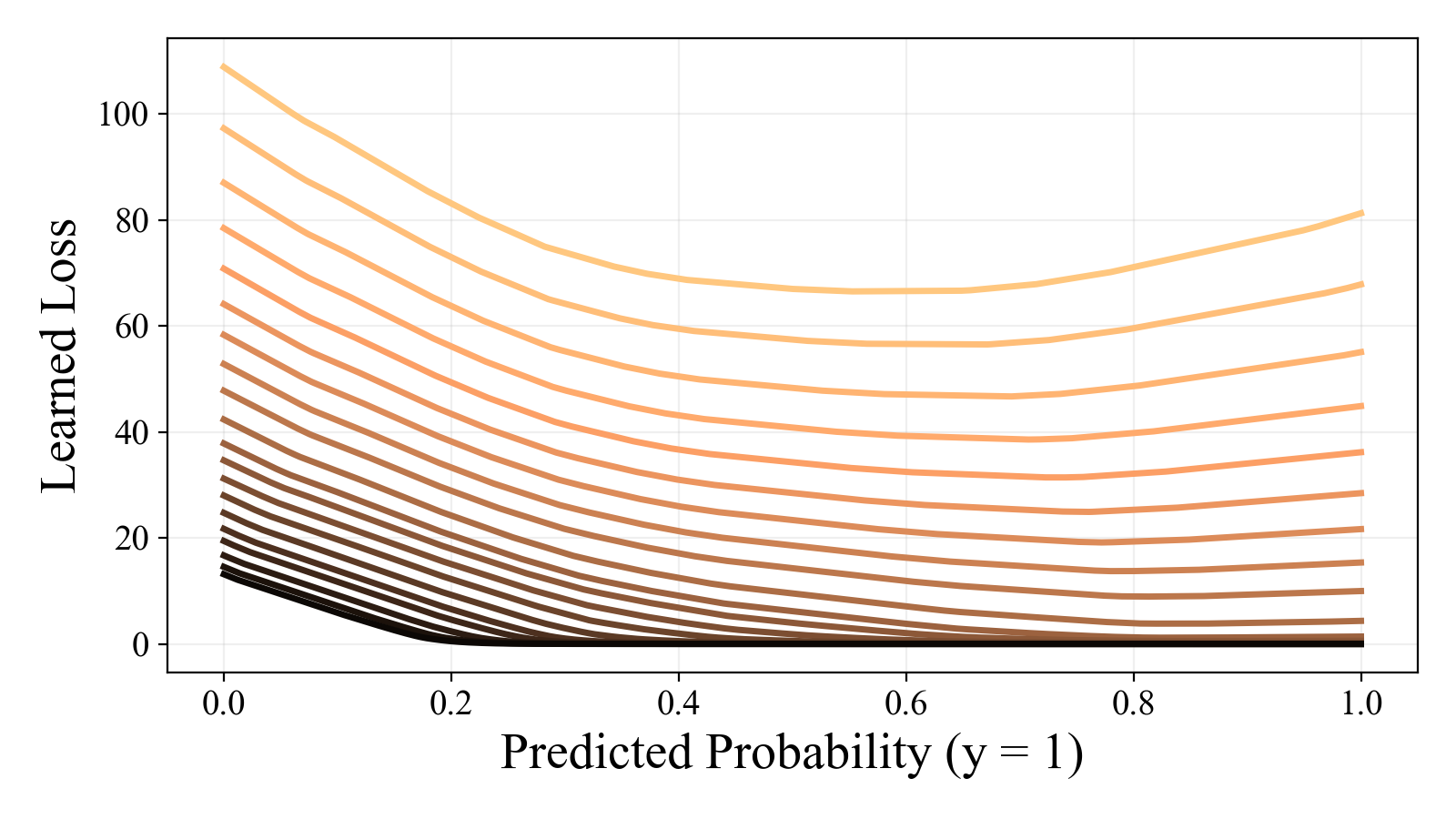}
    \end{subfigure}%
    \hspace{2mm}
    \begin{subfigure}{0.48\textwidth}
        \centering
        \includegraphics[width=1\textwidth]{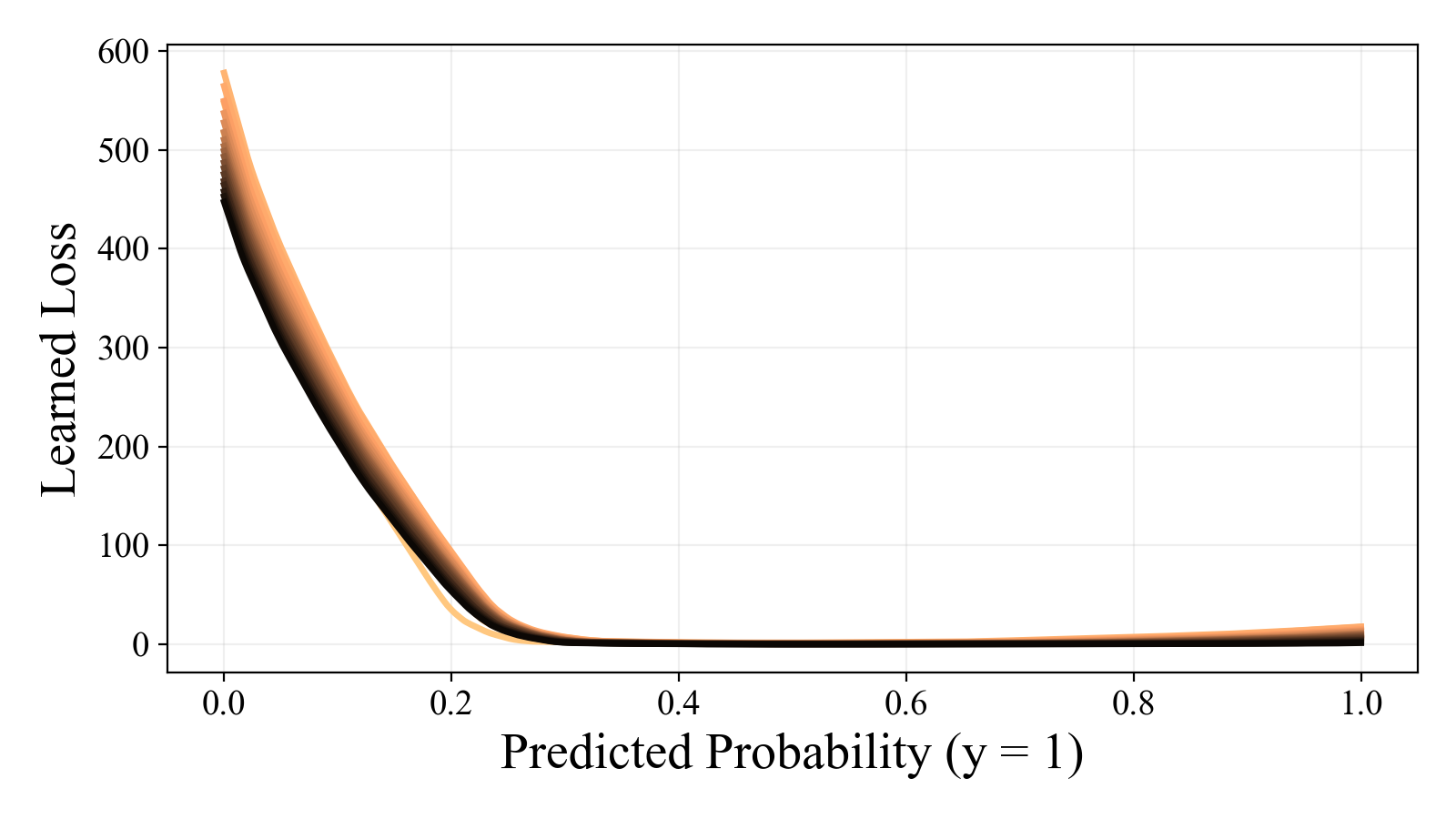}
    \end{subfigure}%

    \vspace{3mm}

    \begin{subfigure}{0.6\textwidth}
        \centering
        \includegraphics[width=1\textwidth]{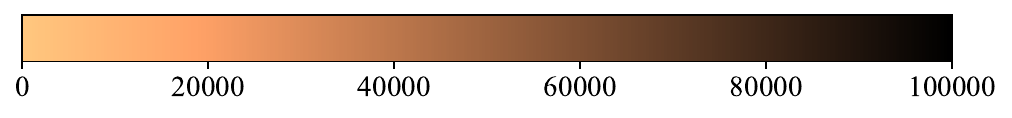}
    \end{subfigure}%
    
\captionsetup{justification=centering}
\caption{Four example loss functions generated by AdaLFL using the network architecture proposed in \protect\citep{bechtle2021meta}, which uses a Softplus activation in the output layer, causing flattening behavior degrading learning performance.}
\label{figure:flat-loss-functions}

\end{figure}

In AdaLFL, we parameterize the loss function using a small feedforward neural network. This representation was chosen due to its high expressiveness and design flexibility. Inspired by \citep{bechtle2021meta, psaros2022meta}, our meta-learned loss function, denoted by $\ell_{\phi}$, has two hidden layers, each with 40 units, and is applied output-wise (making it invariant to the number of outputs).
\begin{equation}
\textstyle \MetaLoss_{\phi}(y, f_{\theta}(x)) = \frac{1}{\mathcal{C}}\sum_{i=0}^{\mathcal{C}}\ell_{\phi}(y_{i}, f_{\theta}(x)_i)
\label{eq:loss-definition}
\end{equation}
Crucially, key design decisions are made regarding the activation functions used in $\ell_{\phi}$ to enforce desirable behavior. In \citep{bechtle2021meta}, ReLU activations are used in the hidden layers, and the smooth Softplus activation is used in the output layer to constrain the loss to be non-negative, \textit{i.e.}, $\ell{\phi}:\mathbb{R}^2 \rightarrow \mathbb{R}_{0}^{+}$, where $\forall y \forall f_{\theta}(x) \MetaLoss_{\phi_t}(y, f_{\theta}(x)) \geq 0$. Unfortunately, this network architecture is prone to \textit{unintentionally} encouraging overly flat regions in the loss function, since all negative inputs to the output layer go to 0, as shown in Figure \ref{figure:flat-loss-functions}. 

Generally, flat regions in the loss function are very detrimental to training as uniform loss is given to non-uniform errors. Removal of the Softplus activation in the output can partially resolve this flatness issue; however, without it, the learned loss functions would violate the typical constraint that a loss function should be at least $\mathcal{C}^1$, \textit{i.e.}, continuous in the zeroth and first derivatives. Alternative smooth activations, such as Sigmoid, TanH, ELU, etc., can be used instead; however, due to their range-bounded limits, they are also prone to encouraging loss functions that have large flat regions when their activations saturate --- a common occurrence when taking gradients through long unrolled optimization paths \citep{antoniou2019train}.

\subsection{Smooth Leaky ReLU}
\label{sec:adalfl-smooth-leaky-relu}

\begin{figure}[]

    \centering
    \begin{subfigure}{0.48\textwidth}
        \centering
        \includegraphics[width=1\textwidth]{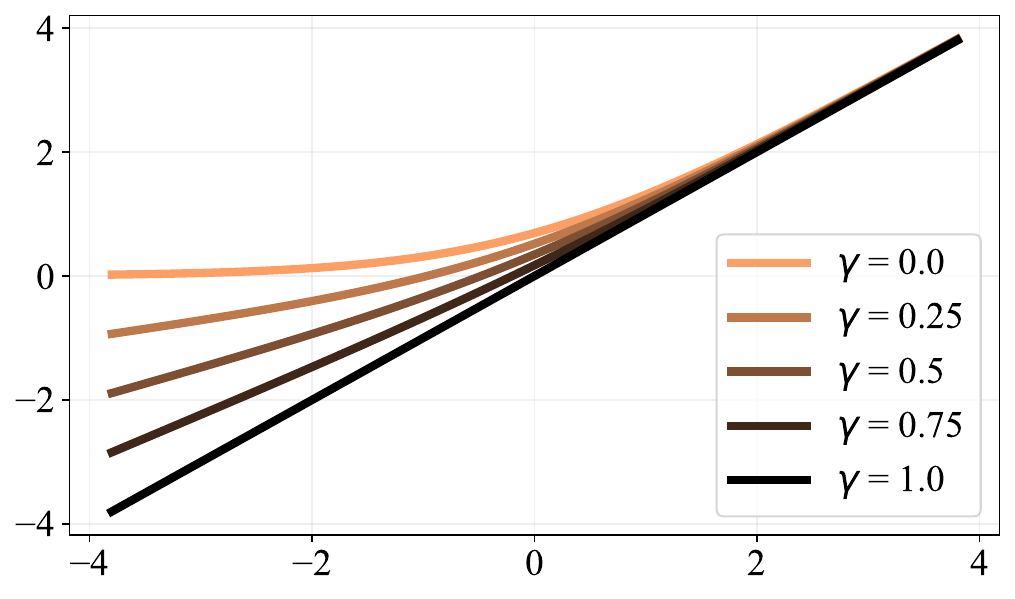}
    \end{subfigure}%
    \hspace{4mm}
    \begin{subfigure}{0.48\textwidth}
        \centering
        \includegraphics[width=1\textwidth]{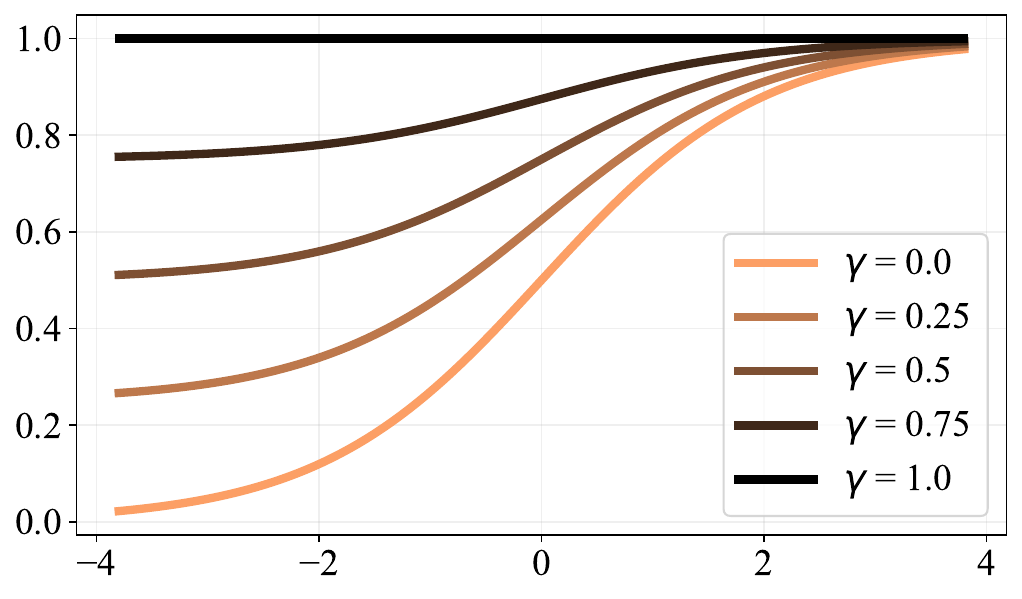}
    \end{subfigure}%

    \begin{subfigure}{0.48\textwidth}
        \centering
        \includegraphics[width=1\textwidth]{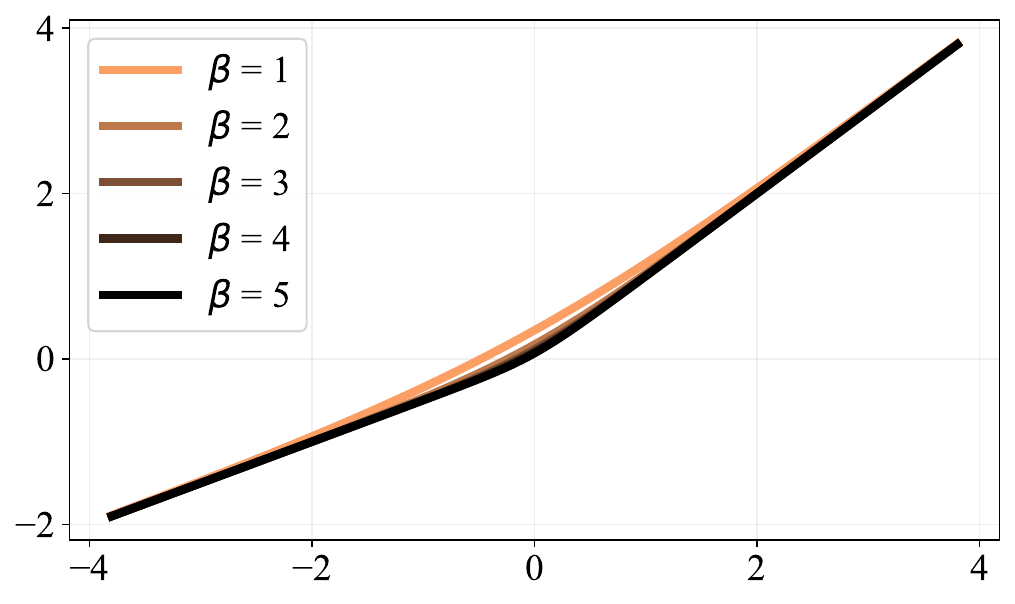}
    \end{subfigure}%
    \hspace{4mm}
    \begin{subfigure}{0.48\textwidth}
        \centering
        \includegraphics[width=1\textwidth]{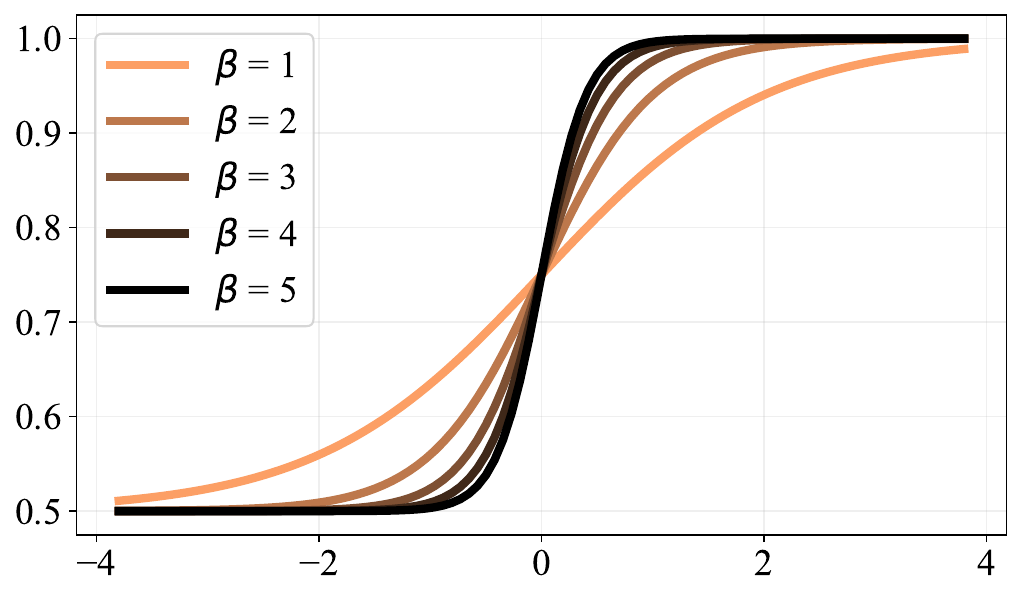}
    \end{subfigure}%

    \begin{subfigure}{0.48\textwidth}
        \centering
        \includegraphics[width=1\textwidth]{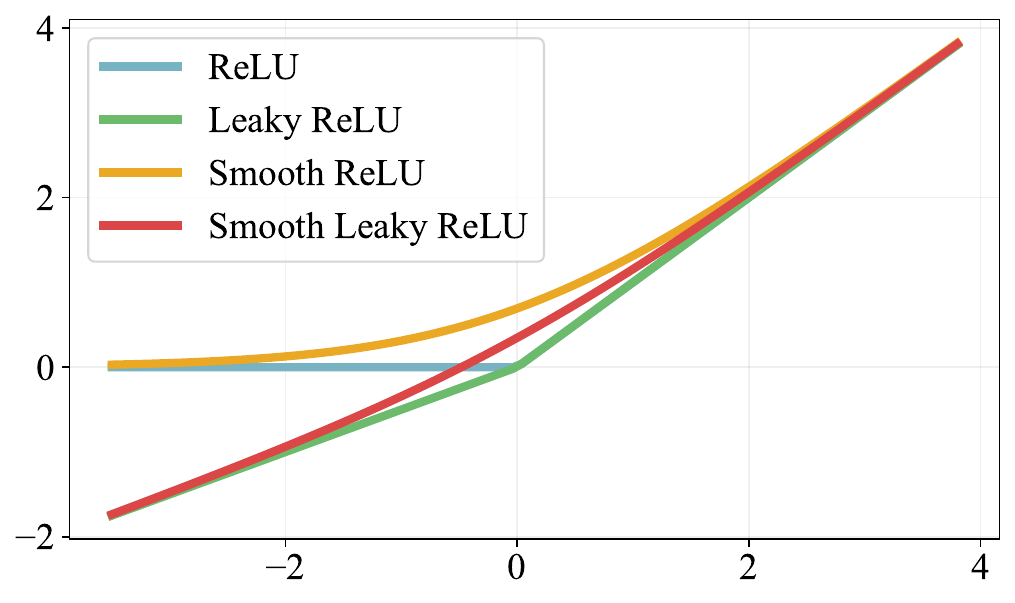}
    \end{subfigure}%
    
\captionsetup{justification=centering}
\caption{The proposed activation function and its corresponding derivatives when shifting $\gamma$ are shown in (a) and (b), respectively. In (c) and (d) the activation function and its derivatives when shifting $\beta$ are shown. Finally, in (c), the smooth leaky ReLU is contrasted with the original smooth and leaky variants of ReLU.}
\label{figure:activation-functions}

\end{figure}

To inhibit the flattening behavior of learned loss functions, a range unbounded activation function should be used. A popular activation function that is unbounded when the leak parameter $\gamma < 0$ is the \textit{Leaky ReLU} \citep{maas2013rectifier}
\begin{align}
\varphi^{Leaky}(x)
&= \max(\gamma \cdot x, x) \\
&= \max(0, x) \cdot (1 - \gamma) + \gamma x.
\label{eq:leakyrelu}
\end{align}
However, it is typically assumed that a loss function should be \textit{at least} $\mathcal{C}^{1}$, \textit{i.e.}, continuous in the zeroth and first derivatives. Fortunately, there is a smooth approximation to the ReLU, commonly referred to as the \textit{SoftPlus} activation function \citep{dugas2000incorporating}, where $\beta$ (typically set to 1) controls the smoothness.
\begin{equation}
\varphi^{Smooth}(x) = \tfrac{1}{\beta} \cdot \log(e^{\beta x} + 1) 
\label{eq:softplus}
\end{equation}
The leaky ReLU is combined with the smooth ReLU by taking the term $max(0, x)$ from Equation \eqref{eq:leakyrelu} and substituting it with the smooth SoftPlus defined in Equation \eqref{eq:softplus} to construct a smooth approximation to the leaky ReLU, which we further refer to as the \textit{Smooth Leaky ReLU} activation function:
\begin{equation}
\varphi^{SLReLU}(x) = \tfrac{1}{\beta} \log(e^{\beta x} + 1) \cdot (1 - \gamma) + \gamma x 
\end{equation}
where the derivative of the smooth leaky ReLU with respect to the input $x$ is
\begin{align}
\frac{d}{dx} \left[ \varphi^{SLReLU}(x) \right]
&= \frac{d}{dx} \left[ \frac{\log(e^{\beta x} + 1) \cdot (1 - y)}{\beta} + \gamma x \right] \\
&= \frac{\frac{d}{dx} [\log(e^{\beta x} + 1)] \cdot (1 - y)}{\beta} + \gamma \\
&= \frac{\frac{d}{dx} [e^{\beta x} + 1] \cdot (1 - y)}{\beta \cdot e^{\beta x} + 1} + \gamma \\
&= \frac{e^{\beta x} \cdot \beta \cdot (1 - y)}{\beta \cdot e^{\beta x} + 1} + \gamma \\
&= \frac{e^{\beta x} (1 - \gamma)}{e^{\beta x} + 1} + \gamma \\
&= \frac{e^{\beta x} (1 - \gamma)}{e^{\beta x} + 1} + \frac{\gamma(e^{\beta x} + 1)}{e^{\beta x} + 1} \\
&= \frac{e^{\beta x} + \gamma}{e^{\beta x} + 1}
\end{align}
The smooth leaky ReLU activation function and its corresponding derivatives are shown in Figure \ref{figure:activation-functions}. The proposed activation function has desirable characteristics for representing a loss function as it is smooth and has linear asymptotic behavior necessary for tasks such as regression, where extrapolation of the learned loss function can often occur. Furthermore, as its output is not bounded when $\gamma > 0$, it does not encourage flatness in the learned loss function. Early iterations of AdaLFL learned $\gamma$ and $\beta$ simultaneously with the network weights $\phi$, however; empirically, we found that setting $\gamma=0.01$ and $\beta=10$ gave adequate inference performance across our experiments. 

\subsection{Offline Loss Function Initialization}
\label{sec:adalfl-offline-optimization}

\begin{algorithm}[t]

\SetAlgoLined
\DontPrintSemicolon
\SetKwInput{Input}{Input}

\caption{Loss Function Initialization (Offline)}
\label{algorithm:offline}

\BlankLine
\Input{
    $\Loss^{meta} \leftarrow$ Task loss function (meta-objective)\;
    \vspace{-2mm}
    \hrulefill
}

\BlankLine
$\MetaLoss_{\phi_{0}} \leftarrow$ Initialize parameters of meta learner\;
\For{$t \in \{0, ... , \mathcal{S}^{init}\}$}{
    $\theta_{0} \leftarrow$ Reset parameters of base learner\;
    \For{$i \in \{0, ... , \mathcal{S}^{inner}\}$}{
        $X$, $y$ $\leftarrow$ Sample from $\Dataset^{train}$\;
        $\MetaLoss^{learned} \leftarrow \MetaLoss_{\phi_{t}}(y, f_{\theta_{i}}(X))$\;
        $\theta_{i + 1} \leftarrow \theta_{i} - \alpha \nabla_{\theta_{i}} \MetaLoss^{learned}$\;
    }
    $X$, $y$ $\leftarrow$ Sample from $\Dataset^{valid}$\;
    $\Loss^{task} \leftarrow \Loss^{meta}(y, f_{\theta_{i + 1}}(X))$\;
    $\phi_{t+1} \leftarrow \phi_{i} - \eta \nabla_{\phi_{t}}\Loss^{task}$\;
}

\BlankLine

\end{algorithm}

One challenge for online loss function learning is achieving a stable and performant initial set of parameters for the learned loss function. If $\phi$ is initialized poorly, too much time is spent on fixing $\phi$ in the early stages of the learning process, resulting in poor base convergence, or in the worst case, $f_{\theta}$ to diverge. To address this, offline loss function learning using \textit{Meta-learning via Learned Loss} (ML$^3$) \citep{bechtle2021meta} is utilized to fine-tune the initial loss function to the base model prior to online loss function learning. The initialization process is summarized in Algorithm \ref{algorithm:offline}, where $\mathcal{S}^{init}=2500$. In AdaLFL's initialization process one step on $\theta$ is taken for each step in $\phi$, \textit{i.e.}, inner gradient steps $\mathcal{S}^{inner}=1$. However, if $\mathcal{S}^{inner} < 1$, implicit differentiation \citep{rajeswaran2019meta,lorraine2020optimizing,gao2022loss} can instead be utilized to reduce the initialization process's memory and computational overhead.

\subsection{Online Loss Function Adaptation}
\label{sec:adalfl-online-optimization}

\begin{algorithm}[t!]

\caption{Loss Function Adaptation (Online)}
\label{algorithm:online}

\SetAlgoLined
\DontPrintSemicolon
\SetKwInput{Input}{Input}

\vspace{1mm}
\Input{
    $\MetaLoss_{\phi} \leftarrow$ Learned loss function (base-objective)\newline
    $\Loss^{meta} \leftarrow$ Task loss function (meta-objective)\;
    \vspace{-2mm}
    \hrulefill
}

\BlankLine
$\theta_{0} \leftarrow$ Initialize parameters of base learner\;
    
\For{$t \in \{0, ... , \mathcal{S}_{train}\}$}{
    $X$, $y$ $\leftarrow$ Sample from $\Dataset^{train}$\;
    $\MetaLoss^{learned} \leftarrow \MetaLoss_{\phi_{t}}(y, f_{\theta_{t}}(X))$\;
    $\theta_{t+1} \leftarrow \theta_{t} - \alpha \nabla_{\theta_{t}} \MetaLoss^{learned}$\;
    $X$, $y$ $\leftarrow$ Sample from $\Dataset^{valid}$\;
    $\Loss^{task} \leftarrow \Loss^{meta}(y, f_{\theta_{t+1}}(X))$\;
    $\phi_{t+1} \leftarrow \phi_{t} - \eta \nabla_{\phi_{t}}\Loss^{task}$\;
}

\BlankLine

\end{algorithm}

\begin{figure}
    
    \centering
    \includegraphics[width=1.06\columnwidth]{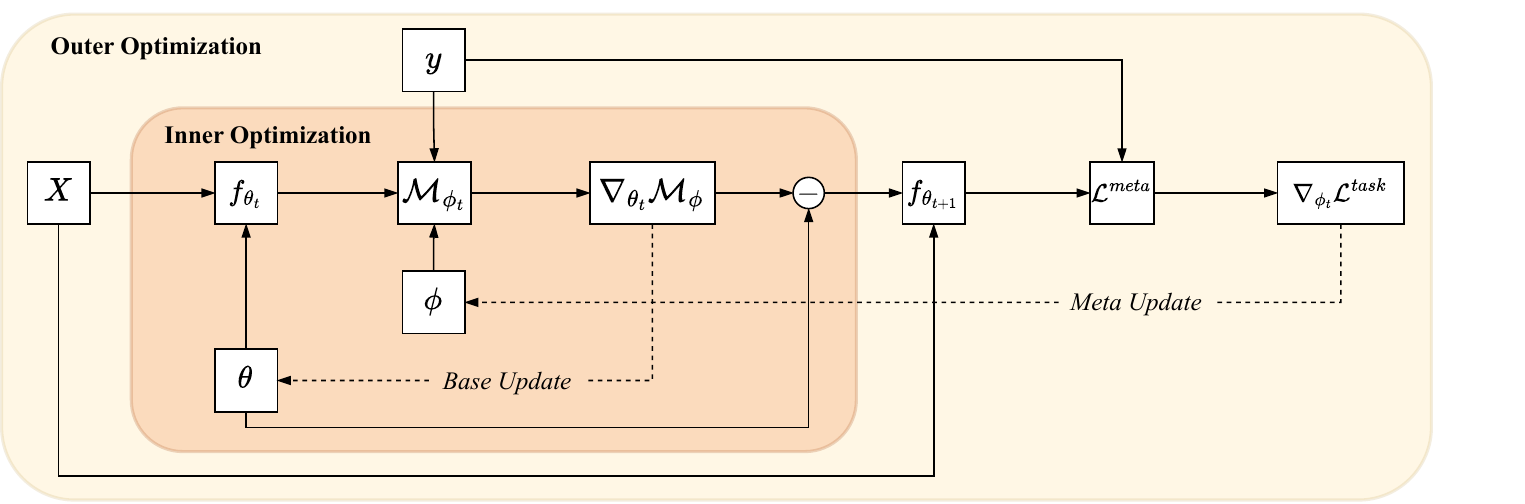}
    \captionsetup{justification=centering}
    \caption{Computational graph of AdaLFL, where $\theta$ is updated using $\MetaLoss_{\phi}$ in the inner loop (\textit{Base Update}). The optimization path is tracked in the computational graph and then used to update $\phi$ based on the meta-objective in the outer loop (\textit{Meta Update}). The dashed lines show the gradients for $\theta$ and $\phi$ with respect to their given objectives.}

\label{fig:computational-graph}
\end{figure}

\noindent
To optimize $\phi$, unrolled differentiation is utilized in the outer loop to update the learned loss function after each update to the base model parameters $\theta$ in the inner loop, which occurs via vanilla backpropagation. This is conceptually the simplest way to optimize $\phi$ as all the intermediate iterates generated by the optimizer in the inner loop can be stored and then backpropagate through in the outer loop \citep{maclaurin2015gradient}. The full iterative learning process is summarized in Algorithm \ref{algorithm:online} and proceeds as follows: perform a forward pass $f_{\theta_{t}}(x)$ to obtain an initial set of predictions. The learned loss function $\MetaLoss_{\phi}$ is then used to produce a base loss value
\begin{equation}
\MetaLoss^{learned} = \MetaLoss_{\phi_{t}}(y, f_{\theta_{t}}(x)).
\label{eq:loss-base}
\end{equation}
Using $\MetaLoss^{learned}$, the current weights $\theta_{t}$ are updated by taking a step in the opposite direction of the gradient of the loss with respect to $\theta_{t}$, where $\alpha$ is the base learning rate. 
\begin{equation}
\begin{split}
\theta_{t+1}
& = \theta_{t} - \alpha \nabla_{\theta_{t}} \MetaLoss_{\phi_{t}}(y, f_{\theta_{t}}(x)) \\
& = \theta_{t} - \alpha \nabla_{\theta_{t}} \mathbb{E}_{X, y} \big[ \MetaLoss_{\phi_{t}}(y, f_{\theta_{t}}(x)) \big]
\end{split}
\label{eq:backward-base}
\end{equation}
which can be further decomposed via the chain rule as shown in Equation \eqref{eq:backward-base-decompose}. Importantly, all the intermediate iterates generated by the (base) optimizer at the $t^{th}$ time-step when updating $\theta$ are stored in memory. 
\begin{equation}
\theta_{t+1} = \theta_{t} - \alpha \nabla_{f} \MetaLoss_{\phi_{t}}(y, f_{\theta_{t}}(x)) \nabla_{\theta_{t}}f_{\theta_{t}}(x)
\label{eq:backward-base-decompose}
\end{equation}
$\phi_{t}$ can now be updated to $\phi_{t+1}$ based on the learning progression made by $\theta$. Using $\theta_{t+1}$ as a function of $\phi_{t}$, compute a forward pass using the updated base weights $f_{\theta_{t+1}}(x)$ to obtain a new set of predictions. The instances can either be sampled from the training set or a held-out validation set. The new set of predictions is used to compute the \textit{task loss} $\Loss^{task}$ to optimize $\phi_{t}$ through $\theta_{t+1}$
\begin{equation}
\Loss^{task} = \Loss^{meta}(y, f_{\theta_{t+1}}(x))
\label{eq:loss-meta}
\end{equation}
where $\Loss^{meta}$ is selected based on the respective application. For example, the squared error loss for the task of regression or the cross-entropy loss for classification. The task loss is a crucial component for embedding the end goal task into the learned loss function. Optimization of the current meta-loss network loss weights $\phi_{t}$ now occurs by taking the gradient of $\Loss^{meta}$, where $\eta$ is the meta learning rate.
\begin{equation}
\begin{split}
\phi_{t+1}
& = \phi_{t} - \eta \nabla_{\phi_{t}}\Loss^{meta}(y, f_{\theta_{t+1}}(x)) \\
& = \phi_{t} - \eta \nabla_{\phi_{t}} \mathbb{E}_{X, y} \big[ \Loss^{meta}(y, f_{\theta_{t+1}}(x)) \big]
\end{split}
\label{eq:backward-meta}
\end{equation}
where the gradient computation is decomposed by applying the chain rule as shown in Equation \eqref{eq:backward-meta-decompose} where the gradient with respect to the meta-loss network weights $\phi_{t}$ requires the updated model parameters $\theta_{t+1}$ from Equation \eqref{eq:backward-base}. 
\small
\begin{align}
\phi_{t+1} 
&= \phi_{t} - \eta \nabla_{f}\Loss^{meta} \nabla_{\theta_{t+1}} f_{\theta_{t+1}} \nabla_{\phi_{t}}\theta_{t+1}(\phi_{t}) \\
&= \phi_{t} - \eta \nabla_{f}\Loss^{meta} \nabla_{\theta_{t+1}} f_{\theta_{t+1}} \nabla_{\phi_{t}}[\theta_{t} - \alpha \nabla_{\theta_{t}}\MetaLoss_{\phi_{t}}]
\label{eq:backward-meta-decompose}
\end{align}
\normalsize
This process is repeated for a fixed number of gradient steps $S_{train}$, which is identical to what would typically be used for training $f_{\theta}$. An overview and summary of the full associated data flow between the inner and outer optimization of $\theta$ and $\phi$, respectively, is given in Figure \ref{fig:computational-graph}.

\section{Background and Related Work}
\label{section:background}

The method that we propose in this chapter addresses the general problem of meta-learning a (base) loss function, \textit{i.e.}, loss function learning. Existing loss function learning methods can be categorized along two key axes, loss function representation, and meta-optimization. Frequently used representations in loss function learning include parametric \citep{gonzalez2020improved,raymond2023learning, raymond2023fast} and nonparametric \citep{liu2021loss,li2022autoloss} genetic programming expression trees. In addition to this, alternative representations such as truncated Taylor polynomials \citep{gonzalez2021optimizing,gao2021searching,gao2022loss} and small feed-forward neural networks \citep{bechtle2021meta} has also been recently explored. Regarding meta-optimization, loss function learning methods have heavily utilized computationally expensive evolution-based methods such as evolutionary algorithms \citep{koza1994genetic} and evolutionary strategies \citep{hansen2001completely}. While more recent approaches have made use of gradient-based approaches unrolled differentiation \citep{maclaurin2015gradient}, and implicit differentiation \citep{lorraine2020optimizing}.

A common trait among these methods is that, in contrast to AdaLFL, they perform \textit{offline} loss function learning, resulting in a severe short-horizon bias towards loss functions which are performant in the early stages of training but often have sub-optimal performance at the end of training. This short-horizon bias arises from how the various approaches compute their respective meta-objectives. In offline evolution-based approaches, the fitness, \textit{i.e.}, meta-objective, is calculated by computing the performance at the end of a partial training session, \textit{e.g.}, $\leq 1000$ gradient steps \citep{gonzalez2021optimizing,raymond2023learning, raymond2023fast}. A truncated number of gradient steps are required to be used as evolution-based methods evaluate the performance of a large number of candidate solutions, typically $L$ loss function over $K$ iterations where $25 \leq L, K \leq 100$. Therefore, performing full training sessions, which can be hundreds, thousands, or even millions of gradient steps for each candidate solution, is infeasible.

Regarding the existing gradient-based approaches, offline unrolled optimization requires the whole optimization path to be stored in memory; in practice, this significantly restricts the number of inner gradient steps before computing the meta-objective to only a small number of steps. Methods such as implicit differentiation can obviate these memory issues; however, it would still require a full training session in the inner loop, which is a prohibitive number of forward passes to perform in tractable time. Furthermore, the dependence of the model parameters on the meta-parameters increasingly shrinks and eventually vanishes as the number of steps increases \citep{rajeswaran2019meta}.

\subsection{Online vs Offline Meta Optimization}
\label{sec:online-vs-offline}

The key algorithmic difference of AdaLFL from prior offline gradient-based methods \citep{bechtle2021meta,gao2022loss} is that $\phi$ is updated after each update to $\theta$ in lockstep in a single phase as opposed to learning $\theta$ and $\phi$ in separate phases. This is achieved by not resetting $\theta$ after each update to $\phi$ (Algorithm \ref{algorithm:offline}, line 3), and consequently, $\phi$ has to adapt to each newly updated time-step such that $\phi = (\phi_0, \phi_1, \dots, \phi_{\mathcal{S}^{train}})$. In offline loss function learning, $\phi$ is learned in the meta-training phase and then is fixed for the full duration of the meta-testing phase where $\theta$ is learned and $\phi = (\phi_0)$. Another crucial difference is that in online loss function learning, there is an implicit tuning of the learning rate schedule, as discussed later in Section \ref{section:implicit-learning-rate-schedule}.

\subsection{Related Paradigms}

Although online loss function learning has not been explored in the meta-learning context, some existing research outside the subfield has previously explored the possibility of adaptive loss functions, such as in \citep{li2019lfs} and \citep{wang2020loss}. However, we emphasize that these approaches are categorically different in that they do not learn the loss function from scratch; instead, they interpolate between a small subset of handcrafted loss functions, updating the loss function after each epoch. Furthermore, in contrast to loss function learning which is both task and model-agnostic, these techniques are restricted to being task-specific, \textit{e.g.}, face recognition only. Finally, this class of approaches does not implicitly tune the base learning rate $\alpha$, as is the case in loss function learning.

\section{Experimental Setup}
\label{section:setup}

In this section, the experimental setup for evaluating AdaLFL is presented. In summary, experiments are conducted across seven open-access datasets and multiple well-established network architectures. The performance of the proposed method is contrasted against the handcrafted cross-entropy loss and AdaLFL's offline counterpart ML$^3$ Supervised \citep{bechtle2021meta}. The experiments were implemented in \texttt{PyTorch} \citep{paszke2017automatic}, and \texttt{Higher} \citep{grefenstette2019generalized}, and the code can be found at \href{http://github.com/Decadz/Online-Loss-Function-Learning}{http://github.com/Decadz/Online-Loss-Function-Learning}.

\subsection{Benchmark Tasks}

Following the established literature on loss function learning, the regression datasets Communities and Crime, Diabetes, and California Housing are used as a simple domain to illustrate the capabilities of the proposed method. Following this classification datasets MNIST \citep{lecun1998gradient}, CIFAR-10, CIFAR-100 \citep{krizhevsky2009learning}, and SVHN \citep{netzer2011reading}, are employed to assess the performance of AdaLFL to determine whether the results can generalize to larger, more challenging tasks. The original training-testing partitioning is used for all datasets, with 10\% of the training instances allocated for validation. In addition, standard data augmentation techniques consisting of normalization, random horizontal flips, and cropping are applied to the training data of CIFAR-10, CIFAR-100, and SVHN during meta and base training.

\subsection{Benchmark Models}

A diverse set of well-established benchmark architectures is utilized to evaluate the performance of AdaLFL. For Communities and Crime, Diabetes, and California Housing a two hidden layer multi-layer perceptron (MLP) taken from \citep{baydin2018hypergradient} is used. For MNIST, logistic regression \citep{mccullagh2019generalized}, the previously mentioned MLP and the LeNet-5 \citep{lecun1998gradient} architecture is used. Following this experiments are conducted on CIFAR-10, VGG-16 \citep{simonyan2014very}, AllCNN-C \citep{springenberg2014striving}, ResNet-18 \citep{he2016deep}, and SqueezeNet \citep{iandola2016squeezenet} are used. For the remaining datasets, CIFAR-100 and SVHN, WideResNet 28-10 and WideResNet 16-8 \citep{zagoruyko2016wide} are employed. 

\subsection{Experimental Setup}

To initialize $\MetaLoss_{\phi}$, $\mathcal{S}^{init}=2500$ steps are taken in offline mode with a meta learning rate of $\eta=1e-3$. In contrast, in online mode, a meta learning rate of $\eta=1e-5$ is used (note, a high meta learning rate in online mode can cause a jittering effect in the loss function, which can cause training instability). The popular Adam optimizer is used in the outer loop for both initialization and online adaptation. 

In the inner loop, all regression models are trained using stochastic gradient descent (SGD) with a base learning rate of $\alpha=0.001$. While the classification models are trained using SGD with a base learning rate of $\alpha=0.01$, and on CIFAR-10, CIFAR-100, and SVHN, Nesterov momentum 0.9, and weight decay 0.0005 are used. The remaining base-model hyper-parameters are selected using their respective values from the literature in an identical setup to \citep{gonzalez2021optimizing}. 

All experimental results reported show the average error rate (classification) or mean squared error (regression) across 10 independent executions on different seeds to analyse algorithmic consistency. Importantly, our experiments control for the base initializations such that all methods get identical initial parameters across the same random seed; thus, any difference in variance between the methods can be attributed to the respective algorithms and their loss functions. Furthermore, the choice of hyper-parameters between ML$^3$ and AdaLFL has been standardized to allow for a fair comparison.

\section{Results and Analysis}
\label{section:results}

\begin{figure*}[]

    \centering
    \begin{subfigure}{0.5\textwidth}
        \centering
        \includegraphics[width=1\textwidth]{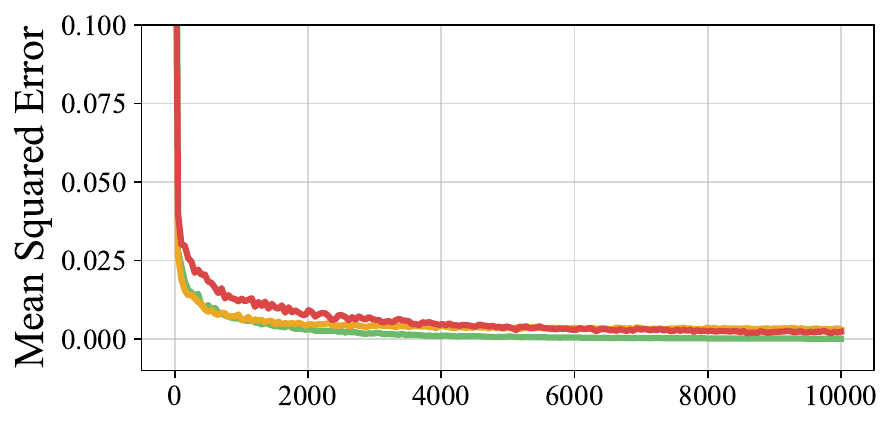}
        \caption{Crime + MLP}
    \end{subfigure}%
    \hfill
    \begin{subfigure}{0.5\textwidth}
        \centering
        \includegraphics[width=1\textwidth]{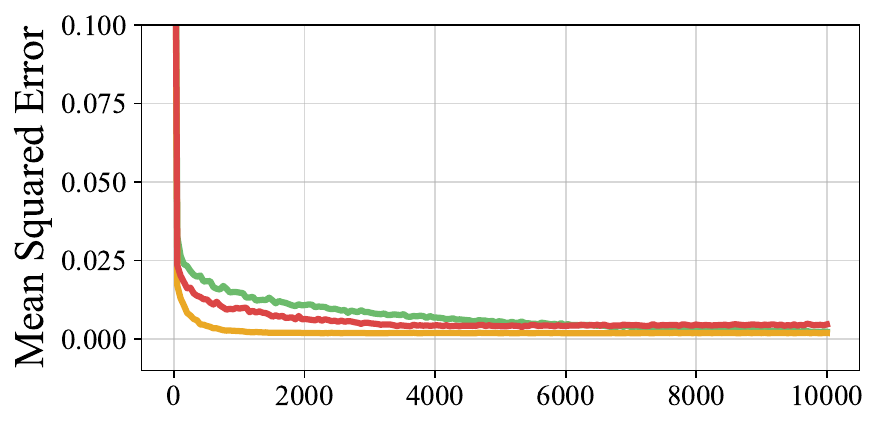}
        \caption{Diabetes + MLP}
    \end{subfigure}%

    \vspace{3mm}

    \centering
    \begin{subfigure}{0.5\textwidth}
        \centering
        \includegraphics[width=1\textwidth]{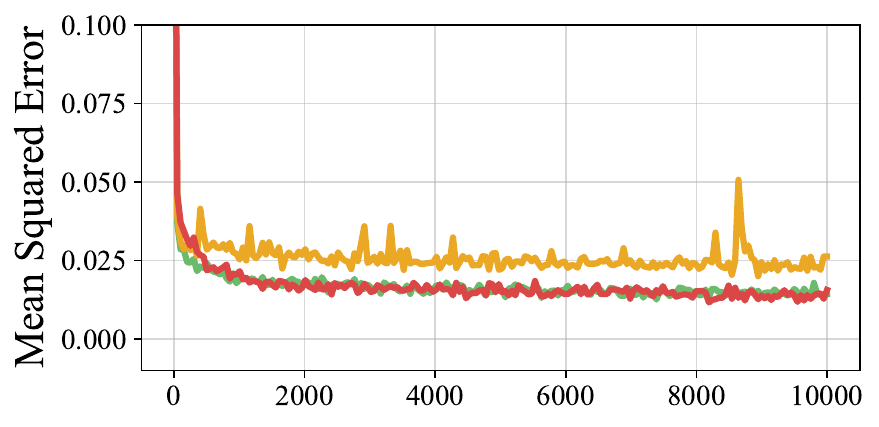}
        \caption{California + MLP}
    \end{subfigure}%
    \begin{subfigure}{0.5\textwidth}
        \centering
        \includegraphics[width=1\textwidth]{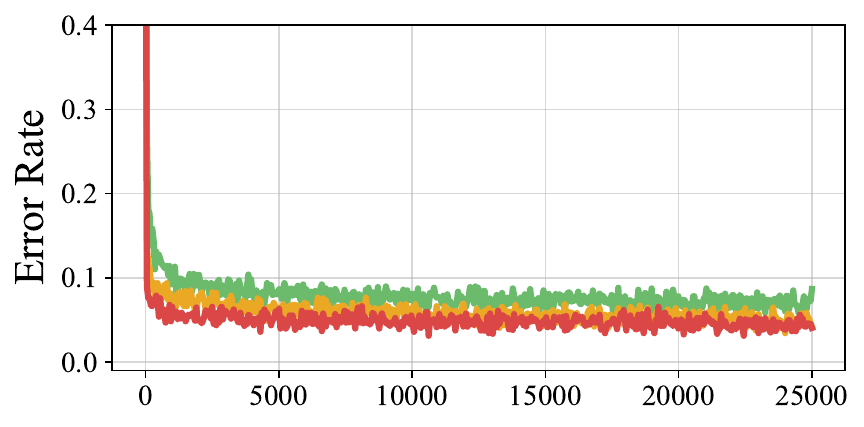}
        \caption{MNIST + Logistic}
    \end{subfigure}%

    \vspace{3mm}    

    \centering
    \begin{subfigure}{0.5\textwidth}
        \centering
        \includegraphics[width=1\textwidth]{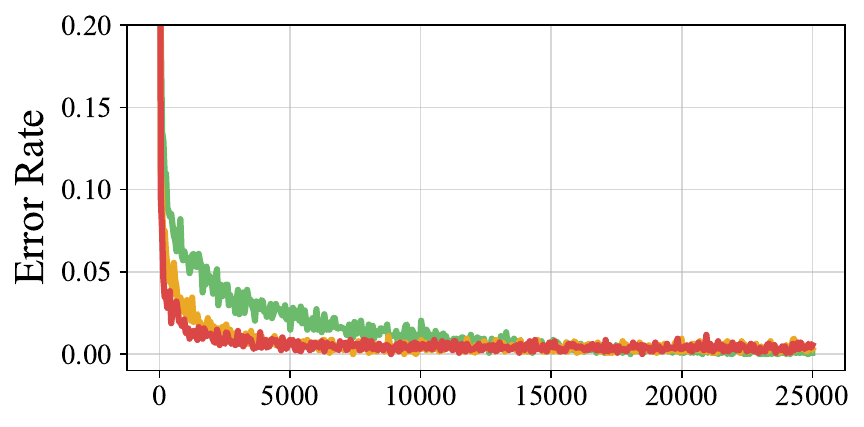}
        \caption{MNIST + MLP}
    \end{subfigure}%
    \hfill
    \begin{subfigure}{0.5\textwidth}
        \centering
        \includegraphics[width=1\textwidth]{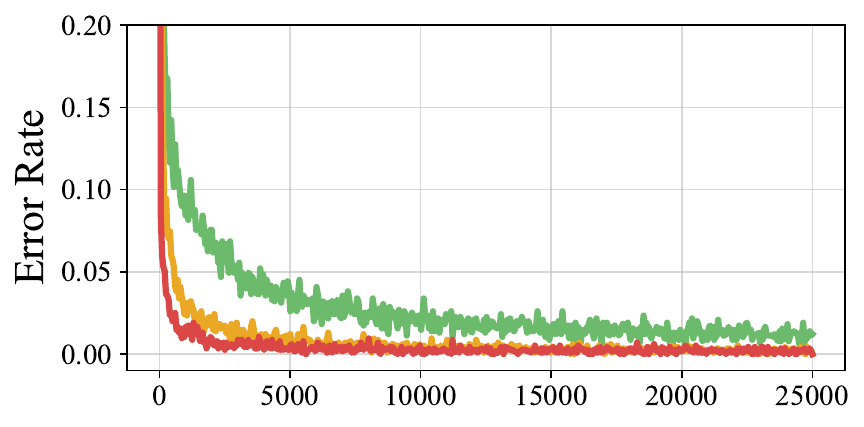}
        \caption{MNIST + LeNet-5}
    \end{subfigure}%
    
    \vspace{3mm}

    \begin{subfigure}{0.5\textwidth}
        \centering
        \includegraphics[width=1\textwidth]{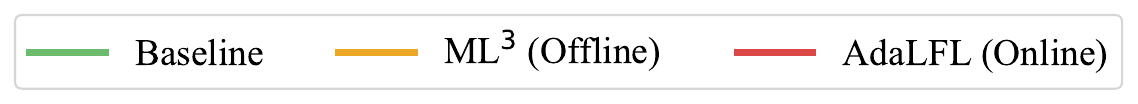}
    \end{subfigure}%
    
\captionsetup{justification=centering}
\caption{Mean learning curves across 10 independent executions of each algorithm on each task + model pair, showing the training mean squared error or error rate (y-axis) against gradient steps (x-axis). Best viewed in color.}

\end{figure*}

\begin{figure*}[]

    \ContinuedFloat
    
    \centering
    \begin{subfigure}{0.5\textwidth}
        \centering
        \includegraphics[width=1\textwidth]{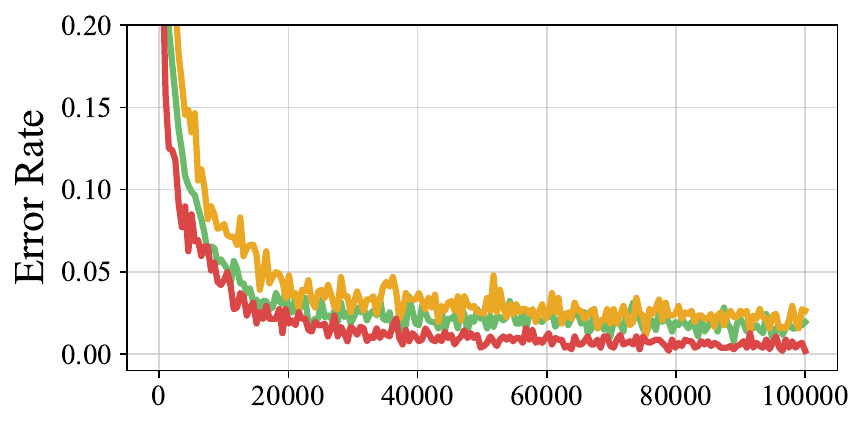}
        \caption{CIFAR-10 + VGG-16}
    \end{subfigure}%
    \hfill
    \begin{subfigure}{0.5\textwidth}
        \centering
        \includegraphics[width=1\textwidth]{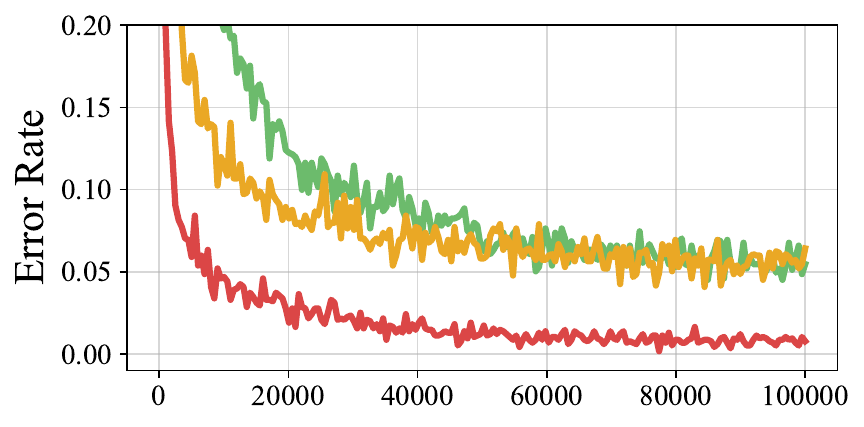}
        \caption{CIFAR-10 + AllCNN-C}
    \end{subfigure}%

    \vspace{3mm} 

    \centering
    \begin{subfigure}{0.5\textwidth}
        \centering
        \includegraphics[width=1\textwidth]{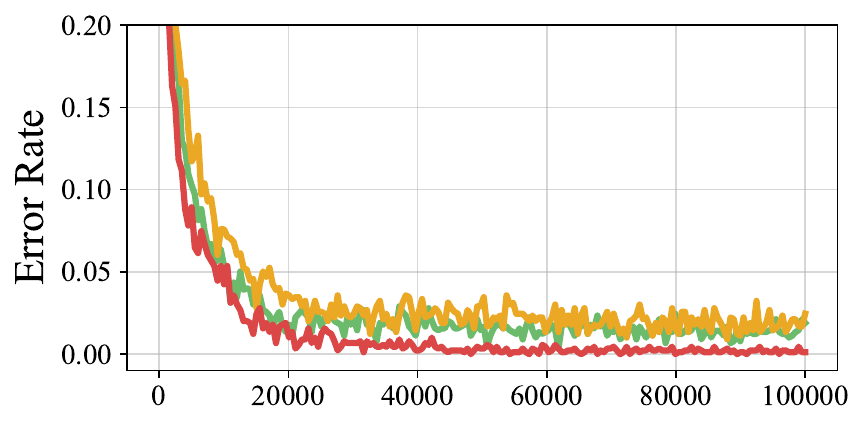}
        \caption{CIFAR-10 + ResNet-18}
    \end{subfigure}%
    \hfill
    \centering
    \begin{subfigure}{0.5\textwidth}
        \centering
        \includegraphics[width=1\textwidth]{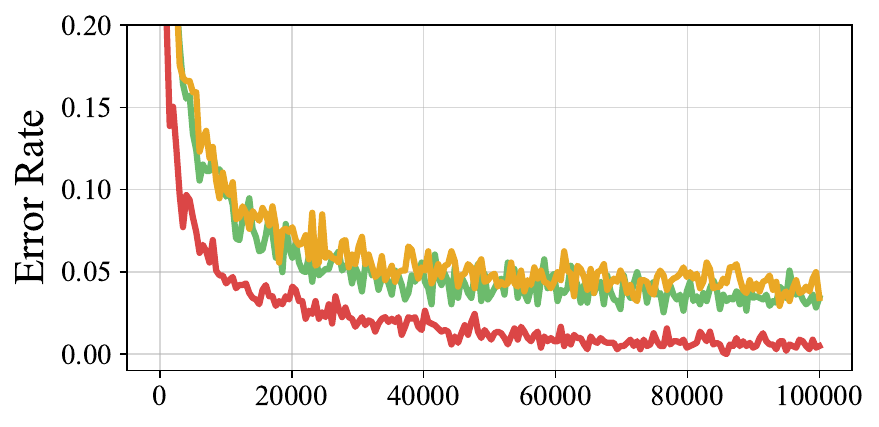}
        \caption{CIFAR-10 + SqueezeNet}
    \end{subfigure}%

    \vspace{3mm} 
    \begin{subfigure}{0.5\textwidth}
        \centering
        \includegraphics[width=1\textwidth]{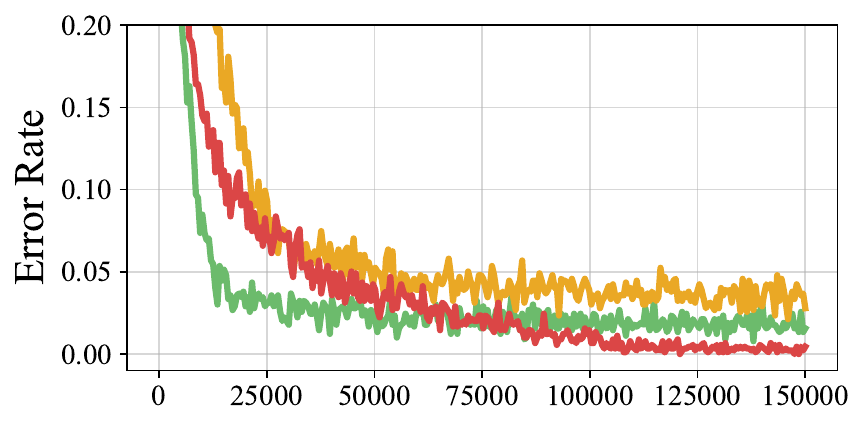}
        \caption{CIFAR-100 + WRN 28-10}
    \end{subfigure}%
    \hfill
    \begin{subfigure}{0.5\textwidth}
        \centering
        \includegraphics[width=1\textwidth]{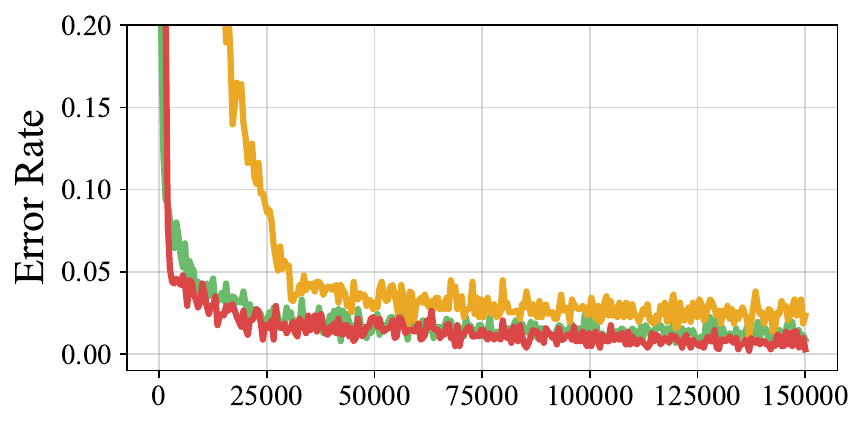}
        \caption{SVHN + WRN 16-8}
    \end{subfigure}%
    
    \vspace{3mm}

    \begin{subfigure}{0.5\textwidth}
        \centering
        \includegraphics[width=1\textwidth]{chapter-5/figures/learning-curves/learning-curve-legend.pdf}
    \end{subfigure}%
    
\captionsetup{justification=centering}
\caption{Mean learning curves across 10 independent executions of each algorithm on each task + model pair, showing the training mean squared error or error rate (y-axis) against gradient steps (x-axis). Best viewed in color.}
\label{fig:adalfl-learning-curves}

\end{figure*}

\subsection{Training Learning Curves}

The results in Figure \ref{fig:adalfl-learning-curves} show the average learning curves of the baseline, ML$^3$, and AdaLFL across 10 executions of each method on each dataset + model pair. The results show that AdaLFL makes clear and consistent gains in convergence speed compared to the baseline and offline loss function learning method ML$^3$, except on the regression datasets due to regularization behavior, and on CIFAR-100 where there was difficulty in achieving a stable initialization. Furthermore, the performance obtained by AdaLFL at the end of training is typically better (lower) than both of the compared methods, suggesting that performance gains are being made in addition to enhanced convergence speeds.

Another key observation is that AdaLFL improves upon the performance of the baseline on the more challenging tasks of CIFAR-10, CIFAR-100, and SVHN, where offline loss functions learning method ML$^3$ consistently performs poorly. Improved performance on these datasets is achieved via AdaLFL adaptively updating the learned loss function throughout the learning process to the changes in the training dynamics. This is in contrast to ML$^3$, where the loss function remains static, resulting in poor performance on tasks where the training dynamics at the beginning of training vary significantly from those at the end of training. 

\subsection{Testing Inference Performance}

\begin{table}[t!]
\centering

\captionsetup{justification=centering}
\caption{Results reporting the mean $\pm$ standard deviation of final inference testing mean squared error or error rate across 10 independent executions of each algorithm on each task + model pair (using no base learning rate scheduler).}

\begin{threeparttable}

\begin{tabular}{p{4cm}ccc}

\hline \noalign{\vskip 1mm}
Task and Model                            &  Baseline  &  ML$^{3}$ (Offline)  &  AdaLFL (Online) \\ \hline \noalign{\vskip 1mm}

\hline \noalign{\vskip 1mm}

\textbf{Crime}                            &                   &                   &                                           \\
MLP \tnote{1}                             & 0.0274$\pm$0.0017 & 0.0270$\pm$0.0025 & \textbf{0.0263$\pm$0.0023}                \\ \noalign{\vskip 1mm}

\hline \noalign{\vskip 1mm} 

\textbf{Diabetes}                         &                   &                   &                                           \\
MLP \tnote{1}                             & 0.0432$\pm$0.0013 & 0.0430$\pm$0.0012 & \textbf{0.0420$\pm$0.0014}                \\ \noalign{\vskip 1mm}

\hline \noalign{\vskip 1mm} 

\textbf{California}                       &                   &                   &                                           \\
MLP \tnote{1}                             & 0.0157$\pm$0.0001 & 0.0276$\pm$0.0058 & \textbf{0.0151$\pm$0.0007}                \\ \noalign{\vskip 1mm}

\hline \noalign{\vskip 1mm} 

\textbf{MNIST}                            &                   &                   &                                           \\
Logistic \tnote{2}                        & 0.0766$\pm$0.0009 & 0.0710$\pm$0.0010 & \textbf{0.0697$\pm$0.0010}                \\ \noalign{\vskip 1mm}
MLP  \tnote{1}                            & 0.0203$\pm$0.0006 & 0.0185$\pm$0.0004 & \textbf{0.0184$\pm$0.0006}                \\ \noalign{\vskip 1mm}
LeNet-5 \tnote{3}                         & 0.0125$\pm$0.0007 & 0.0094$\pm$0.0005 & \textbf{0.0091$\pm$0.0004}                \\ \noalign{\vskip 1mm}

\hline \noalign{\vskip 1mm} 

\textbf{CIFAR-10}                         &                   &                   &                                           \\
VGG-16 \tnote{4}                          & 0.1036$\pm$0.0049 & 0.1024$\pm$0.0055 & \textbf{0.0903$\pm$0.0032}                \\ \noalign{\vskip 1mm}
AllCNN-C \tnote{5}                        & 0.1030$\pm$0.0062 & 0.1015$\pm$0.0055 & \textbf{0.0835$\pm$0.0050}                \\ \noalign{\vskip 1mm}
ResNet-18 \tnote{6}                       & 0.0871$\pm$0.0057 & 0.0883$\pm$0.0041 & \textbf{0.0788$\pm$0.0035}                \\ \noalign{\vskip 1mm}
SqueezeNet \tnote{7}                      & 0.1226$\pm$0.0080 & 0.1162$\pm$0.0052 & \textbf{0.1083$\pm$0.0049}                \\ \noalign{\vskip 1mm}

\hline \noalign{\vskip 1mm} 

\textbf{CIFAR-100}                        &                   &                   &                                           \\           
WRN 28-10 \tnote{8}                       & 0.3046$\pm$0.0087 & 0.3108$\pm$0.0075 & \textbf{0.2668$\pm$0.0283}                \\ \noalign{\vskip 1mm}

\hline \noalign{\vskip 1mm} 

\textbf{SVHN}                             &                   &                   &                                           \\ 
WRN 16-8 \tnote{8}                        & 0.0512$\pm$0.0043 & 0.0500$\pm$0.0034 & \textbf{0.0441$\pm$0.0014}                \\ \noalign{\vskip 1mm}

\hline \noalign{\vskip 1mm} 
\end{tabular}

\begin{tablenotes}\centering
\tiny Network architecture references: \item[1] \citep{baydin2018hypergradient} \item[2] \citep{mccullagh2019generalized} \item[3] \citep{lecun1998gradient} \item[4] \citep{simonyan2014very} \item[5] \citep{springenberg2014striving} \item[6] \citep{he2016deep} \item[7] \citep{iandola2016squeezenet} \item[8] \citep{zagoruyko2016wide}
\end{tablenotes}
\end{threeparttable}

\label{table:inference-results}
\end{table}

The final inference testing results reporting the average mean squared error or error rate across 10 independent executions of each method are shown in Table \ref{table:inference-results}. The results show that AdaLFL's meta-learned loss functions produce superior inference performance when used in training compared to the baseline on all the tested problems. A further observation is that the gains achieved by AdaLFL are consistent and stable. Notably, in most cases, lower variability than the baseline is observed, as shown by the relatively small standard deviation in error rate across the independent runs. 

Regarding the regression results, AdaLFL is shown to consistently achieve strong out-of-sample performance, compared to the baseline and ML$^3$. As for the classification tasks, on MNIST AdaLFL obtains improved performance compared to the baseline, and similar performance to ML$^3$, suggesting that the training dynamics at the beginning of training are similar to those at the end; hence the modest difference in results. While on the more challenging tasks of CIFAR-10, CIFAR-100, and SVHN, AdaLFL produced significantly better results than ML$^3$, demonstrating the scalability of the proposed loss function learning approach. The results attained by AdaLFL are promising given that the models tested were designed and optimized around the baseline loss functions. Larger performance gains may be attained using models designed specifically around meta-learned loss function \citep{kim2018auto, elsken2020meta, ding2022learning}. Thus future work will explore learning the loss function in tandem with the network architecture.

\subsection{Runtime Analysis}

\begin{table}[t!]
\centering

\captionsetup{justification=centering}
\caption{Average runtime of the entire learning process for each benchmark method. Each algorithm is run on a single Nvidia RTX A5000, and results are reported in hours.}

\begin{tabular}{p{6.5cm}ccc}
\hline
\noalign{\vskip 1mm}
Task and Model               & Baseline     & Offline          & Online            \\ \hline \noalign{\vskip 1mm} \hline \noalign{\vskip 1mm}

Crime + MLP                 & 0.01         & 0.03             & 0.05              \\ \noalign{\vskip 1mm}

\hline \noalign{\vskip 1mm} 

Diabetes + MLP               & 0.01         & 0.03             & 0.05              \\ \noalign{\vskip 1mm}

\hline \noalign{\vskip 1mm} 

California + MLP             & 0.01         & 0.03             & 0.05              \\ \noalign{\vskip 1mm}

\hline \noalign{\vskip 1mm} 

MNIST + Logistic             & 0.06         & 0.31             & 0.55              \\ \noalign{\vskip 1mm}
MNIST + MLP                  & 0.06         & 0.32             & 0.56              \\ \noalign{\vskip 1mm}
MNIST + LeNet-5              & 0.10         & 0.38             & 0.67              \\ \noalign{\vskip 1mm}

\hline \noalign{\vskip 1mm} 

CIFAR-10 + VGG-16            & 1.50         & 1.85             & 5.56              \\ \noalign{\vskip 1mm}
CIFAR-10 + AllCNN-C          & 1.41         & 1.72             & 5.53              \\ \noalign{\vskip 1mm}
CIFAR-10 + ResNet-18         & 1.81         & 2.18             & 8.38              \\ \noalign{\vskip 1mm}
CIFAR-10 + SqueezeNet        & 1.72         & 2.02             & 7.88              \\ \noalign{\vskip 1mm}

\hline \noalign{\vskip 1mm} 

CIFAR-100 + WRN 28-10        & 8.81         & 10.3             & 50.49             \\ \noalign{\vskip 1mm}

\hline \noalign{\vskip 1mm} 

SVHN + WRN 16-8              & 7.32          & 7.61             & 24.75              \\ \noalign{\vskip 1mm} \hline \noalign{\vskip 1mm} 
\end{tabular}

\label{table:adalfl-runtime}
\end{table}

The average runtime of the entire learning process of all benchmark methods on all tasks is reported in Table \ref{table:adalfl-runtime}. Notably, there are two key reasons why the computational overhead of AdaLFL is not as bad as it may at first seem. First, the time reported for the baseline does not include the implicit cost of manual hyper-parameter selection and tuning of the loss function, as well as the initial learning rate and learning rate schedule, which is needed before training to attain reasonable performance \citep{goodfellow2016deep}. Second, a large proportion of the cost of AdaLFL comes from storing a large number of intermediate iterates needed for the outer loop. However, the intermediate iterates stored in this process are identical to those used in other popular meta-learning paradigms \citep{andrychowicz2016learning,finn2017model}. Consequently, future work can explore combining AdaLFL with other optimization-based meta-learning methods with minimal overhead cost, similar to \citep{li2017meta,park2019meta,flennerhag2019meta,baik2020meta,baik2021meta}.

\subsection{Visualizing Learned Loss Functions}

\begin{figure}[]
\centering

    \centering
    \begin{subfigure}{0.5\textwidth}
        \centering
        \includegraphics[width=1\textwidth]{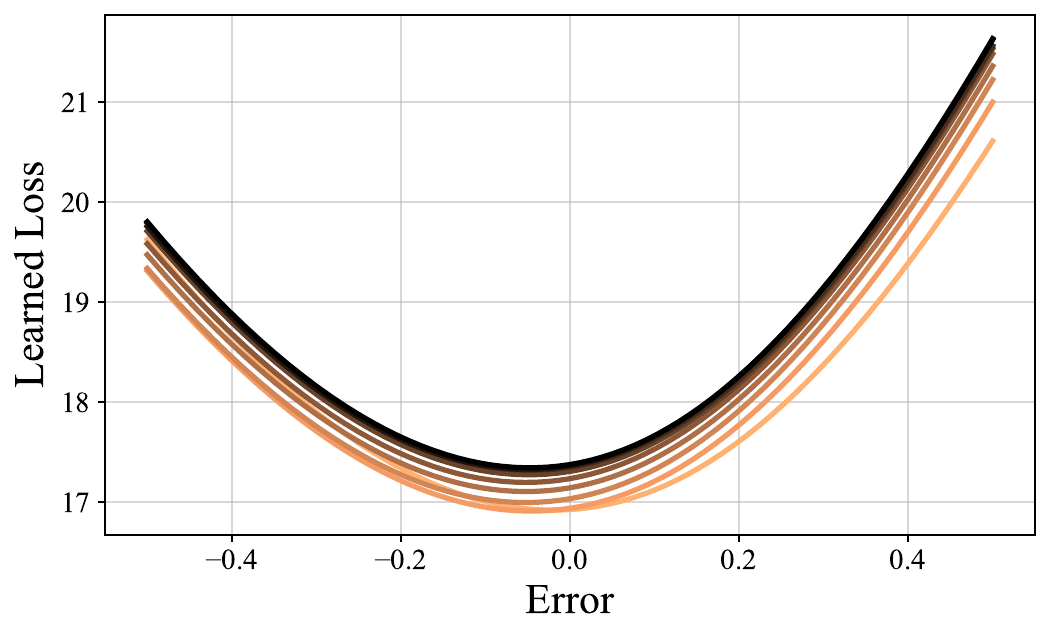}
    \end{subfigure}%
    \hfill
    \begin{subfigure}{0.5\textwidth}
        \centering
        \includegraphics[width=1\textwidth]{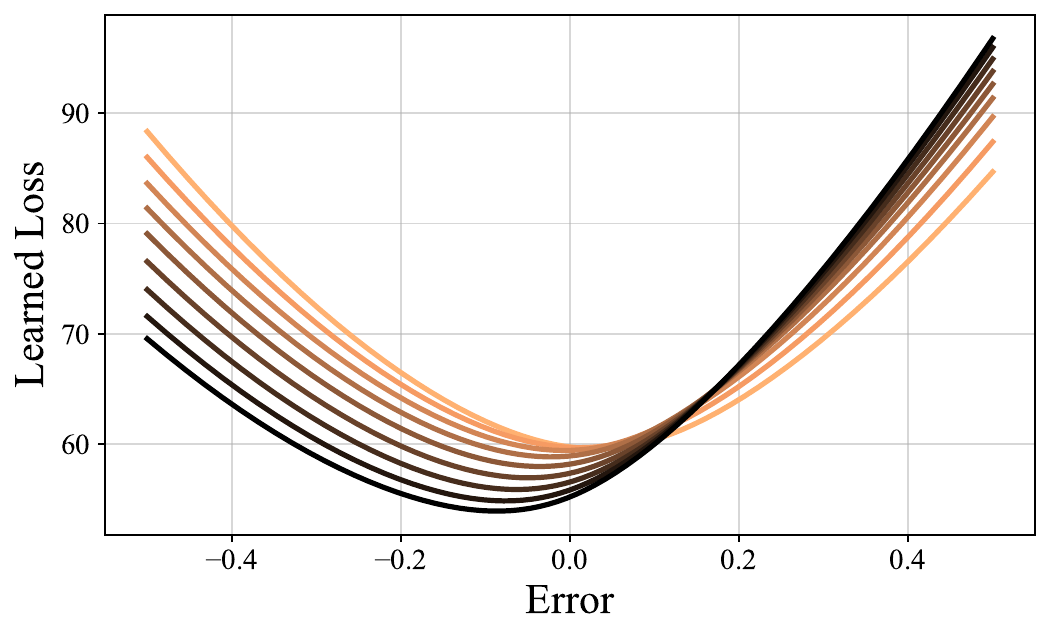}
    \end{subfigure}%
    
    \vspace{2mm}

    \begin{subfigure}{0.5\textwidth}
        \centering
        \includegraphics[width=1\textwidth]{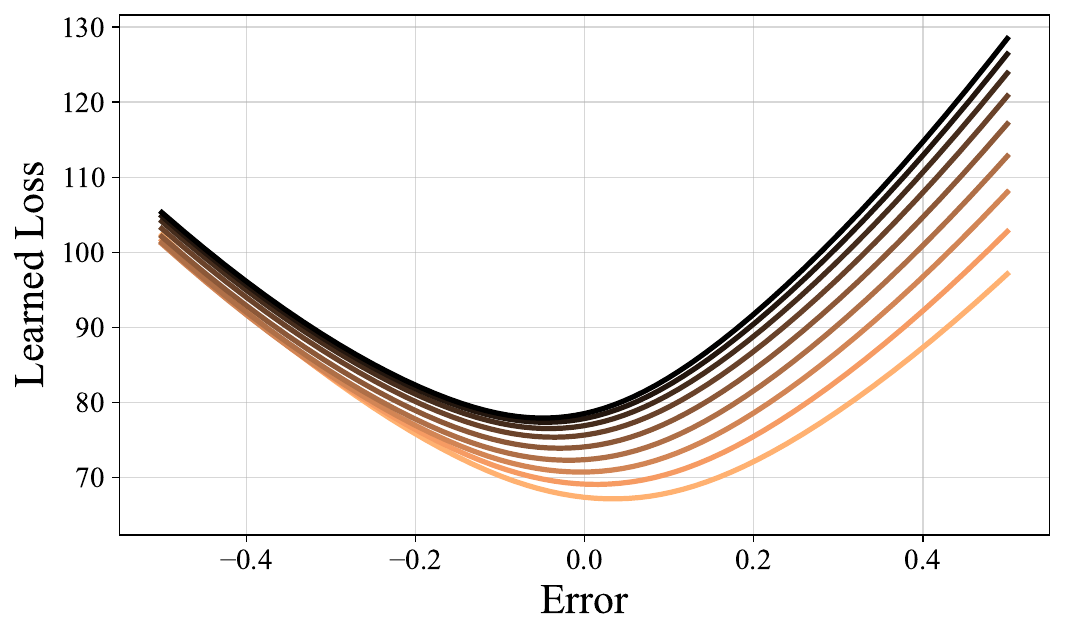}
    \end{subfigure}%
    \hfill
    \begin{subfigure}{0.5\textwidth}
        \centering
        \includegraphics[width=1\textwidth]{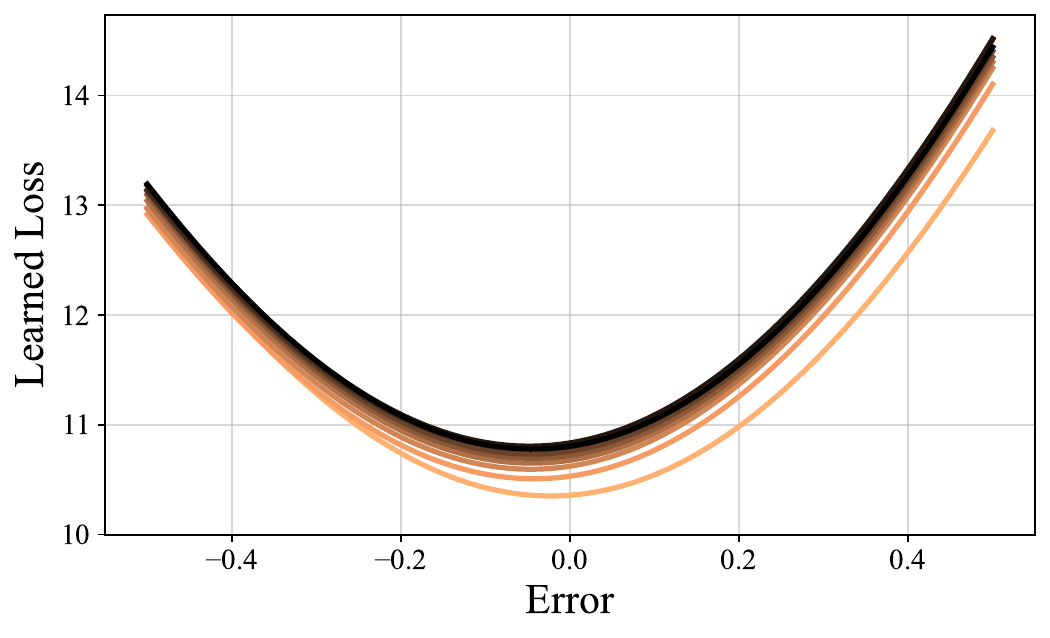}
    \end{subfigure}%

    \vspace{2mm}

    \begin{subfigure}{0.5\textwidth}
        \centering
        \includegraphics[width=1\textwidth]{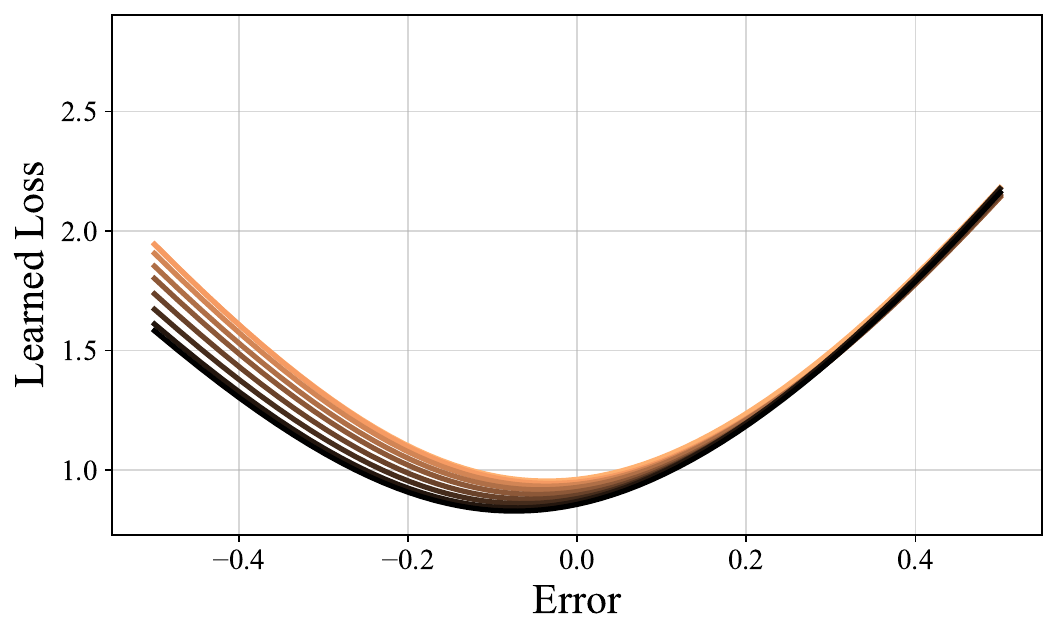}
    \end{subfigure}%
    \hfill
    \begin{subfigure}{0.5\textwidth}
        \centering
        \includegraphics[width=1\textwidth]{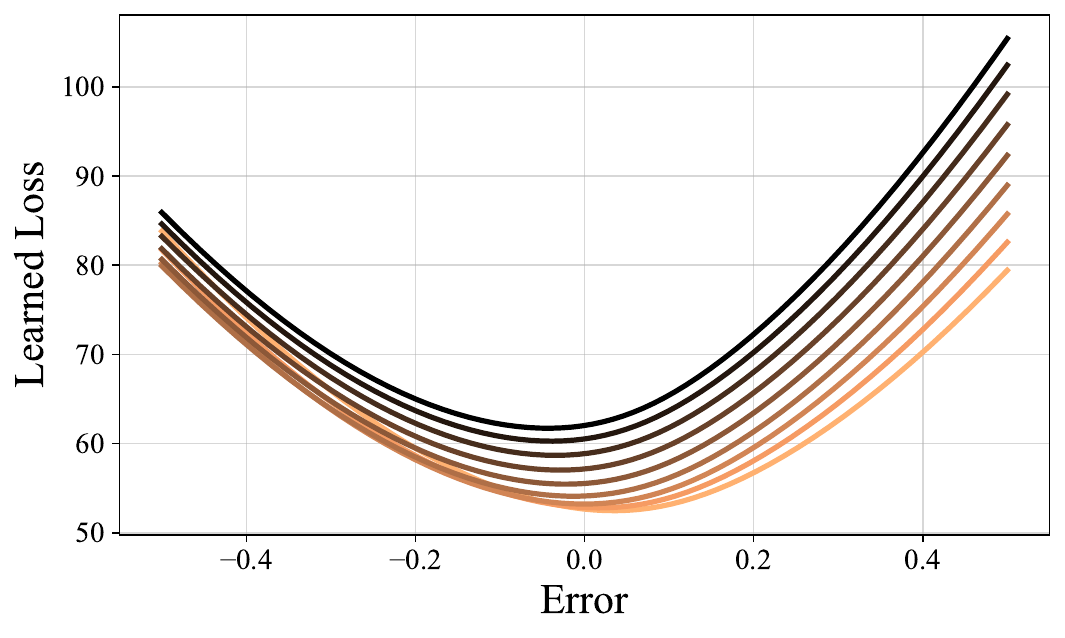}
    \end{subfigure}%

    \vspace{2mm}

    \begin{subfigure}{0.5\textwidth}
        \centering
        \includegraphics[width=1\textwidth]{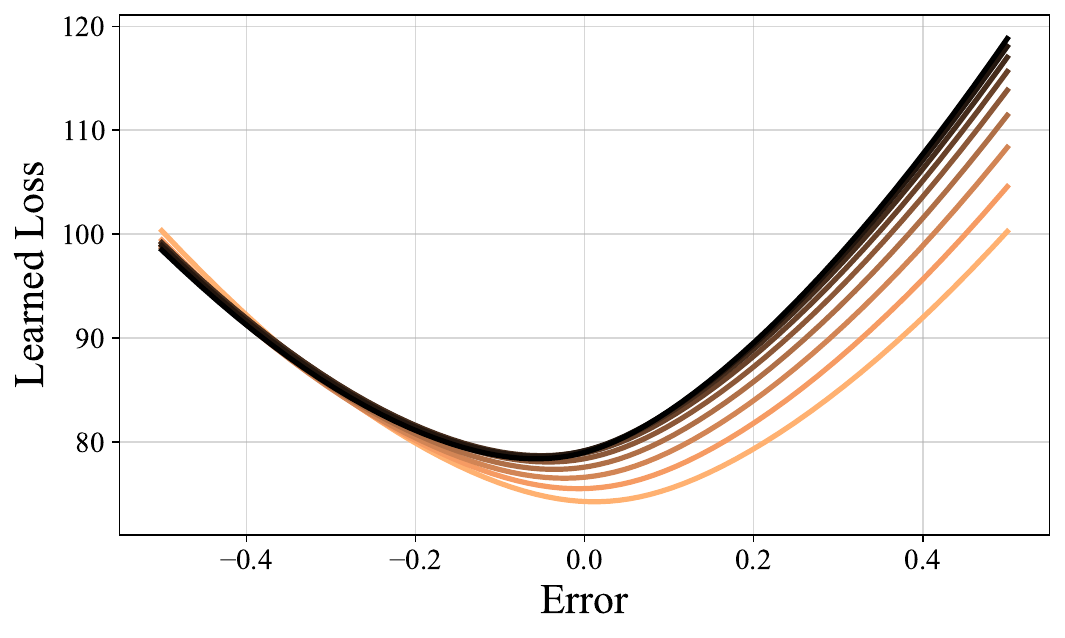}
    \end{subfigure}%
    \hfill
    \begin{subfigure}{0.5\textwidth}
        \centering
        \includegraphics[width=1\textwidth]{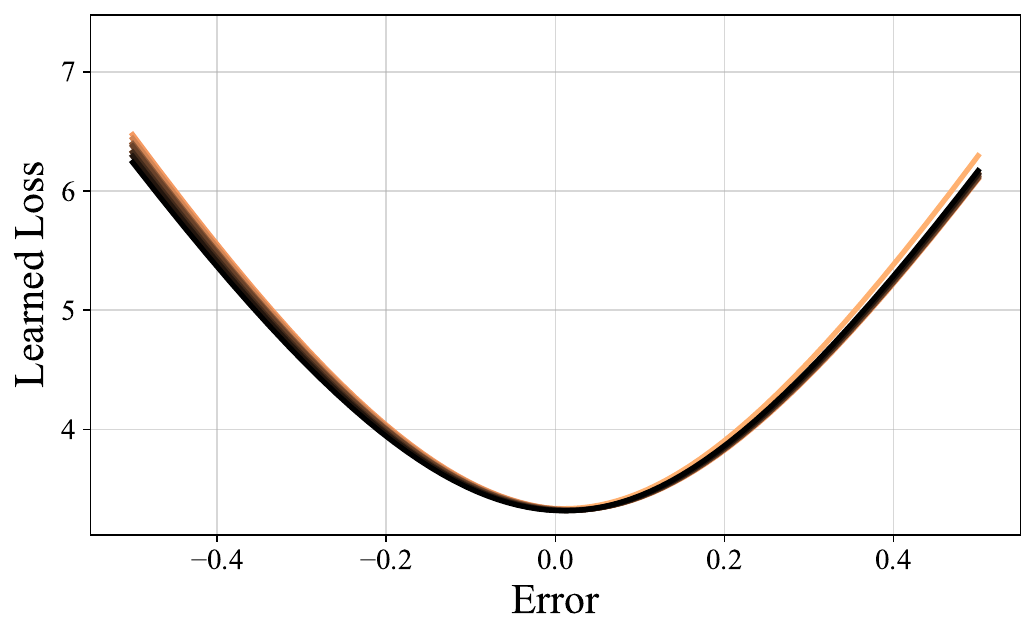}
    \end{subfigure}%
    
    \vspace{2mm}

    \begin{subfigure}{0.6\textwidth}
        \centering
        \includegraphics[width=1\textwidth]{chapter-5/figures/loss-functions/cbar-horizontal-10000.pdf}
    \end{subfigure}%

\captionsetup{justification=centering}
\caption{Regression loss functions generated by AdaLFL on the Communities and Crime, Diabetes, and California Housing datasets, where each plot represents a loss function, and the color represents the current gradient step.}
\label{figure:loss-functions-1}

\end{figure}

\begin{figure}[]
\centering

    \centering
    \begin{subfigure}{0.5\textwidth}
        \centering
        \includegraphics[width=1\textwidth]{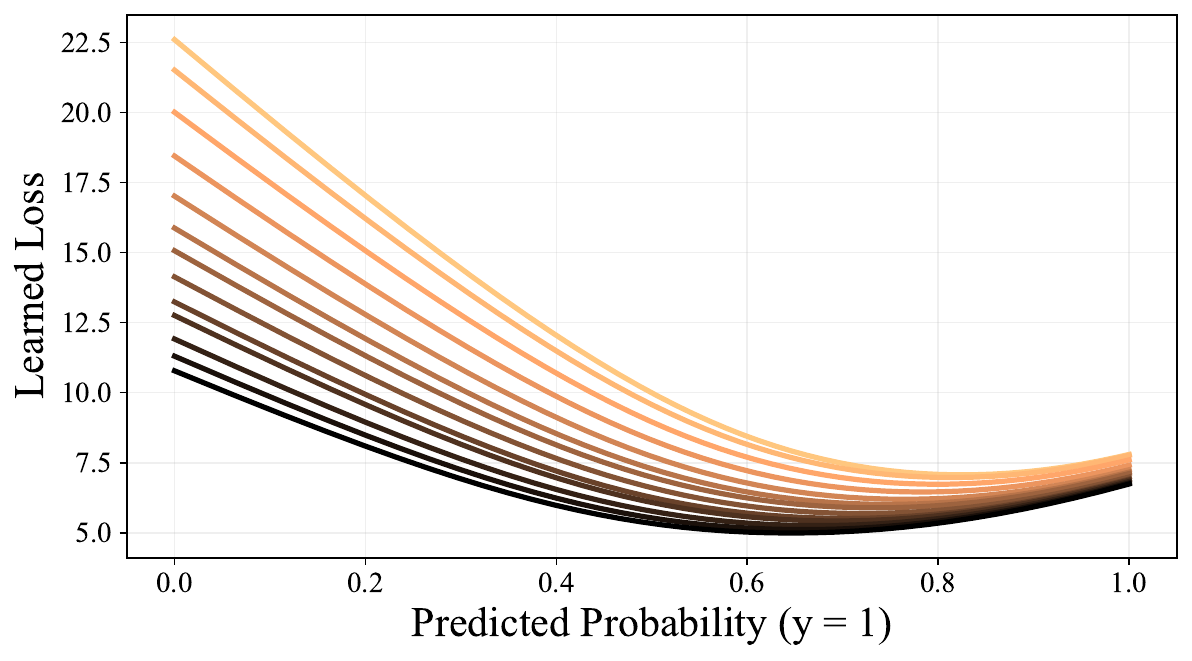}
    \end{subfigure}%
    \hfill
    \begin{subfigure}{0.5\textwidth}
        \centering
        \includegraphics[width=1\textwidth]{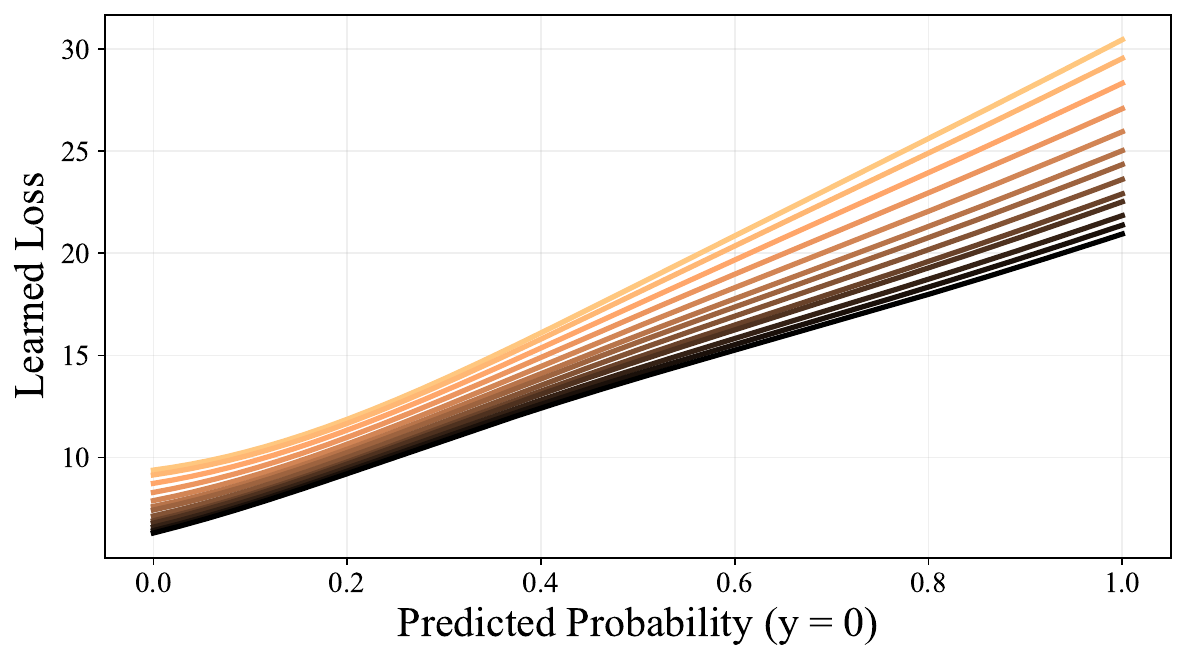}
    \end{subfigure}%

    \vspace{2mm}
    
    \begin{subfigure}{0.5\textwidth}
        \centering
        \includegraphics[width=1\textwidth]{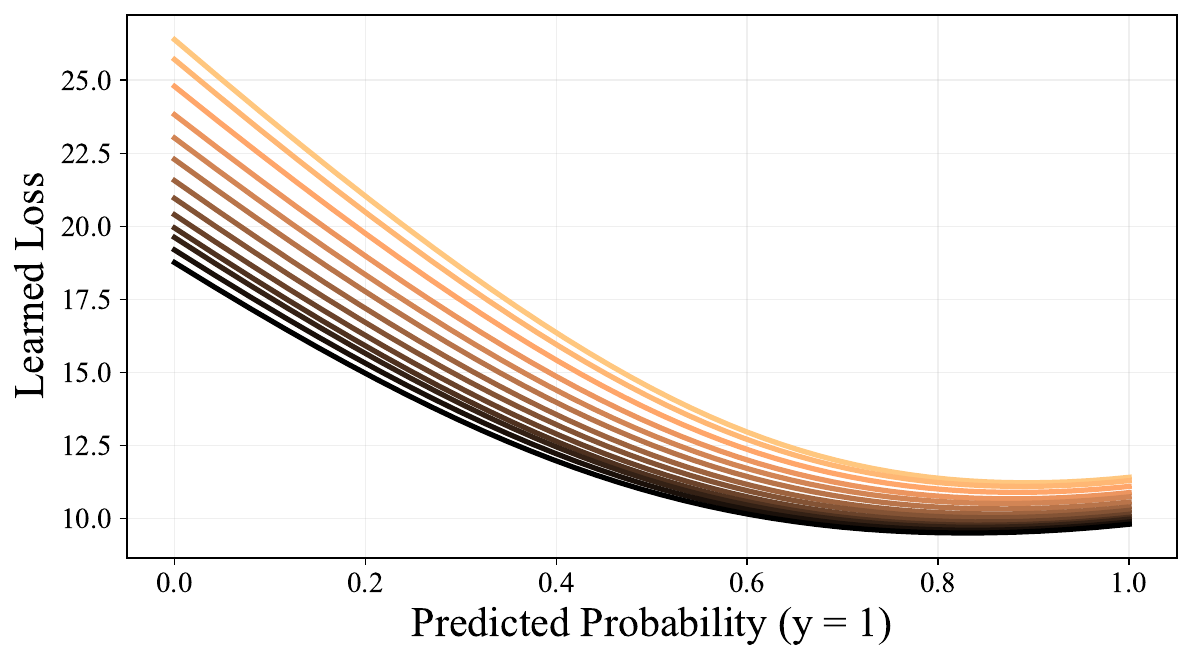}
    \end{subfigure}%
    \hfill
    \begin{subfigure}{0.5\textwidth}
        \centering
        \includegraphics[width=1\textwidth]{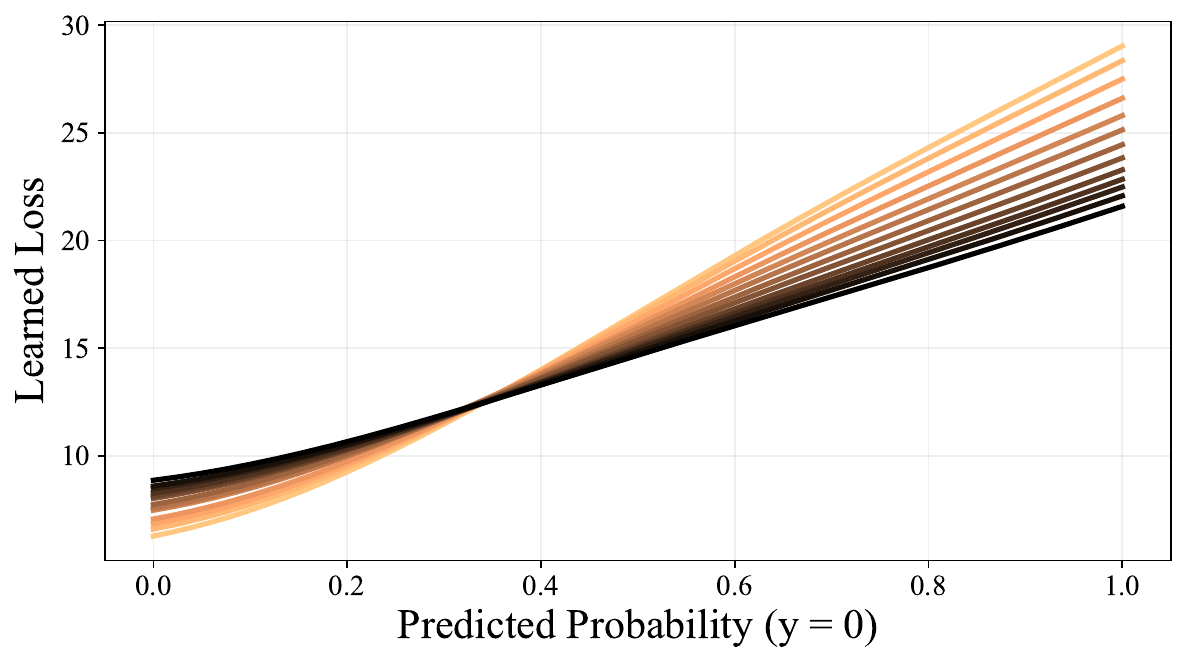}
    \end{subfigure}%
    
    \vspace{2mm}

    \begin{subfigure}{0.5\textwidth}
        \centering
        \includegraphics[width=1\textwidth]{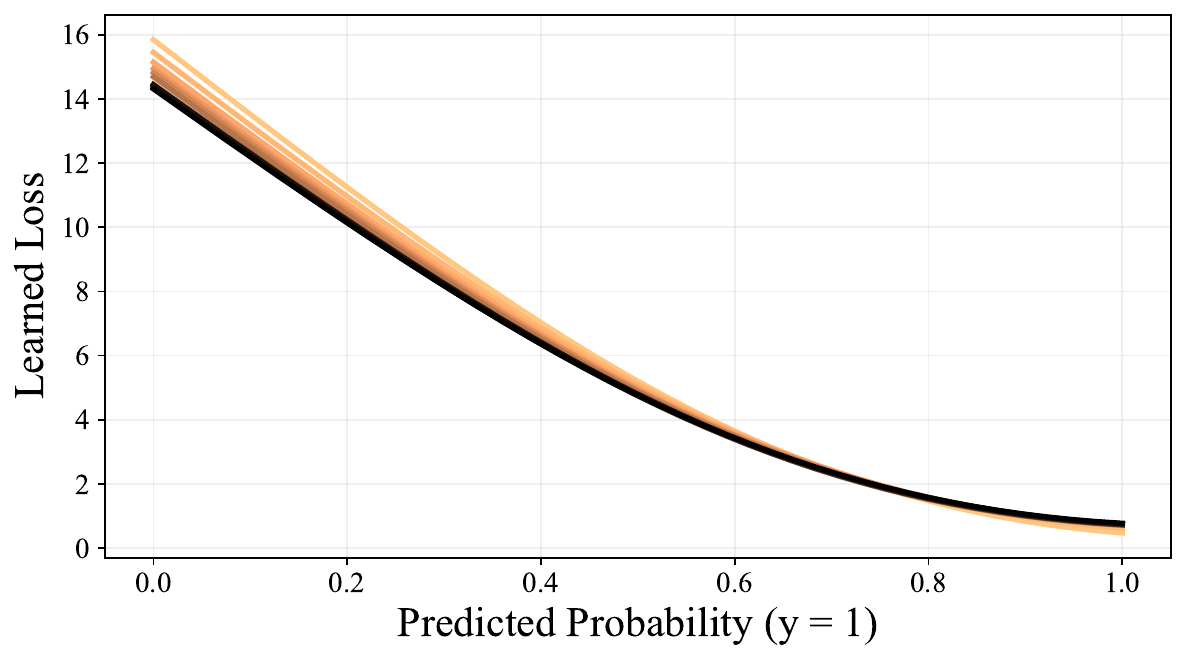}
    \end{subfigure}%
    \hfill
    \begin{subfigure}{0.5\textwidth}
        \centering
        \includegraphics[width=1\textwidth]{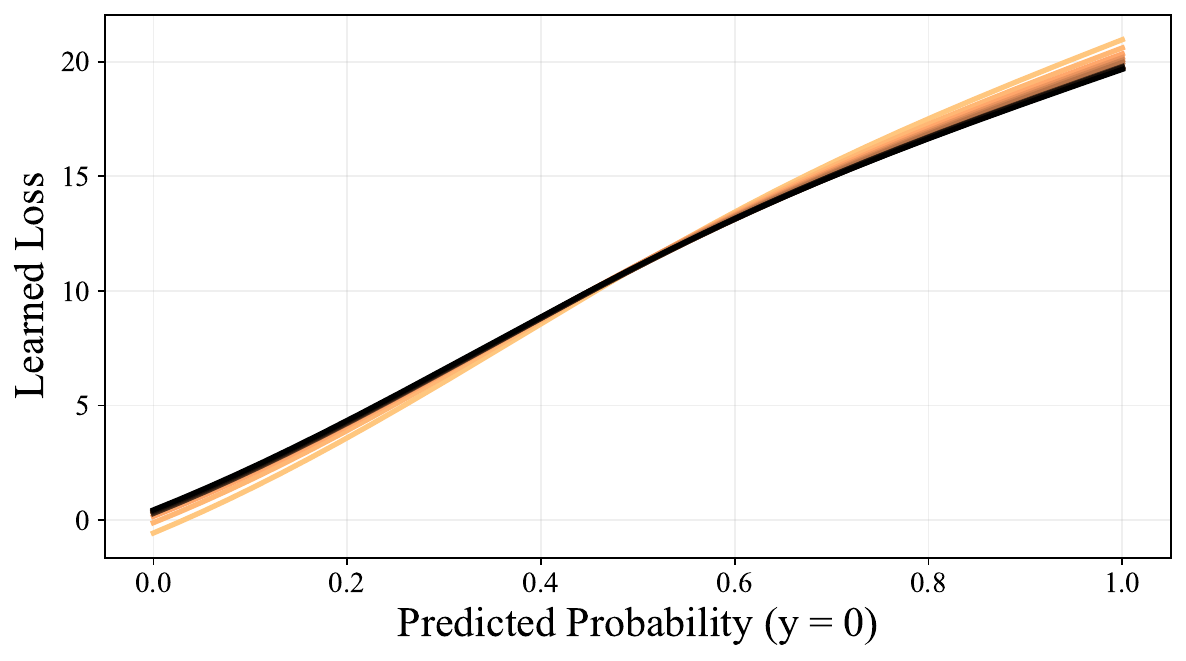}
    \end{subfigure}%
    
    \vspace{2mm}

    \begin{subfigure}{0.5\textwidth}
        \centering
        \includegraphics[width=1\textwidth]{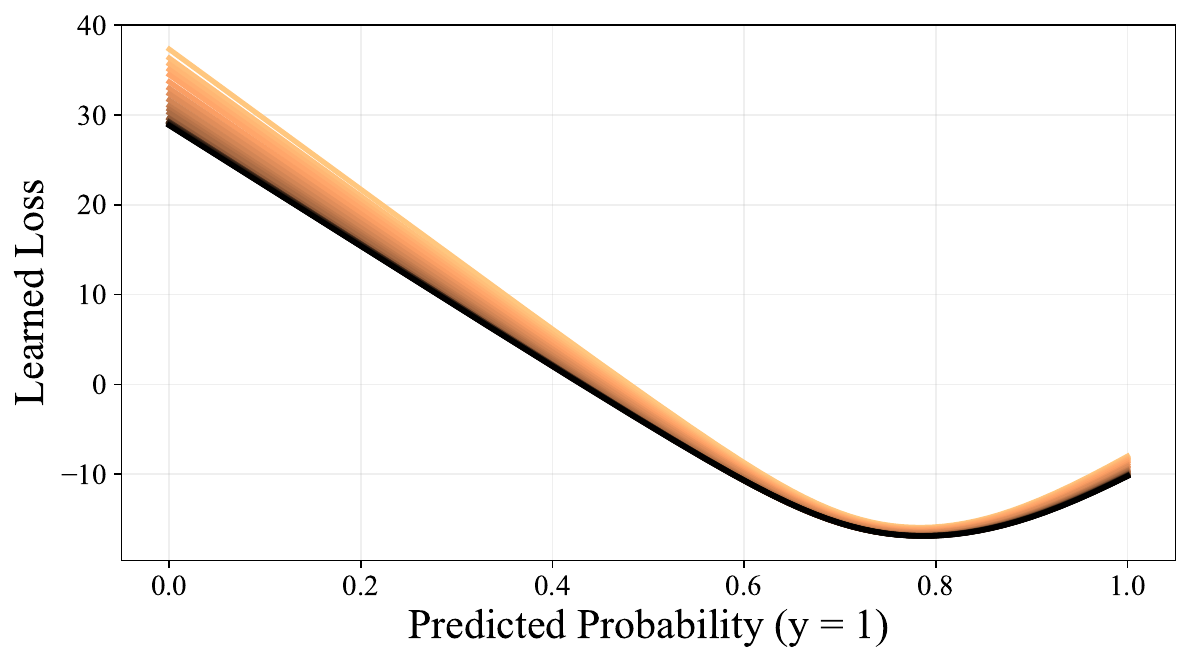}
    \end{subfigure}%
    \hfill
    \begin{subfigure}{0.5\textwidth}
        \centering
        \includegraphics[width=1\textwidth]{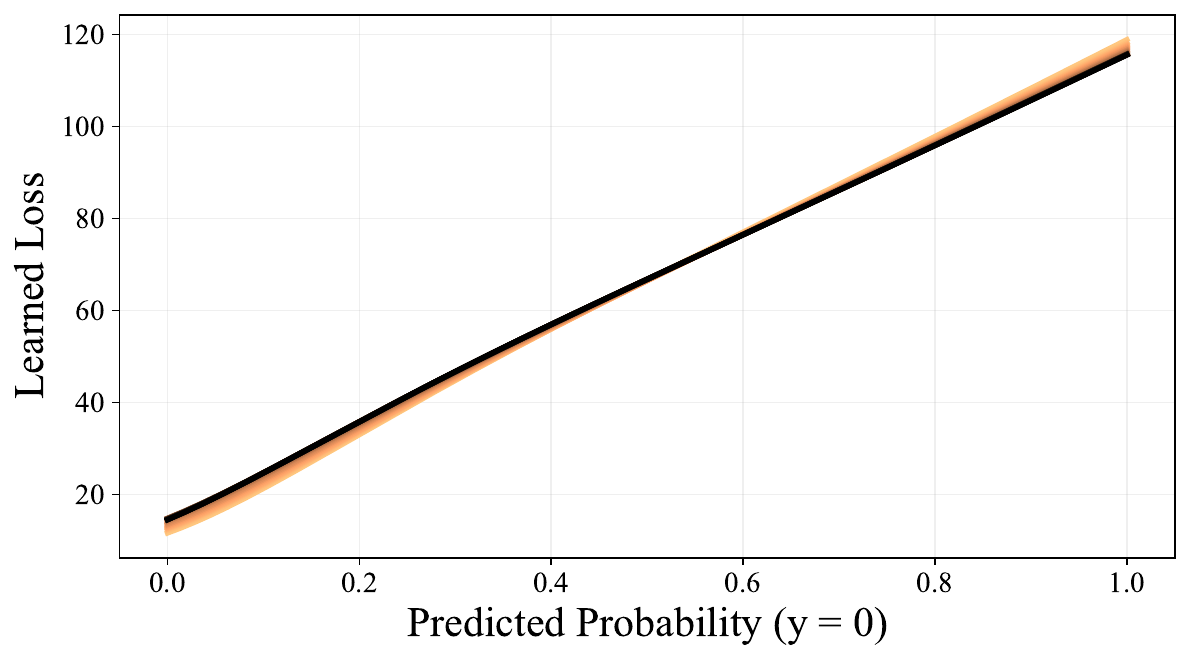}
    \end{subfigure}%
    
    \vspace{2mm}

    \begin{subfigure}{0.6\textwidth}
        \centering
        \includegraphics[width=1\textwidth]{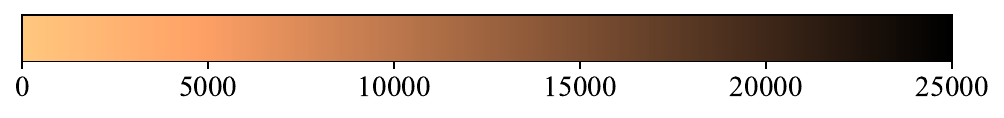}
    \end{subfigure}%

\captionsetup{justification=centering}
\caption{Classification loss functions generated by AdaLFL on the MNIST dataset, where each row represents a loss function, and the color represents the gradient step.}
\label{figure:loss-functions-2}

\end{figure}

\begin{figure}[]
\centering

    \centering
    \begin{subfigure}{0.5\textwidth}
        \centering
        \includegraphics[width=1\textwidth]{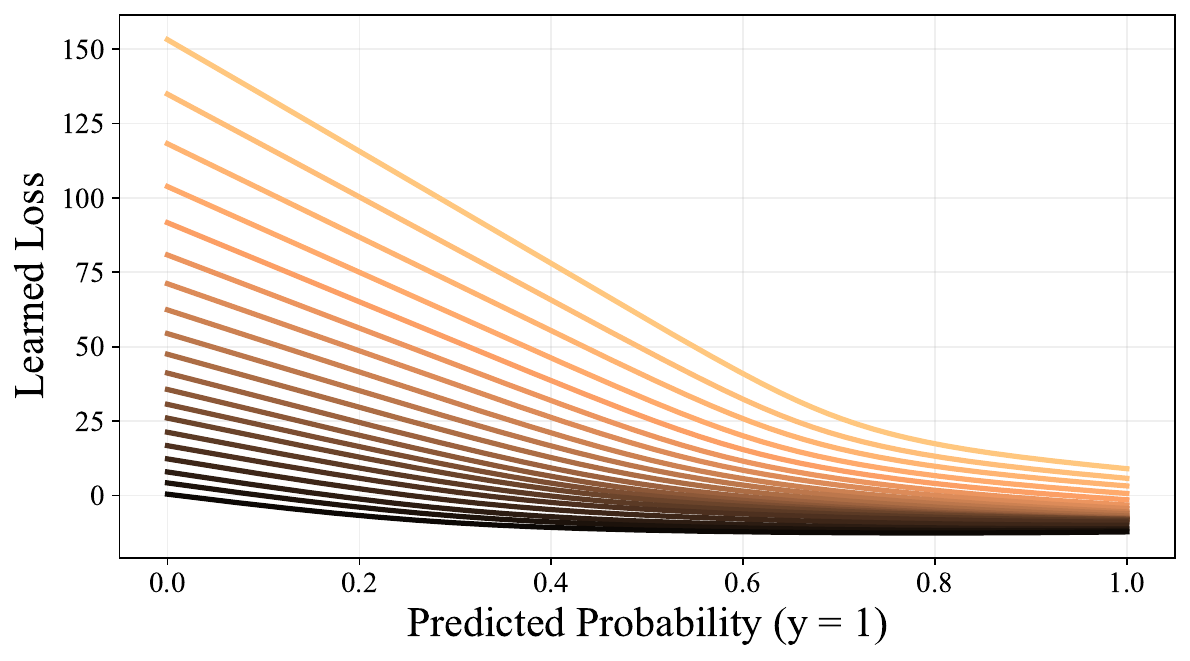}
    \end{subfigure}%
    \hfill
    \begin{subfigure}{0.5\textwidth}
        \centering
        \includegraphics[width=1\textwidth]{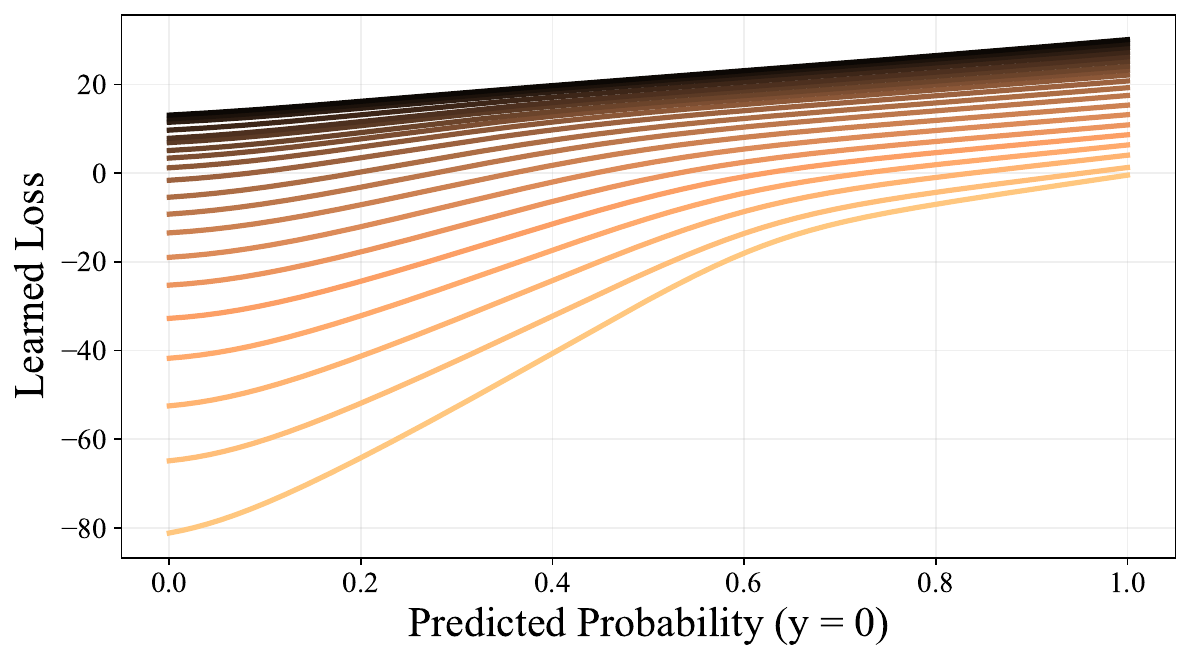}
    \end{subfigure}%

    \vspace{2mm}

    \begin{subfigure}{0.5\textwidth}
        \centering
        \includegraphics[width=1\textwidth]{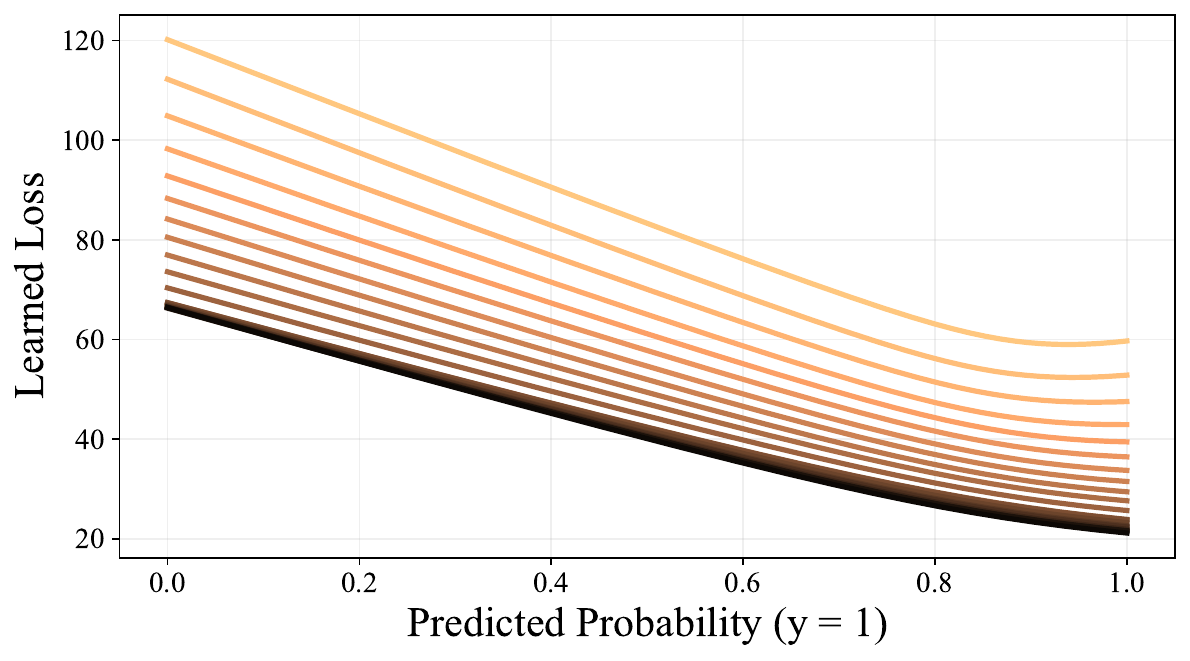}
    \end{subfigure}%
    \hfill
    \begin{subfigure}{0.5\textwidth}
        \centering
        \includegraphics[width=1\textwidth]{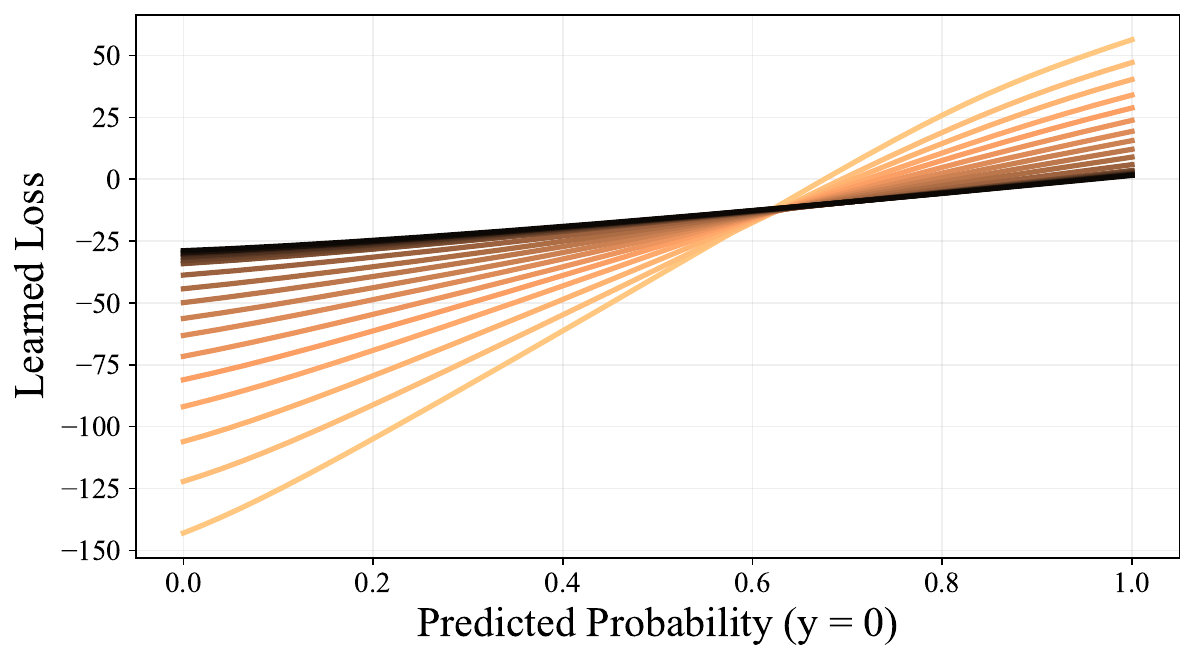}
    \end{subfigure}%

    \vspace{2mm}

    \begin{subfigure}{0.5\textwidth}
        \centering
        \includegraphics[width=1\textwidth]{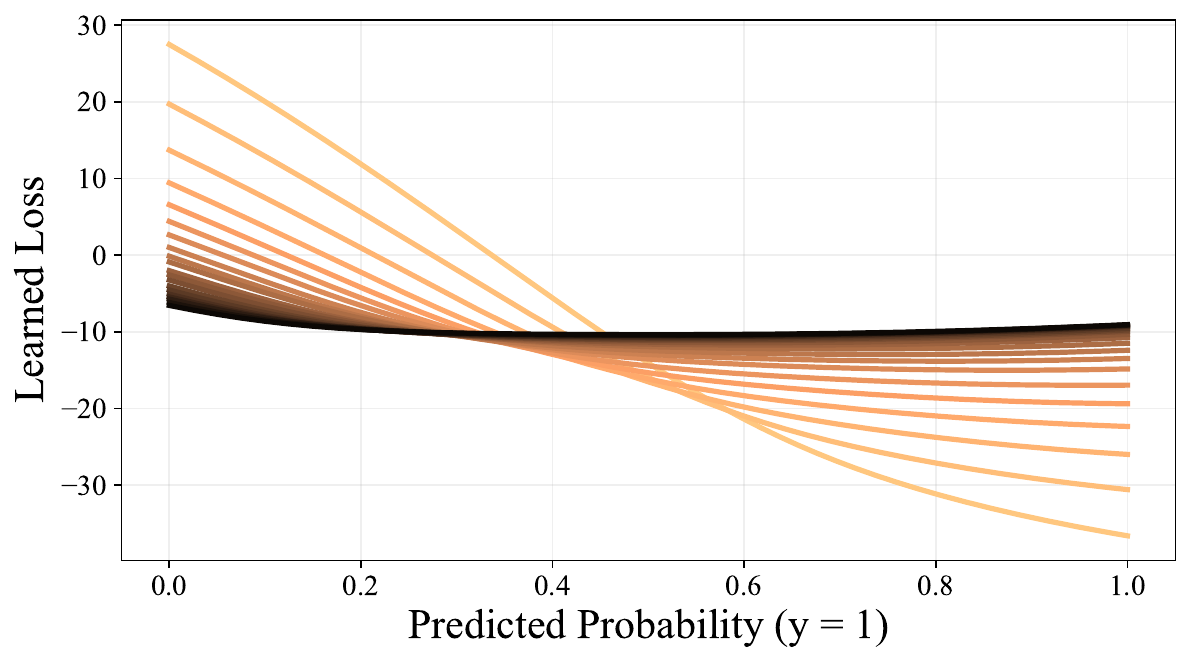}
    \end{subfigure}%
    \hfill
    \begin{subfigure}{0.5\textwidth}
        \centering
        \includegraphics[width=1\textwidth]{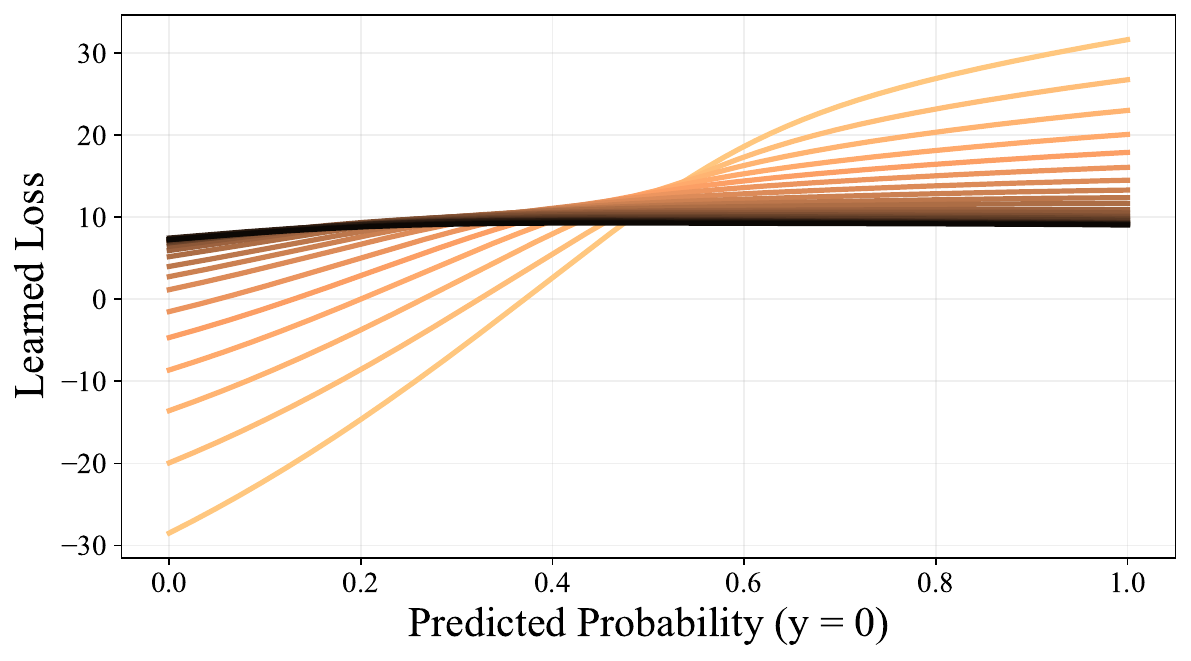}
    \end{subfigure}%

    \vspace{2mm}

    \begin{subfigure}{0.5\textwidth}
        \centering
        \includegraphics[width=1\textwidth]{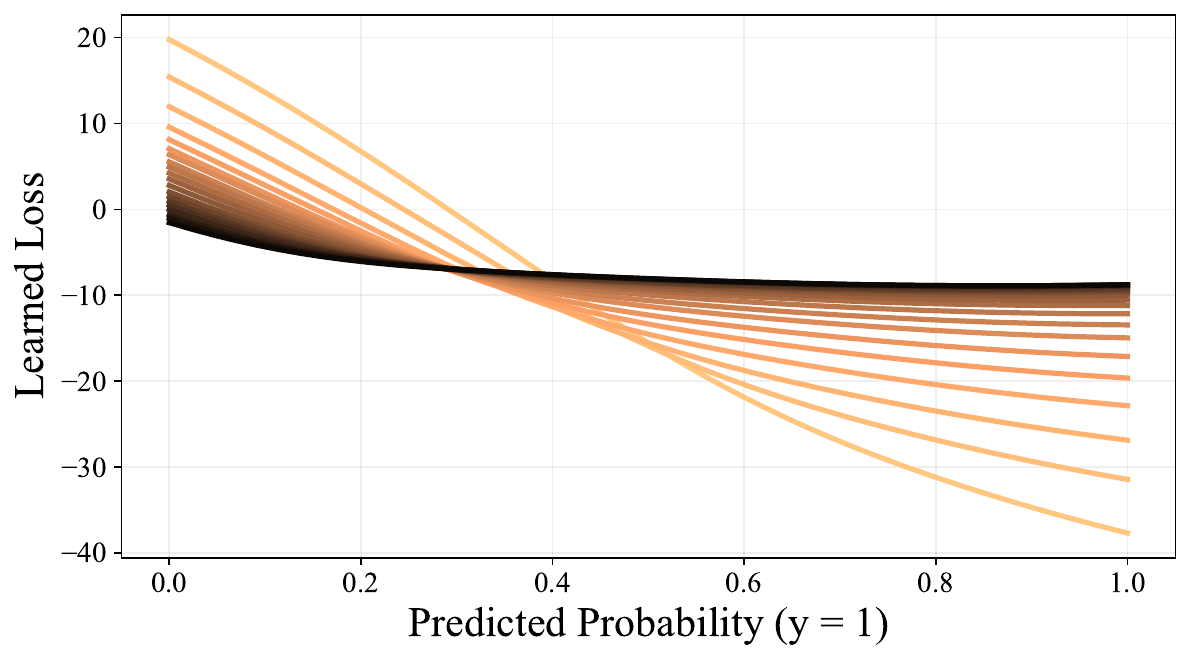}
    \end{subfigure}%
    \hfill
    \begin{subfigure}{0.5\textwidth}
        \centering
        \includegraphics[width=1\textwidth]{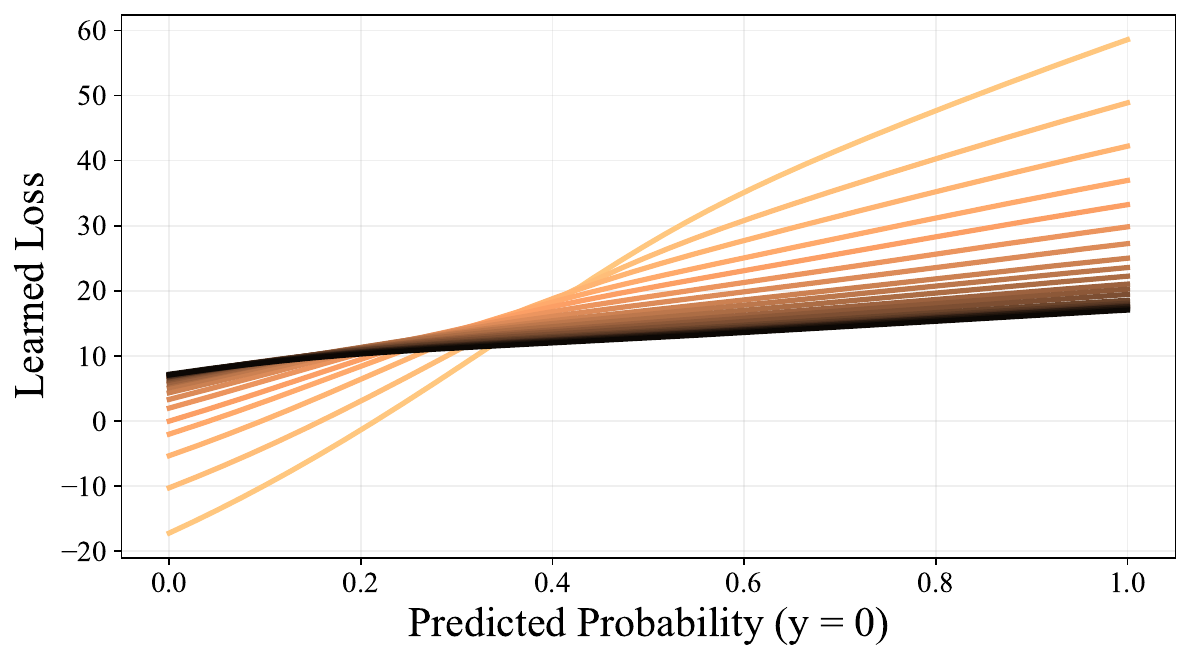}
    \end{subfigure}%

    \vspace{2mm}

    \begin{subfigure}{0.6\textwidth}
        \centering
        \includegraphics[width=1\textwidth]{chapter-5/figures/loss-functions/cbar-horizontal-100000.pdf}
    \end{subfigure}%

\captionsetup{justification=centering}
\caption{Classification loss functions generated by AdaLFL on the CIFAR-10 dataset, where each row represents a loss function, and the color represents the gradient step.}
\label{figure:loss-functions-3}

\end{figure}

To better understand why AdaLFL is so performant, an assortment of learned regression and classification loss functions are presented in Figures \ref{figure:loss-functions-1}, \ref{figure:loss-functions-2}, and \ref{figure:loss-functions-3}. First, examining the learned regression losses in Figures \ref{figure:loss-functions-1}, we observe that AdaLFL can learn a diverse set of adaptive loss functions (note the varying y-axes). While some of these loss functions take on forms similar to the $L2$ or pseudo-Huber loss, many start symmetric around 0 and later shift to become slightly asymmetric. We hypothesize that this phenomenon is due to the learned loss functions adapting to inliers or outliers in the data, and calibrating the pressure of accurately making predictions on those instances.

In regards to the learned classification loss functions shown in Figures \ref{figure:loss-functions-2} and \ref{figure:loss-functions-3}, we see a common pattern in the loss functions of attributing strong penalties for severe misclassification at the start of the learning process, and then gradually pivoting towards a more moderate or minor penalty as learning progressed --- a sort of adaptive focal loss. This behavior enables fast and efficient learning early on and reduces the sensitivity of the base model to outliers in the later stages of the learning process. Furthermore, we can also see in Figure \ref{figure:loss-functions-2} some behavior (in the left plots where $y=1$) which is reminiscent of the label smoothing regularization seen and analyzed in the previous chapter.

\subsection{Implicit Learning Rate Scheduling}
\label{section:implicit-learning-rate-schedule}

In offline loss function learning, it is known from \citep{raymond2023fast, raymond2023learning} that there is implicit initial learning rate tuning of $\alpha$ when meta-learning a loss function since $\exists\alpha\exists\phi:\theta - \alpha\nabla_{\theta}\Loss^{meta} \approx \theta - \nabla_{\theta}\MetaLoss_{\phi}$, where the learned loss function not only learns shape but also learns scale. Consequently, an emergent behavior, unique to online loss function learning, is that the adaptive loss function generated by AdaLFL implicitly embodies multiple different learning rates throughout the learning process hence often causing a fine-tuning of the fixed learning rate or a predetermined learning rate schedule. Analyzing the learned loss functions in Figure \ref{figure:loss-functions-2}, it can be observed that the scale of the learned loss function changes, confirming that implicit learning rate scheduling is occurring over time.

\subsection{Implicit Early Stopping Regularization}
\label{section:implicit-early-stopping}

Another unique property observed in the loss functions generated by AdaLFL is that often once base convergence is achieved the learned loss function will \textit{intentionally} start to flatten or take on a parabolic form, see Figures \ref{figure:loss-functions-2} and \ref{figure:loss-functions-3}. This is implicitly a type of early stopping, also observed in related paradigms such as in hypergradient descent \citep{baydin2018hypergradient}, which meta-learns base learning rates. In hypergradient descent the learned learning rate has been observed to oscillate around 0 near the end of training, at times becoming negative, essentially terminating training. Implicit early stopping is beneficial as it is known to have a regularizing effect on model training \citep{yao2007early}.

\section{Further Analysis}

\subsection{Learned Loss Function vs Learned Learning Rate}

\begin{algorithm}[t]

\SetAlgoLined
\DontPrintSemicolon
\SetKwInput{Input}{Input}

\caption{Learning Rate Initialization (Offline)}
\label{algorithm:meta-lr-offline}

\BlankLine
\Input{
    $\Loss^{meta} \leftarrow$ Task loss function (meta-objective)\;
    \vspace{-2mm}
    \hrulefill
}

\BlankLine
$\alpha_{0} \leftarrow$ Initialize the base learning rate\;
\For{$t \in \{0, ... , \mathcal{S}^{init}\}$}{
    $\theta_{0} \leftarrow$ Reset parameters of base learner\;
    \For{$i \in \{0, ... , \mathcal{S}^{inner}\}$}{
        $X$, $y$ $\leftarrow$ Sample from $\Dataset^{train}$\;
        $\theta_{i + 1} \leftarrow \theta_{i} - \alpha \nabla_{\theta_{i}}\Loss^{meta}(y, f_{\theta_{i}}(X))$\;
    }
    $X$, $y$ $\leftarrow$ Sample from $\Dataset^{valid}$\;
    $\Loss^{task} \leftarrow \Loss^{meta}(y, f_{\theta_{i + 1}}(X))$\;
    $\alpha_{t+1} \leftarrow \alpha_{t} - \eta\nabla_{\alpha_{\phi_{t}}}\Loss^{task}$\;
}

\BlankLine

\end{algorithm}

\begin{algorithm}[t!]

\caption{Learning Rate Adaptation (Online)}
\label{algorithm:meta-lr-online}

\SetAlgoLined
\DontPrintSemicolon
\SetKwInput{Input}{Input}

\vspace{1mm}
\Input{
    $\Loss^{meta} \leftarrow$ Task loss function (meta-objective) \newline
    $\alpha_{0} \leftarrow$ Initial learned learning rate\;
    \vspace{-2mm}
    \hrulefill
}
\BlankLine
$\theta_{0} \leftarrow$ Initialize parameters of base learner\;
    
\For{$t \in \{0, ... , \mathcal{S}_{train}\}$}{
    $X$, $y$ $\leftarrow$ Sample from $\Dataset^{train}$\;
    $\theta_{t+1} \leftarrow \theta_{t} - \alpha \nabla_{\theta_{t}} \Loss^{meta}(y, f_{\theta_{t}}(X))$\;
    $X$, $y$ $\leftarrow$ Sample from $\Dataset^{valid}$\;
    $\Loss^{task} \leftarrow \Loss^{meta}(y, f_{\theta_{t+1}}(X))$\;
    $\alpha_{t+1} \leftarrow \alpha_{t} - \eta \nabla_{\alpha_{t}}\Loss^{task}$\;
}

\BlankLine

\end{algorithm}

\begin{table}[htb!]
\centering

\captionsetup{justification=centering}
\caption{Results comparing meta-learning a learning rate schedule vs an adaptive loss function. Reporting the mean $\pm$ standard deviation of final inference testing performance across 10 independent executions of each algorithm on each task + model pair.}

\begin{threeparttable}

\begin{tabular}{p{3cm}>{\centering\arraybackslash}p{4cm}>{\centering\arraybackslash}p{4cm}}

\hline \noalign{\vskip 1mm}
Task and Model  &  Meta-LR (Online)  &  AdaLFL (Online)
\\ \hline \noalign{\vskip 1mm} \hline \noalign{\vskip 1mm}

\textbf{Crime}              &                   &                                     \\
MLP \tnote{1}               & 0.0274$\pm$0.0018 & \textbf{0.0263$\pm$0.0023}          \\ \noalign{\vskip 1mm}

\hline \noalign{\vskip 1mm} 

\textbf{Diabetes}           &                   &                                     \\
MLP \tnote{1}               & 0.0463$\pm$0.0013 & \textbf{0.0420$\pm$0.0014}          \\ \noalign{\vskip 1mm}

\hline \noalign{\vskip 1mm} 

\textbf{California}         &                   &                                     \\
MLP \tnote{1}               & 0.0154$\pm$0.0004 & \textbf{0.0151$\pm$0.0007}          \\ \noalign{\vskip 1mm}

\hline \noalign{\vskip 1mm} 

\textbf{MNIST}              &                   &                                     \\
Logistic \tnote{2}          & 0.0756$\pm$0.0008 & \textbf{0.0697$\pm$0.0010}          \\ \noalign{\vskip 1mm}
MLP \tnote{1}               & 0.0192$\pm$0.0007 & \textbf{0.0184$\pm$0.0006}          \\ \noalign{\vskip 1mm}
LeNet-5 \tnote{3}           & 0.0097$\pm$0.0013 & \textbf{0.0091$\pm$0.0004}          \\ \noalign{\vskip 1mm}

\hline \noalign{\vskip 1mm} 

\textbf{CIFAR-10}          &                   &                                     \\
VGG-16 \tnote{4}           & 0.0966$\pm$0.0087 & \textbf{0.0903$\pm$0.0032}          \\ \noalign{\vskip 1mm}
AllCNN-C \tnote{5}         & \textbf{0.0656$\pm$0.0017} & 0.0835$\pm$0.0050          \\ \noalign{\vskip 1mm}
ResNet-18 \tnote{6}        & 0.0866$\pm$0.0056 & \textbf{0.0788$\pm$0.0035}          \\ \noalign{\vskip 1mm}
SqueezeNet \tnote{7}       & 0.1173$\pm$0.0065 & \textbf{0.1083$\pm$0.0049}          \\ \noalign{\vskip 1mm}

\hline \noalign{\vskip 1mm} 

\textbf{CIFAR-100}         &                   &                                     \\   
WRN 28-10 \tnote{8}        & \textbf{0.2288$\pm$0.0019} & 0.2668$\pm$0.0283          \\ \noalign{\vskip 1mm}

\hline \noalign{\vskip 1mm} 

\textbf{SVHN}              &                   &                                     \\                                WRN 16-8 \tnote{8}         & \textbf{0.0367$\pm$0.0007} & 0.0441$\pm$0.0014          \\ \noalign{\vskip 1mm}

\hline \noalign{\vskip 1mm} 
\end{tabular}

\begin{tablenotes}\centering
\tiny Network architecture references: \item[1] \citep{baydin2018hypergradient} \item[2] \citep{mccullagh2019generalized} \item[3] \citep{lecun1998gradient} \item[4] \citep{simonyan2014very} \item[5] \citep{springenberg2014striving} \item[6] \citep{he2016deep} \item[7] \citep{iandola2016squeezenet} \item[8] \citep{zagoruyko2016wide}
\end{tablenotes}
\end{threeparttable}

\label{table:learned-learning-rate}
\end{table}

Given the close relationship between meta-learning a loss function and learning a scalar learning rate/learning rate schedule, it is important to compare and contrast the performance of AdaLFL to a method for meta-learning a base learning rate schedule. To construct a fair comparison, we use an identical meta-learning setup to AdaLFL, which allows us to control for the meta-optimization algorithms (offline and online unrolled differentiation) and their respective hyperparameters. The algorithm, which we further refer to as \textit{Meta-LR} is presented in Algorithms \ref{algorithm:meta-lr-offline} and \ref{algorithm:meta-lr-online}, and uses an offline initialization process to find the best initial learning rate, followed by an online process to adapt the learning rate at each gradient step in lockstep with the base model's parameters.

The results of our experiments are presented in Table \ref{table:learned-learning-rate}, which shows that on most of the tested tasks, AdaLFL can obtain superior performance compared to Meta-LR. This empirically suggests that meta-learned loss functions can learn meta-information that cannot be directly captured through only meta-learning a base learning rate, a result similar to the one found in \citep{raymond2023learning}. Interestingly, on a few of the datasets we find that Meta-LR can achieve improved performance, which we hypothesize is due to the learning rate settings (taken from prior works \citep{gonzalez2021optimizing, raymond2023learning}) used in our experiments being sub-optimal, resulting in explicitly learning the learning rate being more effective.

\subsection{Ablation of Loss Function Representations}

\begin{table*}[t!]
\centering

\captionsetup{justification=centering}
\caption{Experimental results exploring alternative loss function representations based on Taylor polynomial parameterizations reporting the mean $\pm$ standard deviation of final inference testing mean squared error or error rate across 10 independent executions of each algorithm on each task + model pair.}

\begin{threeparttable}

\begin{tabular}{lccc}

\hline \noalign{\vskip 1mm}
Task and Model  &  Quadratic-TP (Online)  &  Cubic-TP (Online)  &  AdaLFL (Online)   
\\ \hline \noalign{\vskip 1mm} \hline \noalign{\vskip 1mm}

\textbf{Crime}             &                   &                   &                                    \\
MLP \tnote{1}              & - & \textbf{0.0254$\pm$0.0015} & 0.0263$\pm$0.0023         \\ \noalign{\vskip 1mm}

\hline \noalign{\vskip 1mm} 

\textbf{Diabetes}          &                   &                   &                                    \\
MLP \tnote{1}              & - & \textbf{0.0418$\pm$0.0041} & 0.0420$\pm$0.0014         \\ \noalign{\vskip 1mm}

\hline \noalign{\vskip 1mm} 

\textbf{California}        &                   &                   &                                    \\
MLP \tnote{1}              & - & 0.0783$\pm$0.0167 & \textbf{0.0151$\pm$0.0007}         \\ \noalign{\vskip 1mm}

\hline \noalign{\vskip 1mm} 

\textbf{MNIST}             &                   &                   &                                    \\
Logistic \tnote{2}         & 0.0810$\pm$0.0281 & 0.0707$\pm$0.0009 & \textbf{0.0697$\pm$0.0010}         \\ \noalign{\vskip 1mm}
MLP \tnote{1}              & 0.0205$\pm$0.0008 & 0.0185$\pm$0.0007 & \textbf{0.0184$\pm$0.0006}         \\ \noalign{\vskip 1mm}
LeNet-5 \tnote{3}          & 0.1357$\pm$0.0728 & 0.0096$\pm$0.0006 & \textbf{0.0091$\pm$0.0004}         \\ \noalign{\vskip 1mm}

\hline \noalign{\vskip 1mm} 

\textbf{CIFAR-10}          &                   &                   &                                    \\
VGG-16 \tnote{4}           & 0.1442$\pm$0.0025 & 0.1439$\pm$0.0027 & \textbf{0.0903$\pm$0.0032}         \\ \noalign{\vskip 1mm}
AllCNN-C \tnote{5}         & 0.1086$\pm$0.0100 & 0.0908$\pm$0.0020 & \textbf{0.0835$\pm$0.0050}         \\ \noalign{\vskip 1mm}
ResNet-18 \tnote{6}        & 0.1133$\pm$0.0033 & 0.1309$\pm$0.0070 & \textbf{0.0788$\pm$0.0035}         \\ \noalign{\vskip 1mm}
SqueezeNet \tnote{7}       & 0.1506$\pm$0.0092 & 0.1367$\pm$0.0041 & \textbf{0.1083$\pm$0.0049}         \\ \noalign{\vskip 1mm}

\hline \noalign{\vskip 1mm} 

\textbf{CIFAR-100}         &                   &                   &                                    \\
WRN 28-10 \tnote{8}        & 0.2952$\pm$0.0220 & 0.2934$\pm$0.0621 & \textbf{0.2668$\pm$0.0283}         \\ \noalign{\vskip 1mm}

\hline \noalign{\vskip 1mm} 

\textbf{SVHN}              &                   &                   &                                    \\
WRN 16-8 \tnote{8}         & 0.0494$\pm$0.0000 & 0.0431$\pm$0.0000 & \textbf{0.0441$\pm$0.0014}         \\ \noalign{\vskip 1mm}

\hline \noalign{\vskip 1mm} 
\end{tabular}

\begin{tablenotes}\centering
\tiny Network architecture references: \item[1] \citep{baydin2018hypergradient} \item[2] \citep{mccullagh2019generalized} \item[3] \citep{lecun1998gradient} \item[4] \citep{simonyan2014very} \item[5] \citep{springenberg2014striving} \item[6] \citep{he2016deep} \item[7] \citep{iandola2016squeezenet} \item[8] \citep{zagoruyko2016wide}
\end{tablenotes}
\end{threeparttable}

\label{table:loss-representation}
\end{table*}

In AdaLFL a two-hidden-layer feedforward neural network is used for the loss function representation, this was inspired by its use in prior studies \citep{bechtle2021meta,psaros2022meta}. We chose this representation as it has more expressive power than both quadratic and cubic Taylor polynomials, which were used in \citep{gonzalez2021optimizing} and \citep{gao2021searching,gao2022loss}, respectively. Although the best representation for learned loss functions is not under investigation; it is important to note that the proposed method of online meta-optimization discussed in Section \ref{sec:adalfl-online-optimization} makes no assumptions about the underlying representation used for the learned loss function. Therefore, alternative representations can be used in AdaLFL. 

In Table \ref{table:loss-representation}, results comparing and contrasting the performance between different learned loss function representations are presented. Specifically, we contrast the performance of AdaLFL which uses a feed-forward neural network (NN) with smooth leaky ReLU activations against the aforementioned quadratic and cubic Taylor polynomials (TP) representation. The results show that the NN representation has on average the best performance and consistency in contrast to quadratic and cubic Taylor polynomials, with better performance and very little variance between independent executions on all datasets except the two small regression datasets Crime and Diabetes. These results demonstrate the superiority of the NN representation for learned loss functions, especially when dealing with relatively large learning tasks where expressive behavior is important. Note, that on the regression datasets, we found that the majority of the quadratic TP experiments diverged, even with hyper-parameter tuning.

\subsection{Ablation of Loss Network Activation Functions}

\begin{table*}[t!]
\centering

\captionsetup{justification=centering}
\caption{Ablating the newly proposed smooth leaky ReLU activation function, reporting the mean $\pm$ standard deviation of final inference testing mean squared error or error rate across 10 independent executions of each algorithm on each task + model pair.}

\begin{threeparttable}

\begin{tabular}{lccc}

\hline \noalign{\vskip 1mm}
Task and Model  &  ML$^3$ ReLU (Offline)  &  ML$^3$ SLReLU (Offline)  &  AdaLFL (Online)   
\\ \hline \noalign{\vskip 1mm} \hline \noalign{\vskip 1mm}

\textbf{Crime}             &                   &                   &                                    \\
MLP \tnote{1}              & 0.0270$\pm$0.0025 & 0.0274$\pm$0.0029 & \textbf{0.0263$\pm$0.0023}         \\ \noalign{\vskip 1mm}

\hline \noalign{\vskip 1mm} 

\textbf{Diabetes}          &                   &                   &                                    \\
MLP \tnote{1}              & 0.0481$\pm$0.0020 & 0.0430$\pm$0.0012 & \textbf{0.0420$\pm$0.0014}         \\ \noalign{\vskip 1mm}

\hline \noalign{\vskip 1mm} 

\textbf{California}        &                   &                   &                                    \\
MLP \tnote{1}              & 0.0346$\pm$0.0087 & 0.0276$\pm$0.0058 & \textbf{0.0151$\pm$0.0007}         \\ \noalign{\vskip 1mm}

\hline \noalign{\vskip 1mm} 

\textbf{MNIST}             &                   &                   &                                    \\
Logistic \tnote{2}         & 0.0782$\pm$0.0117 & 0.0710$\pm$0.0015 & \textbf{0.0697$\pm$0.0010}         \\ \noalign{\vskip 1mm}
MLP \tnote{1}              & 0.0167$\pm$0.0021 & 0.0185$\pm$0.0004 & \textbf{0.0184$\pm$0.0006}         \\ \noalign{\vskip 1mm}
LeNet-5 \tnote{3}          & 0.0095$\pm$0.0006 & 0.0094$\pm$0.0005 & \textbf{0.0091$\pm$0.0004}         \\ \noalign{\vskip 1mm}

\hline \noalign{\vskip 1mm} 

\textbf{CIFAR-10}          &                   &                   &                                    \\
VGG-16 \tnote{4}           & 0.1034$\pm$0.0058 & 0.1024$\pm$0.0055 & \textbf{0.0903$\pm$0.0032}         \\ \noalign{\vskip 1mm}
AllCNN-C \tnote{5}         & 0.1087$\pm$0.0174 & 0.1015$\pm$0.0055 & \textbf{0.0835$\pm$0.0050}         \\ \noalign{\vskip 1mm}
ResNet-18 \tnote{6}        & 0.0972$\pm$0.0259 & 0.0883$\pm$0.0041 & \textbf{0.0788$\pm$0.0035}         \\ \noalign{\vskip 1mm}
SqueezeNet \tnote{7}       & 0.1282$\pm$0.0086 & 0.1162$\pm$0.0052 & \textbf{0.1083$\pm$0.0049}         \\ \noalign{\vskip 1mm}

\hline \noalign{\vskip 1mm} 

\textbf{CIFAR-100}         &                   &                   &                                    \\   
WRN 28-10 \tnote{8}        & 0.3114$\pm$0.0063 & 0.3108$\pm$0.0075 & \textbf{0.2668$\pm$0.0283}         \\ \noalign{\vskip 1mm}

\hline \noalign{\vskip 1mm} 

\textbf{SVHN}              &                   &                   &                                    \\
WRN 16-8 \tnote{8}         & 0.0500$\pm$0.0034 & 0.0502$\pm$0.0032 & \textbf{0.0441$\pm$0.0014}         \\ \noalign{\vskip 1mm}

\hline \noalign{\vskip 1mm} 
\end{tabular}

\begin{tablenotes}\centering
\tiny Network architecture references: \item[1] \citep{baydin2018hypergradient} \item[2] \citep{mccullagh2019generalized} \item[3] \citep{lecun1998gradient} \item[4] \citep{simonyan2014very} \item[5] \citep{springenberg2014striving} \item[6] \citep{he2016deep} \item[7] \citep{iandola2016squeezenet} \item[8] \citep{zagoruyko2016wide}
\end{tablenotes}
\end{threeparttable}

\label{table:loss-network-activation}
\end{table*}

An important difference between AdaLFL's neural network representation and prior neural network-based learned loss function representation such as the one used in ML$^3$, is the use of the smooth leaky ReLU activation functions presented in Section \ref{sec:adalfl-smooth-leaky-relu}. This new activation function resolves the issue with the prior network design, which caused unintentional flat regions in the loss function; however, it remains to be seen how much of the performance improvement can be attributed to the newly proposed smooth leaky ReLU activation function vs the newly proposed online optimization algorithm.

In Table \ref{table:loss-network-activation}, results are presented comparing and contrasting the performance between offline loss function learning (\textit{i.e.}, ML$^3$) with the standard ReLU + SoftPlus network architecture, and the new smooth leaky ReLU network architecture. The results show that on most tasks the new activation function improves performance compared to the conventional architecture used in ML$^3$. However, this performance improvement is not a significant contributing factor compared to the change in optimization algorithm, \textit{i.e.}, going from offline to online meta-learning.

This result is due to offline loss function learning algorithms not being prone to unintentional flattening since they are optimized from a random initialization using only one or a few training steps with the base model. As the base model is never close to converging, the loss function is unlikely to be flat since that would in most cases result in a very bad loss at the end of the short training trajectory as measured by $\Loss^{meta}$. Hence, the new smooth leaky ReLU activation function only modestly improves the final learning performance as shown in Table \ref{table:loss-network-activation}.

\subsubsection{Ablation of Meta-Objective}

\begin{figure}[t]

    \centering
    \begin{subfigure}{0.5\textwidth}
        \centering
        \includegraphics[width=1\textwidth]{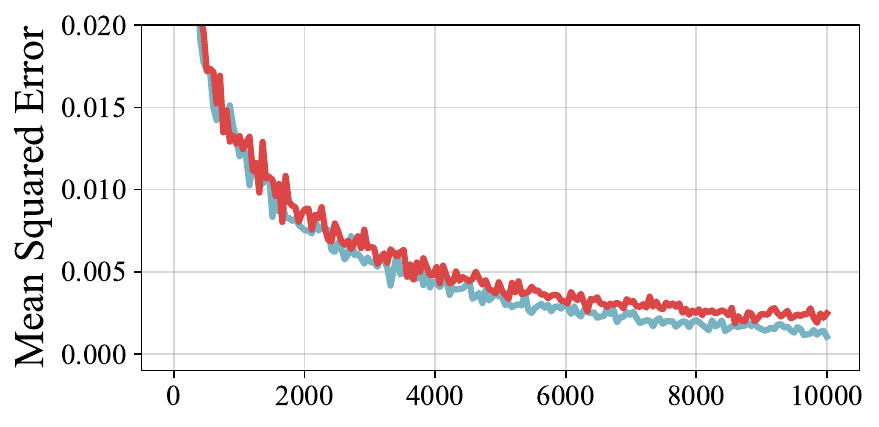}
        \caption{Crime + MLP}
    \end{subfigure}%
    \hfill
    \begin{subfigure}{0.5\textwidth}
        \centering
        \includegraphics[width=1\textwidth]{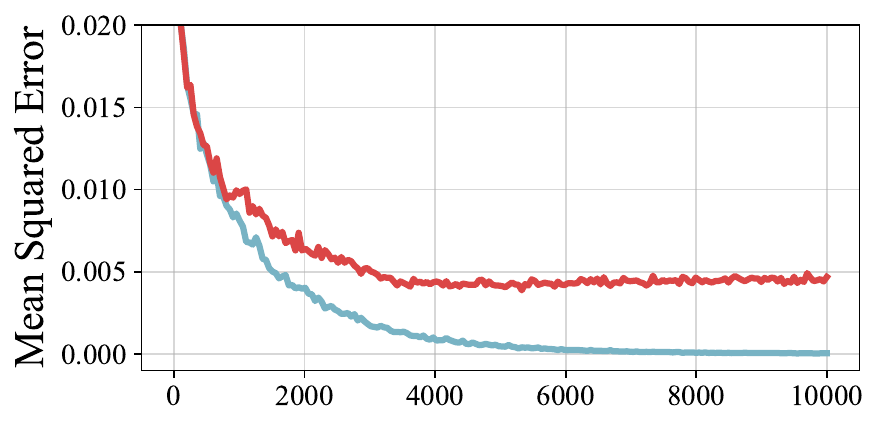}
        \caption{Diabetes + MLP}
    \end{subfigure}%

    \vspace{3mm} 

    \begin{subfigure}{0.5\textwidth}
        \centering
        \includegraphics[width=1\textwidth]{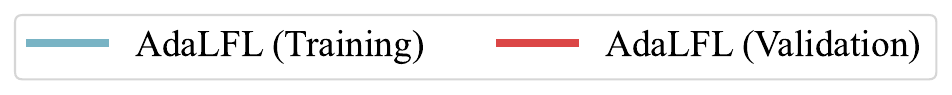}
    \end{subfigure}%
    
\captionsetup{justification=centering}
\caption{Mean learning curves across 10 independent executions of AdaLFL showing the training mean squared error (y-axis) against gradient steps (x-axis) when taking meta gradient steps on the meta-training (blue) vs meta-validation (red) set.}
\label{figure:meta-objective-learning-curves}

\end{figure}

When computing the meta-objective of AdaLFL $\Loss^{task} = \Loss^{meta}(y, f_{\theta_{t+1}}(x))$, the instances can either be sampled from $\Dataset^{train}$ or $\Dataset^{valid}$. Consequently, the learned loss function can either optimize for in-sample performance or out-of-sample generalization, respectively. This behavior is shown on the Communities and Crime and Diabetes dataset, where as shown in Figure \ref{figure:meta-objective-learning-curves}, optimizing the meta-objective using training samples results in the training error quickly approaching 0. In contrast, when using the validation samples the training error does not converge as quickly and to as low of a training error value. This behavior is a form of regularization since, as shown in Table \ref{table:training-vs-validation-results}, the final inference testing error is superior when using validation samples on both the Communities and Crime and Diabetes datasets. This is an important discovery as it suggests that loss function learning can induce a form of regularization, similar to the findings in \citep{gonzalez2021optimizing}, \citep{gonzalez2020effective}, and \citep{raymond2023learning}. 

\begin{table}
\centering

\captionsetup{justification=centering}
\caption{Results reporting the mean $\pm$ standard deviation of final inference testing mean squared error across 10 independent executions (using no learning rate scheduler).}

\begin{tabular}{p{3cm}cc}

\hline \noalign{\vskip 1mm}
Task and Model  &  AdaLFL (Training)   &  AdaLFL (Validation) 
\\ \hline \noalign{\vskip 1mm} \hline \noalign{\vskip 1mm}

Crime + MLP & 0.0267$\pm$0.0022 & \textbf{0.0265$\pm$0.0021}    \\ \noalign{\vskip 1mm}

\hline \noalign{\vskip 1mm} 

Diabetes + MLP & 0.0468$\pm$0.0016 & \textbf{0.0420$\pm$0.0014}   \\ \noalign{\vskip 1mm}

\hline \noalign{\vskip 1mm} 

\end{tabular}

\label{table:training-vs-validation-results}
\end{table}

\subsection{Ablation of Inner Gradient Steps}

In ML$^3$, \citep{bechtle2021meta} suggested taking only one inner step, \textit{i.e.}, setting $\mathcal{S}^{inner}=1$ in Algorithm 1. A reasonable question to ask is whether increasing the number of inner steps to extend the horizon of the meta-objective past the first step will reduce the disparity in performance between ML$^3$ and AdaLFL. To answer this question, experiments are performed on CIFAR-10 AllCNN-C with ML$^3$ setting $\mathcal{S}^{inner} = \{1, 5, 10, 15, 20\}$. The results reported in Table \ref{table:inner-steps} show that increasing the number of inner steps in ML$^3$ up to the limit of what is feasible in memory on a consumer GPU does \textit{not} resolve the short horizon bias present in offline loss function learning. Furthermore, the results show that increasing the number of inner steps only results in marginal improvements in the performance over $\mathcal{S}^{inner}=1$. Hence, offline loss function learning methods that seek to obviate the memory issues of unrolled differentiation to allow for an increased number of inner steps, such as \citep{gao2022loss}, which uses implicit differentiation, are still prone to a kind of short-horizon bias.

\begin{table}[t!]
\centering
\captionsetup{justification=centering}
\caption{Results reporting the mean $\pm$ standard deviation of testing error rates when using an increasing number of inner gradient steps $\mathcal{S}^{inner}$ with ML$^3$.}

\begin{tabular}{p{5cm}c}
\hline
\noalign{\vskip 1mm}
Method                                      & CIFAR-10 + AllCNN-C      \\ \hline \noalign{\vskip 1mm} \hline \noalign{\vskip 1mm}
ML$^3$ ($\mathcal{S}^{inner} = 1$)          & 0.1015$\pm$0.0055      \\ \noalign{\vskip 1mm}
ML$^3$ ($\mathcal{S}^{inner} = 5$)          & 0.0978$\pm$0.0052      \\ \noalign{\vskip 1mm}
ML$^3$ ($\mathcal{S}^{inner} = 10$)         & 0.0985$\pm$0.0050      \\ \noalign{\vskip 1mm}
ML$^3$ ($\mathcal{S}^{inner} = 15$)         & 0.0989$\pm$0.0049      \\ \noalign{\vskip 1mm}
ML$^3$ ($\mathcal{S}^{inner} = 20$)         & 0.0974$\pm$0.0061      \\ \noalign{\vskip 1mm} \hline \noalign{\vskip 1mm}
AdaLFL (Online)                             & \textbf{0.0835$\pm$0.0050}      \\ \noalign{\vskip 1mm} \hline \noalign{\vskip 1mm} 
\end{tabular}

\label{table:inner-steps}
\end{table}

\section{Chapter Summary}

In this chapter, the first fully online approach to loss function learning has been proposed. The proposed technique, \textit{Adaptive Loss Function Learning} (AdaLFL), infers the base loss function directly from the data and adaptively trains it with the base model parameters simultaneously using unrolled differentiation. The results showed that models trained with our method have enhanced convergence capabilities and inference performance compared with the \textit{de facto} standard squared error and cross-entropy loss, and offline loss function learning method ML$^3$. Further analysis of the learned loss functions identified common patterns in the shape of the learned loss function and revealed unique emergent behavior present only in adaptively learned loss functions; namely, implicit tuning of the learning rate schedule and early stopping. Finally, while this chapter has solely focused on meta-learning the loss function in isolation to better understand and analyze its properties, we believe that further benefits can be fully realized upon being combined with existing optimization-based meta-learning techniques. In the next chapter, we will explore this idea further.

\chapter{Meta-Learning Neural Procedural Biases}\label{chapter:npbml}

\textit{In this chapter, we design a method for meta-learning the procedural biases of a deep neural network, which we further refer to as Neural Procedural Bias Meta-Learning (NPBML). Our approach aims to consolidate recent advancements in meta-learned initializations, optimizers, and loss functions by learning them simultaneously and making them adapt to each individual task to maximize the strength of the learned inductive biases. This imbues each learning task with a unique set of procedural biases which is specifically designed and selected to attain strong learning performance in only a few gradient steps. The experimental results show that by meta-learning the procedural biases of a neural network, we can induce strong inductive biases towards a distribution of learning tasks, enabling robust learning performance across many well-established few-shot learning benchmarks.}

\section{Chapter Overview}
\label{section:npbml-introduction}

Humans have an exceptional ability to learn new tasks from only a few example instances. We can often quickly adapt to new domains effectively by building upon and utilizing past experiences of related tasks, leveraging only a small amount of information about the target domain. The field of meta-learning \citep{schmidhuber1987evolutionary, vanschoren2018meta, peng2020comprehensive, hospedales2020meta} explores how deep learning techniques, which often require thousands or even millions of observations to achieve competitive performance, can acquire such a capability. In meta-learning, the learning process is often framed as a bilevel optimization problem \citep{bard2013practical, maclaurin2015gradient, grefenstette2019generalized, lorraine2020optimizing}. The outer optimization aims to learn the underlying regularities across a distribution of related tasks and embed them into the inductive biases of a learning algorithm. Consequently, in the inner optimization, the learning algorithm is utilized to quickly adapt to new learning tasks using only a few example instances.

Model-Agnostic Meta-Learning (MAML) \citep{finn2017model} and its variants \citep{nichol2018reptile, rajeswaran2019meta, song2019maml, triantafillou2020meta}, are a popular approach to meta-learning. In MAML, the outer optimization aims to learn the underlying regularities across a set of related tasks and embed them into a shared parameter initialization. This initialization is then used in the inner optimization's learning algorithm to encourage fast adaptation to new tasks. While successful, these methods resort to simple gradient descent using the cross-entropy loss for classification or squared loss for regression for the inner learning algorithm. Consequently, subsequent research has extended MAML to meta-learn additional components, such as the learning rate \citep{behl2019alpha, baik2020meta}, gradient-based optimizer \citep{li2017meta, lee2018gradient, simon2020modulating, flennerhag2020meta, kang2023meta}, loss function \citep{antoniou2019learning}, and more \citep{antoniou2019train, baik2023meta}. This enables the meta-learning algorithm to induce stronger inductive biases on the learning algorithm, further enhancing performance.

\begin{figure}[t]
\centering
\includegraphics[width=1\textwidth]{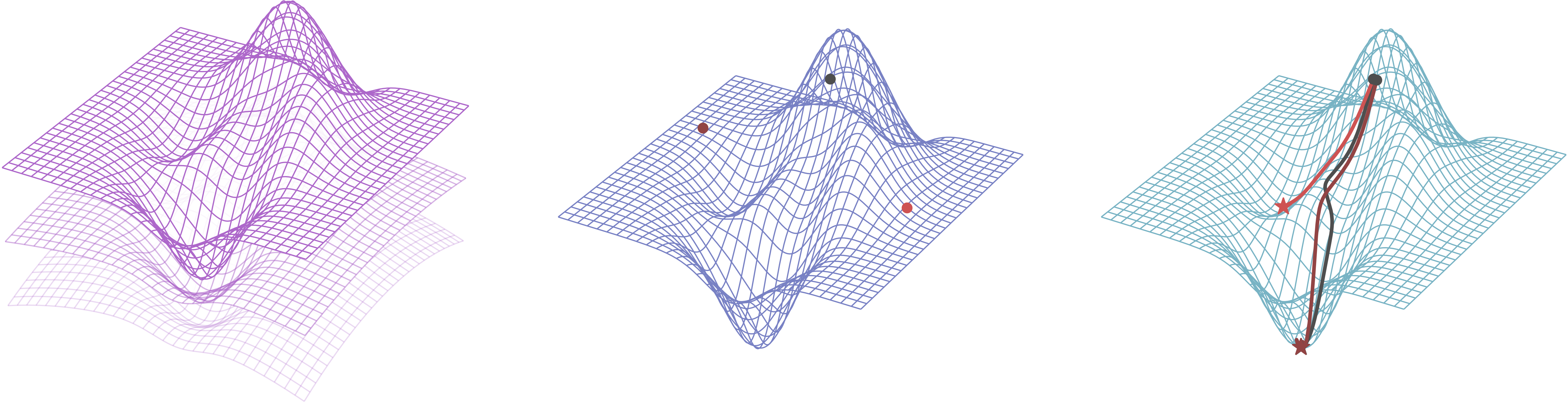}
\captionsetup{justification=centering}
\caption{In NPBML, the procedural biases of a deep neural network are meta-learned. This involves meta-learning three components: the loss function (left), the parameter initialization (center), and the optimizer (right). By meta-learning these components, a strong inductive bias towards fast adaptation can be induced into the learning algorithm.}
\label{fig:npbml-loss-landscapes}
\end{figure}

In this chapter, we propose \textit{Neural Procedural Bias Meta-Learning} (NPBML), a novel gradient-based framework for meta-learning task-adaptive procedural biases for deep neural networks. Procedural biases are the inductive biases that determine the order of traversal over the search space \citep{gordon1995evaluation}, they play a central determining role in the convergence, sample efficiency, and generalization of a learning algorithm. As we will show, the procedural biases are primarily encoded into three fundamental components of a learning algorithm: the loss function, the optimizer, and the parameter initialization. These components define the geometry of the loss landscape, determine the starting point in this space, and guide the optimization process towards the optimum, respectively, as visualized in Figure \ref{fig:npbml-loss-landscapes}. Therefore, we aim to meta-learn these three components to maximize the learning performance when using only a few gradient steps.

To achieve this ambitious goal, we first consolidate three related research areas into one unified end-to-end framework: MAML-based learned initializations \citep{finn2017model}, preconditioned gradient descent methods \citep{lee2018gradient, flennerhag2020meta, kang2023meta}, and meta-learned loss functions \citep{antoniou2019learning, baik2021meta, bechtle2021meta, raymond2023learning, raymond2023online}. We then demonstrate how these meta-learned components can be made task-adaptive through feature-wise linear modulation (FiLM) \citep{perez2018film} to facilitate downstream task-specific specialization towards each task. The proposed NPBML framework is highly flexible and general. As we will show, many existing gradient-based meta-learning approaches arise as special cases of NPBML. To validate the effectiveness of NPBML, we empirically evaluate our proposed algorithm on four well-established few-shot learning benchmarks. The results show that NPBML consistently outperforms many state-of-the-art gradient-based meta-learning algorithms.

\section{Model-Agnostic Meta-Learning}
\label{section:npbml-background}

We begin with a brief overview of meta-learning in the context of few-shot supervised learning. Here we introduce the relevant notation and give an overview of MAML, which will assist in the exposition of our method.

\subsection{Problem Setup}

In few-shot meta-learning, we are given access to a collection of tasks $\{\Task_1, \Task_2, \dots\}$, otherwise known as a meta-dataset, where each task is assumed to be drawn from a task distribution $p(\Task)$. Each task $\Task_i$ contains a support set $\Dataset^S$, and a query set $\Dataset^Q$ (\textit{i.e.}, a training set and a testing set), where $\Dataset^S \cap \Dataset^Q = \emptyset$. Each of these sets contains a set of input-output pairs $\{(x_1, y_1), (x_2, y_2), \dots\}$. Let $x \in X$ and $y \in Y$ denote the inputs and outputs, respectively. In few-shot meta-learning, the goal is to learn a model of the form $f_{\theta}(x):X \rightarrow Y$, where $\theta$ are the model parameters. The primary challenge of few-shot learning is that $f_{\theta}$ must be able to quickly adapt to any new task $\Task_i \sim p(\Task)$ given only a very limited number of instances. For example, in an $N$-way $K$-shot few-shot classification task, $f_{\theta}$ is only given access to $K$ labeled examples of $N$ distinct classes.

\subsection{MAML Overview}

MAML \citep{finn2017model} is a highly influential and seminal method for gradient-based few-shot meta-learning. In MAML, the outer learning objective aims to meta-learn a shared parameter initialization $\bm{\theta}$ over a distribution of related tasks $p(\Task)$. This shared initialization embeds prior knowledge learned from past learning experiences into the learning algorithm such that when a new unseen task is sampled fast adaptation can occur. The outer optimization prototypically occurs by minimizing the sum of the final losses $\sum_{\Task_i \sim p(\Task)}\Loss^{meta}$ on the query set at the end of each task's learning trajectory using gradient descent as follows:
\begin{equation}\label{eq:maml-meta-update}
    \bm{\theta}_{new} = \bm{\theta} - \eta \nabla_{\bm{\theta}} \sum_{\Task_{i} \sim p(\Task)} \left[\Loss^{meta} \Big(\Dataset^{Q}_{i}, \theta_{i, j}(\bm{\theta}) \Big)\right]
\end{equation}
where $\theta_{i, j}(\bm{\theta})$ refers to the adapted model parameters on the $i^{th}$ task at the $j^{th}$ iterations of the inner update rule $\Update^{MAML}$, and $\eta$ is the meta learning rate. The inner optimization for each task starts at the meta-learned parameter initialization $\bm{\theta}$ and applies $\Update^{MAML}$ to the parameters $J$ times:
\begin{equation}\label{eq:maml-base-update}
    \theta_{i, j}(\bm{\theta}) = \bm{\theta} - \alpha \sum_{j=0}^{J-1} \left[ \Update^{MAML} \Big(\Task_{i}, \theta_{i, j}(\bm{\theta}) \Big) \right].
\end{equation}
In the original version of MAML, the inner update rule $\Update^{MAML}$ resorts to simple Stochastic Gradient Descent (SGD) minimizing the loss $\Loss^{base}$, typically set to the cross-entropy or squared loss, with a fixed learning rate $\alpha$ across all tasks
\begin{equation}\label{eq:maml-sgd-update}
    \Update^{MAML} \Big(\Task_{i}, \theta_{i, j}(\bm{\theta}) \Big) := \theta_{i, j} - \alpha \nabla_{\theta_{i, j}} \left[\Loss^{base} \Big(\Dataset^{S}_{i}, \theta_{i, j}\Big) \right].
\end{equation}
This approach assumes that all tasks should use the same fixed learning rule $\Update$ in the inner optimization. Consequently, this greatly limits the performance that can be achieved when taking only a small number of gradient steps with few labeled samples available.

\section{Neural Procedural Bias Meta-Learning}
\label{section:npbml-method}

In this work, we propose \textit{Neural Procedural Bias Meta-Learning} (NPBML), a novel framework that replaces the fixed inner update rule in MAML, \textit{i.e.}, Equation \eqref{eq:maml-sgd-update}, with a meta-learned task-adaptive learning rule. This modifies the inner update rule in three key ways: 
\begin{enumerate}

    \item An optimizer is meta-learned by leveraging the paradigm of preconditioned gradient descent. This involves meta-learning a parameterized preconditioning matrix $P_{\Warp}$ with meta-parameters $\Warp$ to warp the gradients of SGD. 

    \item The conventional loss function $\Loss^{base}$ (\textit{i.e.}, the standard cross-entropy or squared loss) used in the inner optimization is replaced with a meta-learned loss function $\MetaLoss_{\phi}$, where $\bm{\phi}$ are the meta-parameters.

    \item The meta-learned initialization, optimizer, and loss function are adapted to each new task using Feature-Wise Linear Modulation $FiLM_{\psi}$, a general-purpose preconditioning method that has learnable meta-parameters $\bm{\psi}$.
    
\end{enumerate}
For clarity, we provide a high-level overview of NPBML and contrast it to MAML, before expanding on each of these new components in Section \ref{sec:npbml-meta-learned-optimizer}, Section \ref{sec:npbml-meta-learned-loss-function}, and Section \ref{sec:npbml-task-adaptation}, respectively. 

\subsection{Proposed Method Overview}

\begin{figure}[t!]
\centering
\centerline{\includegraphics[width=1\textwidth]{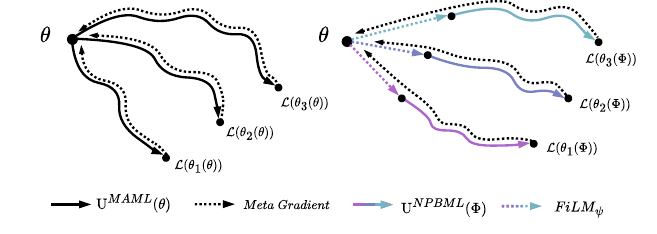}}
\captionsetup{justification=centering}
\caption{In MAML, the update rule $\Update^{MAML}$ optimizes the base model parameters from a shared initialization using simple SGD minimizing $\Loss^{base}$. In contrast, NPBML adapts the model parameters from a task-adapted initialization using $\Update^{NPBML}$, a task-adaptive update rule employing a meta-learned preconditioning matrix $P_{\Warp}$ and loss function $\MetaLoss_{\phi}$.}
\label{fig:maml-vs-npbml}
\end{figure}

The central goal of NPBML is to meta-learn a task-adaptive parameter initialization, optimizer, and loss function. This changes the outer optimization previously seen in Equation \eqref{eq:maml-meta-update} to the following, where $\bm{\Phi} = \{\bm{\theta}, \Warp, \bm{\phi}, \bm{\psi}\}$ refers to the set of meta-parameters:
\begin{equation}\label{eq:npbml-meta-update}
    \bm{\Phi}_{new} = \bm{\Phi} - \eta \nabla_{\Phi} \sum_{\Task_{i} \sim p(\Task)} \left[ \mathcal{L}^{meta} \Big( \Dataset^{Q}_{i}, \theta_{i, j}(\bm{\Phi}) \Big) \right].
\end{equation}
Unlike MAML which employs a fixed update rule $\Update^{MAML}$ for all tasks, simple SGD using $\Loss^{base}$, NPBML uses a fully meta-learned update rule
\begin{equation}\label{eq:npbml-base-update}
    \theta_{i, j}(\bm{\Phi}) = \bm{\theta}(\bm{\psi}) - \alpha \sum_{j=0}^{J-1} \left[ \Update^{NPBML} \Big(\Task_{i}, \theta_{i, j}(\bm{\Phi}) \Big) \right]
\end{equation}
which adjusts $\theta_{i, j}$ in the direction of the negative gradient of a meta-learned loss function $\MetaLoss_{\phi}$. Additionally, the gradient is warped via a meta-learned preconditioning matrix $P_{\Warp}$ as follows:
\begin{equation}\label{eq:npbml-pgd-update}
    \Update^{NPBML} \Big(\Task_{i}, \theta_{i, j}(\bm{\Phi}) \Big) := \theta_{i, j} - \alpha P_{(\Warp, \bm{\psi})} \nabla_{\theta_{i, j}} \left[\MetaLoss_{(\bm{\phi}, \bm{\psi})} \Big(\Dataset^{S}_{i}, \theta_{i, j} \Big) \right]
\end{equation}
where both $\MetaLoss_{\phi}$ and $P_{\Warp}$ are adapted to each task using $FiLM_{\psi}$. This new task-adaptive learning rule empowers each task with a unique set of procedural biases enabling strong and robust learning performance on new unseen tasks in $p(\Task)$ using only a few gradient steps as depicted in Figure \ref{fig:maml-vs-npbml}.

\subsection{Meta-Learned Optimizer}
\label{sec:npbml-meta-learned-optimizer}

In meta-learning, there are two prevalent approaches to meta-learning an optimizer. The first approach directly parameterizes an update rule via the memory of a recurrent neural network \citep{andrychowicz2016learning, ravi2017optimization, chen2022learning}. While these methods can theoretically represent any learning rule, as recurrent neural networks are universal function approximators, they lack an important inductive bias as to what constitutes a reasonable update rule, making them difficult to train and hard to generalize to new tasks \citep{flennerhag2020meta}. In contrast, Preconditioned Gradient Descent (PGD) methods employ a gradient preconditioner $P_{\Warp}$ that rescales the geometry of the parameter space by modifying the update rule as follows:
\begin{equation}\label{eq:preconditioned-gradient-descent}
    \theta_{new} = \theta - \alpha P_{\Warp} \nabla_{\theta} \MetaLoss_{(\bm{\phi}, \bm{\psi})}.
\end{equation}
In traditional optimizer design, setting $P$ to different values yields various well-known optimizers. For instance, when $P = \bm{I}$ (identity matrix) we obtain SGD, when $P = \bm{F}^{-1}$ (inverse Fisher information matrix) we recover natural gradient descent \citep{amari1998natural}, and when $P=\bm{H}^{-1}$ (inverse Hessian matrix) we derive Newton's method.

In contrast to traditional optimizer design, recent meta-learning literature has explored meta-learning $P_{\Warp}$ concurrently with $\bm{\theta}$. In this context, $P_{\Warp}$ can be materialized in various forms, with varying levels of expressiveness. On the low end\footnote{Methods such as Alpha MAML \citep{behl2019alpha}, which meta-learns a scalar learning rate, and MAML++ \citep{antoniou2019train}, which meta-learns a layer-wise learning rate, can be regarded as even lower-end optimizers, where the form of preconditioning is a constant and a block constant diagonal matrix, respectively.}, MetaSGD \citep{li2017meta} meta-learns a parameter-wise learning rate, equivalent to learning a diagonal matrix preconditioning. While on the moderate to high-end, methods such as T-Net \citep{lee2018gradient} learn a block diagonal matrix preconditioning, and WarpGrad \citep{flennerhag2020meta} learns a full matrix preconditioning, via introducing linear and non-linear projection layers into the model, respectively. The most suitable representation for $P_{\Warp}$ is highly dependent on the number of tasks available to meta-learn on, with few tasks available, highly expressive preconditioning methods are prone to meta-overfitting. 

In this work, we take inspiration from T-Nets \citep{lee2018gradient} and insert linear projection layers $\Warp$ into our model $f_{(\bm{\theta}, \Warp, \bm{\psi})}$. The model, composed of an encoder $z$ and a classification head $h$,
\begin{equation}\label{eq:base-model-definition}
    f_{(\bm{\theta, \Warp, \psi})} = z_{(\bm{\theta, \Warp, \psi})} \circ h_{\theta}
\end{equation}
is interleaved with linear projection layers $\Warp$ between each layer in the $L$ layer encoder (where $\sigma$ refers to the non-linear activation functions):
\begin{equation}\label{eq:base-encoder-definition}
    z_{(\bm{\theta}, \Warp, \bm{\psi})}(x_i) = \sigma^{(L)}(\bm{\theta}^{(L)} \Warp^{(L)} (\dots \sigma^{(1)}(\Warp^{(1)}\bm{\theta}^{(1)}x) \dots )).
\end{equation}
As described in Equations \eqref{eq:npbml-meta-update}--\eqref{eq:npbml-pgd-update}, $\Warp$ is meta-learned in the outer loop and held fixed in the inner loop such that preconditioning of the gradients occurs. This form of precondition defines $P_{\Warp}$ as a block-diagonal matrix, where each block is defined by the expression $(\Warp\Warp^{\Transpose})$, as shown in \citep{lee2018gradient}. For a simple model where $f_{(\theta, \Warp)}(x) = \Warp\theta x$, the update rule becomes: 
\begin{equation}\label{eq:tnet-preconditioning}
    \theta_{new} = \theta - \alpha (\Warp\Warp^{\Transpose}) \nabla_{\theta} \MetaLoss_{(\bm{\phi}, \bm{\psi})}.
\end{equation}
We leverage this style of parameterization for $P_{\Warp}$ due to its relative simplicity and high expressive power. However, we emphasize that the NPBML framework is highly general, and other forms of preconditioning, such as those presented in \citep{lee2018gradient,park2019meta,flennerhag2020meta,simon2020modulating}, could be used instead.

\subsection{Meta-Learned Loss Function}
\label{sec:npbml-meta-learned-loss-function}

Unlike MAML, which defines the inner optimization’s loss function $\Loss^{base}$ to be the cross-entropy or squared loss for all tasks, NPBML uses a meta-learned loss function $\MetaLoss_{(\bm{\phi}, \bm{\psi})}$ that is learned in the outer optimization. In contrast to handcrafted loss functions, which typically consider only the ground truth label $y$ and the model predictions $f_{(\theta, \Warp, \bm{\psi})}(x)$, meta-learned loss functions can, in principle, be conditioned on any task-related information \citep{bechtle2021meta, baik2021meta}. In NPBML, the meta-learned loss function is conditioned on three distinct sources of task-related information, which are subsequently processed by three small feed-forward neural networks:
\begin{equation}\label{eq:npbml-base-loss-function}
    \MetaLoss_{(\bm{\phi}, \bm{\psi})} = \Loss^{S}_{(\bm{\phi}, \bm{\psi})} + \Loss^{Q}_{(\bm{\phi}, \bm{\psi})} + \Regularize_{(\bm{\phi}, \bm{\psi})}.
\end{equation}
First, $\Loss^{S}: \mathbb{R}^{2N + 1} \rightarrow \mathbb{R}^{1}$ is an inductive loss function conditioned on task-related information derived from the support set; namely, the one-hot encoded ground truth target and model predictions, and the corresponding loss calculated using $\Loss^{base}$. Next, $\Loss^{Q}: \mathbb{R}^{2N + 1} \rightarrow \mathbb{R}^{1}$ is a transductive loss function conditioned on task-related information derived from the query set. Here, we give $\Loss^{Q}$ access to the model predictions on the query set, embeddings (\textit{i.e.}, relation scores) from a pre-trained relation network \citep{sung2018learning}, and the corresponding loss between the model predictions and embeddings using $\Loss^{base}$. Note that similar embedding functions have previously been used in \citep{rusu2019metalearning, antoniou2019learning}. Finally, the adapted model parameters $\theta_{i, j}$ are used as inputs to meta-learn a weight regularizer $\Regularize: \mathbb{R}^{4L} \rightarrow \mathbb{R}^{1}$. To improve efficiency, we condition $\Regularize$ on the mean, standard deviation, L$1$, and L$2$ norm of each layer’s weights, as opposed to $\theta_{i, j}$ directly.

By employing a meta-learned loss function, NPBML seamlessly utilizes task-specific information not traditionally considered by MAML or its variants in the inner loop. In this particular instantiation of NPBML, we use the unlabelled query instances to extend NPBML to perform transductive inference \citep{vapnik2006transductive}. Many other possibilities exist, and we refer readers to \citep{bechtle2021meta} for further examples of how auxiliary task-related information can be utilized by meta-learned loss functions in domains such as few-shot regression and reinforcement learning.

\subsection{Task-Adaptive Modulation}
\label{sec:npbml-task-adaptation}

\begin{figure}[t!]
\centering
\centerline{\includegraphics[width=1\textwidth]{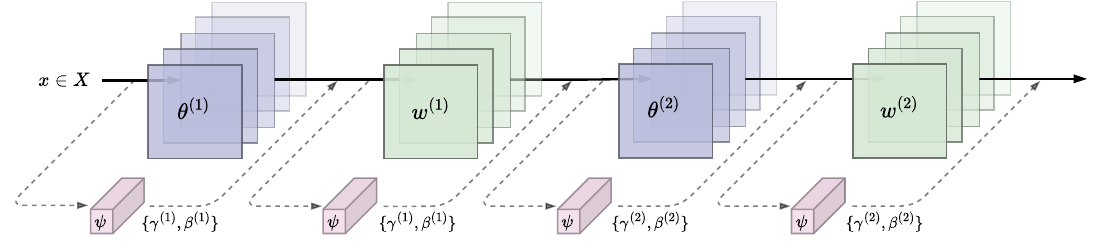}}
\captionsetup{justification=centering}
\caption{An example of a two-layer convolutional neural network in NPBML, where layers $\theta^{(1)}$ and $\theta^{(2)}$, are interleaved with warp preconditioning layers $\Warp^{(1)}$ and $\Warp^{(2)}$. Both types of layers are modulated in the inner loop using feature-wise linear modulation layers to induce task adaptation.}
\label{fig:adaconvnet}
\end{figure}

\begin{figure}[t!]
\centering
\includegraphics[width=0.8\textwidth]{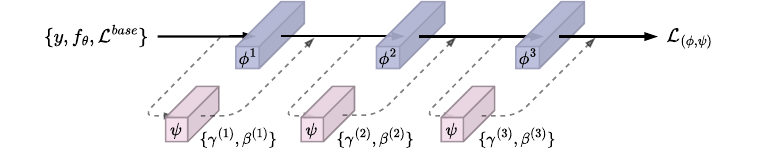}
\captionsetup{justification=centering}
\caption{An example of a meta-learned loss function in NPBML, represented as a composition of feed-forward (linear) layers. These layers are modulated using feature-wise linear modulation, resulting in a task-adaptive meta-learned loss function.}
\label{fig:adalossnet}
\end{figure}

Although all tasks in few-shot learning are assumed to be sampled from the same task distribution $p(\Task)$, the optimal parameter initialization, optimizer, and loss function may differ between tasks. Therefore, in NPBML the meta-learned components $\bm{\Phi}$ are modulated for each new task, providing each task with a unique set of task-adaptive procedural biases. To achieve this, Feature-wise Linear Modulation (FiLM) layers \citep{perez2018film, dumoulin2018feature} are inserted into both the encoder $z_{(\bm{\theta}, \Warp, \bm{\psi})}$ and the meta-learned loss function $M_{(\bm{\phi}, \bm{\psi})}$ as shown in Figures \ref{fig:adaconvnet} and \ref{fig:adalossnet}. $FiLM_{\psi}$ is defined as follows, where $\gamma$ and $\beta$ are the scaling and shifting vectors, respectively, and $\psi$ are the meta-learnable FiLM parameters:
\begin{equation}
    FiLM_{\psi}(x) = (\gamma_{\psi}(x) + \bm{1}) \odot x + \beta_{\psi}(x).
\end{equation}
Affine transformations conditioned on some input have become increasingly popular, being used by several few-shot learning works to make the learned component adaptive \citep{oreshkin2018tadam, jiang2018learning, vuorio2019multimodal, zintgraf2019fast, baik2021meta, baik2023meta}. Furthermore, FiLM layers can help alleviate issues related to batch normalization \citep{ioffe2015batch}, which have been empirically observed to cause training instability due to different distributions of features being passed through the same model in few-shot learning \citep{de2017modulating, antoniou2019train}. In our work, we have found that conditioning the FiLM on the output activations of the previous layers is an effective way to achieve task adaptability. This form of conditioning is in essence a simplified version of that used in CNAPs \citep{requeima2019fast}; however, we have omitted the use of global embeddings as we found it was not necessary for our method.

\subsection{Initialization}

Due to the large number of learnable meta-parameters, initialization becomes an important and necessary aspect to consider. Here we detail how to initialize each of the meta-learned components, \textit{i.e.}, $\bm{\Phi}_0 = \{\bm{\theta_0}, \Warp_0, \bm{\phi}_0, \bm{\psi}_0 \}$, in NPBML. Firstly, we pre-train the encoder weights $\bm{\theta}_0$ prior to meta-learning, following many recent methods in few-shot learning \citep{qiao2018few, rusu2019metalearning, requeima2019fast, ye2020few, ye2021unicorn}, see Appendix \ref{section:npbml-experimental-settings} for more details. For the linear projection layers $\Warp$, we leverage the fact that in PGD, setting $P_{\Warp}$ to the identity $\bm{I}$ recovers SGD. Therefore, we set $\smash{\forall l \in \{1, \dots, L\}: \Warp^{(l)} = \bm{I}}$; note for convolutional layers this corresponds to Dirac initialization. Regarding the meta-learned loss function $\MetaLoss_{(\bm{\phi}, \bm{\psi})}$, the weights $\bm{\phi}$ at the start of meta-training are randomly initialized $\smash{\bm{\phi}_0 \sim \mathcal{N}(0, 1e^{-2})}$; therefore, the $\mathbb{E}[\MetaLoss_{(\bm{\phi}_{0}, \bm{\psi}_{0})}] = 0$, assuming an identity output activation. Consequently, the definition of the meta-learned loss function in Equation \eqref{eq:npbml-base-loss-function} can be modified to
\begin{equation}
    \MetaLoss_{(\bm{\phi}, \bm{\psi})} = \Loss^{base} + \Loss^{S}_{(\bm{\phi}, \bm{\psi})} + \Loss^{Q}_{(\bm{\phi}, \bm{\psi})} + \Regularize_{(\bm{\phi}, \bm{\psi})}
\end{equation}
such that the meta-learned loss function approximately recovers the base loss function at the start of meta-training, \textit{i.e.}, $\MetaLoss_{(\bm{\phi}_{0}, \bm{\psi}_{0})} \approx \Loss^{base}$. Finally, the FiLM layers in NPBML are initialized using a similar strategy taking advantage of the fact that when $\bm{\psi}_{0} \sim \mathcal{N}(0, 1e-2)$ the $\mathbb{E}[\gamma_{\bm{\psi}_{0}}(x)] = \mathbb{E}[\beta_{\bm{\psi}_{0}}(x)] = 0$; consequently, $FiLM_{\bm{\psi}_{0}}(x) \approx x$. When initialized in this manner, the update rule for NPBML at the start of meta-training closely approximates the update rule of MAML.
\begin{equation}
    \Update^{MAML} \Big(\Task_{i}, \theta_{i, j}(\bm{\theta_0}) \Big) \approx \mathbb{E}\left[\Update^{NPBML} \Big(\Task_{i}, \theta_{i, j}(\bm{\Phi_0}) \Big)\right]
\end{equation}

\subsection{Implicit Meta-Learning}

In NPBML, the parameter initialization, gradient-based optimizer, and loss function are explicitly meta-learned in the outer loop. Critically, we make a novel observation that many other key procedural biases are also implicitly learned by meta-learning these three fundamental components (hence the name given to our algorithm). For example, consider the scalar learning rate $\alpha$, which is implicitly meta-learned since
\begin{equation}\label{eq:npbml-implicit-scalar-lr-tuning}
    \exists\alpha\exists\bm{\phi}: \theta_{i, j} - \alpha \nabla_{\theta_{i, j}} \Loss^{base} \approx \theta_{i, j} - \nabla_{\theta_{i, j}} \MetaLoss_{\bm{\phi}},
\end{equation}
and if $\bm{\phi}$ is made to adapt on each inner step as done in \citep{baik2021meta, raymond2023online}, then by extension NPBML also learns a learning rate schedule. Another straightforward related observation is that NPBML implicitly learns a layer-wise learning rate $\smash{\{\alpha^{(1)}, \dots, \alpha^{(L)}\}}$, since for each block $\smash{\{\Warp^{(1)}, \dots, \Warp^{(L)}\}}$ in the block diagonal preconditioning matrix $P_{\Warp}$ the following holds:
\begin{equation}\label{eq:npbml-implicit-layer-wise-lr-tuning}
    \forall l (\exists \Warp^{(l)} \exists \alpha^{(l)}): (\Warp^{(l)} (\Warp^{(l)})^{\Transpose}) \nabla_{\theta^{(l)}} \MetaLoss_{\bm{\phi}} \approx \alpha^{(l)} \nabla_{\theta^{(l)}} \MetaLoss_{\bm{\phi}}.
\end{equation}
There are also less obvious connections that can be drawn. For example, through a transitive relationship, NPBML implicitly learns early stopping when the implicitly learned learning rate approaches zero, as discussed in \citep{baydin2018hypergradient}. Furthermore, since there is a linear scaling rule between the batch size and the learning rate \citep{smith2017don, smith2017bayesian, goyal2017accurate}, NPBML implicitly learns the regularization behavior of the batch size hyperparameter. Another non-trivial example is label smoothing regularization \citep{muller2019does}, which, as proven in \citep{gonzalez2020effective}, can be implicitly induced when meta-learning a loss function.

\section{Background and Related Work}
\label{section:npbml-related-work}

Meta-learning approaches to few-shot learning aim to equip models with the ability to quickly adapt to new tasks given only a limited number of examples by leveraging prior learning experiences across a distribution of related tasks. These approaches are commonly partitioned into three categories: (1) metric-based methods, which aim to learn a similarity metric for efficient class differentiation \citep{koch2015siamese, vinyals2016matching, sung2018learning, snell2017prototypical}; (2) memory-based methods, which utilize architectures that store training examples in memory or directly encode fast adaptation algorithms in the model weights \citep{santoro2016meta, ravi2017optimization}; and (3) optimization-based methods, which aim to learn an optimization algorithm specifically designed for fast adaptation and few-shot learning \citep{finn2017model}. This work explored the latter approach by meta-learning a gradient update rule.

MAML \citep{finn2017model}, a highly flexible task and model-agnostic method for meta-learning a parameter initialization from which fast adaptation can occur. Many follow-up works have sought to enhance MAML’s performance by addressing limitations in the outer-optimization algorithm, such as the memory and compute efficiency \citep{nichol2018reptile, rajeswaran2019meta, raghu2019rapid, oh2020boil}, or the meta-level overfitting \citep{flennerhag2018transferring, rusu2019metalearning}. Relatively fewer works have focused on enhancing the inner-optimization update rule, as is done in our work. Most MAML variants continue to use an inner update rule consisting of SGD with a fixed learning rate minimizing a loss function such as the cross-entropy or squared loss.

Of the works that have explored improving the inner update rule, the vast majority focus on improving the optimizer. For example, early MAML-based methods such as \citep{behl2019alpha, antoniou2019train, li2017meta} explored meta-learning the scalar, layer-wise, and parameter-wise learning rates, respectively. More recent methods have explored more powerful parameterization for the meta-learned optimizer through the utilization of preconditioned gradient descent methods \citep{lee2018gradient, park2019meta, flennerhag2020meta, simon2020modulating, kang2023meta}, which rescale the geometry of the parameter space by modifying the update rule with a learned preconditioning matrix. While these methods have advanced MAML-based few-shot learning, they often lack a task-adaptive property, falsely assuming that all tasks should use the same optimizer.

A small number of recent works have also investigated replacing the inner loss function (e.g., cross-entropy loss) with a meta-learned loss function. In \citep{antoniou2019learning}, a fully transductive loss function represented as a dilated convolutional neural network is meta-learned. Meanwhile, in \citep{baik2021meta}, a set of loss functions and task adapters are meta-learned for each step taken in the inner optimization. Although these approaches have shown a lot of promise, their potential has yet to be fully realized, as they have not yet been meta-learned in tandem with the optimizer as we have done in this work.

\section{Experimental Setup}
\label{section:npbml-experiments}

In this section, we detail the experimental setup and settings used in our evaluation. To evaluate the performance of NPBML, experiments are performed on four well-established few-shot learning datasets: \textit{mini}-Imagenet \citep{ravi2017optimization}, \textit{tiered}-ImageNet \citep{ren18fewshotssl}, CIFAR-FS \citep{bertinetto2018meta}, and FC-100 \citep{oreshkin2018tadam}. For each dataset, experiments are performed using both 5-way 1-shot and 5-way 5-shot configurations. Results are also reported on both the 4-CONV \citep{finn2017model, zintgraf2019fast, flennerhag2020meta, kang2023meta} and ResNet-12 \citep{he2016deep, baik2020meta, baik2021meta} network architectures. For further details not mentioned in this section, please refer to Algorithms \ref{algorithm:npbml-outer-optimization} and \ref{algorithm:npbml-inner-optimization}, as well as our code made available at: \url{https://github.com/Decadz/Neural-Procedural-Bias-Meta-Learning}.

\begin{algorithm}[t]

\caption{Meta-Learning (Outer Loop)}
\label{algorithm:npbml-outer-optimization}

\SetAlgoLined
\DontPrintSemicolon
\SetKwInput{Input}{Input}
\BlankLine
\Input{
    $\Loss^{meta} \leftarrow$ Meta loss function \newline
    $p(\Task) \leftarrow$ Task distribution \newline
    $\eta \leftarrow$ Meta learning rate\;
    \vspace{-2mm}
    \hrulefill
}

\BlankLine

$\bm{\Phi_{0}} \leftarrow$ Initialize meta-parameters $\{\bm{\theta}, \bm{\omega}, \bm{\phi}, \bm{\psi}\}$\;
\For{$t \in \{0, ... , \mathcal{S}^{meta}\}$}{
    $\Task_{0}, \Task_{1}, \dots , \Task_{B}$ $\leftarrow$ Sample tasks from $p(\Task)$\;
    \For{$t \in \{0, ... , \mathcal{S}^{meta}\}$}{
        $\Dataset_{i}^{S} = \{(x_{i}^{s}, y_{i}^{s})\}^{S}_{s=0}$ $\leftarrow$ Sample support from $\Task_{i}$\;
        $\Dataset_{i}^{Q} = \{(x_{i}^{q}, y_{i}^{q})\}^{Q}_{q=0}$ $\leftarrow$ Sample query from $\Task_{i}$\;
        $\theta_{i, j} \leftarrow$ Base-Learning using Algorithm (\ref{algorithm:npbml-inner-optimization})\;
    }
    $\bm{\Phi_{t+1}} \leftarrow \bm{\Phi_{t}} - \eta \frac{1}{B}\nabla_{\bm{\Phi_{t}}} \sum_{i}\Loss^{meta}_{i}(\Dataset^{Q}_{i}, \theta_{i, j})$\;
}
\textbf{return} $\bm{\Phi_{t}}$

\BlankLine

\end{algorithm}

\begin{algorithm}[t!]

\caption{Base-Learning (Inner Loop)}
\label{algorithm:npbml-inner-optimization}

\SetAlgoLined
\DontPrintSemicolon
\SetKwInput{Input}{Input}
\BlankLine
\Input{
    $\Loss^{base} \leftarrow$ Base loss function \newline
    $\Dataset^{S}_{i}, \Dataset^{Q}_{i} \leftarrow$ Support and query sets \newline
    $\bm{\Phi} \leftarrow$ Meta parameters $\{\bm{\theta}, \bm{\omega}, \bm{\phi}, \bm{\psi}\}$ \newline
    $\alpha \leftarrow$ Base learning rate \newline
    $g \leftarrow$ Relation network\;
    \vspace{-2mm}
    \hrulefill
}

\BlankLine

$\theta_{i, 0} \leftarrow$ Initialize base weights with $\boldsymbol{\theta}$

\For{$j \in \{0, ... , \mathcal{S}^{base}\}$}{
    $\hat{y}^{S}_{i}, \hat{y}^{Q}_{i} \leftarrow f_{(\theta_{i, j}, \Warp, \bm{\psi})}(x^{S}_{i} \cup x^{Q}_{i})$\;
    $\Loss^{base}_{i, j} \leftarrow \frac{1}{|\Dataset^{S}|} \sum \Loss^{base}(y^{S}_{i}, \hat{y}^{S}_{i})$\;
    $\Loss^{S}_{i, j} \leftarrow \frac{1}{|\Dataset^{S}|} \sum \Loss_{(\bm{\phi}, \bm{\psi})}^{S}(y^{S}_{i}, \hat{y}^{S}_{i})$\;
    $\Loss^{Q}_{i, j} \leftarrow \frac{1}{|\Dataset^{Q}|} \sum \Loss_{(\bm{\phi}, \bm{\psi})}^{Q}(g(x^{Q}_{i}), \hat{y}^{Q}_{i})$\;
    $\Regularize_{i, j} \leftarrow \Regularize_{(\bm{\phi}, \bm{\psi})}(\theta_{i, j})$\;
    $\MetaLoss_{(\bm{\phi}, \bm{\psi})} \leftarrow \Loss^{base}_{i, j} + \Loss^{S}_{i, j} + \Loss^{Q}_{i, j} + \Regularize_{i, j}$\;
    $\theta_{i, j+1} \leftarrow \theta_{i, j} - \alpha P \nabla_{\theta_{i, j}} \MetaLoss_{(\bm{\phi}, \bm{\psi})}$\;
}
\textbf{return} $\theta_{i, j}$

\BlankLine

\end{algorithm}

\subsection{Benchmark Tasks}\label{section:npbml-dataset-configurations}

The few-shot learning experiments follow a prototypical episodic learning setup \citep{ravi2017optimization}, where each dataset contains a set of non-overlapping meta-training, meta-validation, and meta-testing tasks. Each task $\Task_{i}$ has a support $\Dataset^{S}$ and query $\Dataset^{Q}$ set (\textit{i.e.}, training and testing set, respectively). For an $N$-way $K$-shot classification task, the support set contains $N \times K$ instances, and the query set contains $N \times 15$ instances, where $N$ refers to the number of randomly sampled classes, and $K$ refers to the number of instances (\textit{i.e.}, shots) available from each of those classes. For example, in a $5$-way $5$-shot classification task, the support set contains $5 \times 5 = |\Dataset^{S}|$ instances to train on, and the query set contains $5 \times 15 = |\Dataset^{Q}|$ instances to validate the performance on.

\clearpage

\subsubsection{ImageNet Datasets}

Two commonly used datasets for few-shot learning are two ImageNet derivatives \citep{deng2009imagenet}: \textit{mini}-ImageNet \citep{ravi2017optimization} and \textit{tiered}-ImageNet \citep{ren18fewshotssl}. Both datasets are composed of three subsets (training, validation, and testing), each of which consists of images with a size of $84 \times 84$. The two datasets differ in regard to how the classes are partitioned into mutually exclusive subsets. 

\vspace{3mm}\noindent\textbf{\textit{mini-}ImageNet} randomly samples $100$ classes from the $1000$ base classes in ImageNet. Following \citep{vinyals2016matching} the sampled classes are partitioned such that $64$ classes are allocated for meta-training, $16$ for meta-validation, and $20$ for meta-testing, where for each class $600$ images are available.

\vspace{3mm}\noindent\textbf{\textit{tiered-}ImageNet} as described in \citep{ren18fewshotssl} alternatively stratifies $608$ classes into $34$ higher-level categories in the ImageNet human-curated hierarchy. The $34$ classes are partitioned into $20$ categories for meta-training, $6$ for meta-validation, and $8$ for meta-testing. Each class in \textit{tiered}-ImageNet has a minimum of $732$ instances and a maximum of $1300$.

\subsubsection{CIFAR-100 Datasets}

Additional experiments are also conducted on CIFAR-FS \citep{bertinetto2018meta} and FC-100 \citep{oreshkin2018tadam}, which are few-shot learning datasets derived from the popular CIFAR-100 dataset \citep{krizhevsky2009learning}. The CIFAR100 dataset contains $100$ classes containing $600$ images each, each with a resolution of $32 \times 32$. 

\vspace{3mm}\noindent\textbf{CIFAR-FS} uses a similar sampling procedure to \textit{mini}-ImageNet \citep{ravi2017optimization}, CIFAR-FS is derived by randomly sampling $100$ classes from the $100$ base classes in CIFAR100. The sampled classes are partitioned such that $60$ classes are allocated for meta-training, $20$ for meta-validation, and $20$ for meta-testing.

\vspace{3mm}\noindent\textbf{FC-100} is obtained by using a dataset construction process similar to \textit{tiered}-ImageNet \citep{ren18fewshotssl}, in which class hierarchies are used to partition the original dataset to simulate more challenging few-shot learning scenarios. In FC100 there are a total of $20$ high-level classes, where $12$ classes are allocated for meta-training, $4$ for meta-validation, and $4$ for meta-testing.

\subsection{Network Architectures}\label{section:npbml-network-architectures}

\vspace{3mm}\noindent\textbf{4-CONV:} Following the encoder architecture settings in \citep{zintgraf2019fast, flennerhag2020meta, kang2023meta}, the network consists of four modules. Each module contains a $3\times3$ convolutional layer with 128 filters, followed by a batch normalization layer, a ReLU non-linearity, and a $2\times2$ max-pooling downsampling layer. Following the four modules, we apply average pooling and flatten the embedding to a size of 128.

\vspace{3mm}\noindent\textbf{ResNet-12:} The encoder $z_{\theta}$ follows the standard network architecture settings used in \citep{baik2020meta, baik2021meta}. The network, similar to 4-CONV, consists of four modules. Each module contains a stack of three $3\times3$ convolutional layers, where each layer is followed by batch normalization and a leaky ReLU non-linearity. A skip convolutional over the convolutional stack is used in each module. Each skip connection contains a $1\times1$ convolutional layer followed by batch normalization. Following the convolutional stack and skip connection, a leaky ReLU non-linearity and $2\times2$ max-pooling downsampling layer are placed at the end of each module. The number of filters used in each module is $[64, 128, 256, 512]$, respectively. Following the four modules, we apply average pooling and flatten the embedding to a size of 512.

\vspace{3mm}\noindent\textbf{Classifier:} As MAML and its variants are sensitive to label permutation during meta-testing, we opt to use the permutation invariant classification head proposed by \citep{ye2021unicorn}. This replaces the typical classification head $h_{\theta} \in \mathbb{R}^{in \times N}$ with a \textit{single-weight vector} $\theta_{head} \in \mathbb{R}^{in \times 1}$ which is meta-learned at meta-training time. During meta-testing the weight vector is duplicated into each output, \textit{i.e.}, $h_{\theta} = \{\theta_{c} = \theta_{head}\}^{N}_{c=1}$, and adapted in an identical fashion to traditional MAML. This reduces the number of parameters in the classification head and makes NPBML permutation invariant similar to MetaOptNet \citep{lee2019meta}, CNAPs \citep{requeima2019fast}, and ProtoMAML \citep{triantafillou2020meta}.

\vspace{3mm}\noindent\textbf{Loss Network:} The meta-learned loss function $\MetaLoss$ is composed of three separate networks $\Loss^{S}$, $\Loss^{Q}$, and $\Regularize$ whose scalar outputs are summed to produce the final output as shown in Equation \eqref{eq:npbml-base-loss-function}. The network architecture for all three networks is identical except for the input dimension of the first layer as discussed in Section \ref{sec:npbml-meta-learned-loss-function}. The network architecture is taken from \citep{bechtle2021meta, psaros2022meta, raymond2023online} and is a small feedforward rectified linear unit neural network with two hidden layers and 40 hidden units in each layer. Note $\Loss^{S}$ and $\Loss^{Q}$ are applied instance-wise and reduced using a mean reduction, such that they are both invariant to the number of instances made available at meta-testing time in the support and query set.

\vspace{3mm}\noindent\textbf{Relation Network:} The transductive loss $L^{Q}$ takes an embedding from a pre-trained relation network \citep{sung2018learning} as one of its inputs, similar to \citep{rusu2019metalearning, antoniou2019learning}. The relation network used in our experiments employs the previously mentioned ResNet-12 encoder to generate a set of embeddings for each instance in $D^{S} \cup D^{Q}$. These embeddings are then concatenated and processed using a relation module. The relation module in our experiments consists of two ResNet blocks with $2\times512$ and $512$ filters, respectively. The output is then flattened and averaged pooled to an embedding size of $512$, which is passed through a final linear layer of size $512\times1$ which outputs each relation score. In regard to training, the relation network is trained using the settings described in \citep{sung2018learning}.

\subsection{Experimental Setup}\label{section:npbml-experimental-settings}

\subsubsection{Pre-training Settings}

In NPBML, the encoder portion of the network $z_{\theta}$ is pre-trained prior to meta-learning, following many recent methods in few-shot learning \citep{qiao2018few, rusu2019metalearning, requeima2019fast, ye2020few, ye2021unicorn}. For both the 4-CONV and ResNet-12 models, we follow the pre-training pipeline used in \citep{ye2020few, ye2021unicorn}. During pre-training, we append the feature encoder $z_{\theta}$ with a temporary classification head $f_{\theta}$ and train it to classify all classes in $\Dataset^{train}$ (e.g., over all 64 classes in the \textit{mini}-ImageNet) using the cross-entropy loss. During pre-training, the model is trained over $200000$ gradient steps using SGD with Nesterov momentum and weight decay with a batch size of $128$. The learning rate, momentum coefficient, and weight decay penalty are set to $0.01$, $0.9$, and $0.0005$, respectively. Additionally, we use a multistep learning rate schedule, decaying the learning rate by a factor of $0.1$ at the following gradient steps $\{100000, 150000, 175000, 190000\}$. Note that in our ResNet-12 experiments on \textit{mini}-ImageNet, CIFAR-FS, and FC-100, we observed some meta-level overfitting, which we hypothesize is due to the expressive network architecture coupled with the relatively small datasets. We found that increasing the pre-training weight decay to $0.01$ in these experiments resolved the issue.

\subsubsection{Meta-Learning Settings}

In both the meta-training and meta-testing phases, we adhere to the standard hyperparameter values from the literature \citep{finn2017model}. In the outer loop, our algorithm is trained over $S^{meta}=30,000$ meta-gradient steps using Adam with a meta learning rate of $\eta=0.00001$. As with previous works, our models are trained with a meta-batch size of 2 for 5-shot classification and 4 for 1-shot classification. In the inner loop, the models are adapted using $S^{base}=5$ base-gradient steps using SGD with Nesterov momentum and weight decay. The learning rate, momentum coefficient, and weight decay penalty are set to $0.01$, $0.9$, and $0.0005$ respectively. During meta-testing, the same inner loop hyperparameter settings are employed, and the final model is evaluated over $600$ tasks sampled from $D^{testing}$. Notably, unlike prior work \citep{antoniou2019train, baik2020meta, baik2021meta, baik2023meta}, we do not ensemble the top 3 or 5 performing models from the same run, as this significantly increases the number of parameters and expressive power of the final models.

\subsubsection{NPBML Settings}

In our experiments, the backbone encoders $z_{\theta}$ (\textit{i.e.}, 4-CONV and ResNet-12) are modified in two key ways during meta-learning: (1) with linear projection (warp) layers $\Warp$ for preconditioning gradients, and (2) feature-wise linear modulation layers $FiLM_{\psi}$ for task adaptation. In all our experiments, we modify only the last (\textit{i.e.}, fourth) module with warp and FiLM layers based on the findings from \citep{raghu2019rapid}, which showed that features near the start of the network are primarily reused and do not require adaptation. Additionally, we freeze all preceding modules in both the inner and outer loops, which significantly reduces the storage and memory footprint of the proposed method, with no noticeable effect on the performance.

Following the recommendations of \cite{flennerhag2020meta}, warp layers are inserted after the main convolutional stack and before the residual connection ends in the ResNet modules. Regarding the FiLM layers, they are inserted after each batch normalization layer in a manner similar to the backbone encoders used in \citep{requeima2019fast}. Where for convolutional layers, we first pool in the spatial dimension to obtain the average activation scores per map, and then flatten in the depth/filter dimension before processing $FiLM_{\psi}$ to obtain $\gamma$ and $\beta$. As discussed in Section \ref{sec:npbml-task-adaptation}, the loss networks $\Loss^{S}$, $\Loss^{Q}$, and $\Regularize$ also use FiLM layers, these are interleaved between each of the linear layers as shown in Figure \ref{fig:adalossnet}.

\section{Results and Analysis}

In this section, we evaluate the performance of the proposed method on a set of well-established few-shot learning benchmarks. The experimental evaluation aims to answer the following key questions: (1) Can NPBML perform well across a diverse range of few-shot learning tasks? (2) Do the novel components meta-learned in NPBML individually enhance performance? (3) To what extent does each component synergistically contribute to the overall performance of the proposed algorithm? 

\subsection{ImageNet Experiments}

\begin{table*}[t!]
\centering
\captionsetup{justification=centering}
\caption{Few-shot classification meta-testing accuracy on 5-way 1-shot and 5-way 5-shot \textit{mini-ImageNet} where $\pm$ represents the 95\% confidence intervals.}
\begin{threeparttable}
\begin{tabular}{lccc}
\hline
\multirow{2}{*}{Method}                             & \multirow{2}{*}{Base Learner} & \multicolumn{2}{c}{\textit{mini-}ImageNet (5-way)}    \\ \cline{3-4} 
                                                    &           & 1-shot           & 5-shot                   \\ \hline \hline \noalign{\vskip 1mm}
MAML \citep{finn2017model}                          & 4-CONV    & 48.70$\pm$1.84\% & 63.11$\pm$0.92\%         \\ \noalign{\vskip 1mm}
MetaSGD \citep{li2017meta}                          & 4-CONV    & 50.47$\pm$1.87\% & 64.03$\pm$0.94\%         \\ \noalign{\vskip 1mm}
T-Net \citep{lee2018gradient}$^{\dagger}$           & 4-CONV    & 50.86$\pm$1.82\% & -                        \\ \noalign{\vskip 1mm}
MAML++ \citep{antoniou2019train}$^{\ddagger}$       & 4-CONV    & 52.15$\pm$0.26\% & 68.32$\pm$0.44\%         \\ \noalign{\vskip 1mm}
SCA \citep{antoniou2019learning}$^{\ddagger}$       & 4-CONV    & 54.84$\pm$0.99\% & 71.85$\pm$0.53\%         \\ \noalign{\vskip 1mm}
WarpGrad \citep{flennerhag2020meta}$^{\dagger}$     & 4-CONV    & 52.30$\pm$0.80\% & 68.40$\pm$0.60\%         \\ \noalign{\vskip 1mm}
ModGrad \citep{simon2020modulating}                 & 4-CONV    & 53.20$\pm$0.86\% & 69.17$\pm$0.69\%         \\ \noalign{\vskip 1mm}  
MeTAL \citep{baik2021meta}$^{\ddagger}$             & 4-CONV    & 52.63$\pm$0.37\% & 70.52$\pm$0.29\%         \\ \noalign{\vskip 1mm}
ALFA \citep{baik2023meta}$^{\ddagger}$              & 4-CONV    & 50.58$\pm$0.51\% & 69.12$\pm$0.47\%         \\ \noalign{\vskip 1mm}
GAP \citep{kang2023meta}$^{\ddagger}$               & 4-CONV    & 54.86$\pm$0.85\% & 71.55$\pm$0.61\%         \\ \noalign{\vskip 1mm}
\hdashline \noalign{\vskip 1mm}
NPBML (Ours)$^{\dagger}$                            & 4-CONV    & \textbf{57.49$\pm$0.83\%} & \textbf{75.01$\pm$0.64\%}         \\ \noalign{\vskip 1mm}
\hline \hline \noalign{\vskip 1mm}
MAML \citep{finn2017model}                          & ResNet-12        & 58.60$\pm$0.42\% & 69.54$\pm$0.38\%         \\ \noalign{\vskip 1mm}
MeTAL \citep{baik2021meta}$^{\ddagger}$             & ResNet-12        & 59.64$\pm$0.38\% & 76.20$\pm$0.19\%         \\ \noalign{\vskip 1mm}
ALFA \citep{baik2023meta}$^{\ddagger}$              & ResNet-12        & 59.74$\pm$0.49\% & 77.96$\pm$0.41\%         \\ \noalign{\vskip 1mm}
\hdashline \noalign{\vskip 1mm}
NPBML (Ours)$^{\dagger}$                            & ResNet-12        & \textbf{61.59$\pm$0.80\%} & \textbf{78.18$\pm$0.60\%}         \\ \noalign{\vskip 1mm} \hline \noalign{\vskip 1mm}

\end{tabular}
\begin{tablenotes}\centering\small
$\dagger$ Denotes a base learner that has been modified to include warp preconditioning layers. \\
$\ddagger$ Reported result uses ensembling of the top-3 or 5 base learner models from one run. \\
\end{tablenotes}
\end{threeparttable}
\label{table:npbml-imagenet-experiments-1}
\end{table*}

\begin{table*}[t!]
\centering
\captionsetup{justification=centering}
\caption{Few-shot classification meta-testing accuracy on 5-way 1-shot and 5-way 5-shot \textit{tiered-ImageNet}, where $\pm$ represents the 95\% confidence intervals.}
\begin{threeparttable}
\begin{tabular}{lccc}
\hline 
\multirow{2}{*}{Method}                             & \multirow{2}{*}{Base Learner}       & \multicolumn{2}{c}{\textit{tiered-}ImageNet (5-way)}  \\ \cline{3-4} 
                                                    &           & 1-shot           & 5-shot                   \\ \hline \hline \noalign{\vskip 1mm}
MAML \citep{finn2017model}                          & 4-CONV    & 50.98$\pm$0.26\% & 66.25$\pm$0.19\%         \\ \noalign{\vskip 1mm}
WarpGrad \citep{flennerhag2020meta}$^{\dagger}$     & 4-CONV    & 57.20$\pm$0.90\% & 74.10$\pm$0.70\%         \\ \noalign{\vskip 1mm}
MeTAL \citep{baik2021meta}$^{\ddagger}$             & 4-CONV    & 54.34$\pm$0.31\% & 70.40$\pm$0.21\%         \\ \noalign{\vskip 1mm}
ALFA \citep{baik2023meta}$^{\ddagger}$              & 4-CONV    & 53.16$\pm$0.49\% & 70.54$\pm$0.46\%         \\ \noalign{\vskip 1mm}
GAP \citep{kang2023meta}$^{\ddagger}$               & 4-CONV    & 57.60$\pm$0.93\% & 74.90$\pm$0.68\%         \\ \noalign{\vskip 1mm}
\hdashline \noalign{\vskip 1mm}
NPBML (Ours)$^{\dagger}$                            & 4-CONV    & \textbf{64.24$\pm$0.97\%} & \textbf{79.17$\pm$0.71\%}         \\ \noalign{\vskip 1mm}
\hline \hline \noalign{\vskip 1mm}
MAML \citep{finn2017model}                          & ResNet-12        & 59.82$\pm$0.41\% & 73.17$\pm$0.32\%         \\ \noalign{\vskip 1mm}
Meta-Curvature \citep{park2019meta}                 & WRN-28-10        & 64.40$\pm$0.10\% & 80.21$\pm$0.10\%         \\ \noalign{\vskip 1mm}  
ModGrad \citep{simon2020modulating}                 & WRN-28-10        & 65.72$\pm$0.21\% & 81.17$\pm$0.20\%         \\ \noalign{\vskip 1mm}  
MeTAL \citep{baik2021meta}$^{\ddagger}$             & ResNet-12        & 63.89$\pm$0.43\% & 80.14$\pm$0.40\%         \\ \noalign{\vskip 1mm}
ALFA \citep{baik2023meta}$^{\ddagger}$              & ResNet-12        & 64.62$\pm$0.49\% & 82.48$\pm$0.38\%         \\ \noalign{\vskip 1mm}
\hdashline \noalign{\vskip 1mm}
NPBML (Ours)$^{\dagger}$                            & ResNet-12        & \textbf{72.22$\pm$0.96\%} & \textbf{85.41$\pm$0.61\%}         \\ \noalign{\vskip 1mm} \hline \noalign{\vskip 1mm}

\end{tabular}
\begin{tablenotes}\centering\small
$\dagger$ Denotes a base learner that has been modified to include warp preconditioning layers. \\
$\ddagger$ Reported result uses ensembling of the top-3 or 5 base learner models from one run. \\
\end{tablenotes}
\end{threeparttable}
\label{table:npbml-imagenet-experiments-2}
\end{table*}

We first assess the performance of NPBML and compare it to a range of MAML-based few-shot learning methods on two popular ImageNet derivatives \citep{deng2009imagenet}: \textit{mini-}ImageNet \citep{ravi2017optimization} and \textit{tiered-}ImageNet \citep{ren18fewshotssl}. The results, presented in Tables \ref{table:npbml-imagenet-experiments-1} and \ref{table:npbml-imagenet-experiments-2}, demonstrate that the proposed method NPBML, which uses a fully meta-learned update rule in the inner optimization, significantly improves upon the performance of MAML-based few-shot learning methods. The proposed method achieves higher meta-testing accuracy in the 1-shot and 5-shot settings using both low-capacity (4-CONV) and high-capacity (ResNet-12) models. 

In contrast to PGD methods Meta-SGD, T-Net, WarpGrad, ModGrad, ALFA, and GAP, which meta-learn an optimizer alongside the parameter initialization, NPBML shows clear gains in generalization performance. This improvement is also evident when compared to SCA and MeTAL, which replace the inner optimization's loss function with a meta-learned loss function. These results empirically demonstrate that meta-learning an optimizer and loss function are complementary and orthogonal approaches to improving MAML-based few-shot learning methods. 

On \textit{tiered}-ImageNet, the larger of the two datasets, we find that the difference between NPBML and its competitors is even more pronounced than on \textit{mini-}ImageNet. This result suggests that when given enough data, NPBML can learn highly expressive inner update rules that significantly enhances few-shot learning performance. However, meta-overfitting can occur on smaller datasets, necessitating regularization techniques. Alternatively, we conjecture that less expressive representations for $P_{\Warp}$ would also reduce meta-overfitting.

\subsection{CIFAR-100 Experiments}

\begin{table*}[t!]
\centering
\captionsetup{justification=centering}
\caption{Few-shot classification meta-testing accuracy on 5-way 1-shot and 5-way 5-shot \textit{CIFAR-FS} where $\pm$ represents the 95\% confidence intervals.}
\begin{threeparttable}
\begin{tabular}{lccc}
\hline 
\multirow{2}{*}{Method}                             & \multirow{2}{*}{Base Learner}    & \multicolumn{2}{c}{CIFAR-FS (5-way)}  \\ \cline{3-4} 
                                                    &        & 1-shot           & 5-shot                     \\ \hline \hline \noalign{\vskip 1mm}
MAML \citep{finn2017model}                          & 4-CONV & 57.63$\pm$0.73\% & 73.95$\pm$0.84\%           \\ \noalign{\vskip 1mm}
BOIL \citep{oh2020boil}                             & 4-CONV & 58.03$\pm$0.43\% & 73.61$\pm$0.32\%           \\ \noalign{\vskip 1mm}
MeTAL \citep{baik2021meta}$^{\ddagger}$             & 4-CONV & 59.16$\pm$0.56\% & 74.62$\pm$0.42\%           \\ \noalign{\vskip 1mm}
ALFA \citep{baik2023meta}$^{\ddagger}$              & 4-CONV & 59.96$\pm$0.49\% & 76.79$\pm$0.42\%           \\ \noalign{\vskip 1mm} 
\hdashline \noalign{\vskip 1mm}
NPBML (Ours)$^{\dagger}$                            & 4-CONV & \textbf{64.90$\pm$0.94}\% & \textbf{79.24$\pm$0.69\%}           \\ \noalign{\vskip 1mm} \hline \hline \noalign{\vskip 1mm}

MAML \citep{finn2017model}                          & ResNet-12     & 63.81$\pm$0.54\% & 77.07$\pm$0.42\%           \\ \noalign{\vskip 1mm}
MeTAL \citep{baik2021meta}$^{\ddagger}$             & ResNet-12     & 67.97$\pm$0.47\% & 82.17$\pm$0.38\%           \\ \noalign{\vskip 1mm}
ALFA \citep{baik2023meta}$^{\ddagger}$              & ResNet-12     & 66.79$\pm$0.47\% & 83.62$\pm$0.37\%           \\ \noalign{\vskip 1mm} 
\hdashline \noalign{\vskip 1mm}
NPBML (Ours)$^{\dagger}$                            & ResNet-12     & \textbf{69.30$\pm$0.91\%} & \textbf{83.72$\pm$0.64\%}           \\ \noalign{\vskip 1mm} \hline \noalign{\vskip 1mm}

\end{tabular}
\begin{tablenotes}\centering\small
$\dagger$ Denotes a base learner that has been modified to include warp preconditioning layers. \\
$\ddagger$ Reported result uses ensembling of the top-3 or 5 base learner models from one run. \\
\end{tablenotes}
\end{threeparttable}
\label{table:npbml-cifar100-experiments-1}
\end{table*}

\begin{table*}[t!]
\centering
\captionsetup{justification=centering}
\caption{Few-shot classification meta-testing accuracy on 5-way 1-shot and \\5-way 5-shot \textit{FC-100} where $\pm$ represents the 95\% confidence intervals.}
\begin{threeparttable}
\begin{tabular}{lccc}
\hline 
\multirow{2}{*}{Method}                             & \multirow{2}{*}{Base Learner}    & \multicolumn{2}{c}{FC-100 (5-way)}  \\ \cline{3-4} 
                                                    &               & 1-shot           & 5-shot                    \\ \hline \hline \noalign{\vskip 1mm}
MAML \citep{finn2017model}                          & 4-CONV & 35.89$\pm$0.72\% & 49.31$\pm$0.47\%          \\ \noalign{\vskip 1mm}
BOIL \citep{oh2020boil}                             & 4-CONV & 38.93$\pm$0.45\% & 51.66$\pm$0.32\%          \\ \noalign{\vskip 1mm}
MeTAL \citep{baik2021meta}$^{\ddagger}$             & 4-CONV & 37.46$\pm$0.39\% & 51.34$\pm$0.25\%          \\ \noalign{\vskip 1mm}
ALFA \citep{baik2023meta}$^{\ddagger}$              & 4-CONV & 37.99$\pm$0.48\% & 53.01$\pm$0.49\%          \\ \noalign{\vskip 1mm} 
\hdashline \noalign{\vskip 1mm}
NPBML (Ours)$^{\dagger}$                            & 4-CONV & \textbf{40.56$\pm$0.76\%} & \textbf{53.48$\pm$0.68\%}          \\ \noalign{\vskip 1mm} \hline \hline \noalign{\vskip 1mm}

MAML \citep{finn2017model}                          & ResNet-12     & 37.29$\pm$0.40\% & 50.70$\pm$0.35\%          \\ \noalign{\vskip 1mm}
MeTAL \citep{baik2021meta}$^{\ddagger}$             & ResNet-12     & 39.98$\pm$0.39\% & 53.85$\pm$0.36\%          \\ \noalign{\vskip 1mm} 
ALFA \citep{baik2023meta}$^{\ddagger}$              & ResNet-12     & 41.46$\pm$0.49\% & 55.82$\pm$0.50\%          \\ \noalign{\vskip 1mm} 
\hdashline \noalign{\vskip 1mm}
NPBML (Ours)$^{\dagger}$                            & ResNet-12     & \textbf{43.63$\pm$0.71\%} & \textbf{59.85$\pm$0.70\%}          \\ \noalign{\vskip 1mm} \hline \noalign{\vskip 1mm}

\end{tabular}
\begin{tablenotes}\centering\small
$\dagger$ Denotes a base learner that has been modified to include warp preconditioning layers. \\
$\ddagger$ Reported result uses ensembling of the top-3 or 5 base learner models from one run. \\
\end{tablenotes}
\end{threeparttable}
\label{table:npbml-cifar100-experiments-2}
\end{table*}

Next, we further validate the effectiveness of NPBML on two popular CIFAR-100 derivatives \citep{krizhevsky2009learning}: CIFAR-FS \citep{ravi2017optimization} and FC-100 \citep{ren18fewshotssl}. The results, presented in Tables \ref{table:npbml-cifar100-experiments-1} and \ref{table:npbml-cifar100-experiments-2}, show that NPBML continues to achieve strong and robust generalization performance across all settings and models. These results are particularly impressive, given that both MeTAL and ALFA ensemble the top 5 performing models from the same run, which significantly increases the model size and capacity. These experimental results reinforce our claim that meta-learning a task-adaptive update rule is an effective approach to improving the performance of MAML-based few-shot learning algorithms.

\subsection{Ablation Studies}

To further investigate the performance of NPBML, we conduct two sets of ablation studies to analyze the effectiveness of each component. These experiments are performed using the 4-CONV network architecture in a 5-way 5-shot setting on the \textit{mini-}ImageNet dataset.

\subsection{Ablation of Meta-Learned Components}

First, we examine the importance of the meta-learned optimizer $P_{\Warp}$, loss function $\MetaLoss_{\bm{\phi}}$, and task-adaptive conditioning method $FILM_{\bm{\psi}}$. The results are presented in Table \ref{table:npbml-ablation-study-1}, and they demonstrate that each of the proposed components clearly and significantly contributes to the performance of NPBML. In (2) MAML is modified to include gradient preconditioning, which increases accuracy by $2.09\%$. Conversely in (3) we modify MAML with our meta-learned loss function, resulting in a $6.37\%$ performance increase. Interestingly, the meta-learned loss function enhances performance by a larger margin; however, this may be due to the relatively simple T-Net style optimizer used in NPBML. This suggests that a more powerful parameterization, such as \citep{flennerhag2018transferring} or \citep{kang2023meta}, may further improve performance. In (4), MAML is modified to include both the optimizer and loss function, resulting in a $7.41\%$ performance increase. This further supports our claim that meta-learning both an optimizer and a loss function are complementary and orthogonal approaches to improving MAML. Finally, in (5), we add our task-adaptive conditioning method, increasing performance by $2.22\%$ over the prior experiment and $9.63\%$ over MAML.

\begin{table*}[t!]
\centering
\captionsetup{justification=centering}
\caption{Ablation study of the meta-learned components in NPBML, reporting the meta-testing accuracy on \textit{mini-ImageNet} 5-way 5-shot. A \checkmark denotes that the component is meta-learned, with variant (1) reducing to MAML, while variant (5) represents our final proposed algorithm.}
\begin{tabular*}{\textwidth}{@{\extracolsep{\fill}} cccccc}
\hline \noalign{\vskip 1mm}
           &  Initialization & Optimizer   & Loss Function & Task-Adaptive & Accuracy             \\ \noalign{\vskip 1mm} 
           \hline \hline \noalign{\vskip 1mm}
(1)        & \checkmark      &             &               &               & 65.38$\pm$0.67\%     \\ \noalign{\vskip 1mm}
(2)        & \checkmark      & \checkmark  &               &               & 67.47$\pm$0.68\%     \\ \noalign{\vskip 1mm}
(3)        & \checkmark      &             & \checkmark    &               & 71.75$\pm$0.69\%     \\ \noalign{\vskip 1mm}
(4)        & \checkmark      & \checkmark  & \checkmark    &               & 72.79$\pm$0.67\%     \\ \noalign{\vskip 1mm}
\hdashline \noalign{\vskip 1mm}
(5)        & \checkmark      & \checkmark  & \checkmark    & \checkmark    & \textbf{75.01$\pm$0.64\%}     \\ \noalign{\vskip 1mm} 
\hline
\end{tabular*}
\label{table:npbml-ablation-study-1}
\end{table*}

\subsection{Ablation of Meta-Learned Loss Function}

The prior ablation study shows that the meta-learned loss function $\smash{\MetaLoss_{\bm{\phi}}}$ is a crucial component in NPBML. Therefore, we further investigate each of the components; namely, the meta-learned inductive and transductive loss functions, and weight regularizer. The results are presented in Table \ref{table:npbml-ablation-study-2}, and surprisingly, they show that each of the components in isolation (7), (8), and (9), improves performance by approximately $5\%$. However, when combined in (10), the total performance increase is $6.37\%$. We hypothesize that this result is a consequence of the implicit meta-learning of the learning rate identified in Equation \eqref{eq:npbml-implicit-scalar-lr-tuning}, which not only holds for $\smash{\MetaLoss_{\bm{\phi}}}$, but also for each of its components, \textit{i.e.}, the equality is also true when $\smash{\MetaLoss_{\bm{\phi}}}$ is replaced with $\smash{\Loss^S_{\bm{\phi}}}$, $\smash{\Loss^Q_{\bm{\phi}}}$, or $\smash{\Regularize_{\bm{\phi}}}$. Since all components share implicit learning rate tuning, the performance gains from this behavior do not accumulate; however, the improvement in (10) is better than each component in isolation indicating that each component provides additional unique benefits to the meta-learning process.

\begin{table*}[t!]
\centering
\captionsetup{justification=centering}
\caption{Ablation study of the meta-learned loss function $\MetaLoss_{\phi}$ in NPBML, reporting \\the meta-testing accuracy on \textit{mini-ImageNet} 5-way 5-shot. Note, variants (6) and \\(10) correspond to variants (1) and (3), respectively in Table \ref{table:npbml-ablation-study-1}.}
\resizebox{\textwidth}{!}{%
\begin{tabular}{cccccc}
\hline \noalign{\vskip 1mm}
           &  Base Loss & Inductive Loss   & Transductive Loss & Weight Regularizer & Accuracy    \\ \noalign{\vskip 1mm} 
           \hline \hline \noalign{\vskip 1mm}
(6)        & \checkmark      &             &               &               & 65.38$\pm$0.67\%     \\ \noalign{\vskip 1mm}
(7)        & \checkmark      & \checkmark  &               &               & 70.68$\pm$0.66\%     \\ \noalign{\vskip 1mm}
(8)        & \checkmark      &             & \checkmark    &               & 70.92$\pm$0.68\%     \\ \noalign{\vskip 1mm}
(9)        & \checkmark      &             &               & \checkmark    & 70.04$\pm$0.65\%     \\ \noalign{\vskip 1mm} 
\hdashline \noalign{\vskip 1mm}
(10)       & \checkmark      & \checkmark  & \checkmark   & \checkmark     & \textbf{71.75$\pm$0.69\%}     \\ \noalign{\vskip 1mm} 
\hline
\end{tabular}
}
\label{table:npbml-ablation-study-2}
\end{table*}

\section{Chapter Summary}
\label{section:npbml-conclusion}

In this chapter, we propose a novel meta-learning framework for learning the procedural biases of a deep neural network. The proposed technique, \textit{Neural Procedural Bias Meta-Learning} (NPBML), consolidates recent advancements in MAML-based few-shot learning methods by replacing the fixed inner update rule with a fully meta-learned update rule. This is achieved by meta-learning a task-adaptive loss function, optimizer, and parameter initialization. The experimental results confirm the effectiveness and scalability of the proposed approach, demonstrating strong few-shot learning performance across a range of popular benchmarks. We believe NPBML provides a principled framework for advancing general-purpose meta-learning in deep neural networks. Looking ahead, numerous compelling future research directions exist, such as developing more powerful parameterizations for the meta-learned optimizer or loss function. We expect that further investigation of this topic will result in more expressive inner update rules, resulting in increased robustness and efficiency within the context of optimization-based meta-learning. Finally, broadening the scope of the proposed framework to encompass the related domains of cross-domain few-shot learning and continual learning may be a promising avenue for future exploration.

\chapter{Conclusions and Future Work}\label{chapter:conclusion}

\textit{In this final chapter, we conclude this thesis by revisiting the objectives and highlighting the main findings from each of the contributions. We discuss how each of the chapters contributed to the progression of the overall goal of this thesis, which was to enhance the learning capabilities of deep neural networks by advancing the emerging field of meta-learning, a field devoted to leveraging past learning experiences to improve the overall learning performance and sample efficiency of deep neural networks. Following this, we close out this thesis by giving guidance and several possible directions for future research into meta-learning loss functions.}

\section{Contributions}

This thesis has achieved the following key research objectives:

\begin{itemize}

    \item Developed a novel framework for meta-learning interpretable symbolic loss functions called \textit{Evolved Model-Agnostic Loss} (EvoMAL). This method consolidated prior approaches to loss function learning by integrating gradient-based and evolution-based methods into a unified framework. We achieved this by using genetic programming to discover the symbolic structure of the loss function and then employed unrolled differentiation to optimize its coefficients. Our experimental results demonstrated that models trained using our meta-learned loss functions showed enhanced sample efficiency and asymptotic performance compared to handcrafted loss functions and previous loss function learning techniques across numerous supervised learning tasks and model architectures. Additionally, our algorithm was shown to have a significantly shorter runtime compared to prior loss function learning algorithms employing a two-stage meta-learning approach to loss functions.
    
    \item Conducted an in-depth analysis of the meta-learned loss functions developed by EvoMAL to shed insight into how and why meta-learned loss functions can attain increased performance compared to their handcrafted counterparts. Contrary to prior hypotheses, our analysis revealed that additional factors beyond those originally conjectured contribute to this performance disparity. Through theoretical examination of the \textit{Absolute Cross-Entropy} (ACE) Loss, a loss function developed by EvoMAL, we identified a form of label smoothing as another significant factor influencing performance enhancement. This finding inspired the development of \textit{Sparse Label Smoothing Regularization} (SparseLSR) loss, a method we proposed that is significantly faster to compute relative to non-sparse label smoothing, with the time and space complexity being reduced from linear to constant time with respect to the number of classes being considered. This analysis provided important insights into meta-learned loss functions.
    
    \item Developed \textit{Adaptive Loss Function Learning} (AdaLFL), the first method for meta-learning adaptive loss functions that dynamically evolve throughout the learning process. This approach challenged the conventional notion that a loss function must be a static function, by demonstrating that the loss function can be inferred directly from data and adaptively trained alongside the base model’s parameters using online unrolled differentiation. Our experimental findings showed that models trained with AdaLFL exhibit enhanced convergence capabilities and inference performance compared with the squared error and cross-entropy loss, and offline loss function learning methods. Additionally, further analysis of the learned loss functions also identified recurring patterns in the shape of the learned loss functions and revealed unique emergent behavior exclusively present in adaptively learned loss functions; namely, implicit tuning of the learning rate schedule and early stopping regularization.
    
    \item Developed \textit{Neural Procedural Bias Meta-Learning} (NPBML), a task-adaptive method for simultaneously meta-learning the parameter initialization, optimizer, and loss function of a deep neural network. This approach integrated advancements in meta-learning loss functions with the ongoing research on meta-learning parameter initialization and gradient-based optimizers. This was achieved by consolidating previous loss function learning methods with MAML-based learned parameter initializations and meta-learned preconditioning layers. Furthermore, task adaptivity was achieved through feature-wise linear modulation layers, which allowed the learned components to adapt to individual tasks. Our experimental results showed that NPBML outperforms many state-of-the-art meta-learning methods across a range of popular few-shot learning benchmarks. This objective demonstrates the potential of embedding strong procedural biases into deep neural networks by meta-learning the parameter initialization, optimizer, and loss function, thus enabling robust learning performance informed by prior learning experiences.
    
\end{itemize}

\section{Conclusions}

In this thesis, we have demonstrated that meta-learned loss functions are an effective way to enhance the performance of deep neural networks. Our findings show that the deep neural networks trained with meta-learned loss functions consistently demonstrate improved generalization, convergence, and sample efficiency compared to those trained with conventional handcrafted loss functions. This section provides a detailed discussion of the main conclusions and key takeaways corresponding to each of the research objectives covered in the four contribution chapters (Chapters \ref{chapter:evomal}, \ref{chapter:theory}, \ref{chapter:adalfl}, and \ref{chapter:npbml}).

\subsection{Meta-Learning Symbolic Loss Function}

In Chapter \ref{chapter:evomal}, we introduced Evolved Model-Agnostic Loss (EvoMAL), an algorithm designed for meta-learning interpretable symbolic loss functions. We emphasize two significant findings from this chapter: (1) the advantages of using a hybrid approach to loss function learning, and (2) the benefits of meta-learning interpretable loss functions over black-box sub-symbolic loss functions.

\subsubsection{Hybrid Approach to Loss Function Learning}

This thesis finds that by employing genetic programming to infer the loss function structure and unrolled differentiation to optimize its coefficients, highly performant symbolic loss functions can be meta-learned on commodity hardware. This approach represents an advancement over previous state-of-the-art loss function learning techniques. Past methods either omitted coefficient optimization, leading to poor convergence and suboptimal final inference performance, or included coefficient optimization but relied on costly evolution-based methods. Methods that rely on evolution-based methods to optimize the coefficients require an exponential number of loss function evaluations and require a supercomputer to finish in tractable runtime. In contrast, EvoMAL can meta-learn loss functions on a single consumer-grade GPU.

\subsubsection{Meta-Learning Interpretable Loss Functions}

Another key finding from this research was the significance of meta-learning interpretable and symbolic loss functions. While some methods opt to reduce the runtime of loss function learning by not learning the structure of the loss function, this renders them challenging, if not impossible to formally study and analyze. In contrast, the loss functions meta-learned by EvoMAL are both interpretable and symbolic, rendering them relatively straightforward to analyze. As demonstrated in the subsequent chapter (Chapter \ref{chapter:theory}), meaningful results can be derived through analysis, and these insights can be leveraged to enhance widely used handcrafted loss functions.

\subsection{Analysis of Meta-Learned Loss Functions}

In Chapter \ref{chapter:theory}, we analyzed the learned loss functions developed by EvoMAL to better understand the mechanisms underlying their superior performance in comparison to handcrafted counterparts. Below, we highlight two significant findings from this analysis: (1) a revised hypothesis regarding what meta-learned loss functions learn, and (2) the deep connections between meta-learned loss functions and label smoothing regularization.

\subsubsection{What do Meta-Learned Loss Functions Learn}

This thesis identifies three key factors contributing to enhanced performance when using meta-learned loss functions for training. Firstly, meta-learned loss functions automatically calibrate the robustness of the loss, such that inliers and outliers are penalized appropriately with respect to the target task. Secondly, meta-learned loss functions not only learn shape they also learn scale; consequently, implicitly tuning the base learning rate of an optimizer. Finally, as shown by our theoretical analysis, meta-learned loss functions often incorporate label smoothing regularization on classification tasks, a powerful strategy for avoiding overconfidence in deep neural networks.

\subsubsection{Sparse Label Smoothing Regularization}

In our theoretical analysis, we examined the Absolute Cross-Entropy (ACE) loss, one of the learned loss functions developed by EvoMAL. The analysis revealed deep connections between the ACE loss and label smoothing regularization. Notably, it inspired a trick we later referred to as the “\textit{redistributed loss trick}”, a method for redistributing the non-target losses into the target greatly reducing the time and memory needed to compute label smoothing regularization. Based on this insight we developed Sparse Label Smoothing Regularization (SparseLSR) loss, a loss function that performs similarly to label smoothing regularization but is sparse and only needs to be computed on the target output. Our results show that the proposed loss function is significantly faster to calculate, with the time and space complexity being reduced from linear to constant with respect to the number of classes being considered.

\subsection{Meta-Learning Adaptive Loss Functions}

In Chapter \ref{chapter:adalfl}, we developed Adaptive Loss Function Learning (AdaLFL), a novel method for meta-learning loss functions that dynamically adapt throughout the learning process. In this chapter, there are two key findings that we would like to highlight: (1) the performance benefits of using adaptive loss functions as opposed to static ones, and (2) the emergent behaviors associated with meta-learning adaptive loss functions.

\subsubsection{Adaptive Loss Function}

This thesis finds that the loss function need not remain static, rather it can dynamically adapt alongside the base model throughout the learning process. This was accomplished by proposing an online unrolled differentiation algorithm for performing meta-optimization. By synchronously updating the meta parameters of the loss functions with those of the base model, this method mitigated the prevalent issue of short-horizon bias observed in offline loss function learning. Short-horizon bias in offline loss function learning causes the performance of meta-learned loss functions to degrade near the end of the learning process. The experimental results confirmed that adaptive loss functions consistently outperform both handcrafted loss functions and offline loss function learning techniques on a range of classification and regression tasks. The findings from this work suggest that adaptive loss functions are a promising direction for improving the performance of deep neural networks.

\subsubsection{Implicit Meta-Learning}

In the analysis of adaptive loss functions developed by AdaLFL, it was determined that although only the loss function was explicitly meta-learned, there existed an implicit (unintentional) meta-learning of other crucial procedural biases. Specifically, the learning rate and its schedule, early stopping regularization, and label smoothing regularization were also observed to be meta-learned. This unanticipated finding suggests that by meta-learning just the loss function, we can indirectly meta-learn other key components of a learning algorithm. Consequently, loss function learning subsumes the associated subfields interested in meta-learning learning rates and their schedules. Furthermore, this work highlights a path forward for meta-learning multiple of the procedural biases of a deep neural network, an area of increasing interest within the meta-learning community.

\subsection{Meta-Learning Neural Procedural Biases}

In Chapter \ref{chapter:npbml}, we proposed Neural Procedural Bias Meta-Learning (NPBML), a method designed to meta-learn the procedural biases of a deep neural network. From this work, there are two significant findings that we would like to highlight: (1) by meta-learning the procedural biases of neural networks we can induce strong inductive biases into our learning algorithm toward a specific distribution of learning tasks, and (2) by making the learned components adapt to each task performance can further be enhanced.

\subsubsection{Meta-Learning Procedural Biases of Deep Neural Networks}

This thesis demonstrates the potential to automate the design and selection of the loss function, optimizer, and parameter initialization simultaneously using optimization-based meta-learning. Our proposed method extends existing MAML-based meta-learning methods by incorporating learning gradient preconditioning layers and the base loss function. These components are optimized end-to-end simultaneously by unrolled differentiation, where the stored optimization path can be reused for each component. By meta-learning the procedural biases of deep neural networks, robust inductive biases can be embedded into the model. These powerful inductive biases learned from past learning experiences enable NPBML to attain very competitive few-shot learning performance. The experimental results show that the proposed method outperforms many state-of-the-art meta-learning methods on a variety of few-shot learning tasks.

\subsubsection{Task Adaptive Meta-Learning}

This thesis finds that an effective way to improve meta-learning performance is to adapt each task's meta-representation, \textit{i.e.}, learned loss function, optimizer, and parameter initialization. This was achieved via the use of feature-wise linear modulation (FiLM) layers, which were preconditioned on task-related information. The parameters of the FiLM layers were meta-learned at the same time as the parameters of the meta-learned representation (in the outer loop). The experimental results show task adaptation of the meta-learned representation is critical for obtaining strong performance in few-shot learning. The findings from this contribution suggest that the more you tailor the meta-representation toward a task the better the corresponding performance will be.

\section{Future Work}

Throughout this thesis, hands-on experience has been gained with various meta-learning algorithms. This section aims to highlight key research directions identified throughout our work. In particular, we aim to bring to the attention of the reader several topics that we believe warrant further investigation and that are likely to be a promising direction of research with high impact.

\subsection{Rethinking the Loss Function}

The loss function is typically considered to be a simple static function $\Loss(y, f_{\theta}(x)): \mathbb{R}^2 \rightarrow \mathbb{R}^1$, that takes as input the ground truth $y$ and the base model predictions $f_{\theta}(x)$ and returns a corresponding value which measures the corresponding accuracy or inaccuracy. However, as demonstrated in this thesis, this need not be the case; the loss function can be far more complex and powerful. For instance, it can be made adaptive (as discussed in Chapter \ref{chapter:adalfl}) or conditioned upon auxiliary task information, such as unlabeled instances or the weights of the base model (as explored in Chapter \ref{chapter:npbml}).

Regarding future research directions, many aspects of meta-learned loss functions remain unexplored, presenting numerous opportunities for creativity and innovation. For instance, learned loss functions could be extended to apply uniquely to each instance or output of a task. Moreover, the meta-representation of the loss function could be significantly enhanced beyond the simple feed-forward networks employed in Chapters \ref{chapter:adalfl} and \ref{chapter:npbml}. As illustrated in \citep{antoniou2019learning}, the authors investigated using a large dilated convolutional neural network \citep{yu2015multi, oord2016wavenet} with DenseNet-style residual connectivity \citep{huang2017densely}. Finally, a meta-learned loss function could be leveraged to enhance model distillation \citep{hinton2015distilling} between a teacher network and a student network by considering the intricate interplay between performance on the target task and behavior distillation between the student and the teacher.

\subsection{Towards Structured Meta-Representations}

In Chapters \ref{chapter:evomal}, \ref{chapter:adalfl}, and \ref{chapter:npbml}, we utilized expressive meta-representations to represent the meta-learned loss functions, such as genetic programming expression trees and neural networks. While these unstructured representations are powerful and can capture many unique behaviors; unfortunately, they have some fundamental drawbacks. Specifically, regulating and constraining the behavior of the loss functions is challenging, as the representation distributes the different behaviors of the loss function across the representation. Due to these behaviors being entangled across many parameters, it is difficult to enforce essential learning constraints, which can sometimes lead to suboptimal performance.

Our research has revealed that the learned loss functions developed by EvoMAL, AdaLFL, and NPBML implicitly learn the learning rate, learning rate schedule, early stopping regularization, label smoothing, and more. While this implicit learning is advantageous, it also increases the number of failure modes of the proposed methods. To mitigate these failure modes, it is imperative to regulate the behavior --- for instance, by bounding the learning rate within a stable range or avoiding early stopping before a predetermined number of gradient steps. However, due to the meta-representation of the loss functions being unstructured enforcing desirable behavior is not straightforward.

A promising direction for future work lies in exploring more structured representations in loss function learning. These representations should aim to express similar shapes as meta-learned loss functions while enabling more explicit control over their behavior. An apt exemplar of this research direction is the general and robust regression loss function proposed by \cite{barron2019general}, which can express many fundamental regression loss functions while also being straightforward to constrain and regulate. For instance, if only one parameter is utilized to control the scale of the loss function, then the feasible region can straightforwardly be constrained via a sigmoidal function during meta-learning.

\subsection{Improving Meta-Optimization}

While not directly within the scope of this thesis, meta-optimization is an extremely important field for advancing meta-learning. Throughout our research, we extensively employed unrolled differentiation \citep{maclaurin2015gradient} to optimize the parameters of our learned loss functions. Unrolled differentiation works by taking a fixed number of steps and then computing the loss with respect to the meta-parameter. This requires differentiating the optimization path by propagating gradients backward through the entire learning trajectory. Despite its power, storing the whole trajectory in memory and differentiating it incurs significant computational and memory costs. To mitigate this challenge, only a few steps are generally taken; this causes bias toward fast adaptation and short horizons, often resulting in suboptimal performance for medium and long-shot learning problems.

Several methods have been proposed to address the limitations of unrolled differentiation, such as truncated unrolled differentiation \citep{shaban2019truncated}, implicit differentiation \citep{lorraine2020optimizing}, EvoGrad \citep{bohdal2021evograd}, etc. However, while these approaches tackle computational complexity and memory issues, they do not resolve the bias toward short horizons. For future research direction, trajectory-agnostic meta-optimization techniques such as those proposed by \cite{flennerhag2018transferring, flennerhag2019meta, flennerhag2024optimistic}, hold immense promise. This is because they decouple the learning objective away from the trajectory. This is vital for enabling meta-learning techniques to be applied in continual \citep{de2021continual, wang2024comprehensive} and life-long learning \citep{parisi2019continual} settings, another important substrate of intelligence closely related to the field of meta-learning.

\subsection{Beyond Meta Learned Loss Functions}

To ensure the long-term success and viability of loss function learning, it is essential for future research to explore the applicability and integration of meta-learned loss functions within the broader context of meta-learning. This is because learning the loss function in isolation is not the ultimate goal of meta-learning as a paradigm. The loss function represents just one of many important components in a learning algorithm. While studying meta-learned loss function isolation is important for understanding their behavior, the true potential of loss function learning can only be fully realized by integrating it with other meta-learning paradigms, such as learned optimizers, parameter initialization, and more.

In Chapter 6, we proposed NPBML, the first method to attempt to combine learned loss functions with learned optimizers and parameter initializations. The proposed method demonstrated impressive few-shot learning performance and demonstrated the potential synergy between meta-learned loss functions and other learned components. However, certain aspects were left unexplored in that work. These include enabling continual and lifelong learning \citep{parisi2019continual, de2021continual}, compatibility with non-convolutional neural networks like recurrent neural networks \citep{rumelhart1986learning} and transformers \citep{vaswani2017attention}, and more. Furthermore, it would be advantageous and interesting to explore alternative application domains such as novel view synthesis, e.g., NeRF \citep{mildenhall2021nerf} and Gaussian Splattering \citep{kerbl20233d}, and reinforcement learning \citep{beck2023survey}.


\addcontentsline{toc}{chapter}{Bibliography}

\bibliographystyle{style}
\bibliography{references}

\end{document}